%% file: neurips_2026.tex
\documentclass{article}

\usepackage{amsmath}
\usepackage{amssymb}
\usepackage{mathtools}
\usepackage{amsthm}
\usepackage{pifont}
\usepackage{multirow}
\usepackage{makecell}
\usepackage{multicol}
\usepackage{enumitem}
\usepackage{arydshln}
\usepackage{algorithm}
\usepackage{algpseudocode}
\usepackage[table]{xcolor}
\usepackage{array}

\definecolor{fatal}{HTML}{aa301a}
\definecolor{ideal}{HTML}{639346}
\definecolor{deadlock}{HTML}{523080}
\definecolor{convergence}{HTML}{335d9f}
\definecolor{core}{HTML}{fecf78}

\makeatletter
\define@key{mcb}{text}{\def\mcb@text{#1}}
\define@key{mcb}{color}{\def\mcb@color{#1}}
\newcommand{\mycolorbox}[1][]{%
  \def\mcb@text{}%
  \def\mcb@color{}%
  \setkeys{mcb}{#1}%
  \ifx\mcb@color\empty
    \mbox{\mcb@text}%
  \else
    \colorbox{\mcb@color}{\mcb@text}%
  \fi
}
\makeatother

\newcommand{\ie}{\textit{i}.\textit{e}.}
\newcommand{\eg}{\textit{e}.\textit{g}.}

\newcommand{\normlp}{\ensuremath{\mathrm{NormLP}}}
\newcommand{\tokenprobe}{\ensuremath{\mathrm{TokenProbe}}}
\newcommand{\thinksep}{\texttt{</think>}}

\usepackage{listings}
\definecolor{rit}{RGB}{241, 105, 33}
\definecolor{pykw}{HTML}{0033B3}
\definecolor{pycom}{HTML}{8C8C8C}
\definecolor{pystr}{HTML}{067D17}
\definecolor{pybg}{HTML}{F7F7F7}

\lstdefinestyle{pythonpseudo}{
  language=Python,
  basicstyle=\ttfamily\footnotesize,
  keywordstyle=\color{pykw}\bfseries,
  commentstyle=\color{pycom}\itshape,
  stringstyle=\color{pystr},
  showstringspaces=false,
  backgroundcolor=\color{pybg},
  breaklines=true,
  frame=single,
  rulecolor=\color{black!30},
  xleftmargin=1em,
  framexleftmargin=0.5em,
  columns=fullflexible,
  keepspaces=true,
  morekeywords={Tensor,torch,nn,F,where,zeros,argmax,clamp,arange,
                mean,std,maximum,minimum,bool,float,inf,True,False,None},
}

\usepackage[most,skins,theorems]{tcolorbox}
\tcbset{
  aibox/.style={
    width=\linewidth,
    top=8pt,
    bottom=4pt,
    colback=blue!6!white,
    colframe=black,
    colbacktitle=black,
    enhanced,
    center,
    attach boxed title to top left={yshift=-0.1in,xshift=0.15in},
    boxed title style={boxrule=0pt,colframe=white,},
  }
}

\newtcolorbox{AIbox}[2][]{aibox,title=#2,#1}

 \usepackage[main, final]{neurips_2026}

\usepackage[utf8]{inputenc}
\usepackage[T1]{fontenc}
\usepackage{hyperref}
\usepackage{url}
\usepackage{tocloft}

\usepackage{booktabs}
\usepackage{amsfonts}
\usepackage{nicefrac}
\usepackage{microtype}
\usepackage{xcolor}
\theoremstyle{definition}
\newtheorem{definition}{Definition}
\title{On the Token Value Inequality \\ in Efficient Reasoning}

\author{%
\thead{
  Runjia Zeng$^{\textbf{1, 6}}$, 
  Hang Hua$^{\textbf{2}}$, 
  Yiyang Liu$^{\textbf{3}}$, 
  Zhiqiang Tao$^{\textbf{1}}$, 
  Ruixiang Tang$^{\textbf{4}}$, \\
  \textbf{Qifan Wang}$^{\textbf{5}}$, 
  \textbf{Cheng Han}$^{\textbf{3}}$, 
  \textbf{Dongfang Liu}$^{\textbf{6} \dagger}$~\hspace{5pt}}\vspace{0.1in}
  \\ 
$^{1}$Rochester Institute of Technology ~\hspace{5pt} $^{2}$MIT-IBM Watson AI Lab \\ $^{3}$University of Missouri-Kansas City ~\hspace{5pt} $^{4}$Rutgers University  \\ $^{5}$Meta AI ~\hspace{5pt} $^{6}$Purdue University ~\hspace{5pt} $^{\dagger}$Corresponding author
}

\begin{document}
\raggedbottom
\addtocontents{toc}{\protect\setcounter{tocdepth}{-1}}

\maketitle

\vspace{-0.4cm}
\begin{abstract}
\vspace{-0.2cm}
Chain-of-Thought reasoning has enabled large language models to achieve substantial performance gains on complex tasks. However, these gains come at the cost of dramatically increased token consumption. This raises a fundamental question: is every token in the reasoning trace equally valuable? We present a diagnostic and optimization framework grounded in a key empirical finding: the value of tokens within a CoT reasoning sequence is highly non-uniform, and this non-uniformity can be effectively characterized by token-level log probability signals. We show that normalized log probability helps distinguish core tokens, which carry structural and decisive reasoning content, from redundant tokens, which are exploratory, low-confidence filler that contributes less directly to the final answer. Building on these findings, we formulate the \textbf{TokenProbe} framework around two empirical findings and one claim: findings identify token value inequality first and then establish TokenProbe as a core-token proxy, and the claim introduces an efficient GRPO objective positing that selectively compressing redundant tokens can yield Pareto improvements in the accuracy–token efficiency space. Empirically, our method preserves reasoning quality while reducing the token usage by 76\% of the baseline. Under matched reasoning-length budgets, we show that it can even outperform strong flagship baselines like Gemini-3.1-Pro. \textcolor{rit}{Homepage: \href{https://runjia.tech/tokenprobe/}{runjia.tech/tokenprobe}}.
\end{abstract}

\vspace{-0.5cm}
\section{Introduction}\label{sec:intro}
\vspace{-0.2cm}
The training objective of an autoregressive language model~\cite{bengio2003neural,radford2018improving,brown2020language} is to maximise the log-likelihood of the observed token sequence, $\mathcal{L} = \sum_{t=1}^{T} \log p_\theta\!\bigl(s_t \mid s_{<t},\, x\bigr)$. Under this objective, when the model faces a complex reasoning problem, directly predicting the answer yields a high-entropy conditional distribution, the model is uncertain about the answer, and $p(y\mid x)$ spreads across many candidates. Chain-of-Thought (CoT) reasoning~\cite{wei2022chain,kojima2022large} addresses this by introducing intermediate reasoning steps $z$ that decompose the prediction as $p_\theta(y\mid x) = \sum_{z} p_\theta(z\mid x) \, p_\theta(y\mid x,z)$. Each reasoning step $z_i$ narrows the solution space conditioned on its predecessors, substantially improving accuracy on multi-step reasoning tasks by explicitly introducing latent variables that reduce the dimensionality of the prediction problem~\cite{zelikman2022star,yao2023tree,hao2024training}.

\begin{figure*}[t]
  \centering
  \vspace{-0.8cm}
  \includegraphics[width=0.99\textwidth]{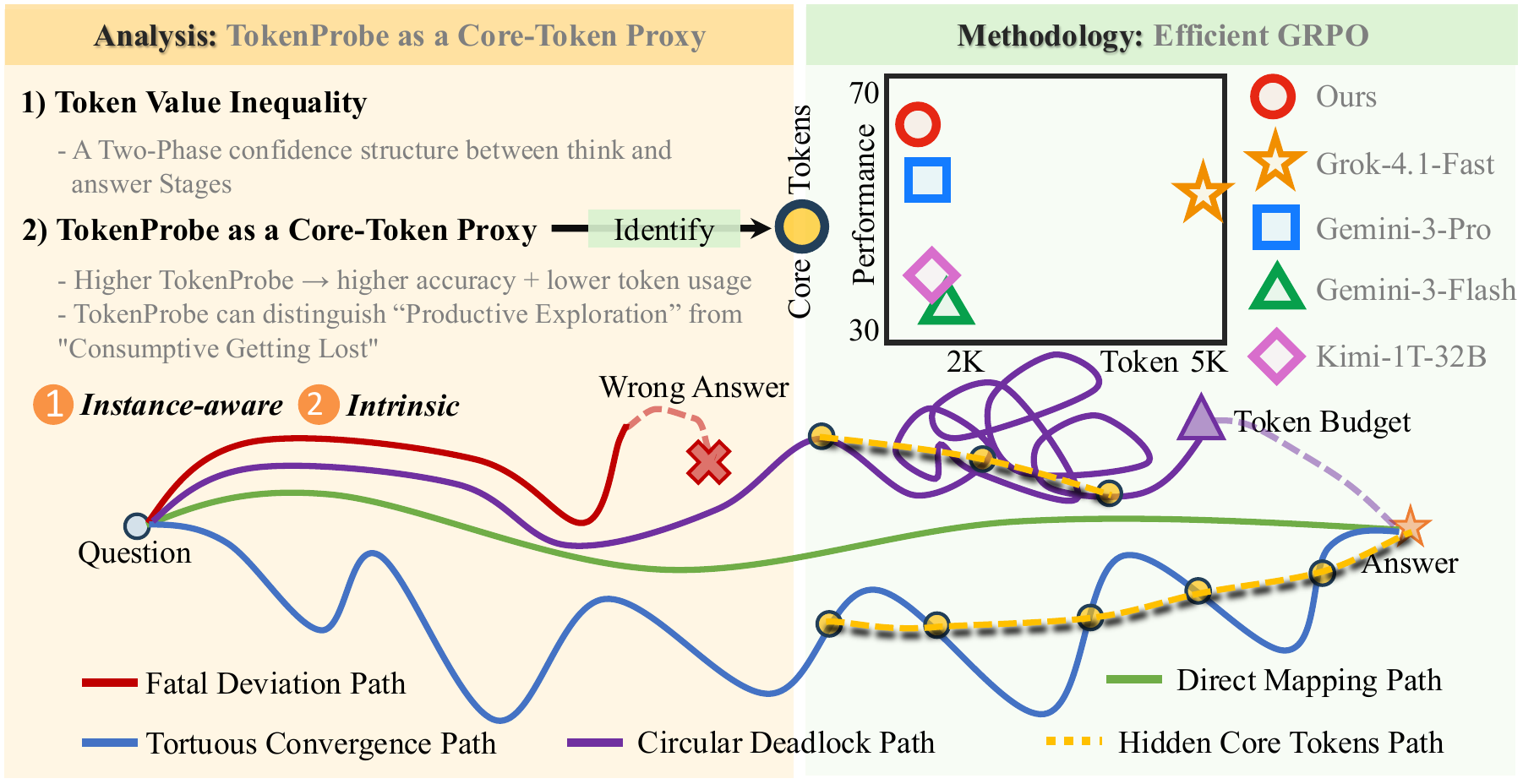}
  \vspace{-0.2cm}
  \caption{\textbf{Overview of TokenProbe: from token value inequality analysis \S\ref{sec: analysis} to efficient reasoning \S\ref{sec: method}}. (a) Probing analysis shows that token contributions are highly unequal, and TokenProbe serves as an proxy for identifying core tokens. (b) Building on this insight, efficient GRPO improves it by preserving hidden core tokens, yielding better reasoning efficiency and Pareto performance.}
  \label{fig:teaser}
  \vspace{-0.6cm}
\end{figure*}

\vspace{-0.1cm}
However, the introduction of latent variables and reasoning steps brings the extreme token usage spurge along the performance gain~\cite{chen2024not,sui2025stop,avogaro2026sparc,arora2025training,sun2025latent} (\ie, enabling CoT on Qwen-4B adds 24.1\% in accuracy at a cost of 4.5$\times$ more tokens). As shown in the Fig. \ref{fig:teaser}, CoT reasoning currently exhibits several characteristic patterns. The \textcolor{ideal}{\textbf{green curve}} represents the ideal reasoning trajectory, where the model moves from the question to the correct answer through a relatively direct semantic mapping. In contrast, the \textcolor{fatal}{\textbf{red curve}} illustrates a fatal deviation path, in which one decisive wrong directional exploration drifts toward an incorrect conclusion. The \textcolor{deadlock}{\textbf{purple curve}} depicts a circular deadlock path, where the model expends substantial token budget while becoming trapped in repetitive or unproductive loops~\cite{holtzman2019curious}. The \textcolor{convergence}{\textbf{blue curve}} corresponds to a tortuous convergence path, where the model eventually reaches the correct answer, but only after a long and unnecessarily oscillatory reasoning process. This raises a central question: \emph{Are these tokens uniformly valuable, especially along the \textcolor{deadlock}{\textbf{purple curve}} and the \textcolor{convergence}{\textbf{blue curve}}?}  If not, it means a substantial fraction of the extra tokens is low-value exploratory redundancy, there exists the possibility of a \textbf{Pareto improvement}: preserving core reasoning content while drastically cutting redundant tokens, thereby greatly reducing the cost of model using since all models are token-based billing under the next token prediction paradigm~\cite{kaplan2020scaling}.

\vspace{-0.1cm}
Answering this question requires a suitable token-granularity diagnostic signal. Concretely, such a signal would arise as an inference-native by-product of generation, without relying on additional reward models~\cite{ouyang2022training,lightman2023let,wang2024math} or human annotation, thus avoiding post-processing overhead and extra forward passes. It would also preferably offer finer granularity, allowing more flexible intra-sequence discrimination than sequence-level or turn-level scores. Motivated by this, we naturally turn to the autoregressive model’s own output, \textbf{token-level log probability}~\cite{wang2025beyond,wang2024chain,yang2025reasonflux}.  It is a direct product of the training objective, already present at every decoding step, and its relative variation within a sequence encodes precisely the model's local confidence dynamics.

\vspace{-0.1cm}
Building on this raw token-level signal, we introduce two derived quantities that play complementary roles. The Normalized Log Probability (\normlp{}) highlights candidate decisive tokens within a response, while the TokenProbe aggregates this signal into a scalar that tracks the model's accuracy and token usage. They jointly drive our efficient GRPO (see \S\ref{sec: method}) and explain \emph{why} it yields Pareto improvements. Two properties make this lever sharp.
\ding{182}~\textbf{Instance-aware.} By normalization, \normlp{} is a relative signal within each response. It identifies which tokens have confidence above the response mean (positive \normlp{}) and which fall below (negative \normlp{}), therefore distinguishing the core tokens flexibly and dynamically (see Tab.~\ref{tab:wait-analysis}), rather than relying on absolute magnitudes. Building on this contrastive signal, we further define the TokenProbe of a response as the ratio of its tokens with positive \normlp{}.
\ding{183}~\textbf{Intrinsic.} For a given model, the TokenProbe is relatively stable and robust to hyperparameter variations (see \S\ref{asec: probing_analysis}). It therefore reflects the model's intrinsic degree of certainty and training characteristics without additional processing or annotations, and in turn helps evaluate the model's reasoning ability (see Fig.~\ref{fig: tokenprobe}), including both its accuracy and its token usage.

\vspace{-0.1cm}
The remainder of the paper is structured as follows. In \S\ref{sec: analysis}, we conduct a systematic probing analysis of token-level log probability, formally introducing the \normlp{} and TokenProbe signals. Through two key findings, we establish that CoT reasoning exhibits a two-phase confidence structure (\textbf{Finding 1}) and that TokenProbe is consistently associated with both reasoning accuracy and token usage (\textbf{Finding 2}), laying the empirical foundation for our approach. Building on these insights, \S\ref{sec: method} presents our efficient GRPO as a \textbf{Claim}, which translates the NormLP-based analysis into practice through two complementary mechanisms, \textit{Selective KL Anchoring}, which focuses regularization on core tokens identified by NormLP, and \textit{Reward Shaping}, which incentivizes dense and concise reasoning trajectories. \S\ref{experiments} reports comprehensive experiments across multiple benchmarks and model families, showing a consistent trend toward Pareto improvement while maintaining accuracy and substantially reducing token usage. We close in \S\ref{sec: discussion} with a detailed discussion of training dynamics, out-of-domain efficiency transfer, and diagnostic ablation.

\vspace{-0.2cm}
\section{Analysis}\label{sec: analysis}
\vspace{-0.2cm}
Since raw log probabilities vary in scale across responses, making direct comparison unreliable, we normalize them within each response to obtain \normlp{}
, and further aggregate it into a single scalar, TokenProbe, that stably reflects the model's intrinsic reasoning confidence. Formally:
\vspace{-0.1cm}
\begin{definition}[Normalized Log Probability]
\label{def:normlp}
Given a response consisting of tokens $\{s_1,\dots,s_T\}$ with per-token
log probabilities $\ell_t = \log p_\theta(s_t\mid s_{<t})$, the
normalized log probability of token $t$ is
\begin{equation}\label{eq:normlp}
  \normlp{}_t = \frac{\ell_t - \bar\ell}{\sigma_\ell + \epsilon},
  \quad\text{where}\quad
  \bar\ell = \frac{1}{T}\sum\nolimits_{t=1}^T \ell_t,\quad
  \sigma_\ell = \sqrt{\frac{1}{T}\sum\nolimits_{t=1}^T(\ell_t-\bar\ell)^2}.
\end{equation}
\end{definition}

\begin{definition}[TokenProbe, Positive Ratio of Normalized Log Probability]
\label{def:TokenProbe}
We define \textbf{TokenProbe} as the ratio of
tokens whose normalized log probability is positive, where $\mathbb{I}[\cdot]$ is the indicator function.
\vspace{-0.2cm}
\begin{equation}\label{eq:TokenProbe}
  \tokenprobe = \frac{1}{T}\sum\nolimits_{t=1}^T \mathbb{I}\!\left[\normlp{}_t > 0\right],
\end{equation}
\end{definition}
\vspace{-0.3cm}

\begin{figure*}[t]
    \centering
    \includegraphics[width=0.98\textwidth]{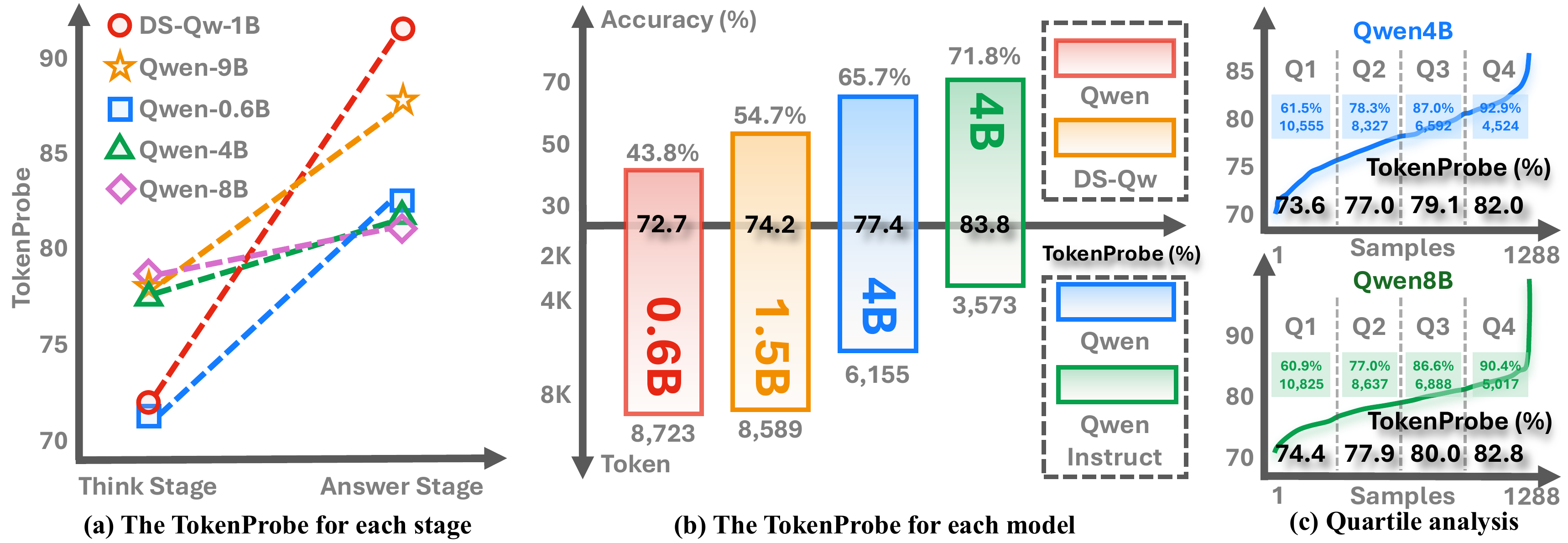}
    \vspace{-0.2cm}
    \caption{\textbf{TokenProbe as an indicator of reasoning efficiency}. (a) TokenProbe increases from the think stage to the answer stage. (b) Across models, higher TokenProbe corresponds to better accuracy and lower token consumption. (c) Quartile analysis shows the same trend at the sample level.}
    \vspace{-0.5cm}
    \label{fig: tokenprobe}
\end{figure*}

Leveraging TokenProbe, we study five models of different scales across four reasoning benchmarks, AIME, MATH \cite{hendrycks2021measuring, lightman2023let}, AMC, and Olympiad-Bench \cite{he2024olympiadbench}, under varied hyperparameter settings (see more in \S\ref{asec: probing_analysis}). We organize the analysis in a coarse-to-fine manner and formulate two central findings supported by the resulting empirical evidence.

\vspace{-0.1cm}
\begin{AIbox}{Finding 1: Token Value Inequality}
$\blacktriangleright$ A two-phase confidence structure between think and answer stages.
\end{AIbox}\label{finding1}

To understand token contributions, we begin by examining the output structure of common reasoning models. The output of a thinking-enabled model~\cite{guo2025deepseek,muennighoff2025s1} is naturally segmented into two phases: \textbf{Think phase}, spanning the content between \texttt{<think>} and \thinksep{}, contains the intermediate reasoning process; and \textbf{Answer phase}, occurring after \thinksep{}, contains the final answer output.

As shown in Fig.~\ref{fig: tokenprobe}(a), the log-probability distributions of the two phases exhibit a pronounced duality: Answer-phase TokenProbe is systematically 3\%--20\% higher than Think-phase TokenProbe. This reveals the underlying structure of CoT reasoning: the Think phase is a high-cost search process (\ie, the model is exploring solution paths), while the Answer phase is a low-cost decision output (\ie, the model has formed a definite conclusion)~\cite{zelikman2022star,hao2024training}.  The ``value'' of reasoning is not uniformly distributed across every token of the Think phase. Part of it serves genuine reasoning functions, while the rest is exploratory noise.
\vspace{-0.1cm}
\begin{AIbox}{Finding 2: TokenProbe as a Core-Token Proxy}
$\blacktriangleright$ Higher TokenProbe → higher accuracy + lower token usage. \\
$\blacktriangleright$ TokenProbe can distinguish "Productive Exploration" from "Consumptive Getting Lost".
\end{AIbox}\label{finding2}
\vspace{-0.1cm}
Building on the finding of token value inequality, we then conduct a finer-grained analysis of TokenProbe to examine its underlying characteristics. More specifically, we define reasoning quality as the ability to achieve higher accuracy under a relatively low yet still reasonable token budget~\cite{jin2024impact,chen2024not}.
Under this view, high-quality reasoning is not merely shorter reasoning, but reasoning that avoids unnecessary detours and self-corrections while preserving performance.

\vspace{-0.1cm}
As shown in Fig.~\ref{fig: tokenprobe}(b), our results suggest that TokenProbe serves as an effective proxy for this notion of reasoning quality. Across four representative models of different scales, we observe a clear pattern: as TokenProbe increases, accuracy consistently rises, while token usage consistently falls. This indicates that higher TokenProbe is associated not with premature truncation, but with more efficient reasoning trajectories that remove unnecessary repeated corrections. Moreover, this pattern is not confined to these four cases. Across all experimental settings we analyze in \S\ref{asec: probing_analysis}, the Pearson correlation between TokenProbe and token usage is consistently negative, whereas the correlation between TokenProbe and accuracy is consistently positive, further supporting that TokenProbe is a useful empirical indicator of both reasoning effectiveness and token efficiency~\cite{wang2025beyond}.

\vspace{-0.1cm}
As shown in Fig.~\ref{fig: tokenprobe}(c), we do a further analysis on Qwen-4B and Qwen-8B. Accuracy increases by 31.4\% from the lowest to the highest quartile, while token usage drops by 57\%. This pattern replicates faithfully on other model scales (see more in \S\ref{asec: probing_analysis}). 

\vspace{-0.1cm}
While Findings 1--2 characterize reasoning quality at the phase and response level, identifying which specific tokens within a trace are productive requires finer-grained analysis. We therefore drill down to the individual token level~\cite{lin2024rho,wang2025beyond}, demonstrating how \normlp{} provides fine-grained semantic discrimination within a reasoning trace (see the complete case in \S\ref{asec: probing_analysis}).
This finding carries the most direct practical implications. As shown in Tab.~\ref{tab:wait-analysis}, we perform a detailed analysis of the commonly used token ``Wait'' at different positions within a single reasoning trace. We find that early occurrences often reflect productive self-correction, whereas late occurrences tend to be wasteful. This distinction is reflected by positive versus negative \normlp{}, thereby providing an empirical foundation for identifying core tokens in the subsequent efficient GRPO framework.

\begin{table}[t]
  \centering
  \caption{%
    Functional analysis of the ``Wait'' token at different positions within
    a single reasoning trace. %
  }
  \vspace{-0.2cm}
  \label{tab:wait-analysis}
  \resizebox{\textwidth}{!}{
  \begin{tabular}{ccll}
    \toprule
    \textbf{Position} & \textbf{\normlp{}} & \textbf{Semantic Role}
      & \textbf{Impact} \\
    \midrule
    Token [54]  (early) & $+0.47$ & First verification discovers
                                     misinterpretation
      & Productive \\
    Token [103] (early) & $+0.22$ & Mild self-correction
      & Productive \\
    Token [575] (mid)  & $-3.87$ & Deep confusion with hesitation
      & Wasteful \\
    Token [1273] (mid-late) & $-4.13$ & Re-questions already-confirmed step
      & Wasteful (+700 tokens) \\
    \bottomrule
  \end{tabular}}
  \vspace{-0.5cm}
\end{table}

\vspace{-0.4cm}
\section{Methodology}\label{sec: method}
\vspace{-0.3cm}
\begin{AIbox}{Claim: Efficient GRPO}
$\blacktriangleright$ Achieving pareto improvement through efficient GRPO guided by the \normlp{} signal.
\end{AIbox}
\vspace{-0.1cm}
Findings 1 and 2 jointly motivate a crucial claim: CoT reasoning contains a large volume of redundant tokens that are identifiable by \normlp{} signals, and these redundant tokens are directly associated with low accuracy and high token consumption. Therefore, if training can guide a model to: (a)~preserve the reasoning density of core tokens, and (b)~compress or eliminate redundant tokens, then it is possible to substantially reduce the token budget of CoT reasoning without significant loss in final-answer accuracy, achieving a Pareto efficiency improvement.

\begin{figure*}[t]
    \centering
    \includegraphics[width=0.98\textwidth]{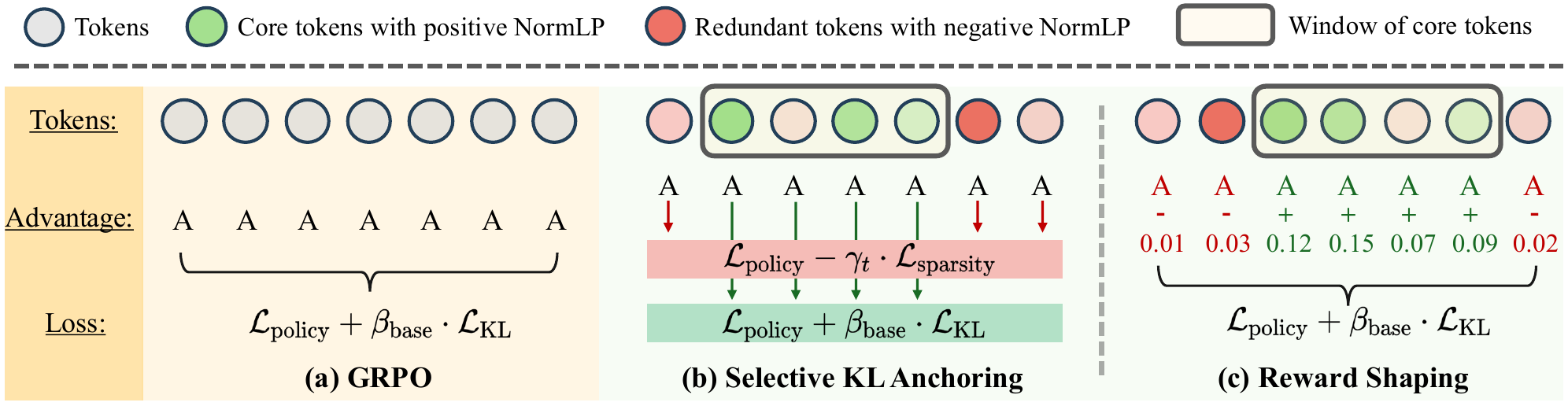}
    \vspace{-0.2cm}
    \caption{\textbf{Comparison of training strategies for efficient reasoning}. (a) Vanilla GRPO treats all tokens equally. (b) \textbf{Selective KL Anchoring (KL)} rescales the KL regularizer per token and imposes a sparsity penalty on over-budget redundant tokens. (c) \textbf{Reward Shaping (RS)} reshapes the token-level advantage, upweighting core tokens and penalising overlong redundant spans.}
    \vspace{-0.4cm}
    \label{fig: method}
\end{figure*}

Based on the analysis above, we build upon standard GRPO~\cite{shao2024deepseekmath,schulman2017proximal,schulman2015high}, which maximises the per-token objective
\begin{equation}
\label{eq: grpo}
\mathcal{L}_{\text{GRPO}} = \mathbb{E}\!\left[\,\frac{\pi_\theta(a_t \mid s_t)}{\pi_{\text{old}}(a_t \mid s_t)}\,\hat{A}_t \;-\; \beta_{\text{base}}\, D_{\text{KL}}\bigl(\pi_\theta \,\|\, \pi_{\text{ref}}\bigr)\,\right],
\end{equation}
where $\hat{A}_t$ is the group-relative advantage and $\beta_{\text{base}}$ is a fixed base KL coefficient~\cite{ziegler2019fine,ouyang2022training}. Guided by the \normlp{} signal, we propose \textbf{two complementary modifications} of Eq.~\ref{eq: grpo} that target different components of the same skeleton in Fig.~\ref{fig: method}. \textbf{Selective KL Anchoring} (\textbf{KL}, \S\ref{sec: kl}) reshapes the \emph{regularization} term so that redundant tokens receive an explicit sparsity pressure once a rollout overruns its budget. \textbf{Reward Shaping} (\textbf{RS}, \S\ref{sec: rs}) reshapes the \emph{advantage} inside the policy term so that core tokens on correct rollouts are upweighted while overlong redundant spans are penalised. The two variants share the same core-token mask and leave the update on wrong-answer rollouts (\ie, $r^{\text{seq}} \leq 0$) untouched (see more discussion in \S\ref{asec: reward_dynamics}). 

To avoid overly sparse rewards and preserve semantic coherence, without introducing additional annotation effort at the turn level, we identify core tokens through a mathematically motivated window selection strategy (see more discussion in \S\ref{asec:window_exploratory_analysis}). The core window is selected based on \normlp{}, with the intuition that tokens in high-density regions are more likely to correspond to essential reasoning steps~\cite{lin2024rho,cui2025entropy}. Specifically, given the normalized scores $\{\normlp{}_t\}_{t=1}^T$, we apply a sliding window of size $W$ and compute the cumulative score for each window as $s_i^{(W)} = \sum_{t=i}^{i+W-1} \normlp{}_t$. Finally, we greedily select the highest-scoring non-overlapping windows and define the corresponding token positions as core tokens (see \S\ref{asec: window}), yielding a binary core mask $CT_t \in \{0,1\}$, where $CT_t = 1$ indicates a core token and $CT_t = 0$ a redundant token.

\subsection{Selective KL Anchoring (KL)}\label{sec: kl}

As illustrated in Fig.~\ref{fig: method}(b), \textbf{KL} keeps the policy term of Eq.~\ref{eq: grpo} intact and makes only the regularizer token-adaptive:
\begin{equation}
\label{eq: kl}
\mathcal{L}_{\text{KL}} = \mathbb{E}\!\left[\,\frac{\pi_\theta(a_t \mid s_t)}{\pi_{\text{old}}(a_t \mid s_t)}\,\hat{A}_t \;\;\boxed{\,-\, \beta_t\, D_{\text{KL}}\bigl(\pi_\theta \,\|\, \pi_{\text{ref}}\bigr) \,-\, \gamma_t\, \log \pi_\theta(a_t \mid s_t)\,}\;\right],
\end{equation}
where the boxed block replaces the base KL term of Eq.~\ref{eq: grpo}. The $\beta_t$ is a per-token stability coefficient~\cite{rafailov2023direct} and $\gamma_t$ activates a log-probability sparsity term on suppressible tokens (see \S\ref{asec: kl}). The two coefficients are dispatched by a simple rule parameterised by the core mask $CT_t$ and the current rollout length $T$ relative to the length limit $L_{\text{limit}}$:

\textbf{Core tokens ($CT_t = 1$).}
We keep the vanilla regularizer (\ie, $\beta_t = \beta_{\text{base}},\, \gamma_t = 0$), anchoring optimisation on reasoning-essential tokens without adding any sparsity pressure.
\textbf{Over-budget redundant tokens ($CT_t = 0,\, T > L_{\text{limit}}$).}
We remove the KL constraint ($\beta_t = 0$) and inject an overflow-proportional sparsity penalty $\gamma_t = \lambda_{\text{decay}} \cdot (T - L_{\text{limit}}) / T$, suppressing further redundant generation once a trace has already exceeded the budget.
\textbf{Within-budget redundant tokens ($CT_t = 0,\, T \leq L_{\text{limit}}$).}
We fall back to the stable setting ($\beta_t = \beta_{\text{base}},\, \gamma_t = 0$), tolerating moderate redundancy rather than prematurely suppressing it.

\subsection{Reward Shaping (RS)}\label{sec: rs}

As illustrated in Fig.~\ref{fig: method}(c), \textbf{RS} keeps the regularizer of Eq.~\ref{eq: grpo} intact and instead reshapes the advantage inside the policy term.
\begin{equation}
\label{eq: rs}
\mathcal{L}_{\text{RS}} = \mathbb{E}\!\left[\,\frac{\pi_\theta(a_t \mid s_t)}{\pi_{\text{old}}(a_t \mid s_t)}\,\boxed{\,\hat{A}_t^{\text{RS}}\,} \;-\; \beta_{\text{base}}\, D_{\text{KL}}\bigl(\pi_\theta \,\|\, \pi_{\text{ref}}\bigr)\,\right],
\end{equation}
where the boxed $\hat{A}_t^{\text{RS}}$ replaces the vanilla $\hat{A}_t$ of Eq.~\ref{eq: grpo} on correct rollouts ($r^{\text{seq}} > 0$) and falls back to $\hat{A}_t$ otherwise. Concretely,
\begin{equation}
\label{eq:rs_adv}
\hat{A}_t^{\mathrm{RS}} = \hat{A}_t + \beta \left( \alpha \,\normlp_t \,\mathbb{I}[CT_t=1] - \lambda \,\frac{\max(0,\,T-L_{\mathrm{limit}})}{T}\,\mathbb{I}[CT_t=0] \right).
\end{equation}
where $\normlp{}_t$ is the z-score normalised log probability, $\alpha$ upweights core tokens along high-reward trajectories, $\lambda$ penalises redundant tokens once a trace runs long (see \S\ref{asec: rs}), and $L_{\text{limit}} = \text{core\_len} \times \text{current\_scale}$ is a length budget progressively tightened through curriculum annealing~\cite{bengio2009curriculum}. We deliberately restrict shaping to correct rollouts (\ie, wrong-answer rollouts retain the vanilla $\hat{A}_t$), since rewarding core tokens on a failed reasoning trace would propagate misleading credit to precisely the tokens that should be re-examined.

\newpage
\vspace{-0.4cm}
\section{Experiments}\label{experiments}
\vspace{-0.1cm}
\begin{table*}[t]
  \centering
  \caption{\textbf{Main results on four in-domain reasoning benchmarks.}
  \textbf{\#Tk.}\ is the average number of generated tokens per response
  (lower is better). \textbf{Acc.}\ is averaged accuracy in \% (higher is better).}
  \resizebox{\textwidth}{!}{%
  \begin{tabular}{l>{\cellcolor[gray]{0.9}}c>{\cellcolor[gray]{0.9}}ccccccccc}
  \toprule
  \multirow{2}{*}{Model} & \multicolumn{2}{c}{\textbf{Average}} & \multicolumn{2}{c}{\textbf{AIME}} & \multicolumn{2}{c}{\textbf{MATH}} & \multicolumn{2}{c}{\textbf{AMC}} & \multicolumn{2}{c}{\textbf{Olympid}} \\
  \cmidrule(lr){2-3} \cmidrule(lr){4-5} \cmidrule(lr){6-7} \cmidrule(lr){8-9} \cmidrule(lr){10-11}
  & \#Tk. & Acc. & \#Tk. & Acc. & \#Tk. & Acc. & \#Tk. & Acc. & \#Tk. & Acc. \\
  \midrule
  \multicolumn{11}{c}{\textbf{Closed-API Reasoning Models}}\\
  \midrule
  GPT-5.4-Mini         & 987 & 69.6 & 1714 & 36.7 & 589 & 88.6 & 1137 & 73.5 & 1232 & 56.6\\
  Claude-4.6-Sonnet &1221&	61.3&	1907&	20.0&	801	&86.0&	1413&	51.8&	1478&	46.1   \\
  Grok-4.1-Fast&  4945&	52.3&	9361&	10.0&	1866&	74.8&	4017&	43.4&	7144	&38.5   \\
  Gemini-3.1-Pro& 1586	&54.4&	2020&	10.0&	1274&	79.2&	1798&	37.3	&1771	&40.1  \\
  Gemini-3-Flash&  1828&	35.1&	2039&	3.3&	1625&	56.0&	1948&	24.1&	1954&	22.4   \\
  \midrule
  \multicolumn{11}{c}{\textbf{Open Source Reasoning Models}}\\
  \midrule
  Qwen-80B-3B~\cite{yang2025qwen3} &1843	&28.3	&2048	&0.0 	&1646&	51.0	&1972	&13.3	&1965	&14.7   \\
  GLM-358B-32B~\cite{glm2024chatglm} &1545&	44.5	&2039&	6.7	&1182&	68.0	&1775&	31.3&	1763&	30.4   \\
  DeepSeek-671B-37B~\cite{liu2024deepseek} &1542	&46.7&	1978&	13.3	&1200&	70.2&	1733&	36.1&	1751&	32.1   \\
  Kimi-1T-32B~\cite{team2025kimi}   &1649&	40.0&	2048&	0.0&	1342&	63.4&	1807&	28.9	&1840&	25.8 \\
  \midrule
  \multicolumn{11}{c}{\textbf{Efficient Reasoning Models}} \\
  \midrule
  DEER~\cite{yang2025dynamic} \textcolor{lightgray}{\scriptsize{[TMLR'25]}} & 1891 & 16.0 & 2622 & 13.3 & 1082 & 50.6 & 2040 & 0.0 & 1821 & 0.0  \\
  NoThink~\cite{ma2025reasoning} \textcolor{lightgray}{\scriptsize{[Arxiv'25]}} & 1156 & 10.3 & 1626 & 3.3 & 738 & 37.8 & 1146 & 0.0 & 1114 & 0.0  \\
  L1~\cite{aggarwal2025l1} \textcolor{lightgray}{\scriptsize{[COLM'25]}} & 4573 & 43.2 & 6058 & 20.0 & 2819 & 71.4 & 4946 &  42.2& 4470 & 39.3  \\
  O1-Pruner~\cite{luo2025o1} \textcolor{lightgray}{\scriptsize{[Arxiv'25]}} & 5248 &  39.5 & 7051 & 10.0 & 3179 &  70.8& 5413 & 42.2 & 5347 & 35.1  \\
  PEAR~\cite{huang2025pear} \textcolor{lightgray}{\scriptsize{[ICLR'26]}} & 4240 & 41.8 & 6030 & 16.7 & 2322 & 69.6 & 4707 & 43.4  & 3901 & 37.3  \\
  ETR~\cite{xiong2026etr} \textcolor{lightgray}{\scriptsize{[ACL'26]}} & 4566 & 42.2 & 6209 & 20.0 & 2799 & 70.4 & 4889 & 39.8 & 4366 & 38.7  \\
  \midrule
  \multicolumn{11}{c}{\textbf{Ours}} \\
  \midrule
  Qwen-0.6B~\cite{yang2025qwen3} & 5890 &40.7  & 7291 & 16.7 & 3951 &  69.8& 6308 & 39.8 & 6010 & 36.6  \\
  \hspace{0.4cm} -   \hspace{0.1cm}w/ KL &  2023&  42.3&  2195& 16.7 &1717  &72.0  & 2120 &  44.6&  2060&  35.9 \\
  \hspace{0.4cm} -   \hspace{0.1cm}w/ RS &  1806&	36.3	&1898&	6.7	&1503&	68.6	&1987	&36.1	&1834	&33.8   \\
  DeepScaleR-1.5B~\cite{luo2025deepscaler} & 4741 & 59.6 &6564  & 26.7 & 2769 & 88.0 & 4790 & 69.9 & 4841 & 53.8 \\
  \hspace{0.4cm} -   \hspace{0.1cm}w/ KL & 1698 &56.1  & 2065 & 20.0 & 1185 & 85.8 & 1816 & 66.3 & 1726 & 52.1  \\
  \hspace{0.4cm} -   \hspace{0.1cm}w/ RS & 1993&	55.9&	2365	&23.3&	1494	&85.4&	2089&	63.9&	2023	&50.8 \\
  Qwen-4B~\cite{yang2025qwen3} & 6132 & 62.4 & 7683 & 30.0 & 4195 & 92.0 & 6274 & 72.3 & 6376 &55.4   \\
  \hspace{0.4cm} -   \hspace{0.1cm}w/ KL & 1886 & 61.0 &  2253&  33.3&  1246& 88.2 & 2092 & 66.3 & 1952 & 56.1  \\
  \hspace{0.4cm} -   \hspace{0.1cm}w/ RS & 2026&	62.3	&2532	&30.0	&1391&	90.2&	2154&	71.1&	2025&	58.1\\
  Phi-Reasoning-4B~\cite{abdin2025phi}  & 5515 & 59.6 & 7237 & 26.7 & 3559 & 87.4 & 5362 & 68.7 & 5901 & 55.6      \\
  \hspace{0.4cm} -   \hspace{0.1cm}w/ KL & 2919  & 62.0 &4190 & 26.7 &1756 & 89.8 & 2992 & 72.2 & 2737 & 59.1    \\
  \hspace{0.4cm} -   \hspace{0.1cm}w/ RS & 2992 & 62.6 & 3997 & 26.7 & 1977 & 89.8 & 2982 & 74.7 & 3011 & 59.1  \\
  Qwen-8B~\cite{yang2025qwen3}  & 6252 &61.9 & 7737 & 33.3 & 4412 & 89.2& 6404 & 71.1 & 6454 & 53.9    \\
  \hspace{0.4cm} -   \hspace{0.1cm}w/ KL & 1476 & 62.7 & 1945 & 30.0 & 1008 & 88.6  & 1486 & 73.5 & 1466 &  58.5 \\
  \hspace{0.4cm} -   \hspace{0.1cm}w/ RS & 2313 & 68.3 & 3276 &33.3 & 1462 & 94.2 & 2265 & 81.9 & 2249 & 63.9  \\
  \bottomrule
  \end{tabular}
  }
  \vspace{-0.5cm}
  \label{tab:main_results}
\end{table*}

\subsection{Experimental Setup}\label{sec: expe_setup}

\vspace{-1mm}
\textbf{Training and Evaluation.}
Following \cite{aggarwal2025l1}, we train on the DeepScaleR-Preview dataset~\cite{luo2025deepscaler}, a 40K math corpus drawn from AIME, AMC, Omni-MATH~\cite{gao2024omni}, and STILL~\cite{min2024imitate}.
We evaluate on four in-domain reasoning benchmarks: AIME~2025, MATH-500~\cite{hendrycks2021measuring,lightman2023let}, AMC, and OlympiadBench~\cite{he2024olympiadbench}.
The full training recipe, per-benchmark rollout budgets, and prompts are deferred to Appx.~\ref{asec: training_settings} and Appx.~\ref{asec: eval_settings}. Training uses a 4{,}096-token rollout cap and evaluation an 8{,}192-token cap for efficient reasoning models and ours. As for the backbone of ours, we train and evaluate our method across five backbones spanning 0.6B--8B parameters: Qwen3-0.6B / 4B / 8B~\cite{yang2025qwen3}, DeepScaleR-1.5B~\cite{luo2025deepscaler}, and Phi-4-mini-reasoning~\cite{abdin2025phi}, covering various model families and scales.

\vspace{-1mm}
\textbf{Baselines.}
We group baselines into three categories (full configuration in Appx.~\ref{asec: baselines}):
(i) \emph{closed-API reasoning models}: Claude-4.6-Sonnet, Grok-4.1-Fast, Gemini-3.1-Pro, Gemini-3-Flash, and GPT-5.4-mini;
(ii) \emph{open-source large reasoning models}: Qwen3-Next-80B-A3B~\cite{yang2025qwen3}, GLM-4.7-358B-32B~\cite{glm2024chatglm}, DeepSeek-V3.2-671B-37B~\cite{liu2024deepseek}, and Kimi-K2-1T-32B~\cite{team2025kimi};
(iii) \emph{efficient reasoning methods}: DEER~\cite{yang2025dynamic}, NoThink~\cite{ma2025reasoning}, L1~\cite{aggarwal2025l1}, O1-Pruner~\cite{luo2025o1}, PEAR~\cite{huang2025pear}, and ETR~\cite{xiong2026etr}.

\vspace{-0.3cm}
\subsection{Main Results}\label{sec: main_res}
\vspace{-0.1cm}

Tab.~\ref{tab:main_results} leads to three key observations organised along the
\emph{same-budget} and \emph{same-accuracy} perspectives. \textbf{First, performance superiority at a matched token budget comparison.}
At a $\sim$2K-token rollout budget, our method matches or surpasses
some closed-API and open-source large reasoners.
Qwen-8B-w/KL reaches $62.7\%$ at 1{,}476 tokens and outperforms
Gemini-3.1-Pro by $+8.3\%$ (\ie, $62.7$ $vs.$ $54.4$ at 1{,}476 $vs.$ 1{,}586 tokens),
and Kimi-1T-32B by $+21.6\%$.
Our narrower point is that TokenProbe training moves open sub-10B models toward better accuracy and token trade-offs at far lower serving cost. See more comparisons in \S\ref{asec: unlimited_token} and \S~\ref{asec: eval_settings}.
\textbf{Second, generality across backbones with drastic token reduction.}
Applied to every backbone, our objectives compress response length by
$1.8$--$4.2\times$ with essentially no accuracy loss.
Qwen-8B-w/KL matches the base Qwen-4B
(\ie, $62.7$ $vs.$ $61.9$, $+0.8\%$) while using $4.2\times$ fewer
tokens (6{,}252 $vs.$ 1{,}476).
The effect is most pronounced on Qwen-8B-w/RS, which
raises accuracy by $+6.4\%$ (\ie, $68.3$ $vs.$ $61.9$) while
cutting tokens from 6{,}252 to 2{,}313 ($2.7\times$).
This suggests that compressing redundant chain-of-thought acts as an implicit regularizer~\citep{setlur2024rewarding} over the reasoning trace. More importantly, improved reasoning efficiency reduces the risk that generation is prematurely constrained by the token budget, thereby increasing the likelihood of reaching a complete respond within the available budget. For example, on AIME, the Qwen-8B base model uses substantially more tokens than our method on average (7,737 $vs.$ 3,276), and similar trends hold for AMC and Olympiad.
\textbf{Third, breaking the accuracy and efficiency trade-off.}
Using Qwen-0.6B as the common backbone (base at $40.7\%$ with 5{,}890
tokens), existing efficient-reasoning methods fall into two failure
modes along a trade-off that our method avoids.
\textit{(i) Aggressive compression at a catastrophic accuracy cost.}
DEER collapses to $16.0\%$ at 1{,}891 tokens and NoThink to $10.3\%$
at 1{,}156 tokens (\ie, $-24.7\%$ and $-30.4\%$ relative to base),
despite using similar or fewer tokens than our run.
\textit{(ii) Accuracy preserved but little compression.} PEAR ($41.8\%$ at 4{,}240 tk) and ETR
($42.2\%$ at 4{,}566 tk) both match base-model accuracy but deliver
only a $1.3$ to $1.4\times$ token reduction.
In contrast, Qwen-0.6B-w/KL attains $42.3\%$ at 2{,}023 tokens
(\ie, $+1.6\%$ over base at $2.9\times$ compression), matching the
strongest training-based baseline while using $2.2$ to $2.3\times$ fewer
tokens and avoiding the accuracy collapse of the inference-time
baselines. Additional analysis is provided in \S\ref{sec: discussion}.

\vspace{-0.3cm}
\subsection{Discussion}\label{sec: discussion}
\begin{figure*}[t]
    \centering
    \includegraphics[width=0.98\textwidth]{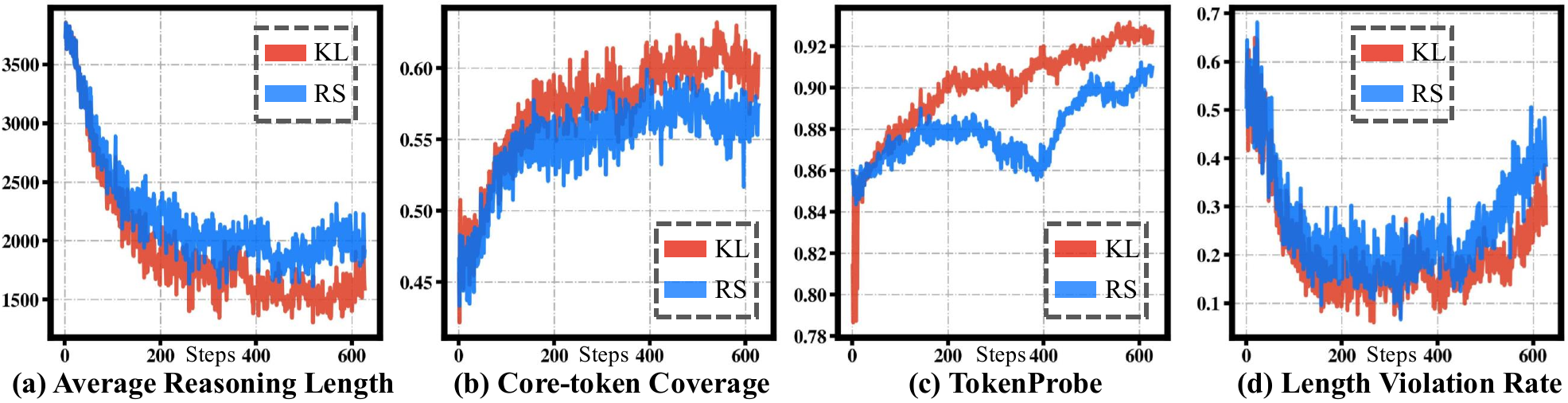}
    \vspace{-0.3cm}
    \caption{\textbf{TokenProbe signals during RL training}.}
    \vspace{-0.5cm}
    \label{fig: study1}
\end{figure*}
\subsubsection{Signals Along RL Training}

To understand \emph{how} our objectives compress reasoning, we track four
diagnostic signals along RL training on Qwen-4B
(Fig.~\ref{fig: study1}). \textbf{(a) Average reasoning length.}
This signal is the mean number of response tokens generated per rollout,
our most direct proxy for training-time compression.
As shown in Fig.~\ref{fig: study1}(a), both KL and RS rapidly pull the
response length from roughly $3.7$K tokens at the start of RL to a much
shorter regime~\cite{arora2025training,aggarwal2025l1}.
KL compresses more aggressively, stabilising at $1.6$ to $1.8$K tokens,
whereas RS remains longer at $1.9$ to $2.0$K tokens.
This indicates that both objectives successfully suppress excessive
deliberation, but with different strength. KL enforces a hard sparsity
pressure, while RS trades compression for a softer reward-shaped
adjustment, which is consistent with the results shown in Tab. \ref{tab:main_results}. \textbf{(b) Core-token coverage.}
This signal is the fraction of generated tokens covered by
TokenProbe-selected core windows
(\ie, the number of selected windows $\times$ window size, normalised
by the average response length), and therefore measures how densely the remaining tokens carry useful signal.
Fig.~\ref{fig: study1}(b) shows that coverage rises steadily under both
objectives, meaning the surviving tokens become progressively denser in
TokenProbe-selected content.
KL reaches a higher final coverage than RS, suggesting that KL removes
redundant context more sharply and leaves a larger fraction of the
response occupied by high-value windows.
RS follows the same upward trend but stays lower, consistent with its
longer generations, improving the density of useful tokens while
retaining more non-core connective or exploratory material.
\textbf{(c) TokenProbe.}
As shown in Fig.~\ref{fig: study1}(c), both methods trace a clear upward
trajectory. KL rises from $\sim$$0.84$ to above $0.92$, while RS
increases more smoothly and eventually approaches the same regime.
The steeper KL curve is consistent with a sharper compression effect,
whereas the smoother RS curve reflects reward-shaped improvement without
fully eliminating exploratory material.
\textbf{(d) Length violation rate.}
This signal is the fraction of valid responses whose length exceeds the
dynamic TokenProbe budget
(\ie, the selected core length multiplied by the current curriculum
scale), diagnosing how well the policy stays within a progressively
tightened budget.
Fig.~\ref{fig: study1}(d) shows that the violation rate first drops
substantially as the model learns to respect the TokenProbe-derived
budget, then rebounds later in training, especially for RS.
This rebound is expected because the budget is progressively tightened through
the curriculum scale while the policy is still adapting. See more in \S\ref{asec: token_limits}.

\begin{figure*}[t]
    \centering
    \includegraphics[width=0.98\textwidth]{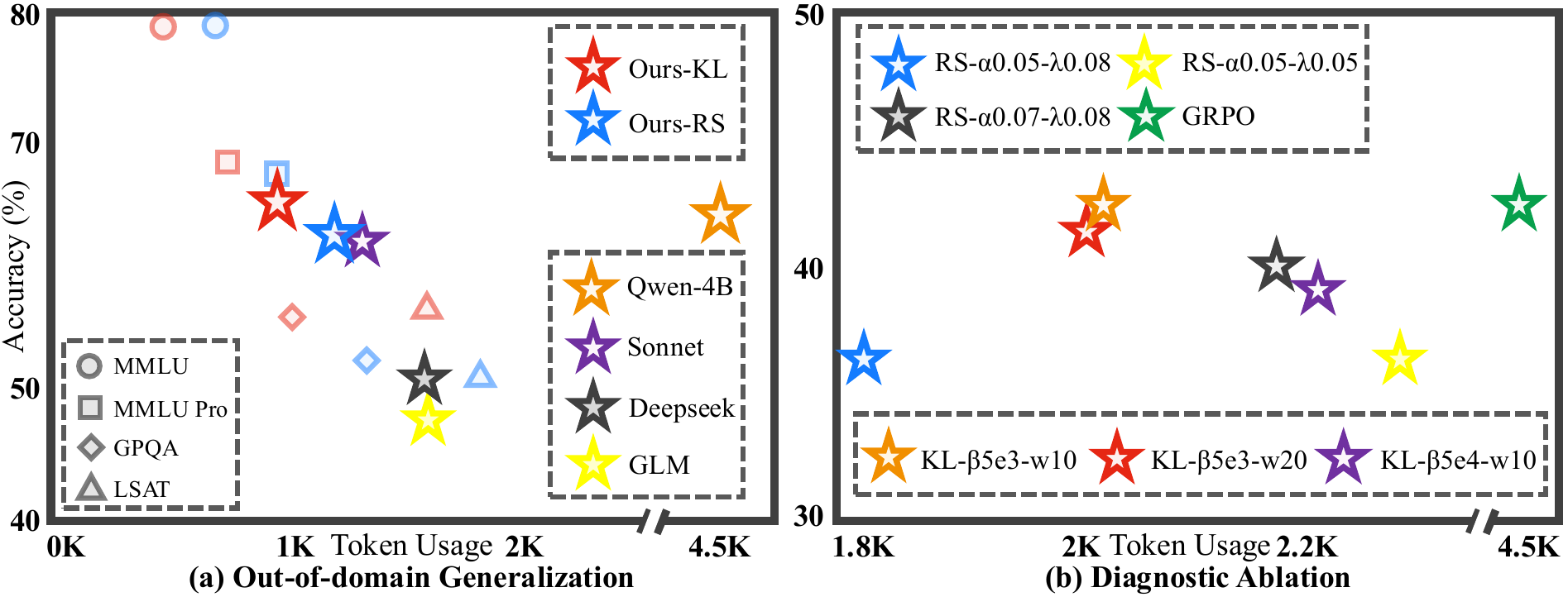}
    \vspace{-0.3cm}
    \caption{\textbf{Transfer and ablation of TokenProbe-based efficient reasoning}. (a) Our variants match accuracy with far fewer tokens. (b) Different hypeparameter settings for KL and RS (more in \S\ref{asec:window_exploratory_analysis}).}
    \vspace{-0.5cm}
    \label{fig: study2}
\end{figure*}

\vspace{-0.2cm}
\subsubsection{Out-of-domain Efficiency Transfer}
\vspace{-0.1cm}
To assess whether the token-efficiency gains transfer beyond mathematics,
we evaluate the Qwen-4B base model, RS, and KL on four additional
benchmarks spanning scientific, legal, and multi-disciplinary reasoning, including
MMLU~\cite{hendrycks2020measuring}, MMLU-Pro~\cite{wang2024mmlu},
LSAT~\cite{zhong2024agieval}, and GPQA~\cite{rein2024gpqa}.
As shown in Fig.~\ref{fig: study2}, both objectives substantially
compress length relative to the base model
(\ie, $4{,}401$ tokens), with RS reduced to $1{,}179$ tokens ($3.7\times$)
and KL to $935$ tokens ($4.7\times$).
Crucially, this compression is not paid for in accuracy.
RS maintains a competitive average of $57.74$ $vs.$ $59.62$
(\ie, $-1.9\%$), while KL actually \emph{improves} over the base model
to $60.61$ (\ie, $+1.0\%$) despite using far fewer tokens.
The effect is most pronounced when compared with state-of-the-art closed-source models such as Sonnet 4.6, where our method achieves higher accuracy while using relatively same tokens ($\sim150\times$ cheaper in Tab.~\ref{tab:api_pricing_closed_models}).
This confirms that the efficiency generalize robustly to diverse reasoning domains (see more discussions in \S\ref{asec: eval_settings}).

\vspace{-0.2cm}
\subsubsection{Diagnostic Ablation}
\vspace{-0.1cm}
We ablate the two hyperparameter groups of our method, the coefficients
$(\beta, w)$ in KL and the
coefficients $(\alpha, \lambda)$ in RS, reporting average token count and accuracy on the four
math benchmarks.
The GRPO baseline reaches $42.3\%$ at $4{,}461$ tokens on average and
serves as the reference point throughout.
\textbf{Coefficients ($\alpha$, $\lambda$) in RS.}
A key design choice is the balance between the core-token reward
$\alpha$ and the redundancy penalty $\lambda$, which governs how
aggressively redundant tokens are suppressed without starving the model
of informative-token credit.
Holding $\alpha{=}0.05$, raising $\lambda$ from $0.05$ to $0.08$ reduces
token usage by $21\%$ (\ie, $2{,}291 \to 1{,}806$) at essentially
unchanged accuracy (\ie, $36.1$ $vs.$ $36.3$), confirming that a
stronger length penalty compresses responses without harming
correctness.
Further raising $\alpha$ to $0.07$ at $\lambda{=}0.08$ improves accuracy
to $39.9\%$ at $2{,}179$ tokens, indicating that a larger core-token
reward encourages more informative reasoning rather than merely shorter
traces.
\textbf{Coefficients  ($\beta$) and window size ($w$) in KL.}
The shaping coefficient $\beta$ controls the magnitude of the per-token
advantage correction, while the window size $w$ controls the
neighbourhood over which TokenProbe density is averaged.
Under a matched window, $\beta{=}5\times 10^{-3}$ outperforms
$\beta{=}5\times 10^{-4}$ (\ie, $42.3$ $vs.$ $39.0$ at $\approx$2{,}200
tokens), indicating that a sufficiently strong shaping signal is
necessary for the token-level guidance to affect policy updates.
Comparing window sizes at $\beta{=}5\times 10^{-3}$, $w{=}10$ yields
$42.3\%$ at $2{,}023$ tokens $vs.$ $41.4\%$ at $2{,}008$ tokens for
$w{=}20$, a marginal difference that suggests the method is
insensitive to moderate window-size variation.

\newpage
\section{Insights and Limitations}\label{sec:additional_insights}
\vspace{-0.2cm}
\textit{We examine whether compressed traces can teach efficient reasoning, where further compression becomes unstable, and how the validity of TokenProbe extends across training regimes and tasks.} These analyses highlight the need to preserve useful self-correction and sufficient intermediate computation while checking that confidence remains aligned with reasoning quality. Readers can inspect these interactions on our \href{https://runjia.tech/tokenprobe/}{\textbf{project homepage}} through interactive core-token window visualizations, case studies displaying every token's \normlp{}, and accuracy--cost Pareto plots spanning the full set of evaluated models and their reported evaluation settings.

\subsection{Unsuccessful Attempts: SFT Distillation}
\vspace{-0.1cm}
We explored SFT on compressed teacher traces to test whether imitation can transfer concise reasoning. Offline trace compression can remove the causal dependencies that make reasoning successful. Useful self-corrections can carry low \normlp{} and be discarded together with redundant detours. The surviving steps are then concatenated across missing context, forcing the student to imitate transitions that no longer contain the information on which the teacher's decisions depended. Our SFT distillation failures are consistent with this disruption of the reasoning process (\S\ref{asec: unsuccessful_attempts}). The usefulness of a token depends on its role within the trajectory, including the revisions and checks needed to preserve coherent reasoning in subsequent steps and maintain a complete path to the final answer.

Online feedback offers a way to learn concision while preserving useful exploration. Complete rollouts retain the context of each decision, and outcome-based updates can reinforce self-correction when it contributes to a solution. This motivates learning compact reasoning through differential reinforcement within intact trajectories. Filtered SFT may still provide a useful initialization, but its suitability depends on whether the resulting policy retains coherent reasoning and a meaningful confidence signal before TokenProbe-guided RL begins (\S\ref{asec:sft_outlook}).

\vspace{-0.2cm}
\subsection{Limits of Token Compression}
\vspace{-0.1cm}
The efficiency gains raise the question of how far compression can proceed before reasoning or training becomes unstable. Stable compression requires preserving sufficient intermediate computation. Excessive shortening pressure can disrupt this computation and drive generation into repetition. In this regime, response length can increase as accuracy collapses, so the failure appears as a joint deterioration of reasoning quality and efficiency (\S\ref{asec: token_limits}). Token-aware optimization must therefore preserve the policy's capacity to complete a solution as the budget tightens, allowing room for useful self-correction and the intermediate steps needed to resolve the problem.

The attainable compression level depends on the checkpoint and the difficulty distribution of the task. Problems requiring more sequential computation leave less room for shortening, while the policy's initialization can affect its stability under optimization. The reported collapse trajectories and supplementary budget sweeps illustrate these dependencies (\S\ref{asec: token_limits}, \S\ref{asec: un_appendix_tables}). They motivate compression schedules that respond to correctness and signs of degeneration, with enough flexibility to retain useful intermediate steps on harder problems throughout training.

\vspace{-0.2cm}
\subsection{Limitations and Future Directions}
\vspace{-0.1cm}
The empirical success of TokenProbe leaves open why confidence tracks reasoning value and how far this relationship extends beyond the evaluated settings. The signal's interpretation depends on the policy that generates the trace. Repetitive errors can be locally predictable, and the confidence--correctness relationship can reverse after a change in training regime (\S\ref{asec: unsuccessful_attempts}, Tab.~\ref{tab:sft_reversal}). Applying TokenProbe therefore requires revalidating its ability to distinguish productive reasoning from learned repetition. A mechanistic account of when \normlp{} identifies reasoning-critical content and a theoretical characterization of its reliability remain open (\S\ref{asec: theoretical_certification}).

Extending the framework beyond mathematics also requires examining whether the diagnostic signal and the efficiency gains survive changes in task structure. The reported efficiency transfer and supplementary TokenProbe diagnostic provide evidence across domains, with broader validation still needed (\S\ref{asec: general_reasoning_efficiency}). Code generation and agentic reasoning would extend this evaluation to settings where executable steps, external feedback, and longer interactions change the role of intermediate tokens. Such studies could clarify how token value depends on task structure and whether TokenProbe can guide efficient reasoning across different computational demands and feedback structures.

\newpage
\section{Related Work}\label{sec:related}
\vspace{-0.2cm}
\subsection{Efficient Reasoning for Large Language Models}
\vspace{-0.1cm}
Efficient reasoning~\cite{sui2025stop} targets the \emph{overthinking} phenomenon~\cite{chen2024not} in long chain-of-thought, where verbose or redundant intermediate steps inflate inference cost. Existing approaches can be broadly categorized into three paradigms: \textit{model-oriented}, \textit{output-oriented}, and \textit{prompt-oriented}.

\textit{Model-oriented} methods fine-tune LLMs for intrinsic conciseness, primarily via RL with length-aware rewards~\cite{aggarwal2025l1,luo2025o1,arora2025training,yeo2025demystifying} or supervised distillation from compressed CoT traces~\cite{han2025token,xia2025beyond}, optionally combined with model merging~\cite{wortsman2022model,wan2024knowledge}. However, these signals are defined at the response level and thus cannot distinguish which tokens within a trace are genuinely informative, often yielding mild compression or accuracy collapse (see Tab.~\ref{tab:main_results}).
\textit{Output-oriented} methods rewrite the output format, compressing CoT into latent representations~\cite{hao2024training,deng2024explicit,cheng2024compressed} on the view that much of the gain comes from hidden computation rather than verbose text~\cite{pfau2024let,goyal2023think}. It normally comes with architectural changes or incurs additional inference cost.
\textit{Prompt-oriented} methods enforce concision at the input level through token-budget prompts~\cite{nayab2024concise}, Chain-of-Draft instructions~\cite{xu2025chain}, or difficulty-aware model routing~\cite{chuang2024learning,miranda2025hybrid}. Their effect is bounded by the underlying model's unchanged policy.

TokenProbe falls within the \textit{model-oriented} paradigm but departs from previous length punishment in two respects. Rather than optimizing a scalar response-length objective, we introduce a \emph{token-level} TokenProbe signal that scores the local importance of every generated token, and inject this signal into both a selective KL anchoring objective (\S\ref{sec: kl}) and a reward-shaping term (\S\ref{sec: rs}). This redirects the compression pressure from \emph{how long} the trace is to \emph{which tokens} carry the reasoning signal, enabling strong compression without accuracy collapse in our five-backbone setting (Tab.~\ref{tab:main_results}).
\vspace{-0.2cm}
\subsection{Variants of Group Relative Policy Optimization}
\vspace{-0.1cm}
Group Relative Policy Optimization (GRPO)~\cite{shao2024deepseekmath} has become the dominant critic-free framework~\cite{ouyang2022training,ahmadian2024back} for training large reasoning models. Existing variants can be broadly categorized into two paradigms: \textit{reward design}, which governs the granularity and source of the supervision signal used to form group-relative advantages, and \textit{optimization design}, which concerns the policy-gradient objective, importance sampling, loss aggregation, and regularization.

On the \textit{reward design} axis, dense signals~\cite{ng1999policy,setlur2024rewarding} shorten the credit-assignment horizon but risk mis-specification and reward hacking. Along action granularity, existing work splits into \emph{token-level} signals derived from policy/reference likelihood ratios, implicit outcome-reward reparameterisation, or token entropy~\cite{chan2024dense,wang2025beyond,lin2024rho,yuan2024free,yang2024qwen2,cui2025entropy}, \emph{step-level} signals from trained PRMs~\cite{lightman2023let,wang2024math,uesato2022solving} or online Monte-Carlo rollouts~\cite{chen2024alphamath,ma2023let,luo2024improve}, and \emph{turn-level} signals for agentic tasks~\cite{zeng2024agenttuning,qin2023toolllm}. On the \textit{optimization design} axis, critic-free variants refine group-relative normalisation and clipping (DAPO~\cite{yu2025dapo}, Dr.\ GRPO~\cite{liu2025understanding}, LitePPO~\cite{liu2025part}, GFPO~\cite{shrivastava2025sample}), replace token-wise importance ratios with sequence- or geometric-level alternatives (GSPO~\cite{zheng2025group}, GMPO~\cite{zhao2025geometric}, CISPO~\cite{chen2025minimax}), introduce off-policy or hybrid objectives (TOPR~\cite{roux2025tapered}, LUFFY~\cite{yan2025learning}, Prefix-RFT~\cite{huang2025blending}), and modulate regularisation through KL penalties~\cite{ziegler2019fine}, entropy control~\cite{cui2025entropy}, and response-level length penalties~\cite{arora2025training,team2025kimi}. However, these length penalties act at the \emph{response} granularity and therefore cannot selectively suppress redundant tokens while preserving reasoning-critical ones.

TokenProbe contributes jointly on both axes. On the \textit{reward-design} side, it replaces response-level length penalties with a \emph{token-level} TokenProbe signal that flags redundant tokens inside every trace. On the \textit{optimization-design} side, it converts this signal into a selective KL anchoring term (\S\ref{sec: kl}) and a reward-shaping correction (\S\ref{sec: rs}) that steer GRPO advantages toward informative tokens while leaving reasoning-critical regions untouched. This token-aware coupling resolves the response-level length coupling that limits current variants, and delivers the $2$-$4\times$ compression observed in Tab.~\ref{tab:main_results}.
\vspace{-0.4cm}
\section{Conclusion}\label{conclusion}
\vspace{-0.2cm}
We present \textbf{TokenProbe}, an inference-native token-level diagnostic and optimization framework for efficient Chain-of-Thought reasoning grounded in token value inequality.
It has three merits.
\textbf{i)} It elucidates token value inequality via inference-native \normlp{} and TokenProbe signals. \textbf{ii)} It offers a GRPO-compatible optimization path through selective KL anchoring and reward shaping. \textbf{iii)} It demonstrates Pareto-improving token efficiency across the tested backbones and reasoning benchmarks.
As a whole, we conclude that the outcomes elucidated in this paper impart essential understandings and thus necessitate further exploration within the field of efficient reasoning.

\clearpage
\section*{Acknowledgements}
This research was supported by the National Science Foundation under Grant No. 2450068. This work used NCSA Delta GPU through allocation CIS250460 from the Advanced Cyberinfrastructure Coordination Ecosystem: Services \& Support (ACCESS) program, which is supported by U.S National Science Foundation grants No. 2138259, No. 2138286, No. 2138307, No. 2137603, and No. 2138296. We acknowledge that this work is supported by Purdue Polytechnic Institute. We also gratefully acknowledge the support of RIT Research Computing \citep{https://doi.org/10.34788/0s3g-qd15}. 

\bibliographystyle{plain}
\bibliography{reference}

\newpage
\addtocontents{toc}{\protect\setcounter{tocdepth}{3}}
\appendix
\renewcommand{\thesection}{S\arabic{section}}
\renewcommand{\thetable}{S\arabic{table}}
\renewcommand{\thefigure}{S\arabic{figure}}
\renewcommand{\theHtable}{S\arabic{table}}
\renewcommand{\theHfigure}{S\arabic{figure}}
\setcounter{table}{0}
\setcounter{figure}{0}
\centerline{\textbf{SUMMARY OF THE APPENDIX}}

This appendix provides additional experimental results and discussion for our NeurIPS 2026 submission, \textit{On the Token Value Inequality in Efficient Reasoning}, organized as follows.

\begingroup
\setcounter{tocdepth}{3}
\renewcommand{\contentsname}{}
\vspace{-2.5em}
\tableofcontents
\endgroup

\newpage
\section{Model Notations}
\label{sec:notations}

Since every model provider follows a different naming strategy (\eg, Gemini uses ``E'' for activated parameters, GPT uses only total parameters, and many other models do not identify any parameter size), we adopt a unified naming convention throughout this paper to ensure consistent model identity. For open-source MoE models, we use the format \texttt{[Model Name]-[Total Params]-[Activated Params]} (\eg, \texttt{Step-Flash-196B-11B}), and for open-source dense models we use \texttt{[Model Name]-[Total Params]} (\eg, \texttt{Qwen-4B}). We generally omit version numbers for open-source models, as we use relatively recent releases whose identifiers are listed in Tab.~\ref{tab:models_opensource}. For closed-source models, we use the format \texttt{[Model Name]-[Version]-[Mode]} (\eg, \texttt{Claude-4.6-Sonnet}, \texttt{Gemini-3-Flash}).

\begin{table}[htbp]
\centering
\small
\setlength{\tabcolsep}{6pt}
\renewcommand{\arraystretch}{1.2}
\caption{Training models used in Table~\ref{tab:main_results}.}
\label{tab:models_training}
\begin{tabular}{l l l}
\toprule
\textbf{Model Name} & \textbf{Identifier} & \textbf{Link} \\
\midrule
Qwen-0.6B       & \texttt{Qwen/Qwen3-0.6B}                      & \href{https://huggingface.co/Qwen/Qwen3-0.6B}{HuggingFace} \\
DeepScaleR-1.5B & \texttt{agentica-org/DeepScaleR-1.5B-Preview} & \href{https://huggingface.co/agentica-org/DeepScaleR-1.5B-Preview}{HuggingFace} \\
Qwen-4B         & \texttt{Qwen/Qwen3-4B}                        & \href{https://huggingface.co/Qwen/Qwen3-4B}{HuggingFace} \\
Phi-Reasoning-4B & \texttt{microsoft/Phi-4-mini-reasoning}                        & \href{https://huggingface.co/microsoft/Phi-4-mini-reasoning}{HuggingFace} \\
Qwen-8B        & \texttt{Qwen/Qwen3-8B}                   & \href{https://huggingface.co/Qwen/Qwen3-8B}{HuggingFace} \\
\bottomrule
\end{tabular}
\vspace{-0.5cm}
\end{table}

\begin{table*}[htbp]
\centering
\footnotesize
\setlength{\tabcolsep}{5pt}
\renewcommand{\arraystretch}{1.2}
\caption{Open-source baseline models used in our experiments.
  Identifiers follow the HuggingFace \texttt{org/model-name} convention.
  For MoE models, parameters are listed as Total\,(Active).}
\label{tab:models_opensource}
\begin{tabular}{p{3.2cm} p{6.0cm} p{2.5cm} l}
\toprule
\textbf{Model Name} & \textbf{HF Identifier} & \textbf{Parameters} & \textbf{Link} \\
\midrule
\multicolumn{4}{c}{\textit{Dense Models}} \\
\midrule
Qwen-4B-Instruct         & \texttt{Qwen/Qwen3-4B-Instruct-2507}                 & 4B    & \href{https://huggingface.co/Qwen/Qwen3-4B-Instruct-2507}{HuggingFace} \\
Qwen-9B                  & \texttt{Qwen/Qwen3.5-9B}                             & 9B    & \href{https://huggingface.co/Qwen/Qwen3.5-9B}{HuggingFace} \\
DS-Qw-1B & \texttt{deepseek-ai/DeepSeek-R1-Distill-Qwen-1.5B}   & 1.5B  & \href{https://huggingface.co/deepseek-ai/DeepSeek-R1-Distill-Qwen-1.5B}{HuggingFace} \\
\midrule
\multicolumn{4}{c}{\textit{Mixture-of-Experts (MoE) Models}} \\
\midrule
GPT-117B-5B              & \texttt{openai/gpt-oss-120b}                         & 117B\,(5B)  & \href{https://huggingface.co/openai/gpt-oss-120b}{HuggingFace} \\
GLM-Air-106B-12B         & \texttt{zai-org/GLM-4.5-Air}                         & 106B\,(12B) & \href{https://huggingface.co/zai-org/GLM-4.5-Air}{HuggingFace} \\
GLM-358B-32B             & \texttt{zai-org/GLM-4.7}                            & 358B\,(32B)   & \href{https://huggingface.co/zai-org/GLM-4.7}{HuggingFace} \\
GLM-744B-40B             & \texttt{zai-org/GLM-5}                              & 744B\,(40B)   & \href{https://huggingface.co/zai-org/GLM-5}{HuggingFace} \\
Step-Flash-196B-11B      & \texttt{stepfun-ai/Step-3.5-Flash}                   & 196B\,(11B) & \href{https://huggingface.co/stepfun-ai/Step-3.5-Flash}{HuggingFace} \\
DeepSeek-671B-37B        & \texttt{deepseek-ai/DeepSeek-V3.2}                   & 671B\,(37B) & \href{https://huggingface.co/deepseek-ai/DeepSeek-V3.2}{HuggingFace} \\
DeepSeek-1.6T-49B        & \texttt{deepseek-ai/DeepSeek-V4-Pro}                 & 1.6T\,(49B) & \href{https://huggingface.co/deepseek-ai/DeepSeek-V4-Pro}{HuggingFace} \\
DeepSeek-284B-13B        & \texttt{deepseek-ai/DeepSeek-V4-Flash}               & 284B\,(13B) & \href{https://huggingface.co/deepseek-ai/DeepSeek-V4-Flash}{HuggingFace} \\
Qwen3-Next-80B-3B   & \texttt{qwen/qwen3-next-80b-a3b-thinking}        & 80B\,(3B) & \href{https://huggingface.co/Qwen/Qwen3-Next-80B-A3B-Thinking}{HuggingFace} \\
Kimi-1T-32B    & \texttt{moonshotai/kimi-k2-thinking}             & 1.1T\,(32B)    &  \href{https://huggingface.co/moonshotai/Kimi-K2-Thinking}{HuggingFace} \\
Hunyuan3-295B-21B    & \texttt{tencent/Hy3-preview}                     & 295B\,(21B)    & \href{https://huggingface.co/tencent/Hy3-preview}{HuggingFace} \\
\bottomrule
\end{tabular}
\vspace{-0.5cm}
\end{table*}

\begin{table*}[htbp]
\centering
\footnotesize
\setlength{\tabcolsep}{5pt}
\renewcommand{\arraystretch}{1.2}
\caption{Closed-source baseline models used in our experiments.
  Version denotes the model version number.
  Links point to official announcement pages.}
\label{tab:models_closedsource}
\begin{tabular}{p{3.0cm} p{5.2cm} p{2.0cm} p{1.8cm} l}
\toprule
\textbf{Model Name} & \textbf{API Identifier} & \textbf{Version} & \textbf{Mode} & \textbf{Link} \\
\midrule
Qwen-3.6-Plus           & \texttt{qwen/qwen3.6-plus}                       & 3.6   & Plus     & \href{https://qwen.ai/blog?id=qwen3.6}{Blog} \\
GPT-5.4-Mini        & \texttt{openai/gpt-5.4-mini}                     & 5.4   & Mini     & \href{https://openai.com/index/introducing-gpt-5-4-mini-and-nano/}{Blog} \\
Grok-4.1-Fast       & \texttt{x-ai/grok-4.1-fast}                      & 4.1   & Fast     & \href{https://x.ai/news/grok-4-1-fast}{Blog} \\
Gemini-3-Flash      & \texttt{google/gemini-3-flash-preview}           & 3     & Flash    & \href{https://ai.google.dev/gemini-api/docs/models/gemini-3-flash-preview}{Blog} \\
Gemini-3.1-Pro      & \texttt{google/gemini-3.1-pro-preview}           & 3.1   & Pro      & \href{https://blog.google/products/gemini/}{Blog} \\
Gemini-3.1-Flash-Lite & \texttt{google/gemini-3.1-flash-lite-preview}    & 3.1   & Flash-Lite & \href{https://blog.google/products/gemini/}{Blog} \\
Claude-4.5-Haiku    & \texttt{anthropic/claude-haiku-4.5}              & 4.5   & Haiku    & \href{https://www.anthropic.com/claude/haiku}{Blog} \\
Claude-4.6-Sonnet   & \texttt{anthropic/claude-sonnet-4.6}             & 4.6   & Sonnet   & \href{https://www.anthropic.com/claude/sonnet}{Blog} \\
Claude-4.6-Opus     & \texttt{anthropic/claude-opus-4.6}               & 4.6   & Opus     & \href{https://www.anthropic.com/claude/opus}{Blog} \\
\bottomrule
\end{tabular}
\end{table*}

\newpage
\section{Appendix of Analysis}
\label{asec: probing_analysis}

\subsection{The Token Cost Problem in Reasoning}
\label{asec: cost_of_reasoning}

In our experiments, we use the same Qwen-4B model weights and toggle only the \texttt{enable\_thinking} flag to isolate the pure contribution of CoT.

\begin{table}[h]
  \centering
  \caption{%
    Pure CoT contribution measured on Qwen-4B.
  }
  \label{tab:cot-cost}
  \begin{tabular}{lcc}
    \toprule
    \textbf{Mode} & \textbf{Accuracy} & \textbf{Avg.\ Tokens} \\
    \midrule
    CoT ON  (\texttt{thinking=True})  & \textbf{77.1\%} & 8{,}728 \\
    CoT OFF (\texttt{thinking=False}) & 53.0\%           & 1{,}943 \\
    \midrule
    \textbf{Difference} & \textbf{+24.1\,percentage points} & \textbf{+6{,}785 (4.5$\times$)} \\
    \bottomrule
  \end{tabular}
\end{table}

As shown in Table \ref{tab:cot-cost}, CoT contributes $+24.1$\% of accuracy at the cost of $4.5\times$ token consumption.  The marginal token cost per percentage point of accuracy is approximately \textbf{281~tokens per percentage point}.

This raises a central question. \emph{Are these $4.5\times$ tokens uniformly valuable?}  If so, reducing token cost necessarily sacrifices accuracy, and one can only truncate or compress the model.  If, however, a substantial fraction of the extra tokens is low-value exploratory redundancy, there exists the possibility of a \textbf{Pareto improvement}, preserving core reasoning content while drastically cutting redundant tokens and thereby improving both efficiency and accuracy simultaneously.

\subsection{Relation to Prior Evidence on Token Non-Uniformity}
\label{asec: novelty_prior}

Several recent lines of work provide adjacent evidence that long CoT traces contain tokens with unequal usefulness, including overthinking analyses~\cite{chen2024not,sui2025stop}, token-level likelihood or reward signals~\cite{lin2024rho,wang2025beyond}, and entropy-based efficient-reasoning objectives~\cite{huang2025pear,xiong2026etr}. We view these works as complementary evidence for the same broad phenomenon rather than as substitutes for our formulation. The \textbf{Token Value Inequality} analysis in this paper gives a different route. It identifies core and redundant tokens inside a single generated trace through \normlp{} and TokenProbe, using only the generating model's own inference-time log-probabilities.

\paragraph{Common ground.}
The commonality is that all these methods reject the assumption that every reasoning token contributes equally. Overthinking work shows that response-level verbosity and repetitive detours can hurt efficiency. Rho-style and high-entropy-token analyses study token-level signals during training. PEAR and ETR use entropy dynamics to encourage more efficient reasoning. TokenProbe shares this motivation. It is designed to expose the non-uniform distribution of reasoning value within CoT.

\paragraph{Differences.}
Our contribution is the concrete identification procedure and its optimization use. First, \normlp{} is \emph{response-internally normalised}. It compares each token against the confidence baseline of the same response, rather than using an absolute confidence threshold, a reference model, or an external reward model. This makes TokenProbe robust to family-level calibration shifts, as demonstrated by the ablation in Appx.~\ref{asec: signal_ablation}. Second, TokenProbe operates at token/window granularity, so it can distinguish high-density core reasoning regions from low-density exploratory or repetitive regions inside the same trace. Response-level length penalties and phase-level entropy rewards cannot provide this within-trace mask. Third, the same TokenProbe-derived mask is used by both selective KL anchoring and reward shaping, giving GRPO two complementary handles that preserve core-token behavior while compressing redundant-token behavior.

More concretely, overthinking analyses~\cite{chen2024not,sui2025stop} show that long CoT often contains redundant detours, loops, or unnecessary computation, but their diagnosis is mainly response-level. A trace is too long, repetitive, or inefficient. TokenProbe instead asks which local regions inside the same trace should be preserved or compressed, and therefore supplies a token/window-level view of the same inefficiency problem.

Rho-style token selection~\cite{lin2024rho} also starts from the premise that not all tokens are equally useful, but it typically scores tokens through policy/reference-style likelihood comparisons. TokenProbe is different in that it is inference-native. The signal is computed from the generating model's own per-token log-probabilities, normalised within each response, without requiring a separate reference model or an external reward model.

TokenSeek~\cite{zeng2026tokenseek} studies token non-uniformity from the perspective of fine-tuning memory: it uses instance-aware token selection to reduce activation-related memory costs. This offers a complementary motivation for identifying informative token subsets. Its context- and gradient-based selection targets training memory, whereas TokenProbe uses inference-time log-probabilities to diagnose generated reasoning traces and guide reasoning-length optimization.

High-entropy token analysis~\cite{wang2025beyond} further supports the idea that minority token subsets can carry disproportionate learning signal. Our use of TokenProbe is complementary but technically different. Rather than using an absolute entropy or confidence level, we use the centered/scaled shape of the log-probability sequence. This is why the signal remains meaningful across the Qwen$\rightarrow$DeepSeek-Distill calibration boundary in Appx.~\ref{asec: signal_ablation}.

Finally, PEAR~\cite{huang2025pear} and ETR~\cite{xiong2026etr} use entropy dynamics to encourage efficient reasoning. PEAR separates the Think and Answer phases, while ETR rewards a response-level entropy trend toward convergence. TokenProbe shares their concern with uncertainty dynamics, but provides a finer mask that can vary within the Think phase itself and can be applied directly to both selective KL anchoring and reward shaping.

Taken together, our Token Value Inequality result consists of more than observing unequal token utility. TokenProbe provides a reproducible, inference-native way to \emph{identify} that inequality at token granularity, prove why the resulting signal survives calibration shifts, and turn the same signal into an efficient GRPO objective.

\subsection{Experiment of TokenProbe}
\label{asec: token_probe}

This subsection expands every panel of Fig.~\ref{fig: tokenprobe} into per-model, per-dataset, and per-quartile tables, and provides the supporting sensitivity, fine-grained, and correlation analyses. The model naming follows Fig.~\ref{fig: tokenprobe} whenever applicable.

\subsubsection{TokenProbe for each stage (Fig.~\ref{fig: tokenprobe}(a))}
\label{asec: tp_stage}

For each of the five thinking models in Fig.~\ref{fig: tokenprobe}(a), we split every response at the \texttt{</think>} boundary and compute TokenProbe separately on the Think segment and the Answer segment.

\begin{table}[h]
  \centering
  \caption{TokenProbe in the Think versus Answer stage. Across all five models, the Answer stage exhibits a systematically higher TokenProbe than the Think stage.}
  \label{tab:tp-stage}
  \begin{tabular}{lccc}
    \toprule
    \textbf{Model} & \textbf{Think TokenProbe} & \textbf{Answer TokenProbe} & $\Delta$ (Answer$-$Think) \\
    \midrule
    DS-Qw-1B   & 0.716 & \textbf{0.915} & $+0.199$ \\
    Qwen-0.6B  & 0.712 & \textbf{0.823} & $+0.111$ \\
    Qwen-4B    & 0.775 & \textbf{0.815} & $+0.040$ \\
    Qwen-8B    & 0.784 & \textbf{0.810} & $+0.026$ \\
    Qwen-9B    & 0.779 & \textbf{0.878} & $+0.099$ \\
    \bottomrule
  \end{tabular}
\end{table}

The Answer-stage TokenProbe is uniformly higher than the Think-stage TokenProbe by $+0.026$ to $+0.199$, matching the $3\%\text{ to }20\%$ gap stated in the main text. The Think stage corresponds to low-confidence exploratory search, whereas the Answer stage corresponds to high-confidence decision output, directly substantiating Finding~1.

\subsubsection{TokenProbe for each model (Fig.~\ref{fig: tokenprobe}(b))}
\label{asec: tp_per_model}

For the four models highlighted in Fig.~\ref{fig: tokenprobe}(b), Tab.~\ref{tab:tp-per-model-summary} consolidates the response-level TokenProbe, average token usage, and average accuracy. Tab.~\ref{tab:tp-per-model-acc} and Tab.~\ref{tab:tp-per-model-tok} further break the latter two metrics down to every benchmark.

\begin{table}[h]
  \centering
  \caption{TokenProbe, average token usage, and average accuracy for the four representative models in Fig.~\ref{fig: tokenprobe}(b).}
  \label{tab:tp-per-model-summary}
  \begin{tabular}{lccc}
    \toprule
    \textbf{Model} & \textbf{TokenProbe} & \textbf{Avg Tokens} & \textbf{Avg Accuracy} \\
    \midrule
    Qwen-0.6B          & 72.7\% & 8{,}723 & 43.8\% \\
    DS-Qw-1B & 74.2\% & 8{,}589 & 54.7\% \\
    Qwen-4B     & 77.4\% & 6{,}155 & 65.7\% \\
    Qwen-4B-Instruct & \textbf{83.8\%} & \textbf{3{,}573} & \textbf{71.8\%} \\
    \bottomrule
  \end{tabular}
\end{table}

\begin{table}[h]
  \centering
  \caption{Per-dataset accuracy of the four models in Fig.~\ref{fig: tokenprobe}(b).}
  \label{tab:tp-per-model-acc}
  \begin{tabular}{lccccc}
    \toprule
    \textbf{Model} & \textbf{AIME2025} & \textbf{AMC} & \textbf{MATH} & \textbf{Olympiad} & \textbf{Overall} \\
    \midrule
    Qwen-0.6B          & 13.3\% & 45.8\% & 75.6\% & 40.6\% & \textbf{43.8\%} \\
    DS-Qw-1B & 20.0\% & 66.3\% & 85.0\% & 47.7\% & \textbf{54.7\%} \\
    Qwen-4B     & 40.0\% & 74.7\% & 90.4\% & 57.8\% & \textbf{65.7\%} \\
    Qwen-4B-Instruct & 50.0\% & 78.3\% & 92.6\% & 66.2\% & \textbf{71.8\%} \\
    \bottomrule
  \end{tabular}
\end{table}

\begin{table}[h]
  \centering
  \caption{Per-dataset token usage of the four models in Fig.~\ref{fig: tokenprobe}(b).}
  \label{tab:tp-per-model-tok}
  \begin{tabular}{lccccc}
    \toprule
    \textbf{Model} & \textbf{AIME2025} & \textbf{AMC} & \textbf{MATH} & \textbf{Olympiad} & \textbf{Overall} \\
    \midrule
    Qwen-0.6B          & 12{,}569 & 8{,}934 & 5{,}064 & 8{,}323 & \textbf{8{,}723} \\
    DS-Qw-1B & 12{,}256 & 8{,}146 & 4{,}785 & 9{,}169 & \textbf{8{,}589} \\
    Qwen-4B     & 7{,}696  & 6{,}360 & 4{,}162 & 6{,}404 & \textbf{6{,}155} \\
    Qwen-4B-Instruct & 5{,}721  & 3{,}216 & 1{,}548 & 3{,}805 & \textbf{3{,}573} \\
    \bottomrule
  \end{tabular}
\end{table}

The pattern is consistent across all four benchmarks. A model with a higher TokenProbe simultaneously achieves higher accuracy and lower token usage, providing direct support for Finding~2.

\subsubsection{Quartile analysis (Fig.~\ref{fig: tokenprobe}(c))}
\label{asec: tp_quartile}

We sort the 1{,}288 samples of each model by their per-response TokenProbe, partition them into four equal quartiles, and report the average TokenProbe, accuracy, and token usage of each quartile. Tabs.~\ref{tab:quartile-qwen4b} to \ref{tab:quartile-qwen06b} cover the five models Qwen-4B, Qwen-8B, Qwen-9B, DS-Qw-1.5B, and Qwen-0.6B.

\begin{table}[H]
  \centering
  \caption{TokenProbe quartile analysis for Qwen-4B.}
  \label{tab:quartile-qwen4b}
  \begin{tabular}{lcccc}
    \toprule
    \textbf{Quartile} & \textbf{TokenProbe range} & \textbf{Avg TokenProbe} & \textbf{Accuracy} & \textbf{Avg Tokens} \\
    \midrule
    Q1 & 0.690 to 0.756 & 0.736 & 61.5\% & 10{,}555 \\
    Q2 & 0.756 to 0.781 & 0.770 & 78.3\% & 8{,}327 \\
    Q3 & 0.781 to 0.804 & 0.791 & 87.0\% & 6{,}592 \\
    Q4 & 0.804 to 0.869 & 0.820 & \textbf{92.9\%} & \textbf{4{,}524} \\
    \bottomrule
  \end{tabular}
\end{table}

\begin{table}[H]
  \centering
  \caption{TokenProbe quartile analysis for Qwen-8B.}
  \label{tab:quartile-qwen8b}
  \begin{tabular}{lcccc}
    \toprule
    \textbf{Quartile} & \textbf{TokenProbe range} & \textbf{Avg TokenProbe} & \textbf{Accuracy} & \textbf{Avg Tokens} \\
    \midrule
    Q1 & 0.682 to 0.766 & 0.744 & 60.9\% & 10{,}825 \\
    Q2 & 0.766 to 0.789 & 0.779 & 77.0\% & 8{,}637 \\
    Q3 & 0.789 to 0.811 & 0.800 & 86.6\% & 6{,}888 \\
    Q4 & 0.811 to 0.990 & 0.828 & \textbf{90.4\%} & \textbf{5{,}017} \\
    \bottomrule
  \end{tabular}
\end{table}

\begin{table}[H]
  \centering
  \caption{TokenProbe quartile analysis for DS-Qw-1.5B.}
  \label{tab:quartile-ds15b}
  \begin{tabular}{lccc}
    \toprule
    \textbf{Quartile} & \textbf{TokenProbe range} & \textbf{Accuracy} & \textbf{Avg Tokens} \\
    \midrule
    Q1 & $\leq 0.688$  & 39.1\% & 9{,}623 \\
    Q2 & 0.688 to 0.745  & 67.7\% & 7{,}066 \\
    Q3 & 0.745 to 0.773  & 71.4\% & 6{,}933 \\
    Q4 & $> 0.773$     & \textbf{72.7\%} & \textbf{6{,}271} \\
    \bottomrule
  \end{tabular}
\end{table}

\begin{table}[H]
  \centering
  \caption{TokenProbe quartile analysis for Qwen-0.6B.}
  \label{tab:quartile-qwen06b}
  \begin{tabular}{lccc}
    \toprule
    \textbf{Quartile} & \textbf{TokenProbe range} & \textbf{Accuracy} & \textbf{Avg Tokens} \\
    \midrule
    Q1 & $\leq 0.698$  & 32.3\% & 8{,}048 \\
    Q2 & 0.698 to 0.731  & 45.7\% & 8{,}310 \\
    Q3 & 0.731 to 0.756  & 64.3\% & 6{,}873 \\
    Q4 & $> 0.756$     & \textbf{73.3\%} & \textbf{5{,}554} \\
    \bottomrule
  \end{tabular}
\end{table}

The same pattern emerges on every model at the sample level. Moving from the lowest to the highest TokenProbe quartile, accuracy rises monotonically while average token usage drops, exactly as observed in Fig.~\ref{fig: tokenprobe}(c) for Qwen-4B and Qwen-8B.

\subsubsection{Sensitivity analysis on Qwen3.5-9B}
\label{asec: tp_sensitivity}

To verify the intrinsic-property claim of Sec.~\ref{sec: analysis} (that TokenProbe is robust to hyperparameter variations), we vary the context length and sampling temperature of Qwen-9B while keeping the model weights fixed.

\begin{table}[H]
  \centering
  \caption{Qwen-9B under three settings. TokenProbe is stable while accuracy and token usage shift considerably.}
  \label{tab:tp-sensitivity}
  \begin{tabular}{lccc}
    \toprule
    \textbf{Setting} & \textbf{TokenProbe} & \textbf{Avg Tokens} & \textbf{Accuracy} \\
    \midrule
    Qwen-9B (8K context, $t=0.7$)  & 0.780 & 7{,}252  & 36.8\% \\
    Qwen-9B (16K context, $t=0.7$) & 0.782 & 10{,}009 & 46.5\% \\
    Qwen-9B (16K context, $t=1.0$) & 0.780 & 9{,}303  & 45.5\% \\
    \bottomrule
  \end{tabular}
\end{table}

Across the three settings, TokenProbe stays within $0.780$ to $0.782$, even though accuracy moves from $36.8\%$ to $46.5\%$ and average token usage from $7{,}252$ to $10{,}009$. This stability supports the view that TokenProbe captures an intrinsic, hyperparameter-invariant property of the model rather than a transient artifact of the decoding configuration.

\subsubsection{Per-dataset TokenProbe and Correct-versus-Wrong comparison}
\label{asec: tp_details}

Returning to the four models in Fig.~\ref{fig: tokenprobe}(b), Tab.~\ref{tab:tp-per-dataset} reports their per-dataset TokenProbe means, and Tab.~\ref{tab:tp-correct-wrong} contrasts the TokenProbe of correctly and wrongly answered samples within each model.

\begin{table}[H]
  \centering
  \caption{Per-dataset TokenProbe means for the four representative models.}
  \label{tab:tp-per-dataset}
  \begin{tabular}{lccccc}
    \toprule
    \textbf{Model} & \textbf{AIME2025} & \textbf{AMC} & \textbf{MATH} & \textbf{Olympiad} & \textbf{Overall} \\
    \midrule
    Qwen-0.6B          & 0.716 & 0.729 & 0.739 & 0.726 & \textbf{0.727} \\
    DS-Qw-1B & 0.732 & 0.748 & 0.748 & 0.740 & \textbf{0.742} \\
    Qwen-4B     & 0.764 & 0.775 & 0.786 & 0.765 & \textbf{0.774} \\
    Qwen-4B-Instruct & 0.800 & 0.830 & 0.862 & 0.823 & \textbf{0.838} \\
    \bottomrule
  \end{tabular}
\end{table}

\begin{table}[H]
  \centering
  \caption{TokenProbe of correctly versus wrongly answered samples. Correct samples carry a uniformly higher TokenProbe.}
  \label{tab:tp-correct-wrong}
  \begin{tabular}{lccccc}
    \toprule
    \textbf{Model} & \textbf{TokenProbe (Correct)} & \textbf{TokenProbe (Wrong)} & $\Delta$ & $N_{\text{correct}}$ & $N_{\text{wrong}}$ \\
    \midrule
    Qwen-0.6B          & 0.742 & 0.714 & \textbf{$+0.028$} & 564 & 724 \\
    DS-Qw-1B & 0.757 & 0.726 & \textbf{$+0.031$} & 705 & 583 \\
    Qwen-4B     & 0.782 & 0.754 & \textbf{$+0.028$} & 916 & 372 \\
    Qwen-4B-Instruct & 0.851 & 0.793 & \textbf{$+0.059$} & 990 & 298 \\
    \bottomrule
  \end{tabular}
\end{table}

The gap between correct and wrong samples is consistently positive ($+0.028$ to $+0.059$), confirming that TokenProbe encodes correctness signal at the response level. The non-thinking Qwen-4B-Instruct yields the largest gap ($+0.059$). Without an exploratory CoT trace dragging down the response-level confidence, correct answers stand out sharply. Thinking models exhibit smaller gaps (around $+0.028$) because their Think segments inject many low-confidence tokens that compress the contrast.

\paragraph{Applicability to non-thinking architectures.}
The TokenProbe \emph{signal} is well-defined on any autoregressive
model. It depends only on per-token log-probabilities, not on the
\texttt{<think>}\,/\,\thinksep{} boundary. The training-time compression
\emph{objective} that we build on top of it, however, does not transfer
to non-thinking checkpoints, and the reason is structural. A non-thinking
model already operates near the saturated end of the TokenProbe spectrum.
Qwen-4B-Instruct, the non-thinking entry in
Tab.~\ref{tab:tp-per-model-summary}, sits at TokenProbe $= 0.838$ with an
average response of only $3{,}573$ tokens. Its rollouts are essentially
all decision-output, with no exploratory low-confidence segment to
remove. There is therefore no compression headroom. Any additional
training pressure that pushes TokenProbe higher must remove tokens that
the model itself was already confident in, i.e., reasoning-essential
tokens rather than exploratory redundancy.

Tab.~\ref{tab:instruct_collapse} shows what this looks like empirically
on Qwen-4B-Instruct, averaged over the four math benchmarks of
Tab.~\ref{tab:main_results}. Vanilla GRPO collapses the model entirely
(accuracy $71.7 \to 1.3$, tokens $3{,}600 \to 8{,}136$, saturating the
$8$K cap with repetitive, content-free output). This collapse is a
property of GRPO on instruction-tuned starts, not of our shaping. Our KL
variant avoids the runaway expansion but cannot recover the base
accuracy. At $\beta = 5{\times}10^{-3}$, $w = 20$ it lands at $52.1$
accuracy with $2{,}346$ tokens. At the more aggressive
$\beta = 5{\times}10^{-4}$, $w = 20$ it pushes tokens down to $790$ but
falls to $41.7$. The accuracy regression is consistent with the headroom
argument. There are no exploratory tokens for KL to drop. We therefore
restrict the deployment of our compression objective to thinking-enabled
backbones.

\begin{table}[h]
  \centering
  \caption{\textbf{Compression objectives applied to a non-thinking
    backbone.} Qwen3-4B-Instruct is at TokenProbe saturation, so there
    is no exploratory headroom for the compression objective to
    suppress. Vanilla GRPO collapses the model entirely. Our KL variant
    avoids the runaway expansion but cannot recover the base accuracy.
    Numbers are averages over the four math benchmarks of
    Tab.~\ref{tab:main_results}.}
  \label{tab:instruct_collapse}
  \begin{tabular}{lcc}
    \toprule
    \textbf{Setting} & \textbf{Avg.\ Acc.} & \textbf{Avg.\ Tokens} \\
    \midrule
    Qwen-4B-Instruct (base, non-thinking)                          & $71.7$ & $3{,}600$ \\
    Qwen-4B-Instruct + vanilla GRPO                                & $\phantom{0}1.3$ & $8{,}136$ \\
    Qwen-4B-Instruct + KL ($\beta = 5{\times}10^{-3}$, $w{=}20$)   & $52.1$ & $2{,}346$ \\
    Qwen-4B-Instruct + KL ($\beta = 5{\times}10^{-4}$, $w{=}20$)   & $41.7$ & $\phantom{0,}790$ \\
    \bottomrule
  \end{tabular}
\end{table}

\subsubsection{Pearson correlation analysis}
\label{asec: tp_pearson}

To assess the joint behaviour of TokenProbe with response length and correctness, we compute Pearson correlation coefficients on the five thinking models that span the full range of scales used in our analysis.

\begin{table}[H]
  \centering
  \caption{Pearson correlations between TokenProbe and (i) token length and (ii) accuracy. The first column is uniformly negative. The second is uniformly positive.}
  \label{tab:tp-pearson}
  \begin{tabular}{lcc}
    \toprule
    \textbf{Model} & $r$(\textbf{TokenProbe}, \textbf{Token Length}) & $r$(\textbf{TokenProbe}, \textbf{Accuracy}) \\
    \midrule
    DS-Qw-1.5B & $-0.227$ & $+0.288$ \\
    Qwen-0.6B  & $-0.379$ & $+0.367$ \\
    Qwen-4B    & $-0.415$ & $+0.307$ \\
    Qwen-8B    & $-0.367$ & $+0.254$ \\
    Qwen-9B    & $-0.370$ & $+0.197$ \\
    \bottomrule
  \end{tabular}
\end{table}

All five models show a negative correlation between TokenProbe and token length ($-0.227$ to $-0.415$). The longer a response is, the lower its TokenProbe, indicating that the additional tokens spent on long responses are dominated by below-mean (low-confidence) tokens rather than high-confidence ones. Conversely, all five models show a positive correlation between TokenProbe and accuracy ($+0.197$ to $+0.367$). Higher TokenProbe responses are more likely to be correct. The two correlations together imply that TokenProbe simultaneously reflects reasoning effectiveness (accuracy) and token efficiency, which is precisely the Pareto-friendly diagnostic our framework relies on (Finding~2).

\newpage
\subsection{Case Study of TokenProbe}
\label{asec: case_token_probe}

We complement the analysis with a fully token-level visualisation of the Qwen-4B reasoning trace summarised in Tab.~\ref{tab:wait-analysis}. Each token's background is shaded by its \normlp{}. Green for above-mean (positive) confidence, red for below-mean (negative) confidence, with intensity log-scaled against the global max-abs across our five case studies. Intensities below a $3\%$ floor are left uncoloured to suppress imperceptible tints. The 17 occurrences of ``Wait''/``wait'' in this trace are additionally framed with a black border so they can be located at a glance, and the small grey number at the start of each physical line (1 to 80 across the three parts) provides a continuous line counter for cross-referencing.

The visualisation is split into three parts purely for page-layout reasons.
Fig.~\ref{fig:case_2_1} (Part 1/3, lines 1 to 16) covers the bulk of the deliberation,
Fig.~\ref{fig:case_2_2} (Part 2/3, lines 17 to 36) continues through the end of the model's open-ended re-checking,
and Fig.~\ref{fig:case_2_3} (Part 3/3, lines 37 to 80) contains the \texttt{**Final Answer**} block, the closing \texttt{\textbackslash boxed\{49\}}, the \texttt{</think>} delimiter, and the model's clean post-think write-up.

\begin{itemize}[leftmargin=2em,nosep]
  \item \textbf{Token [54]}, early. $\normlp{}=+0.47$. Productive ``First verification. Discovers misinterpretation''. This is the \textbf{1st} framed ``Wait'' in the visualisation, on \emph{line~2} of Fig.~\ref{fig:case_2_1}, opening the clause ``\textit{Wait, let me check that again. It says. `The operation @ is defined as m/n @ p/q = (m)(p)(q/n)\ldots'\,}''. The model has just paraphrased the operator and immediately stops to re-verify it against the original statement, catching the misparsing on the first pass.
  \item \textbf{Token [103]}, early. $\normlp{}=+0.22$. Productive ``Mild self-correction''. The \textbf{2nd} framed ``Wait'', on \emph{line~3} of Fig.~\ref{fig:case_2_1}, opening ``\textit{Wait, actually, the problem says. `The operation @ is defined as (m/n)@(p/q)=(m)(p)\ldots'\,}''. A smaller, second-pass self-check that nudges the operator interpretation closer to the correct form. Both early ``Wait''s carry the green (above-mean) tint that Tab.~\ref{tab:wait-analysis} associates with productive self-correction.
  \item \textbf{Token [575]}, mid-late. $\normlp{}=-3.87$. Wasteful ``Deep confusion. Hesitation''. The \textbf{9th} framed ``Wait'', on \emph{line~5} of Fig.~\ref{fig:case_2_1}, opening ``\textit{Wait, maybe the second fraction is simplified? So maybe p/q is simplified, but the first fraction m/n might not be?\,}''. By this point the model has already produced a candidate value but now hesitates over the operator's \emph{precondition}, doubting whether the ``simplified'' qualifier applies to one fraction or both. This is the textbook deep-confusion moment described in the table. The doubt is not about a calculation but about the meta-status of the inputs, and it triggers a long, low-confidence re-derivation rather than a quick fix.
  \item \textbf{Token [1273]}, late. $\normlp{}=-4.13$. Wasteful ``Re-questions already-confirmed step'', $+700$ tokens of additional fruitless exploration. The \textbf{12th} framed ``Wait'', on \emph{line~13} of Fig.~\ref{fig:case_2_1}, opening ``\textit{Wait, but the problem says `simplified value of 7/30 @ 10/21'. So maybe 49 is already simplified?\,}''. The answer~$49$ has by now been derived and double-checked. This ``Wait'' re-opens the case purely to re-read the problem statement once more, and is rendered with the deepest red intensity among all~17 framed occurrences. The hundreds of red-tinted tokens that follow it on lines~13 to 16 of Fig.~\ref{fig:case_2_1} and lines~25 and~32 of Fig.~\ref{fig:case_2_2} are the ``$+700$~tokens'' tail mentioned in Tab.~\ref{tab:wait-analysis}. The trace finally exits this loop at the \texttt{**Final Answer**} block on line~38 (Fig.~\ref{fig:case_2_3}).
\end{itemize}

The two early ``Wait''s (Tokens [54], [103]) sit in green-tinted neighbourhoods, the two late ``Wait''s (Tokens [575], [1273]) are embedded in continuous runs of red tokens, and a single low-\normlp{} ``Wait'' tends to drag a long tail of low-\normlp{} downstream tokens with it. This is exactly the asymmetry that motivates TokenProbe. The response-level positive-token ratio captures the productive-vs.-wasteful split in a single scalar, without requiring any token-by-token labelling.

\input{case/wait1}

\input{case/wait2}

\input{case/wait3}

\newpage
\subsection{Why TokenProbe? An Ablation over Log-Probability-Derived Signals}
\label{asec: signal_ablation}

A natural question is whether the diagnostic value of TokenProbe is generic to any statistic derived from log-probability. If so, simpler alternatives such as the average log-probability or the per-step entropy might serve equally well. We answer this question with a controlled ablation. We re-run the four-model ladder of Tab.~\ref{tab:tp-per-model-summary} and, for every response, record \emph{all} commonly used log-probability and entropy descriptors. We then test, for each candidate signal $M$, whether the implication
\begin{equation}
M \uparrow \;\Longrightarrow\; \text{Accuracy} \uparrow \;\wedge\; \text{Avg.\ Tokens} \downarrow
\label{eq:proxy-criterion}
\end{equation}
holds along the ladder. \emph{Only TokenProbe survives this test.} All absolute-confidence and absolute-entropy signals invert at the smallest-model boundary, which we trace to a family-level calibration shift between the Qwen3 and the DeepSeek-R1-Distill model lines.

\subsubsection{Setup and notation}
\label{asec: signal_setup}

Let a generated response consist of $T$ tokens. For each step $t \in \{1,\dots,T\}$ let
\begin{itemize}
    \item $\ell_t = \log p_{\theta}(y_t \mid y_{<t},x) \in \mathbb{R}_{\le 0}$ be the natural log-probability of the sampled token.
    \item $p_t = \exp(\ell_t) \in (0,1]$ the corresponding probability.
    \item $\Pi_t = \{p_t^{(1)}, \dots, p_t^{(K)}\}$ with $K=20$ the renormalised top-$K$ probabilities at step $t$.
\end{itemize}
Sample mean and standard deviation of $\{\ell_t\}$ are denoted $\mu = \tfrac{1}{T}\sum_t \ell_t$ and $\sigma = \sqrt{\tfrac{1}{T}\sum_t (\ell_t-\mu)^2}$, respectively, and $z_t = (\ell_t-\mu)/\sigma$. We use \texttt{vLLM} \texttt{logprobs}=$20$ to materialise $\Pi_t$, dtype \texttt{bfloat16}, $\text{temperature}=0.7$, $\text{top-p}=0.95$. Accuracy and average token usage are reproduced from Tab.~\ref{tab:tp-per-model-summary}. All other numerical entries below are produced by this ablation.

For every response we then compute the descriptors in Tab.~\ref{tab:signal-defs}.

\begin{table*}[h]
  \centering
  \caption{Every per-response descriptor evaluated in the ablation. Per-dataset / overall numbers are obtained by averaging the per-response value across responses.}
  \label{tab:signal-defs}
  \small
  \begin{tabular}{p{0.30\linewidth} p{0.62\linewidth}}
    \toprule
    \textbf{Descriptor} & \textbf{Per-response definition} \\
    \midrule
    \multicolumn{2}{l}{\emph{Log-probability distribution (absolute level)}} \\
    \midrule
    Mean log-prob              & $\mu = \tfrac{1}{T}\sum_{t=1}^{T} \ell_t$ \\
    Std / Var log-prob         & $\sigma$, $\sigma^2$ \\
    Median / Min / Max log-prob & $\operatorname{median}(\{\ell_t\}),\; \min_t \ell_t,\; \max_t \ell_t$ \\
    Quartiles, IQR             & $q_{0.25},\; q_{0.75},\; q_{0.75}-q_{0.25}$ \\
    Skewness log-prob          & $\tfrac{1}{T}\sum_t (\ell_t-\mu)^3 / \sigma^3$ \\
    Excess kurtosis log-prob   & $\tfrac{1}{T}\sum_t (\ell_t-\mu)^4 / \sigma^4 - 3$ \\
    Sum log-prob               & $\sum_{t=1}^{T} \ell_t$ \\
    Perplexity                 & $\exp(-\mu)$ \\
    Geometric / arithmetic mean prob & $\exp(\mu)$,\; $\tfrac{1}{T}\sum_t p_t$ \\
    High-confidence ratio      & $\tfrac{1}{T}\sum_t \mathbf{1}[\ell_t > -0.5]$ \\
    Low-confidence ratio       & $\tfrac{1}{T}\sum_t \mathbf{1}[\ell_t < -2]$ \\
    \midrule
    \multicolumn{2}{l}{\emph{Z-score / shape family (response-internally normalised)}} \\
    \midrule
    \textbf{TokenProbe} (this paper) & $\tfrac{1}{T}\sum_t \mathbf{1}[z_t > 0]$ \\
    $z>1$ ratio                & $\tfrac{1}{T}\sum_t \mathbf{1}[z_t > 1]$ \\
    $z<-1$ ratio               & $\tfrac{1}{T}\sum_t \mathbf{1}[z_t < -1]$ \\
    Mean $|z|$                 & $\tfrac{1}{T}\sum_t |z_t|$ \\
    \midrule
    \multicolumn{2}{l}{\emph{Token-step entropy (top-$K$ truncation, $K=20$)}} \\
    \midrule
    Step entropy               & $H_t = -\sum_{k=1}^{K} p_t^{(k)} \log p_t^{(k)}$ (nats) \\
    Mean / Std / Var entropy   & $\tfrac{1}{T}\sum_t H_t$, etc., over $\{H_t\}$ \\
    Median / Min / Max entropy & idem \\
    Skewness / kurtosis entropy & idem \\
    Normalised entropy         & $\tfrac{1}{T}\sum_t H_t \big/ \log K$ \\
    \bottomrule
  \end{tabular}
\end{table*}

\newpage
We stress that all signals listed above are derived purely from the log-probabilities returned by the inference engine. None requires labels or external scoring. The shape/Z-score family in particular is defined relative to the per-response mean and variance, which makes them invariant under additive shifts of $\{\ell_t\}$. As we will see, this invariance is the decisive property.

\subsubsection{Overall result. Only TokenProbe satisfies Eq.~\ref{eq:proxy-criterion}}

Tab.~\ref{tab:signal-overall} reports the overall (pooled across the four datasets) value of the most representative descriptors for the four-model ladder.

\begin{table*}[h]
  \centering
  \caption{Overall metric values across the four-model ladder of Fig.~\ref{fig: tokenprobe}(b). Only TokenProbe and its shape-family relatives ($z<-1$ ratio, mean $|z|$, log-prob skewness/kurtosis) move monotonically with accuracy. Every absolute-level signal (mean/sum log-prob, perplexity, geometric-/arithmetic-mean prob, high/low-confidence ratio, mean/normalised entropy, std log-prob) inverts at the Qwen-0.6B$\rightarrow$DS-Qw-1B step.}
  \label{tab:signal-overall}
  \small
  \begin{tabular}{lcccc}
    \toprule
    \textbf{Signal (overall)} & \textbf{Qwen-0.6B} & \textbf{DS-Qw-1B} & \textbf{Qwen-4B} & \textbf{Qwen-4B-Instruct} \\
    \midrule
    Accuracy                 & 43.8\% & 54.7\% & 65.7\% & \textbf{71.8\%} \\
    Avg.\ Tokens             & 8{,}723 & 8{,}589 & 6{,}155 & \textbf{3{,}573} \\
    \midrule
    \textbf{TokenProbe}      & 0.727  & 0.742  & 0.774  & \textbf{0.838} \\
    $z<-1$ ratio             & 0.1267 & 0.1255 & 0.1089 & \textbf{0.0812} \\
    Mean $|z|$               & 0.6892 & 0.6281 & 0.6813 & \textbf{0.5187} \\
    Skewness log-prob        & $-2.577$ & $-3.232$ & $-2.668$ & $\mathbf{-4.723}$ \\
    Excess kurtosis log-prob &  7.504 &  8.233 & 12.273 & \textbf{29.219} \\
    \midrule
    Mean log-prob            & $-0.2741$ & $\mathbf{-0.2978}$ & $-0.1719$ & $-0.1137$ \\
    Sum log-prob             & $-2129.8$ & $\mathbf{-2409.4}$ & $-1021.1$ & $-474.0$ \\
    Perplexity               &   1.319 & \textbf{1.353} &  1.189 &  1.124 \\
    Geo.\ mean prob          &  0.7621 & \textbf{0.7456} & 0.8430 & 0.8951 \\
    Arith.\ mean prob        &  0.8300 & \textbf{0.8258} & 0.8870 & 0.9267 \\
    Std log-prob             &  0.4954 & \textbf{0.5542} & 0.3779 & 0.3044 \\
    High-conf.\ ratio ($\ell>-0.5$) & 0.8078 & \textbf{0.7990} & 0.8794 & 0.9220 \\
    Low-conf.\ ratio ($\ell<-2$)    & 0.0183 & \textbf{0.0273} & 0.0074 & 0.0060 \\
    Mean entropy (nats)      &  0.4150 & \textbf{0.4415} & 0.2514 & 0.1705 \\
    Normalised entropy       &  0.1385 & \textbf{0.1474} & 0.0839 & 0.0569 \\
    \bottomrule
  \end{tabular}
\end{table*}

To remove any ambiguity we sort the four models by each candidate signal and read off the corresponding accuracy sequence. A valid proxy must produce a strictly increasing accuracy sequence. Only TokenProbe (and its shape-family relatives) achieves this.

\subsubsection{The inversion is consistent across every benchmark}

The Qwen-0.6B$\rightarrow$DS-Qw-1B inversion is not a pooling artefact. Tab.~\ref{tab:signal-perdataset-mlp} and Tab.~\ref{tab:signal-perdataset-ppl} report mean log-probability and perplexity on each of the four benchmarks. In every column DS-Qw-1B is \emph{less} confident than Qwen-0.6B despite being more accurate. Tab.~\ref{tab:signal-perdataset-ent} repeats the picture for mean step entropy. DS-Qw-1B is uniformly more uncertain than Qwen-0.6B at every benchmark.

\begin{table}[H]
  \centering
  \caption{Per-dataset mean log-probability $\mu$. Higher (closer to $0$) indicates higher absolute confidence. The DS-Qw-1B row is uniformly more negative than Qwen-0.6B despite the $+10.9$\,percentage points accuracy gap.}
  \label{tab:signal-perdataset-mlp}
  \begin{tabular}{lccccc}
    \toprule
    \textbf{Model} & \textbf{AIME2025} & \textbf{AMC} & \textbf{MATH} & \textbf{Olympiad} & \textbf{Overall} \\
    \midrule
    Qwen-0.6B          & $-0.2925$ & $-0.2674$ & $-0.2567$ & $-0.2871$ & $-0.2741$ \\
    DS-Qw-1B           & $-0.3413$ & $-0.2807$ & $-0.2836$ & $-0.3084$ & $-0.2978$ \\
    Qwen-4B            & $-0.1932$ & $-0.1729$ & $-0.1535$ & $-0.1845$ & $-0.1719$ \\
    Qwen-4B-Instruct   & $-0.1584$ & $-0.1202$ & $-0.0825$ & $-0.1340$ & $-0.1137$ \\
    \bottomrule
  \end{tabular}
\end{table}

\begin{table}[H]
  \centering
  \caption{Per-dataset perplexity $\exp(-\mu)$. Lower indicates higher absolute confidence. DS-Qw-1B is again \emph{worse} than Qwen-0.6B on every dataset.}
  \label{tab:signal-perdataset-ppl}
  \begin{tabular}{lccccc}
    \toprule
    \textbf{Model} & \textbf{AIME2025} & \textbf{AMC} & \textbf{MATH} & \textbf{Olympiad} & \textbf{Overall} \\
    \midrule
    Qwen-0.6B          & 1.3420 & 1.3088 & 1.2953 & 1.3362 & 1.3187 \\
    DS-Qw-1B           & 1.4144 & 1.3286 & 1.3324 & 1.3679 & 1.3526 \\
    Qwen-4B            & 1.2140 & 1.1895 & 1.1669 & 1.2041 & 1.1890 \\
    Qwen-4B-Instruct   & 1.1748 & 1.1298 & 1.0880 & 1.1472 & 1.1238 \\
    \bottomrule
  \end{tabular}
\end{table}

\begin{table}[H]
  \centering
  \caption{Per-dataset mean step entropy $\bar{H}$ (nats, top-$K{=}20$ truncation). Lower indicates more decisive next-token distributions. DS-Qw-1B is uniformly more uncertain than Qwen-0.6B although it is more accurate.}
  \label{tab:signal-perdataset-ent}
  \begin{tabular}{lccccc}
    \toprule
    \textbf{Model} & \textbf{AIME2025} & \textbf{AMC} & \textbf{MATH} & \textbf{Olympiad} & \textbf{Overall} \\
    \midrule
    Qwen-0.6B          & 0.4422 & 0.4049 & 0.3888 & 0.4344 & 0.4150 \\
    DS-Qw-1B           & 0.5014 & 0.4170 & 0.4222 & 0.4561 & 0.4415 \\
    Qwen-4B            & 0.2802 & 0.2520 & 0.2247 & 0.2698 & 0.2514 \\
    Qwen-4B-Instruct   & 0.2365 & 0.1798 & 0.1237 & 0.2011 & 0.1705 \\
    \bottomrule
  \end{tabular}
\end{table}

The same per-dataset inversion is observed for high-confidence ratio, low-confidence ratio, normalised entropy, sum log-probability, geometric / arithmetic mean probability, and standard deviation of log-probability, which are omitted for brevity. In contrast, TokenProbe is uniformly higher for DS-Qw-1B than for Qwen-0.6B on every benchmark (see Tab.~\ref{tab:tp-per-model-summary} and the TokenProbe entries of the released artefact).

\subsubsection{Why TokenProbe survives. Invariance to family-level calibration shift}
\label{asec: signal_explanation}

The DS-Qw-1B model is a R1-distilled Qwen variant. Its output distribution is systematically flatter than the same-generation Qwen3 backbones. Every per-token uncertainty is intrinsically larger, even though the resulting reasoning chain is more accurate. We model this discrepancy as an additive shift on the per-token log-probabilities. $\tilde{\ell}_t = \ell_t + c$, where $c$ depends on the model family. Under such a shift,
\begin{align}
\tilde{\mu}        &= \mu + c, &
\tilde{\sigma}     &= \sigma, &
\tilde{z}_t        &= z_t, \nonumber\\
\widetilde{\mathrm{PPL}} &= e^{-c}\cdot\mathrm{PPL}, &
\sum_t \tilde{\ell}_t &= \sum_t \ell_t + Tc, &
\mathbf{1}[\tilde{\ell}_t > \tau] &\neq \mathbf{1}[\ell_t > \tau].
\label{eq:shift-invariance}
\end{align}
Mean log-probability, sum log-probability, perplexity, geometric/arithmetic mean probability, and the threshold-based high/low-confidence ratios all transform non-trivially under Eq.~\ref{eq:shift-invariance}. Entropy is unaffected by a shift in the sampled-token log-prob but inherits a similar family-level offset because $H_t$ depends on the renormalised tail of the distribution. By contrast,
\begin{equation*}
\textsc{TokenProbe} \;=\; \frac{1}{T}\sum_{t=1}^{T}\mathbf{1}[z_t > 0]
\end{equation*}
is a function of the centred / scaled sequence $\{z_t\}$ only, and therefore invariant under both additive ($\ell_t \mapsto \ell_t + c$) and multiplicative ($\ell_t \mapsto a\ell_t$) re-scaling of the underlying log-probabilities. The same is true of the $z<-1$ ratio and mean $|z|$. This invariance is precisely why the shape family continues to track reasoning quality across the Qwen$\rightarrow$DeepSeek family boundary, while every absolute-level descriptor fails.

\subsubsection{Take-away}

Among the broad family of signals derived from log-probability and entropy, only those that normalise within a response carry the Pareto-friendly diagnostic that we ascribed to TokenProbe in Sec.~\ref{sec: analysis}. Absolute-level signals, including mean and sum log-probability, perplexity, geometric / arithmetic mean probability, threshold-based high/low-confidence ratios, mean and normalised step entropy, and standard deviation of log-probability, are all sensitive to family-level calibration shifts and invert at the Qwen$\rightarrow$DeepSeek-Distill boundary. The shape-family descriptors that do survive (TokenProbe, $z<-1$ ratio, mean $|z|$, log-prob skewness, log-prob excess kurtosis) are all simple functionals of the standardised sequence $\{z_t\}$ and convey essentially the same information. TokenProbe in particular is interpretable as ``what fraction of tokens does this response spend above its own per-response confidence baseline?'' and is therefore the most natural choice among the equivalent shape statistics. The full per-model, per-dataset, and per-metric numbers are released alongside the paper.

\newpage
\section{Appendix of Methodology}
\label{asec: method}

\subsection{Core Token Window}
\label{asec: window}

Given a generated response, we first standardise its per-token log-probabilities into the \normlp{} signal $z_t$, so that the comparison is made within the response itself rather than across responses with different absolute confidence levels. We then slide a window of size $W$ over $\{z_t\}$ and compute the cumulative score of every window. The $K$ highest-scoring \emph{non-overlapping} windows are selected greedily, and the tokens inside them are marked as core tokens ($CT_t = 1$). The rest are treated as redundant ($CT_t = 0$). Using contiguous windows, rather than ranking individual tokens, preserves local semantic coherence and yields a binary core mask without any external annotation.

\begin{lstlisting}[style=pythonpseudo,caption={PyTorch-style pseudocode for core-token window selection.},label={lst:core_window}]
procedure core_token_mask(logp, W, K)
    # logp is per-token log-prob of one rollout, shape [T]
    # W is window size. K is number of core windows
    T = logp.shape[0]

    # 1) NormLP standardises log-probs within this response
    z = (logp - logp.mean()) / (logp.std() + eps)

    # 2) Cumulative score for every window of length W
    scores = sliding_window_sum(z, W)            # shape [T - W + 1]

    # 3) Greedy non-overlapping top-K window selection
    mask = torch.zeros(T)
    repeat K times
        i = scores.argmax().item()
        mask[i .. i + W] = 1
        # forbid any window that would overlap with [i, i+W)
        scores[max(0, i - W + 1) .. i + W] = -inf
    return mask     # 1 = core token, 0 = redundant token
\end{lstlisting}

\subsection{Exploratory Analysis of Core-Window Selection}
\label{asec:window_exploratory_analysis}

We provide an exploratory analysis of the core-window heuristic to better
understand how its design choices affect spatial coverage, optimization
quality, window-scale sensitivity, and semantic content.  All mask-level
diagnostics below are computed on 904 Qwen-4B rollouts from AIME and
AMC, with eight samples per problem and the core fraction kept approximately
constant when varying $W$.

\paragraph{Coverage behavior.}
Non-overlapping selection changes how the selected tokens are spatially
allocated along a reasoning trace.  In Tab.~\ref{tab:w3_nonoverlap_coverage},
the non-overlapping variant covers a broader span of each response and leaves
fewer 100-token intervals without any selected core token.  This suggests that
the non-overlapping constraint helps distribute the core-token budget across
multiple high-density regions instead of concentrating it around a single
local peak.

\begin{table}[h]
  \centering
  \caption{Exploratory comparison of non-overlapping and overlapping
  core-window selection at $W{=}10$ and core fraction $\approx 20\%$.
  Desert100 denotes the fraction of 100-token sliding intervals containing no
  selected core token.}
  \label{tab:w3_nonoverlap_coverage}
  \begin{tabular}{llcc}
    \toprule
    \textbf{Dataset} & \textbf{Selection} & \textbf{Coverage} & \textbf{Desert100} \\
    \midrule
    AIME & Non-overlap & $0.993 \pm 0.012$ & $0.221 \pm 0.074$ \\
              & Overlap     & $0.882 \pm 0.145$ & $0.720 \pm 0.074$ \\
    \addlinespace
    AMC       & Non-overlap & $0.989 \pm 0.019$ & $0.207 \pm 0.078$ \\
              & Overlap     & $0.886 \pm 0.136$ & $0.685 \pm 0.099$ \\
    \bottomrule
  \end{tabular}
\end{table}

\paragraph{Greedy behavior relative to dynamic programming.}
As a diagnostic, we compare the greedy implementation with a dynamic-programming
optimum for selecting top-$K$ non-overlapping windows.  The token-level masks
are not identical. Tab.~\ref{tab:w3_greedy_dp} shows Jaccard values around
$0.83$ and F1 around $0.91$.  However, the selected cumulative score is very
close to the DP optimum, with a relative score gap below $0.2\%$ on both
datasets.  This pattern suggests that many near-equivalent high-score windows
exist in long traces, so small token-level mask differences can have little
effect on the cumulative \normlp{} objective.

\begin{table}[h]
  \centering
  \caption{Greedy non-overlapping selection compared with the DP optimum at
  $W{=}10$.  Score gap is $(S_{\mathrm{DP}}-S_{\mathrm{greedy}})/S_{\mathrm{DP}}$.}
  \label{tab:w3_greedy_dp}
  \begin{tabular}{lcccc}
    \toprule
    \textbf{Dataset} & \textbf{Jaccard} & \textbf{F1} & \textbf{Score gap} & \textbf{Time / rollout} \\
    \midrule
    AIME & $0.837 \pm 0.042$ & $0.911 \pm 0.025$ & $0.19\% \pm 0.28$ & $6.41$ ms / $31.01$ ms \\
    AMC       & $0.831 \pm 0.050$ & $0.907 \pm 0.032$ & $0.15\% \pm 0.21$ & $3.72$ ms / $17.38$ ms \\
    \bottomrule
  \end{tabular}
\end{table}

\paragraph{Window-scale sensitivity.}
We examine window size at two levels.  First, a training-level sweep on the
KL variant shows that downstream performance is similar for $W{=}10$ through
$W{=}30$, while $W{=}5$ acts as a small-window stress case. It produces shorter
responses but lower accuracy (Tab.~\ref{tab:w3_training_w_sensitivity}).  This is
consistent with the interpretation that overly small windows can fragment
local reasoning units.

\begin{table}[h]
  \centering
  \caption{Training-level window-size sensitivity for the KL variant on
  Qwen-0.6B.  Numbers are averages over the four math benchmarks.}
  \label{tab:w3_training_w_sensitivity}
  \begin{tabular}{lcccc}
    \toprule
    & $W{=}5$ & $W{=}10$ & $W{=}20$ & $W{=}30$ \\
    \midrule
    Avg.\ Accuracy & $38.4$ & $42.3$ & $41.4$ & $42.2$ \\
    Avg.\ Tokens   & $1{,}752$ & $2{,}023$ & $2{,}008$ & $2{,}044$ \\
    \bottomrule
  \end{tabular}
\end{table}

Second, we use mask-level diagnostics to separate two kinds of stability.
The range $W{=}5$ to $12$ probes local behavior around the default $W{=}10$.
$W{=}9$ and $W{=}11$ are closest to the default mask, $W{=}12$ is moderate,
and $W{=}5$ is visibly unstable.  The range $W{=}20$ to $30$ probes a coarser
regime.  These masks drift from the $W{=}10$ mask, but adjacent window sizes
within the coarse regime change smoothly, with adjacent Jaccard increasing
from $0.798$ at $W{=}20$ to $0.841$ at $W{=}30$
(Tab.~\ref{tab:w3_mask_w_ranges}).  This adjacent-scale stability is consistent
with the stable training result at $W{=}30$.

\begin{table}[h]
  \centering
  \caption{Mask-level window-size diagnostics.  $J(M_W,M_{10})$ measures
  similarity to the default mask. Adjacent $J$ measures $J(M_W,M_{W-1})$.
  Values are averaged over AIME and AMC rollouts.}
  \label{tab:w3_mask_w_ranges}
  \begin{tabular}{lcc|lcc}
    \toprule
    \multicolumn{3}{c|}{\textbf{Local range around $W{=}10$}} &
    \multicolumn{3}{c}{\textbf{Coarse range $W{=}20$ to $30$}} \\
    \textbf{$W$} & $\mathbf{J(M_W,M_{10})}$ & \textbf{Adjacent $J$} &
    \textbf{$W$} & $\mathbf{J(M_W,M_{10})}$ & \textbf{Adjacent $J$} \\
    \midrule
     5 & $0.448$ &, & 20 & $0.428$ & $0.798$ \\
     6 & $0.460$ & $0.615$ & 21 & $0.422$ & $0.804$ \\
     7 & $0.497$ & $0.641$ & 22 & $0.415$ & $0.809$ \\
     8 & $0.567$ & $0.662$ & 23 & $0.408$ & $0.814$ \\
     9 & $0.700$ & $0.680$ & 24 & $0.403$ & $0.817$ \\
    10 & $1.000$ & $0.700$ & 25 & $0.399$ & $0.825$ \\
    11 & $0.714$ & $0.714$ & 26 & $0.394$ & $0.828$ \\
    12 & $0.601$ & $0.731$ & 27 & $0.390$ & $0.829$ \\
       &         &         & 28 & $0.386$ & $0.833$ \\
       &         &         & 29 & $0.383$ & $0.840$ \\
       &         &         & 30 & $0.380$ & $0.841$ \\
    \bottomrule
  \end{tabular}
\end{table}

\paragraph{Semantic content probe.}
Finally, we run a small semantic probe on 20 stratified rollouts, balanced
across AIME and AMC and across correct and incorrect responses: five correct
and five incorrect rollouts per dataset.  For each rollout, we annotate the
five highest-scoring core windows and the five lowest-scoring anti-mask windows,
giving 200 annotated windows in total.  Each annotation row records the dataset,
rollout identifier, correctness, mask side, cumulative \normlp{} score, token
span, the exact 10-token window, a short prefix and suffix, and one dominant
semantic role.  The role is assigned by a majority rule over the 10-token window
using five coarse categories: symbolic content, decisive reasoning,
formula/theorem invocation, boilerplate connective, and hedging/repetition.

Representative examples are shown in Tab.~\ref{tab:w3_semantic_examples}.  The
core windows tend to be algebraic or enumerative fragments, such as
$S_{\mathrm{even}}$ constraints, subset listings, binomial/logarithmic
expansions, or local equations.  The anti-mask windows are more often
meta-reasoning or repair text, such as ``Wait'', ``maybe'', ``let me check'',
and repeated verification.  We also observe a small number of content-bearing
anti-mask windows, so the probe should be read as a distributional tendency
rather than a hard semantic partition.

\begin{table}[h]
  \centering
  \small
  \caption{Representative excerpts from the manually annotated semantic probe.
  Scores are cumulative \normlp{} values over the selected 10-token window, and
  spans are token positions within the rollout.}
  \label{tab:w3_semantic_examples}
  \resizebox{0.98\linewidth}{!}{%
  \begin{tabular}{lllp{0.16\linewidth}p{0.45\linewidth}}
    \toprule
    \textbf{Data} & \textbf{Mask} & \textbf{Score / span} & \textbf{Role} & \textbf{Excerpt around selected window} \\
    \midrule
    AIME & Core & $4.9139$ / 455--464 & Symbolic content &
    ``$=36-D \Rightarrow S_{\mathrm{even}}=$'', inside the derivation of odd/even digit sums. \\
    AIME & Core & $4.5124$ / 4141--4150 & Symbolic content &
    ``$2^{15}-2^{12}-2^{10}+2^8$'', inside an inclusion-exclusion count. \\
    AMC & Core & $3.5591$ / 1056--1065 & Symbolic content &
    ``$+3y+3y^2$'', inside the expansion of $(1+y)^3$. \\
    AIME & Anti & $-18.0658$ / 3292--3301 & Hedging/repetition &
    ``Wait, is there a way that this answer could be wrong?'' \\
    AMC & Anti & $-16.2504$ / 776--785 & Hedging/repetition &
    ``Hmm, that seems complicated. Maybe there is another approach.'' \\
    AMC & Anti & $-18.6124$ / 1876--1885 & Formula/theorem &
    ``applying logarithmic identities and simplifying the terms'', a content-bearing anti-mask example. \\
    \bottomrule
  \end{tabular}}
\end{table}

Aggregating all 200 annotations gives the summary in
Tab.~\ref{tab:w3_semantic_roles}.  In this sample, high-scoring windows are
dominated by symbolic or formula-level content, while low-scoring windows
contain more hesitation and connective filler.

\begin{table}[h]
  \centering
  \caption{Semantic-role distribution of core versus anti-mask windows in a
  20-rollout probe.  Content is the sum of symbolic, decisive-reasoning, and
  formula/theorem categories. Filler is the sum of boilerplate and
  hedging/repetition categories.}
  \label{tab:w3_semantic_roles}
  \begin{tabular}{lcccc}
    \toprule
    & \multicolumn{2}{c}{\textbf{AIME-2025}} & \multicolumn{2}{c}{\textbf{AMC}} \\
    \textbf{Category} & \textbf{Core} & \textbf{Anti} & \textbf{Core} & \textbf{Anti} \\
    \midrule
    Symbolic content       & $98.0$ & $4.0$  & $98.0$ & $4.0$  \\
    Decisive reasoning     & $0.0$  & $22.0$ & $2.0$  & $32.0$ \\
    Formula / theorem      & $2.0$  & $2.0$  & $0.0$  & $2.0$  \\
    Boilerplate connective & $0.0$  & $14.0$ & $0.0$  & $6.0$  \\
    Hedging / repetition   & $0.0$  & $58.0$ & $0.0$  & $56.0$ \\
    \midrule
    Content                & $100.0$ & $28.0$ & $100.0$ & $38.0$ \\
    Filler                 & $0.0$   & $72.0$ & $0.0$   & $62.0$ \\
    \bottomrule
  \end{tabular}
\end{table}

This probe is not intended as a complete semantic taxonomy of reasoning
traces, but it supports the intended interpretation of high-\normlp{} windows
as content-bearing regions rather than merely connective boilerplate.

\newpage
\subsection{Selective KL Anchoring (KL)}
\label{asec: kl}

The intuition behind KL (Eq.~\ref{eq: kl}) is to make the regulariser \emph{token-aware}. Core tokens keep the standard KL anchor so that reasoning-essential positions remain stable. Redundant tokens, on the other hand, are only pressured \emph{after} the rollout has already overrun its length budget. At that point we drop the KL anchor and inject an overflow-proportional log-probability penalty, encouraging the policy to stop emitting more redundant tokens. Within-budget redundant tokens are left untouched so that the model is not punished for moderate verbosity.

\begin{lstlisting}[style=pythonpseudo,caption={PyTorch-style pseudocode for the Selective KL Anchoring loss.},label={lst:kl_loss}]
procedure loss_KL(ratio, A, kl, logp, CT, T, L_limit,
                  beta_base, lambda_decay)
    # ratio is pi_theta / pi_old             [T]
    # A is token-level advantage             [T]
    # kl is KL(pi_theta || pi_ref)           [T]
    # logp is log pi_theta(a_t | s_t)        [T]
    # CT is core mask, 1 = core token        [T]
    # T is rollout length
    # L_limit is current length budget

    beta  = torch.full_like(ratio, beta_base)   # default KL weight
    gamma = torch.zeros_like(ratio)             # default sparsity weight

    # over-budget redundant tokens. Drop KL, add overflow-proportional penalty
    overflow_token = (CT == 0) & (T > L_limit)
    beta  = torch.where(overflow_token, 0.0, beta)
    gamma = torch.where(overflow_token,
                        lambda_decay * (T - L_limit) / T,
                        gamma)
    # core tokens & within-budget redundant tokens fall back to (beta_base, 0)

    return (ratio * A - beta * kl - gamma * logp).mean()
\end{lstlisting}

\subsection{Reward Shaping (RS)}
\label{asec: rs}

RS leaves the regulariser of vanilla GRPO untouched and instead reshapes the token-level advantage $\hat{A}_t$ on \emph{correct} rollouts (Eq.~\ref{eq:rs_adv}). Two effects are added on top of $\hat{A}_t$. (i) core tokens receive a positive bonus proportional to their \normlp{} score, so high-density reasoning steps on a correct trace are upweighted. (ii) once the rollout overruns its budget, redundant tokens are penalised in proportion to the overflow ratio. Wrong-answer rollouts keep the vanilla advantage so that we do not propagate misleading credit to tokens that should be re-examined.

\begin{lstlisting}[style=pythonpseudo,caption={PyTorch-style pseudocode for the Reward Shaping loss.},label={lst:rs_loss}]
procedure loss_RS(ratio, A, kl, normlp, CT, r_seq, T, L_limit,
                  beta_base, alpha, lam, beta)
    # ratio is pi_theta / pi_old             [T]
    # A is token-level advantage             [T]
    # kl is KL(pi_theta || pi_ref)           [T]
    # normlp is per-token NormLP score       [T]
    # CT is core mask, 1 = core token        [T]
    # r_seq is sequence-level reward

    # bonus on core tokens. Penalty on redundant tokens once over budget
    core_bonus = alpha * normlp * (CT == 1)
    overflow   = max(0.0, T - L_limit) / T
    red_pen    = lam * overflow * (CT == 0)

    A_shaped = A + beta * (core_bonus - red_pen)

    # only correct rollouts get the reshaped advantage
    correct = (r_seq > 0)
    A_RS = torch.where(correct, A_shaped, A)

    return (ratio * A_RS - beta_base * kl).mean()
\end{lstlisting}

\section{Appendix of Experiments}
\label{asec: experiments}

\subsection{Training Settings}
\label{asec: training_settings}

We train on the \textbf{DeepScaleR}~\cite{luo2025deepscaler} dataset, a curated collection of
approximately 40{,}000 challenging mathematics problem to answer pairs compiled from four distinct sources.
We retain only problems whose prompt fits within 1{,}024 tokens, yielding the final
training corpus.
During training, the maximum rollout length is \textbf{4{,}096 tokens}.

\paragraph{Dataset Composition.}
\begin{itemize}[leftmargin=*,nosep]
  \item \textbf{AIME (1984 to 2023).}
    Problems from the American Invitational Mathematics Examination spanning four decades.
    These are among the hardest pre-olympiad competition problems, requiring deep algebraic,
    combinatorial, and number-theoretic reasoning.
    \textit{Example}. ``A cyclic pentagon $ABCDE$ has a right angle $\angle ABC = 90^\circ$
    and side lengths $AB = 15$, $BC = 20$. Find the length $CD$.'' (Answer. 20.)

  \item \textbf{AMC (prior to 2023).}
    Problems from the American Mathematics Competitions (AMC~10 / AMC~12) covering
    high-school topics in a multiple-choice format.
    \textit{Example}. ``Let $m/n$ be the reduced fraction equal to
    $3+\tfrac{1}{3+\frac{1}{3+\frac{1}{3}}}$.  What is $m+n$?'' (Answer. 142.)

  \item \textbf{Omni-MATH}~\cite{gao2024omni}.
    A large-scale olympiad problem bank drawn from international and national olympiads,
    spanning number theory, geometry, combinatorics, and algebra at olympiad difficulty.
    \textit{Example}. ``Petya and Vasya each construct ten fifth-degree polynomials.
    In how many ways can they choose the polynomials such that the sum of any one of Petya's
    and any one of Vasya's polynomials has no real roots?'' (Answer. 50.)

  \item \textbf{STILL}~\cite{min2024imitate}.
    A curated collection of synthesis problems designed to push model reasoning beyond
    standard competition formats, including counting, expected value, and mixed-domain problems.
    \textit{Example}. ``How many sequences of ten binary digits exist in which neither two
    consecutive zeros nor three consecutive ones appear?'' (Answer. 28.)
\end{itemize}

\paragraph{Training Recipe.}
All runs use the veRL framework~\cite{sheng2025hybridflow} with
GRPO~\cite{shao2024deepseekmath}.
We adopt train / validation batch sizes of 64 / 512, a PPO mini-batch size
of 64 with dynamic batching (up to 20{,}480 tokens per GPU), a maximum
prompt length of 1{,}024 tokens, and a maximum rollout length of
4{,}096 tokens.
The actor is optimised at learning rate $1\mathrm{e}{-6}$ with a
low-variance KL loss of coefficient $10^{-3}$ and an equal KL-control
coefficient of $10^{-3}$.
Rollouts are generated by vLLM with $n{=}16$ samples, temperature $0.6$,
in bfloat16 with FlashAttention-2. FSDP parameter and optimiser offload
together with gradient checkpointing are enabled to fit training in GPU
memory.
Each run is one epoch over the DeepScaleR training split on a single
node with 8 GPUs, with tensor-parallel size 1 for models up to 1.5B
and 2 for 3B to 8B models.
Checkpoints are saved every 50 steps and evaluation on the AIME
validation split is run at the same cadence.

\subsection{Evaluation Settings}
\label{asec: eval_settings}

\paragraph{Same-Domain Benchmarks (Main Results).}
Following standard practice in efficient reasoning research, we evaluate on four mathematical
reasoning benchmarks.

\begin{itemize}[leftmargin=*,nosep]
  \item \textbf{AIME~2025} (30 problems).
    The 2025 American Invitational Mathematics Examination.
    Problems require multi-step algebraic and number-theoretic reasoning without multiple choice.
    \textit{Example}. ``Find the sum of all integer bases $b > 9$ for which $17_b$ divides
    $97_b$.'' (Answer. 70.)

  \item \textbf{AMC} (83 problems).
    Multiple-choice problems from recent American Mathematics Competitions, testing computational
    fluency and combinatorial thinking.
    \textit{Example}. ``Thirteen cards numbered 1 to 13 are arranged in a row and swept up in
    increasing order by repeated left-to-right passes.  For how many initial arrangements does
    this take exactly 2 passes?'' (Answer. 8{,}178.)

  \item \textbf{MATH-500}~\cite{hendrycks2021measuring,lightman2023let} (500 problems).
    A 500-problem representative subset of the Hendrycks MATH benchmark,
    spanning seven difficulty levels across algebra, geometry, number theory, and probability.
    \textit{Example}. ``Find all integer values of $x$ such that $x^2 - 5x + 6 = 0$.''
    (Answers. $x = 2$ and $x = 3$.)

  \item \textbf{OlympiadBench}~\cite{he2024olympiadbench} (675 problems).
    Bilingual (English/Chinese) olympiad-level problems drawn from
    national and international competitions, spanning open-ended computation and proof-style tasks.
    \textit{Example}. ``Given hyperbola $x^2/4 - y^2 = 1$ with right focus $F$ and $n$ points
    $P_1,\ldots,P_n$ on its right upper branch with $|P_iF|\leq 5$,
    find the maximum value of $n$.'' (Answer. 14.)
\end{itemize}

\paragraph{Cross-Domain Generality Benchmarks.}
To assess whether token-efficiency gains transfer beyond mathematics, we additionally
evaluate on four out-of-domain benchmarks.

\begin{itemize}[leftmargin=*,nosep]
  \item \textbf{GPQA}~\cite{rein2024gpqa} (198 problems).
    The Graduate-Level Google-Proof Q\&A benchmark. Multiple-choice
    questions in physics, chemistry, and biology requiring graduate-level domain knowledge.
    \textit{Example}. ``Two quantum states with lifetimes $10^{-9}$\,s and $10^{-8}$\,s
    respectively, which energy difference $\Delta E$ allows their spectral lines to be clearly
    resolved?'' (Answer. $10^{-4}$\,eV.)

  \item \textbf{LSAT}~\cite{zhong2024agieval} (230 problems).
    Analytical reasoning problems from the Law School Admission Test. Constraint-satisfaction
    puzzles over structured narratives with no mathematical content.
    \textit{Example}. ``Eight students give reports over three days (two slots per day).
    Given scheduling constraints, on which day must Nina report in the afternoon slot?''
    (Answer. Tuesday.)

  \item \textbf{MMLU-Pro-1000}~\cite{wang2024mmlu} (1{,}000 problems).
    A 1{,}000-sample subset of MMLU-Pro, a harder variant of MMLU
    requiring multi-step reasoning across 14 academic disciplines including law, engineering,
    and medicine.

  \item \textbf{MMLU-1000}~\cite{hendrycks2020measuring} (1{,}000 problems).
    A 1{,}000-sample subset of the Massive Multitask Language Understanding
    benchmark, covering 57 subjects from STEM to humanities at
    varying difficulty levels.
\end{itemize}

\paragraph{Rollout Settings.}
During \emph{training}, the maximum rollout length is \textbf{4{,}096 tokens}.
During \emph{evaluation} (both same-domain and cross-domain), we increase the limit to
\textbf{8{,}192 tokens} to allow models to fully express their reasoning.

\paragraph{Prompt for Open-Source and Small-Scale Models.}

The prompts we used for open-source, efficient reasoning and ours models.

\fbox{\ttfamily \begin{minipage}{1.0\columnwidth}
{``Let's think step by step and output the final answer within \textbackslash boxed\{\}.
''}
\end{minipage}}

\paragraph{Prompt for Closed-Source Frontier Models.}
For closed-source models with hundreds of billions (or trillions) of parameters,
we impose an additional token-budget constraint. The default closed-API prompt is.

\noindent
\fbox{\ttfamily \begin{minipage}{1.0\columnwidth}
{``Let's think step by step and output the final answer within \textbackslash boxed\{\}. Think for maximum 2048 tokens.
''}
\end{minipage}}

\noindent
This design is a \textbf{cost-control proxy}, not perfect budget matching. Most frontier
rows use a soft 2K instruction, while our open models use a hard 8K generation cap.
In the revision we add one direct 8K check for Gemini-3.1-Pro. We also supplement the additional API spend across the four math benchmarks at 8K output
length would have been substantial under the per-million-token pricing in
Tab.~\ref{tab:api_pricing_closed_models}.
The Gemini result shows that a frontier API can substantially outperform our sub-10B
models under the same nominal budget. We therefore emphasize our main advantage as a lower-cost open alternative (Tab.~\ref{tab:api_pricing_closed_models}).

\begin{table}[h]
  \centering
  \caption{\textbf{Gemini-3.1-Pro budget check.}
  The main table keeps the original 2K closed-API comparison. This appendix table reports the single newly added 8K Gemini run.}
  \resizebox{0.98\linewidth}{!}{%
  \begin{tabular}{lcccccccccc}
  \toprule
  \multirow{2}{*}{Model} & \multicolumn{2}{c}{\textbf{Average}} & \multicolumn{2}{c}{\textbf{AIME}} & \multicolumn{2}{c}{\textbf{MATH}} & \multicolumn{2}{c}{\textbf{AMC}} & \multicolumn{2}{c}{\textbf{OlympiadBench}} \\
  \cmidrule(lr){2-3} \cmidrule(lr){4-5} \cmidrule(lr){6-7} \cmidrule(lr){8-9} \cmidrule(lr){10-11}
  & \#Tk. & Acc. & \#Tk. & Acc. & \#Tk. & Acc. & \#Tk. & Acc. & \#Tk. & Acc. \\
  \midrule
  Gemini-3.1-Pro (2K) & 1{,}586 & 54.4 & 2{,}020 & 10.0 & 1{,}274 & 79.2 & 1{,}798 & 37.3 & 1{,}771 & 40.1 \\
  Gemini-3.1-Pro (8K) & 2{,}581 & 88.0 & 5{,}163 & 80.0 & 1{,}385 & 99.4 & 2{,}532 & 96.4 & 3{,}358 & 79.0 \\
  \bottomrule
  \end{tabular}
  }
  \label{tab:gemini_8k_budget_check}
\end{table}

\begin{table}[h]
  \centering
  \caption{\textbf{API Pricing of Closed Models.}
  Prices are reported in USD per million tokens.
 Ratios compare output price against
  our Qwen-4B serving estimate of \(\$0.10\)/M output tokens.}
  \resizebox{0.98\linewidth}{!}{%
  \begin{tabular}{lcccc}
  \toprule
  \textbf{Model} & \textbf{Input} & \textbf{Output} & \textbf{Output / Qwen-4B} & \textbf{Source} \\
  \midrule
  GPT-5.4-Mini & \(\$0.75\) & \(\$4.50\) & \(45\times\) & \href{https://developers.openai.com/api/docs/models/gpt-5.4-mini}{OpenAI} \\
  Gemini-3.1-Pro Preview (\(\leq\)200K prompt) & \(\$2.00\) & \(\$12.00\) & \(120\times\) & \href{https://ai.google.dev/gemini-api/docs/pricing}{Google} \\
  Claude-4.6-Sonnet & \(\$3.00\) & \(\$15.00\) & \(150\times\) & \href{https://docs.anthropic.com/en/docs/about-claude/models}{Anthropic} \\
  Our Qwen-4B serving estimate & N/A & \(\$0.10\) & \(1\times\) & Estimate \\
  \bottomrule
  \end{tabular}
  }
  \label{tab:api_pricing_closed_models}
\end{table}

\begin{figure}[h]
  \centering
  \includegraphics[width=\linewidth]{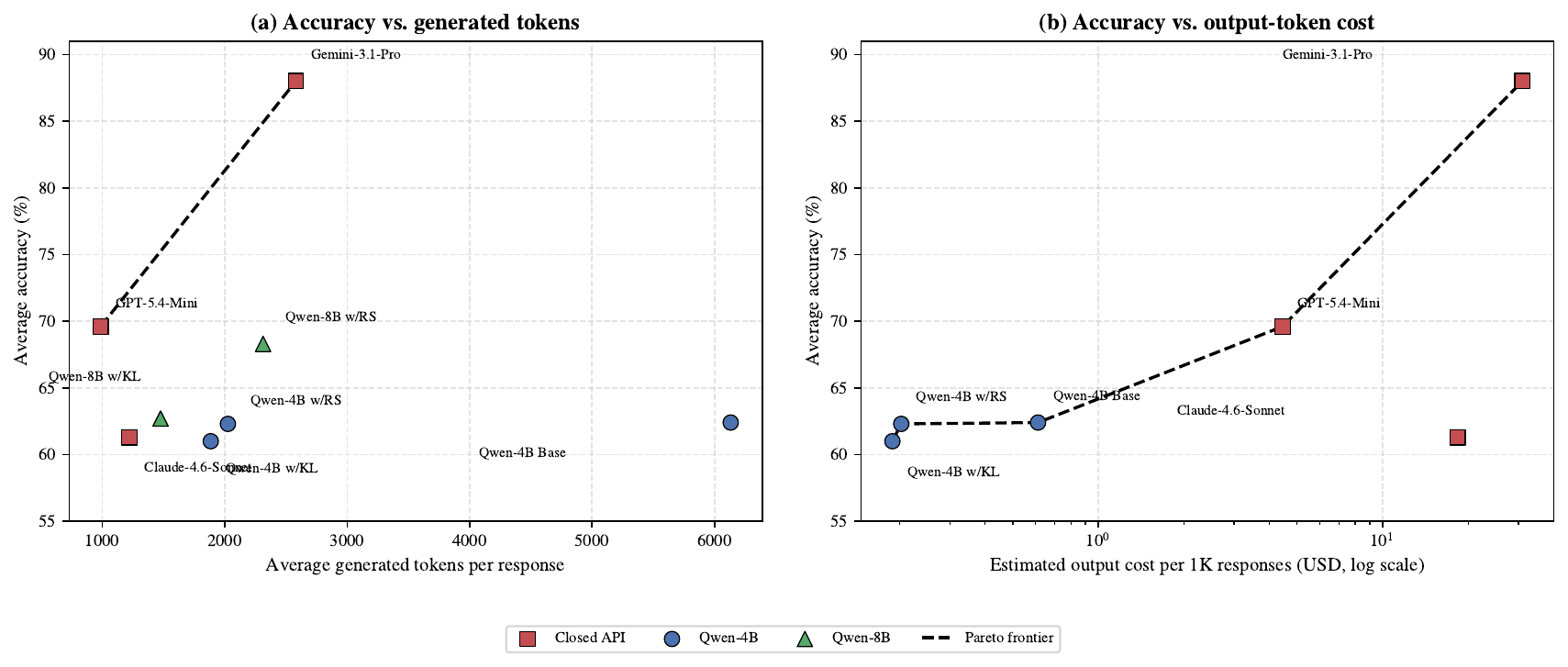}
  \caption{\textbf{Pareto view of closed-API comparison.}
  Left panel shows the accuracy and token trade-off. Right panel shows the accuracy and cost trade-off using output-token pricing.
  The added Gemini-3.1-Pro 8K point shows that frontier APIs can dominate the accuracy and token frontier, while Qwen-4B variants remain far cheaper to serve.}
  \label{fig:closed_api_pareto}
\end{figure}

Fig.~\ref{fig:closed_api_pareto} summarizes the intended interpretation of this comparison.
On accuracy versus generated tokens, the 8K Gemini point lies above the small open models,
so we do not claim absolute parity with frontier APIs. On accuracy versus estimated output
cost, however, the Qwen-4B variants occupy the low-cost regime. Their output-token price is
two orders of magnitude below Gemini-3.1-Pro and Claude-4.6-Sonnet. This is the practical
trade-off our method targets.

\subsection{Baselines}
\label{asec: baselines}

We compare against six recent methods for efficient reasoning, spanning both training-based
and inference-time (no-training) approaches.

\paragraph{Training-Based Methods.}

\begin{itemize}[leftmargin=*,nosep]
  \item \textbf{ETR}~\cite{xiong2026etr} (Entropy Trend Reward).
    Provides auxiliary reward based on the \emph{trend} of the model's token-distribution entropy
    during reasoning. When the model produces a correct answer and its entropy decreases
    monotonically toward the \texttt{</think>} boundary, an additional reward bonus is granted.
    This encourages convergence toward the answer rather than indefinite exploration.
    Implemented via veRL~\cite{sheng2025hybridflow} / GRPO~\cite{shao2024deepseekmath} on the DeepMath training set.

  \item \textbf{L1}~\cite{aggarwal2025l1} (Length-Controlled RL).
    Trains the model to respect a user-specified token budget via a length-coordinated reward.
    Two inference variants exist. \textit{L1-Exact} targets a specific token count, and
    \textit{L1-Max} enforces a maximum budget.
    The budget is encoded as a \texttt{num\_tokens} field in the training data.
    Implemented via veRL~\cite{sheng2025hybridflow} / PPO~\cite{schulman2017proximal}.

  \item \textbf{O1-Pruner}~\cite{luo2025o1}.
    Constructs offline per-question reference statistics (average generation length and accuracy)
    using a frozen reference model.  A length-coordinated reward $R_{LH}$ penalises responses
    that exceed the reference length without a proportional accuracy gain, compressing redundant
    reasoning steps.  The original paper does not use veRL. We re-implement it within the
    veRL framework.

  \item \textbf{PEAR}~\cite{huang2025pear} (Phase-based Entropy Awareness Reward).
    Divides each response into a \emph{thinking phase} (tokens before \texttt{</think>}) and an
    \emph{answer phase} (tokens after).  Entropy-based reward signals encourage exploration
    during thinking and rapid convergence during answering.
    Implemented as a standard veRL training run with a custom reward manager.
\end{itemize}

\paragraph{Inference-Time Methods (No Training Required).}

\begin{itemize}[leftmargin=*,nosep]
  \item \textbf{DEER}~\cite{yang2025dynamic} (Dynamic Early Exit Reasoning).
    Monitors model confidence after generating a \emph{trial answer} mid-sequence.
    If the log-probability of the trial answer exceeds a threshold, generation terminates
    early, discarding further reasoning tokens.  No fine-tuning is required. The method
    intervenes purely at the decoding stage.

  \item \textbf{NoThink}~\cite{ma2025reasoning}.
    Suppresses explicit chain-of-thought by prefilling an empty \texttt{<think></think>}
    block, directing the model to answer immediately.  Accuracy is recovered via best-of-$N$
    sampling with majority vote, or through budget-forcing with a token limit.
    No fine-tuning is required.
\end{itemize}

\paragraph{Baseline Training Configuration.}
For the four training-based baselines (ETR, L1, PEAR, O1-Pruner) we
reuse the common recipe of Appx.~\ref{asec: training_settings}
(batch 64, lr $1\mathrm{e}{-6}$, KL $10^{-3}$, $n{=}16$ rollouts,
4K rollout length, 1 epoch on DeepScaleR, 8 GPUs per node), swapping
in each method's own reward function and data preprocessing.
L1 additionally prepends a per-sample ``Think for $N$ tokens.''
instruction with $N \sim \mathcal{U}(100, 2048)$.
O1-Pruner first generates $K{=}2$ reference rollouts per question
(temperature 0.6, top-$p$ 0.95, 2K tokens) to compute the
reference-length statistic required by its length-harmonising
reward. Its train batch is reduced to 8.
DEER and NoThink are inference-only and run on the base checkpoint
with temperature 0.6, top-$p$ 0.95, and an 8K generation budget
(DEER. Confidence threshold 0.9, think-ratio 0.5. NoThink. Best-of-$N$
voting with $N{=}8$).
All baselines are evaluated with the same script and generation
settings as our method (Appx.~\ref{asec: eval_settings}).

\subsection{Capacity-dependent Behavior of KL and RS}
\label{asec: kl_rs_capacity}

The KL and RS variants show different relative behavior across backbones,
but the difference is structured rather than random. Tab.~\ref{tab:kl_rs_capacity}
reorganizes the main-result rows by model scale and reports the average
accuracy gap between RS and KL. The gap changes from $-6.0$ percentage
points on Qwen-0.6B to $+5.6$ percentage points on Qwen-8B, an
$11.6$-point swing. With the two 4B entries treated as tied in model
scale, the rank correlation between parameter count and the RS-KL gap is
near-perfect (Spearman $\rho=0.97$), and the Pearson correlation between
$\log$ parameter count and the RS-KL gap is $0.95$.

\begin{table}[h]
  \centering
  \caption{\textbf{Capacity-dependent behavior of KL and RS.}
  Accuracy and token counts are averaged over the four in-domain reasoning
  benchmarks in Tab.~\ref{tab:main_results}. Positive RS-KL means RS is
  more accurate than KL.}
  \resizebox{0.85\linewidth}{!}{%
  \begin{tabular}{lccccccc}
  \toprule
  \textbf{Backbone} & \textbf{Params} & \textbf{Base Acc.} & \textbf{KL Acc.} & \textbf{RS Acc.} & \textbf{RS-KL} & \textbf{KL \#Tk.} & \textbf{RS \#Tk.} \\
  \midrule
  Qwen-0.6B          & 0.6B & 40.7 & 42.3 & 36.3 & $-6.0$ & 2{,}023 & 1{,}806 \\
  DeepScaleR-1.5B    & 1.5B & 59.6 & 56.1 & 55.9 & $-0.2$ & 1{,}698 & 1{,}993 \\
  Qwen-4B            & 4B   & 62.4 & 61.0 & 62.3 & $+1.3$ & 1{,}886 & 2{,}026 \\
  Phi-Reasoning-4B   & 4B   & 59.6 & 62.0 & 62.6 & $+0.6$ & 2{,}919 & 2{,}992 \\
  Qwen-8B            & 8B   & 61.9 & 62.7 & 68.3 & $+5.6$ & 1{,}476 & 2{,}313 \\
  \bottomrule
  \end{tabular}
  }
  \label{tab:kl_rs_capacity}
\end{table}

This pattern follows from where the two variants intervene in the GRPO
objective. KL changes only the regularization side. Core tokens keep the
standard KL anchor, while over-budget redundant tokens lose that anchor and
receive a sparsity pressure. The policy-gradient term itself is unchanged,
making KL a conservative, low-variance intervention. RS instead reshapes
the advantage used by the policy-gradient term. It reinforces high-\normlp{}
core tokens on correct rollouts and penalizes redundant tokens when the
rollout exceeds the budget. This supplies a stronger learning signal, but
also injects additional variance directly into the policy update.

\paragraph{Conclusion and recommendation.}
The apparent KL/RS inconsistency is therefore better interpreted as a
capacity-dependent trade-off. Small models benefit from the safer KL
intervention. On Qwen-0.6B, KL preserves and slightly improves accuracy
while RS over-corrects and loses $4.4$ points relative to the base model.
Large models can absorb the stronger RS signal and convert it into accuracy.
On Qwen-8B, RS improves the base model by $6.4$ points and outperforms KL by
$5.6$ points, albeit with more tokens than KL. As a practical rule of thumb,
we recommend KL for models at or below roughly 1B parameters, either variant
for the 1B to 4B transition region depending on whether compression or
accuracy is preferred, and RS for models of 4B parameters or larger when the
primary goal is accuracy improvement rather than maximal token compression.

\subsection{Reward Dynamics During RL Training}
\label{asec: reward_dynamics}

\begin{figure}[h]
  \centering
  \includegraphics[width=0.78\linewidth]{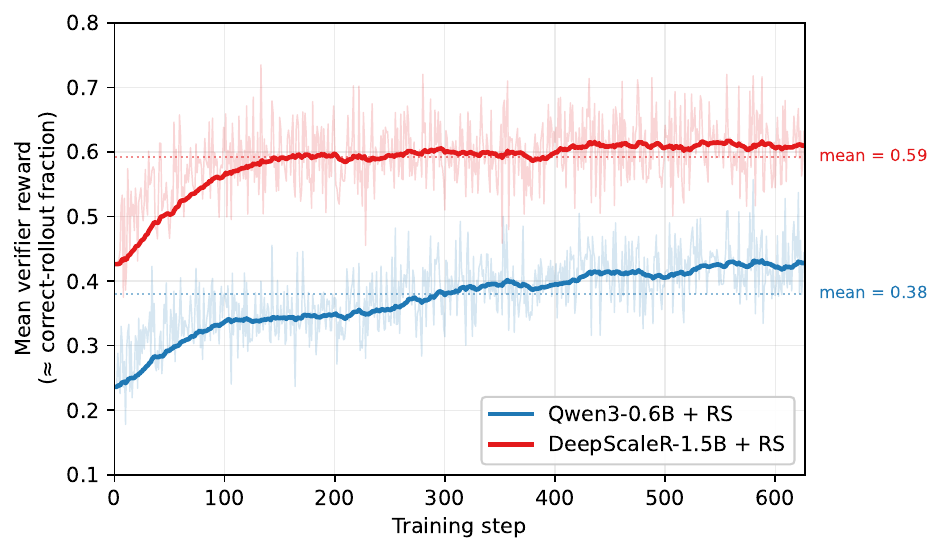}
  \caption{\textbf{Reward mean during
    RS training.} Mean verifier reward per training step, with EWMA smoothing
    (halflife $20$). The lighter background trace is the unsmoothed signal.
    Across the entire run, DeepScaleR-1.5B operates above
    $\mathrm{reward\_mean} \approx 0.59$ on average, while Qwen-0.6B
    operates at $\approx 0.38$, a $1.56\times$ gap. This is the density of
    rollouts on which RS actually fires.}
  \label{fig: reward_dynamics}
\end{figure}

The capacity-dependent pattern in Tab.~\ref{tab:kl_rs_capacity} admits a
second, complementary explanation that becomes visible once we look at
training-time reward signals. Recall that the RS objective
(Eq.~\ref{eq:rs_adv}) reshapes the advantage \emph{only} on correct rollouts
($r^{\text{seq}} > 0$), so the density of the RS shaping signal at every
training step is exactly the per-step correct-rollout fraction.
Fig.~\ref{fig: reward_dynamics} compares this signal between Qwen3-0.6B
and DeepScaleR-1.5B, both trained with the same RS configuration
($\alpha = 0.05$, $\lambda = 0.08$).

Three quantitative observations stand out. First, the gap is structural,
not transient. At the start of training Qwen-0.6B already sits at a
correct-rollout fraction of $0.256$ versus $0.450$ for DeepScaleR-1.5B
(first-10-step averages), and at the end of training the same ordering
holds at $0.427$ versus $0.612$ (last-50-step averages). The smaller
backbone never catches up. Second, the overall ratio of correct-rollout
fractions across the run is $0.593 / 0.380 \approx 1.56$, meaning RS
shaping fires on roughly $1.6\times$ more tokens per training step on the
larger backbone. Third, even Qwen3-0.6B's \emph{final} correct fraction
($0.43$) is below DeepScaleR-1.5B's \emph{initial} fraction ($0.45$). The
small backbone spends its entire training budget at a regime the larger
backbone has already exited.

This signal-density argument complements the variance-vs-regularizer
explanation given earlier. KL operates on the regularization side and
fires on every token regardless of $r^{\text{seq}}$, so its effective
density is unaffected by correctness rate. RS fires only on a minority of
small-backbone rollouts, leaving the majority of training updates governed
by the unmodified GRPO advantage. The combination of (a) reduced shaping
density and (b) higher gradient variance on small backbones is consistent
with the $-6.0$-point RS-KL gap on Qwen-0.6B in
Tab.~\ref{tab:kl_rs_capacity}, and with the smooth crossover into
RS-favoring territory as backbones grow.

\subsection{Computational Overhead of TokenProbe Operations}
\label{asec: overhead}

The TokenProbe-specific operations added to the GRPO training loop are
(i) the per-token \normlp{} z-score over each rollout's log-probabilities
returned by vLLM, and (ii) the greedy non-overlapping top-$K$ core-window
selection of Listing~\ref{lst:core_window}. Both run on CPU on
already-collected log-probabilities \emph{after} each rollout completes.
Neither runs on the inference path of the trained model.

We measure the wall-clock cost of these two operations as a function of
trace length $T$ on a single CPU thread.
Tab.~\ref{tab:tokenprobe_overhead} reports the per-trace timings and
amortizes them against a representative GRPO step time of $30$\,s with
$16$ rollouts per step (rollout dominates step time, while the policy update
itself is only a fraction of vLLM generation).

\begin{table}[h]
  \centering
  \caption{\textbf{Computational overhead of TokenProbe operations.}
    \normlp{} (z-score) and greedy non-overlapping window selection both
    run on CPU on log-probabilities that vLLM already returns. Per-call
    timings averaged over $2{,}000$ trials. Per-step amortization assumes
    a $30$\,s GRPO step with $16$ rollouts. \normlp{} is essentially
    constant-time across $T$ (single z-score over the array). Window
    selection grows linearly in $T \cdot K$ but stays in the
    single-millisecond regime even on $16$K traces. Inference-time
    overhead is exactly zero. The trained checkpoint is decoded by an
    unmodified vLLM \texttt{generate} call.}
  \label{tab:tokenprobe_overhead}
  \resizebox{0.85\linewidth}{!}{%
  \begin{tabular}{rrrrr}
    \toprule
    Trace length $T$ & \normlp{} (\textmu s) & Window sel.\ (\textmu s) & Total (\textmu s) & Frac.\ of GRPO step \\
    \midrule
    $1{,}024$  & $25.3$ & $126.6$    & $151.8$   & $0.009\,\%$ \\
    $2{,}048$  & $26.2$ & $280.6$    & $306.8$   & $0.018\,\%$ \\
    $4{,}096$  & $29.1$ & $629.0$    & $658.2$   & $0.037\,\%$ \\
    $8{,}192$  & $33.9$ & $1{,}475.9$ & $1{,}509.8$ & $0.082\,\%$ \\
    $16{,}384$ & $44.3$ & $3{,}850.6$ & $3{,}894.8$ & $0.210\,\%$ \\
    \bottomrule
  \end{tabular}
  }
\end{table}

The combined \normlp{} and window-selection cost stays below $0.21\%$ of
one GRPO step even on the longest $16$K traces we tested, and below
$0.09\%$ at the $8$K trace length used in our training runs. Because the
operations run only during training and not at inference time, the
$3$ to $4\times$ token-count reductions reported in
Tab.~\ref{tab:main_results} translate one-to-one into wall-clock speedups
when serving the trained model.

\subsection{Hard-Truncation Strawman Baseline}
\label{asec: hard_truncation}

A natural question is whether the token reductions reported in
Tab.~\ref{tab:main_results} could be obtained without our method by
simply hard-capping the unmodified base model at the same output budget.
We test this directly. We run \texttt{Qwen/Qwen3-4B} (base, no training,
no re-prompting) with $\mathrm{max\_tokens} = 2{,}048$ on the four
in-domain reasoning benchmarks of Tab.~\ref{tab:main_results}, holding
all other decoding parameters identical (thinking mode, $T = 0.6$,
top-$p = 0.95$, bf16). We compare the resulting accuracy and token usage against
the same base model evaluated at the original $8$K budget (the
\texttt{Qwen-4B} row of Tab.~\ref{tab:main_results}), and against our
trained KL/RS variants on the same backbone.

\begin{table}[h]
  \centering
  \caption{\textbf{Hard-truncation strawman vs.\ trained variants on
    Qwen3-4B.} Hard-capping the unmodified base model at $2{,}048$ tokens
    collapses average accuracy by $30.5$ points relative to the same
    model at $8$K, and truncates $93.3\%$ of problems at the budget
    boundary. Our trained KL/RS variants reach the same
    $\sim$2K-token output budget while retaining accuracy within $1.4$
    points of the unconstrained $8$K base, demonstrating that the token
    reductions in Tab.~\ref{tab:main_results} are not equivalent to
    output clipping. \textbf{\#Tk.} is the average generated token count
    (lower is better). \textbf{Acc.} is averaged accuracy in \% (higher
    is better). \textbf{\%Trunc.} is the fraction of rollouts that hit
    the $2{,}048$-token cap. The base$@8$K and our trained variants are
    reproduced from Tab.~\ref{tab:main_results}.}
  \label{tab:hard_truncation_baseline}
  \resizebox{\textwidth}{!}{%
  \begin{tabular}{lccccccccccc}
  \toprule
  \multirow{2}{*}{\textbf{Setting}}
    & \multicolumn{3}{c}{\textbf{Average}}
    & \multicolumn{2}{c}{\textbf{AIME-2025}}
    & \multicolumn{2}{c}{\textbf{MATH-500}}
    & \multicolumn{2}{c}{\textbf{AMC}}
    & \multicolumn{2}{c}{\textbf{Olympiad-Bench}} \\
  \cmidrule(lr){2-4} \cmidrule(lr){5-6} \cmidrule(lr){7-8} \cmidrule(lr){9-10} \cmidrule(lr){11-12}
    & \#Tk. & Acc. & \%Trunc.
    & \#Tk. & Acc. & \#Tk. & Acc. & \#Tk. & Acc. & \#Tk. & Acc. \\
  \midrule
  Qwen-4B base @ $8$K cap                       & $6{,}132$ & $62.4$ & N/A & $7{,}683$ & $30.0$ & $4{,}195$ & $92.0$ & $6{,}274$ & $72.3$ & $6{,}376$ & $55.4$ \\
  Qwen-4B base @ $2$K hard cap (\textbf{new})   & $2{,}018$ & $\mathbf{31.94}$ & $93.3\%$  & $2{,}048$ & $10.0$ & $1{,}941$ & $59.0$ & $2{,}040$ & $32.5$ & $2{,}041$ & $26.2$ \\
  $\Delta$ (hard cap $-$ $8$K base)             & $-4{,}114$ & $\mathbf{-30.46}$ & N/A & $-5{,}635$ & $-20.0$ & $-2{,}254$ & $-33.0$ & $-4{,}234$ & $-39.8$ & $-4{,}335$ & $-29.2$ \\
  \midrule
  Qwen-4B + KL (ours)                           & $1{,}886$ & $61.0$ & N/A & $2{,}253$ & $33.3$ & $1{,}246$ & $88.2$ & $2{,}092$ & $66.3$ & $1{,}952$ & $56.1$ \\
  Qwen-4B + RS (ours)                           & $2{,}026$ & $62.3$ & N/A & $2{,}532$ & $30.0$ & $1{,}391$ & $90.2$ & $2{,}154$ & $71.1$ & $2{,}025$ & $58.1$ \\
  \bottomrule
  \end{tabular}
  }
\end{table}

Hard truncation collapses average accuracy from $62.4$ to $31.94$
($-30.46$ points) at an average output of $2{,}018$ tokens, with
$93.3\%$ of rollouts hitting the $2{,}048$-token cap (essentially $100\%$
on AIME and AMC, where reasoning chains rarely complete within $2$K
tokens). The single benchmark on which truncation is partially survivable
is MATH-500, whose problems often admit short canonical solutions. Even
there the drop is $33$ points. By contrast, our trained KL and RS variants
on the same Qwen-4B backbone reach the same $\sim$2K-token output budget
while retaining $61.0$ and $62.3$ average accuracy respectively, within
$1.4$ points of the unconstrained $8$K base. The $+29$ to $30$ point gap
between Qwen-4B + KL/RS and Qwen-4B base @ $2$K under matched output
budget is direct evidence that token reduction by training is not
equivalent to token reduction by clipping. Our methods produce reasoning
trajectories that are intrinsically efficient, whereas hard truncation
externally clips trajectories that the base model would have continued.

\subsection{Case Studies}

To complement the quantitative comparisons in the preceding sections, we present five paired case studies drawn from the evaluation outputs of our KL-shaped 8B checkpoint and the matched unshaped 8B base, run on identical prompts under an 8K token budget. We surface two failure modes of the base model that the KL-shaped model consistently avoids. \textbf{(A)} exhausting the 8K budget without producing an extractable answer, and \textbf{(B)} reaching the same correct answer but spending 3 to 6\,$\times$ more tokens. For readability each case is laid out on its own page. The green box reports metadata (prompt, ground truth, per-model token counts and correctness), the two gray boxes contain head to tail excerpts of the two responses, and the body text below them analyses what each model did differently.

\input{case/compare}

\subsection{Scenario Analysis Under an Unlimited Token Budget}
\label{asec: unlimited_token}

To examine how token-budget constraints affect reasoning performance, we conduct a \emph{token budget gradient analysis} on six models originally evaluated without fixed token limits: Step-Flash-196B-11B, Hunyuan3-295B-21B, Qwen-3.6-Plus, GLM-Air-106B-12B, Qwen3-Next-80B-3B, and Gemini-3-Flash.
These models span both open-source MoE architectures and closed-API services.
We impose token budgets of 1K, 2K, 4K, 8K, 16K, and 32K on top of the unlimited (256K) baseline, and measure the resulting accuracy degradation across all four mathematical reasoning benchmarks.

\paragraph{Evaluation Protocol.}
For the unlimited setting, we set the maximum completion length to 256K tokens, which in practice imposes no effective constraint on any model.
For each simulated budget level $B \in \{$1K, 2K, 4K, 8K, 16K, 32K$\}$, we apply the following post-hoc protocol to the unlimited generations.
(1)~if a sample's \texttt{completion\_tokens} exceeds~$B$, we mark it as incorrect and cap its token count at~$B$ for the purpose of computing average token usage.
(2)~otherwise, we retain the original correctness label and token count unchanged.
This post-hoc approach ensures fair comparison across budget levels, since all evaluations are derived from the same set of underlying model generations, isolating the pure effect of the token budget constraint from any variation in decoding behavior.

\begin{figure*}[t]
  \centering
  \includegraphics[width=\textwidth]{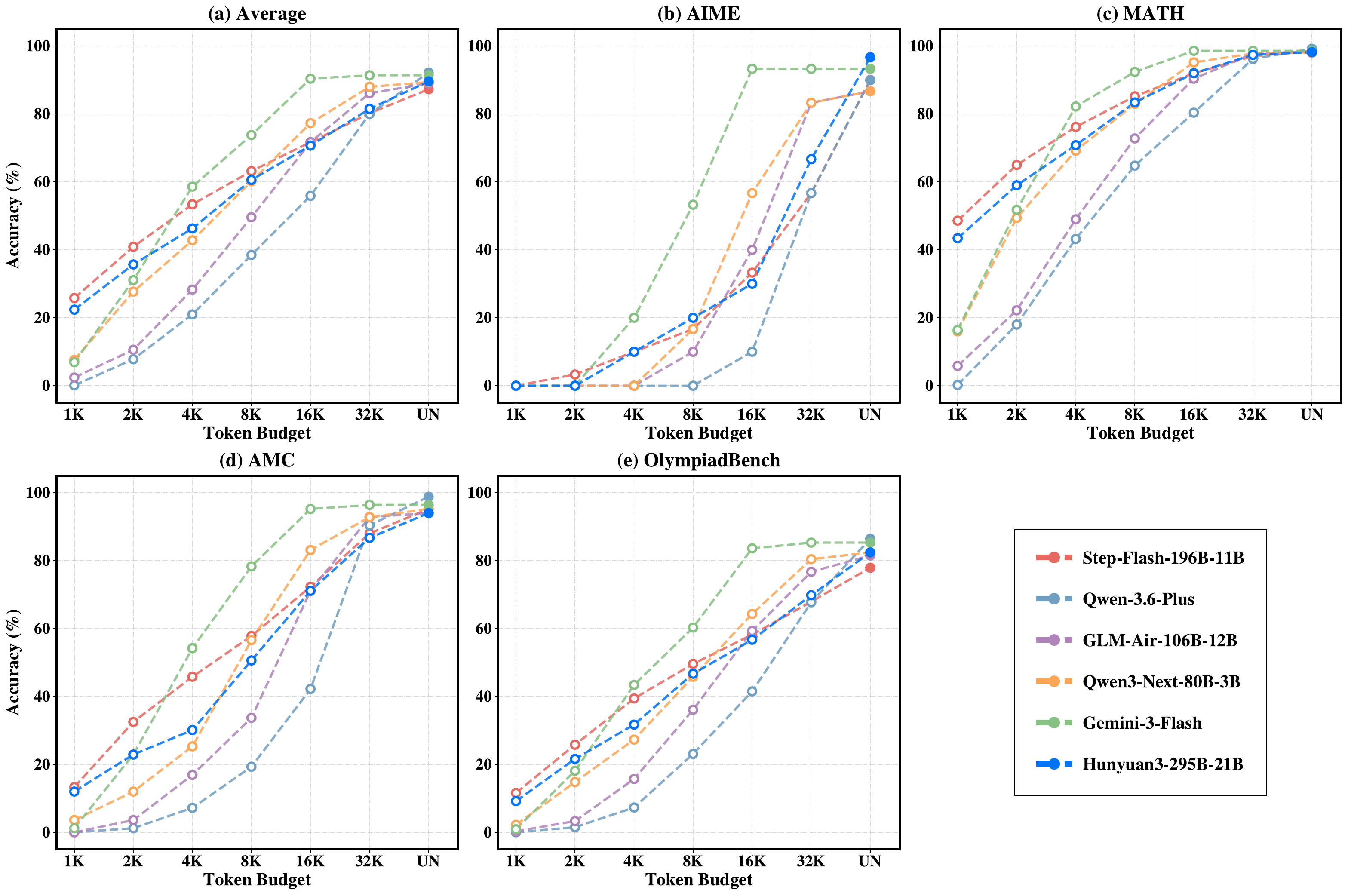}
  \caption{\textbf{Token budget gradient analysis for six unlimited reasoning models.}
  Each subplot shows accuracy (\%) as a function of the imposed token budget (1K, 2K, 4K, 8K, 16K, 32K, and unlimited at 256K).
  Solid markers denote the unlimited setting. Hollow markers denote simulated budget constraints.
  Dashed lines connect data points for each model.
  (a)~Average across all four benchmarks.
  (b)~AIME.
  (c)~MATH-500.
  (d)~AMC.
  (e)~OlympiadBench.}
  \label{fig:unlimited_gradient}
\end{figure*}

\begin{table*}[h]
  \centering
  \caption{\textbf{Unlimited setting (256K budget).}}
  \resizebox{\textwidth}{!}{%
  \begin{tabular}{lcccccccccc}
  \toprule
  \multirow{2}{*}{Model} & \multicolumn{2}{c}{\textbf{Average}} & \multicolumn{2}{c}{\textbf{AIME}} & \multicolumn{2}{c}{\textbf{MATH}} & \multicolumn{2}{c}{\textbf{AMC}} & \multicolumn{2}{c}{\textbf{OlympiadBench}} \\
  \cmidrule(lr){2-3} \cmidrule(lr){4-5} \cmidrule(lr){6-7} \cmidrule(lr){8-9} \cmidrule(lr){10-11}
  & \#Tk. & Acc. & \#Tk. & Acc. & \#Tk. & Acc. & \#Tk. & Acc. & \#Tk. & Acc. \\
  \midrule
  Qwen-3.6-Plus       & 18{,}099 & 92.2 & 30{,}755 & 90.0 & 9103 & 99.2 & 19{,}836 & 98.8 & 23{,}987 & 86.4 \\
  Gemini-3-Flash      & 4844     & 91.4 & 8332     & 93.3 & 2830 & 98.6 & 5049     & 96.4 & 6156     & 85.3 \\
  Qwen3-Next-80B-3B   & 8619     & 89.4 & 16{,}463 & 86.7 & 4202 & 98.0 & 9279     & 95.2 & 11{,}462 & 82.4 \\
  GLM-Air-106B-12B    & 12{,}125 & 89.0 & 22{,}320 & 86.7 & 6564 & 98.4 & 13{,}338 & 94.0 & 15{,}643 & 81.5 \\
  Step-Flash-196B-11B & 13{,}907 & 87.3 & 31{,}432 & 90.0 & 4502 & 98.6 & 12{,}583 & 95.2 & 20{,}258 & 77.9 \\
  Hunyuan3-295B-21B   & 12{,}384 & 89.6 & 26{,}957 & 96.7 & 4372 & 98.2 & 11{,}832 & 94.0 & 17{,}740 & 82.4 \\
  \bottomrule
  \end{tabular}
  }
  \label{tab:unlimited_256k}
\end{table*}

\begin{table*}[h]
  \centering
  \caption{\textbf{Token budget gradient results at 2K.}}
  \resizebox{\textwidth}{!}{%
  \begin{tabular}{lcccccccccc}
  \toprule
  \multirow{2}{*}{Model} & \multicolumn{2}{c}{\textbf{Average}} & \multicolumn{2}{c}{\textbf{AIME}} & \multicolumn{2}{c}{\textbf{MATH}} & \multicolumn{2}{c}{\textbf{AMC}} & \multicolumn{2}{c}{\textbf{OlympiadBench}} \\
  \cmidrule(lr){2-3} \cmidrule(lr){4-5} \cmidrule(lr){6-7} \cmidrule(lr){8-9} \cmidrule(lr){10-11}
  & \#Tk. & Acc. & \#Tk. & Acc. & \#Tk. & Acc. & \#Tk. & Acc. & \#Tk. & Acc. \\
  \midrule
  Qwen-3.6-Plus       & 1998 & 7.8  & 1980 & 0.0  & 1961 & 18.0 & 2041 & 1.2  & 2020 & 1.5  \\
  Gemini-3-Flash      & 1836 & 31.1 & 2048 & 0.0  & 1647 & 51.8 & 1910 & 22.9 & 1958 & 18.1 \\
  Qwen3-Next-80B-3B   & 1832 & 27.7 & 2048 & 0.0  & 1636 & 49.4 & 1957 & 12.0 & 1952 & 14.8 \\
  GLM-Air-106B-12B    & 1979 & 10.6 & 2048 & 0.0  & 1891 & 22.2 & 2034 & 3.6  & 2035 & 3.3  \\
  Step-Flash-196B-11B & 1571 & 40.9 & 2019 & 3.3  & 1213 & 65.0 & 1755 & 32.5 & 1794 & 25.8 \\
  Hunyuan3-295B-21B   & 1587 & 35.7 & 2048 & 0.0  & 1310 & 59.0 & 1770 & 22.9 & 1749 & 21.6 \\
  \bottomrule
  \end{tabular}
  }
  \label{tab:unlimited_2k}
\end{table*}

\paragraph{Analysis.}
As shown in Fig.~\ref{fig:unlimited_gradient} and Tables~\ref{tab:unlimited_256k} to \ref{tab:unlimited_1k}, all six models perform strongly under the unlimited setting, with average accuracies ranging from 87.3\% (Step-Flash-196B-11B) to 92.2\% (Qwen-3.6-Plus).
When the token budget is tightened, however, accuracy drops substantially and the degradation rate varies markedly across models.
This pattern highlights a practical tension in deploying reasoning models: the latent reasoning traces that support high accuracy also incur substantial token cost, while hard token budgets introduced for cost control can directly reduce reasoning quality.

\textbf{Token efficiency determines resilience to budget constraints.}
The six models span a wide range of token consumption at unlimited, from 4{,}844 (Gemini-3-Flash) to 18{,}099 (Qwen-3.6-Plus) average tokens.
This spread is closely reflected in their sensitivity to tighter budgets.
Gemini-3-Flash, consuming roughly one-quarter the tokens of Qwen-3.6-Plus while achieving a comparable 91.4\% accuracy (Tab.~\ref{tab:unlimited_256k}), retains 91.4\% at 32K, 90.4\% at 16K, and still 73.8\% at 8K (Tab.~\ref{tab:unlimited_8k}).
In contrast, Qwen-3.6-Plus drops from 92.2\% to 80.0\% at 32K, 55.9\% at 16K, and 38.5\% at 8K.
Qwen3-Next-80B-3B (8{,}619 tokens, moderate usage) shows a more gradual degradation pattern (88.0\% at 32K, 77.3\% at 16K, 60.2\% at 8K), whereas the more verbose GLM-Air-106B-12B (12{,}125 tokens) and Step-Flash-196B-11B (13{,}907 tokens) degrade more steeply.
Taken together, these comparisons suggest that robustness under budget constraints is more closely aligned with unlimited token usage than with unlimited accuracy, so models with more concise reasoning traces tend to degrade more gracefully.

\textbf{Harder problems suffer disproportionately from budget constraints.}
Across all six models, MATH-500 is the most resilient to budget cuts. Even at 1K, Step-Flash-196B-11B retains 48.6\% accuracy (Tab.~\ref{tab:unlimited_1k}), and Gemini-3-Flash achieves 92.4\% at 8K and 82.2\% at 4K (Fig.~\ref{fig:unlimited_gradient}(c)).
AIME, the most challenging benchmark, is the most sensitive. At 4K, only Gemini-3-Flash, Step-Flash-196B-11B, and Hunyuan3-295B-21B achieve nonzero accuracy, while the remaining models score 0.0\% (Tab.~\ref{tab:unlimited_4k}). At 1K, all six models score 0.0\% on AIME.
This ordering is consistent with both problem difficulty and the length of reasoning required: harder competition problems tend to require longer chains of thought, and these chains are the first to be truncated under tight token budgets.
AMC and OlympiadBench fall between these extremes, with OlympiadBench showing the sharpest degradation among the mid-difficulty benchmarks. For instance, Qwen-3.6-Plus drops from 86.4\% at unlimited to 23.1\% at 8K and 0.0\% at 1K on OlympiadBench (Fig.~\ref{fig:unlimited_gradient}(e)).

\textbf{High unlimited accuracy does not guarantee robustness.}
Qwen-3.6-Plus achieves the highest unlimited accuracy (92.2\%) yet drops to 0.1\% average accuracy at 1K, with 0.0\% on AIME, AMC, and OlympiadBench (Tab.~\ref{tab:unlimited_1k}).
Its AMC performance collapses from 98.8\% at unlimited to just 1.2\% at 2K (Tab.~\ref{tab:unlimited_2k}).
In contrast, Step-Flash-196B-11B has the lowest unlimited accuracy (87.3\%) but still achieves 48.6\% on MATH-500 at 1K, the highest among all models at this budget. This is because Step-Flash generates shorter completions on easier problems (4{,}502 tokens on MATH vs.\ 31{,}432 on AIME at unlimited), so its MATH performance is less affected by truncation.
Similarly, Qwen3-Next-80B-3B maintains 69.2\% on MATH at 4K (Tab.~\ref{tab:unlimited_4k}) despite dropping to 0.0\% on AIME at the same budget, indicating that degradation is driven by the mismatch between token demand and budget rather than by model capability alone.
These findings motivate token-efficient optimization: models that learn to reason more concisely without sacrificing accuracy should be more robust to realistic budget constraints.

\section{Unsuccessful Attempts. SFT Distillation Path}
\label{asec: unsuccessful_attempts}

Before arriving at the RL-based optimization approach, we explored whether Supervised Fine-Tuning (SFT) with LoRA~\cite{hu2022lora} could serve as a simpler alternative for improving token efficiency.
The motivation is that the rollout process in Reinforcement Learning is, operationally, a form of inference.
We hypothesized that by distilling the reasoning traces generated during inference from a larger teacher model and using LoRA to fine-tune a smaller student model, we could transfer efficient reasoning behavior.
Since token-level log probability is an inherent training signal in SFT, and the next-token prediction loss~\cite{radford2018improving} directly optimizes the likelihood of the training tokens, we expected this approach to align naturally with our \normlp{}-based framework.

\textbf{Our results show that the SFT distillation path is not a viable standalone solution.}
It substantially degrades accuracy (\ie, $-33$ to $-51$\, accuracy drop) and disrupts the semantic validity of log-probability signals, rendering the \normlp{} framework inapplicable to the resulting models.

\subsection{Experimental Design}
\label{asec:sft_design}

We conduct a full factorial experiment with 2~base models $\times$ 2~distillation data sources = 4~settings.

\begin{table}[h]
  \centering
  \caption{SFT distillation experiment matrix.}
  \label{tab:sft_matrix}
  \begin{tabular}{lllll}
    \toprule
    \textbf{Setting} & \textbf{Base Model} & \textbf{Distill Source} & \textbf{Think Tags} \\
    \midrule
    S1 & Qwen3-4B~\cite{yang2025qwen3} & Phi-4-Reasoning~\cite{abdin2025phi}  & Yes  \\
    S2 & Qwen3-4B~\cite{yang2025qwen3} & Qwen3-30B-A3B~\cite{yang2025qwen3}  & No  \\
    S3 & Qwen3-8B~\cite{yang2025qwen3} & Phi-4-Reasoning~\cite{abdin2025phi}  & Yes  \\
    S4 & Qwen3-8B~\cite{yang2025qwen3} & Qwen3-30B-A3B~\cite{yang2025qwen3}  & No \\
    \bottomrule
  \end{tabular}
\end{table}

\paragraph{Training configuration (shared across all 4 settings).}
LoRA rank\,=\,16, alpha\,=\,32, targeting all linear layers. 3~epochs. Learning rate\,=\,$2\!\times\!10^{-5}$. Max sequence length\,=\,8{,}192. Global batch size\,=\,96.

\paragraph{Distillation data.}
The training data is constructed by applying \normlp{}-based logprob filtering to reasoning traces from two teacher models.
The Phi-4-Reasoning data preserves \texttt{<think>} tags and has no degenerate opening patterns, with 99.9\% of samples containing \texttt{\textbackslash boxed\{\}} answers.
The Qwen3-30B-A3B data lacks think tags entirely, contains 70.6\% degenerate opening patterns (\eg, ``This is a complex or challenging question\ldots''), and only 81.9\% contain boxed answers.

\begin{table}[h]
  \centering
  \caption{Distillation data quality comparison.}
  \label{tab:sft_data}
  \begin{tabular}{lcc}
    \toprule
    \textbf{Feature} & \textbf{Phi4 Data} & \textbf{Qwen Data} \\
    \midrule
    Sample count & 25{,}695 & 29{,}761 \\
    Contains \texttt{<think>} tags & 100.0\% & 0.0\% \\
    Degenerate opening patterns & 0.0\% & 70.6\% \\
    Contains \texttt{\textbackslash boxed\{\}} & 99.9\% & 81.9\% \\
    Average compression ratio & 0.375 & 0.291 \\
    \bottomrule
  \end{tabular}
\end{table}

\paragraph{Evaluation.}
All models are evaluated using vLLM~\cite{kwon2023efficient} with LoRA online loading, \texttt{max\_tokens}=8{,}192, temperature=0.6, top\_p=0.95, on four benchmarks. AIME\,2025 (30 problems), AMC (83 problems), MATH~\cite{hendrycks2021measuring} (500 problems), and Olympiad Bench~\cite{he2024olympiadbench} (675 problems).

\subsection{Results}
\label{asec:sft_results}

\subsubsection{Accuracy}

\begin{table}[h]
  \centering
  \caption{SFT distillation accuracy vs.\ baselines. All 4 SFT settings suffer 33 to 51\,percentage points accuracy degradation. $\Delta$~denotes the drop from the corresponding base models.}
  \label{tab:sft_accuracy}
  \resizebox{\linewidth}{!}{%
  \begin{tabular}{lcccccc}
    \toprule
    \textbf{Setting} & \textbf{AIME} & \textbf{AMC} & \textbf{MATH} & \textbf{Olympiad} & \textbf{Overall} & $\Delta$\textbf{(percentage points)} \\
    \midrule
    Qwen-4B think & 60.0\% & 84.3\% & 95.8\% & 68.4\% & \textbf{77.1\%} & N/A \\
    Qwen-8B think & 46.7\% & 80.7\% & 94.4\% & 68.3\% & \textbf{72.5\%} & N/A \\
    Qwen-4B nothink & 20.0\% & 57.8\% & 85.0\% & 49.0\% & 53.0\% & N/A \\
    \midrule
    S1 (4B + Phi4) & 3.3\% & 28.9\% & 58.2\% & 25.0\% & 37.6\% & $-39.5$ \\
    S2 (4B + Qwen) & 3.3\% & 18.1\% & 42.0\% & 16.6\% & 26.2\% & $-50.9$ \\
    \textbf{S3 (8B + Phi4)} & 13.3\% & 24.1\% & 60.2\% & 26.4\% & 39.1\% & $-33.4$ \\
    S4 (8B + Qwen) & 6.7\% & 25.3\% & 48.2\% & 21.5\% & 31.8\% & $-40.7$ \\
    \bottomrule
  \end{tabular}}
\end{table}

The best SFT setting, S3 with Qwen3-8B + Phi4 data, achieves only 39.1\% overall accuracy, a $-33.4$\,percentage points drop from the 8B baseline (72.5\%).
The worst setting, S2 with Qwen3-4B + Qwen data, drops by $-50.9$\,percentage points.
On the hardest benchmark (AIME\,2025), all SFT settings collapse to 3 to 13\% accuracy versus 47 to 60\% for the baselines.

\subsubsection{Token Usage and Efficiency}

\begin{table}[h]
  \centering
  \caption{Token usage and efficiency comparison. Efficiency is measured as accuracy\,\%~/~kilo-tokens. S3 is the only SFT setting exceeding the RL baseline's efficiency, but at $-33.4$\,percentage points accuracy cost.}
  \label{tab:sft_efficiency}
  \begin{tabular}{lccc}
    \toprule
    \textbf{Setting} & \textbf{Accuracy} & \textbf{Avg.\ Tokens} & \textbf{Efficiency (acc\%/ktok)} \\
    \midrule
    Qwen3-4B think (RL) & \textbf{77.1\%} & 8{,}728 & 8.83 \\
    Qwen3-8B think (RL) & \textbf{72.5\%} & 9{,}244 & 7.84 \\
    \midrule
    S1 (4B + Phi4) & 37.6\% & 4{,}822 & 7.80 \\
    S2 (4B + Qwen) & 26.2\% & 7{,}258 & 3.62 \\
    \textbf{S3 (8B + Phi4)} & 39.1\% & 4{,}161 & 9.39 \\
    S4 (8B + Qwen) & 31.8\% & 6{,}127 & 5.18 \\
    \midrule
    Qwen3-4B nothink & 53.0\% & 1{,}943 & \textbf{27.28} \\
    Qwen3-4B-Instruct nothink & 71.8\% & 3{,}573 & \textbf{20.10} \\
    \bottomrule
  \end{tabular}
\end{table}

\textbf{No SFT setting falls on the accuracy and token Pareto frontier.}
The nothink baseline (53.0\% accuracy, 1{,}943 tokens) Pareto-dominates all SFT settings, achieving higher accuracy with fewer tokens.
The Instruct nothink baseline (71.8\% accuracy, 3{,}573 tokens) further dominates, achieving nearly double the accuracy of S3 with comparable token usage.
Thus, under the evaluated settings, SFT distillation offers neither a clean efficiency advantage over disabling Chain-of-Thought nor a competitive accuracy-preserving alternative to the RL path.

\subsection{Conclusion and Analysis}
\label{asec:sft_analysis}

\subsubsection{Pathological TokenProbe Reversal}

In RL-trained baseline models, higher TokenProbe (the proportion of tokens with above-average log probability) correlates positively with correctness ($r \approx +0.25$ to $+0.31$).
SFT distillation reverses this relationship in all four settings.

\begin{table}[h]
  \centering
  \caption{TokenProbe reversal in SFT models.  The sign of both correlations flips, indicating that the logprob signal no longer carries the same semantic meaning after SFT distillation.}
  \label{tab:sft_reversal}
  \resizebox{\textwidth}{!}{%
  \begin{tabular}{lcccc}
    \toprule
    \textbf{Model} & \textbf{Correct TokenProbe} & \textbf{Incorrect TokenProbe} & $r(\text{TokenProbe}, \text{correct})$ & $r(\text{tokens}, \text{TokenProbe})$ \\
    \midrule
    4B think (RL baseline) & 0.784 & 0.762 & $+0.307$ & $-0.415$ \\
    8B think (RL baseline) & 0.790 & 0.774 & $+0.254$ & $-0.367$ \\
    \midrule
    S1 (4B + Phi4) & 0.810 & 0.891 & $\mathbf{-0.404}$ & $\mathbf{+0.858}$ \\
    S2 (4B + Qwen) & 0.800 & 0.878 & $\mathbf{-0.464}$ & $\mathbf{+0.599}$ \\
    S3 (8B + Phi4) & 0.789 & 0.883 & $\mathbf{-0.525}$ & $\mathbf{+0.866}$ \\
    S4 (8B + Qwen) & 0.782 & 0.852 & $\mathbf{-0.422}$ & $\mathbf{+0.492}$ \\
    \bottomrule
  \end{tabular}}
\end{table}

In RL baselines, correct samples have higher TokenProbe than incorrect ones (the model is more confident when right).
In all four SFT settings, incorrect samples have \textit{substantially higher} TokenProbe than correct ones, meaning the model is most confident when wrong.
The correlation $r(\text{TokenProbe}, \text{correctness})$ shifts from $+0.3$ (RL) to $-0.5$ (SFT), a total swing of 0.8.
Similarly, $r(\text{tokens}, \text{TokenProbe})$ reverses from $-0.4$ to $+0.9$.
In RL models, longer responses tend to carry lower confidence, consistent with genuine exploration; in SFT models, longer responses instead have \textit{higher} confidence, consistent with degeneration loops that fill the token budget with repetitive high-confidence patterns.

\subsubsection{Degeneration Loop Microstructure}

We inspect the logprob microstructure to understand the mechanism behind TokenProbe reversal.

\begin{table}[h]
  \centering
  \caption{Degeneration loop statistics. Incorrect samples exhibit flattened logprob distributions, rare zero-crossings, and extremely long positive runs, all of which are hallmarks of repetitive degeneration.}
  \label{tab:sft_degeneration}
  \begin{tabular}{llcccc}
    \toprule
    \textbf{Setting} & \textbf{Category} & \textbf{lp\_std} & \textbf{ZC/100tok} & \textbf{Max pos.\ run} & \textbf{2nd-half TokenProbe} \\
    \midrule
    S1 & Correct ($n$=10) & 0.398 & 22.8 & 124 & 0.854 \\
    S1 & Incorrect ($n$=30) & 0.240 & 12.0 & 7{,}522 & 0.920 \\
    \midrule
    S3 & Correct ($n$=11) & 0.440 & 24.0 & 74 & 0.824 \\
    S3 & Incorrect ($n$=29) & 0.276 & 12.6 & 7{,}811 & 0.915 \\
    \midrule
    S4 & Correct ($n$=7) & 0.454 & 24.6 & 545 & 0.823 \\
    S4 & Incorrect ($n$=33) & 0.353 & 15.0 & 7{,}096 & 0.911 \\
    \bottomrule
  \end{tabular}
\end{table}

The statistics in Tab.~\ref{tab:sft_degeneration} reveal three recurring signatures.
\begin{enumerate}[nosep]
  \item \textbf{Logprob flattening}. Incorrect samples' logprob standard deviation is only 60 to 78\% of correct samples', indicating near-uniform confidence across all tokens, which is a signature of repetitive pattern generation.
  \item \textbf{Monotonic positive runs}. The longest contiguous segment of above-average logprob in incorrect samples reaches 7{,}000 to 7{,}800 tokens. The model sustains ``full confidence'' for thousands of consecutive tokens while producing entirely wrong output.
  \item \textbf{Second-half TokenProbe escalation}. Incorrect samples' TokenProbe in the second half reaches 0.91 to 0.92, compared to 0.82 to 0.85 for correct samples, indicating that degeneration becomes more pronounced as generation proceeds.
\end{enumerate}

\subsubsection{Detailed Case Studies}
\label{asec:sft_cases}

We present three case studies that illustrate distinct failure modes. All examples use the same prompts across settings.

\paragraph{Case Study 1. MATH \#12, Four-Way Comparison.}
\textit{Problem}. ``The proper divisors of 12 are 1, 2, 3, 4 and 6. A proper divisor of an integer $N$ is a positive divisor of $N$ that is less than $N$. What is the sum of the proper divisors of the sum of the proper divisors of 284?''
\textit{Ground truth}. 284.

\begin{table}[h]
  \centering
  \caption{MATH \#12 across all 4 SFT settings. The same problem elicits four completely different failure modes.}
  \label{tab:case_math12}
  \begin{tabular}{lccp{6cm}}
    \toprule
    \textbf{Setting}  & \textbf{Tokens} & \textbf{TokenProbe} & \textbf{Behavior} \\
    \midrule
    S1 (4B+Phi4)  & 451 & 0.885 & Compact think section. Factorization $\to$ sum $\to$ correct answer $\boxed{284}$ \\
    S2 (4B+Qwen)  & 8{,}192 & 0.894 & Empty think $\to$ correct reasoning initially, then enters infinite loop. ``284's sum of proper divisors is 220. 220's sum of proper divisors is 284.'' repeated until truncation \\
    S3 (8B+Phi4)  & 484 & 0.872 & Think section present, correct approach, but arithmetic error. Sums 220's divisors to 234 instead of 284 $\to$ $\boxed{234}$ \\
    S4 (8B+Qwen)  & 8{,}192 & 0.948 & Empty think $\to$ ``This is a complex or challenging question\ldots'' $\to$ brute-force enumeration checking every integer from 4 to 365 $\to$ truncated \\
    \bottomrule
  \end{tabular}
\end{table}

Together, the four outcomes illustrate the range of failure modes induced by SFT distillation.

\textbf{S1} (the only correct setting) produces a compact 451-token solution using the think section for quick factorization.

\textbf{S2} demonstrates \textit{degeneration loops}. The model correctly computes that 284 and 220 are amicable numbers, but then enters an infinite loop alternating between the two facts. Its second-half TokenProbe is 1.000 (every token in the second half is above the mean logprob), showing that the model remains highly confident while producing an endless repetition that consumes all 8{,}192 tokens without ever outputting a boxed answer.

\textbf{S3} illustrates a subtler failure: \textit{loss of self-correction capability}. The model's approach is correct (find divisors of 220 and sum them), but it makes an arithmetic error ($1+2+4+5+10+11+20+22+44+55+110 = 284$, not 234). An RL-trained model would typically catch such errors through ``Wait'' self-check loops. The SFT model lacks this self-correction behavior because distillation removed the ``Wait'' tokens from the training data (they fall in low-\normlp{} regions and are filtered out).

\textbf{S4} shows \textit{complete strategic collapse}. Unable to use prime factorization (which requires the structured reasoning destroyed by distillation), the model resorts to brute-force trial division, checking \textit{every integer} from 4 to 365 as a potential divisor of 284. After 8{,}192 tokens, it has only reached divisor candidate 365 and is truncated without producing an answer. Its TokenProbe of 0.948 (highest among all four settings) reflects the mechanically repetitive nature of brute-force enumeration, where every ``$N$. $N \times 2 = M$, too big, nope'' follows the identical pattern with maximal confidence.

\paragraph{Case Study 2. MATH \#377, Extreme Degeneration.}
\textit{Problem}. ``The superfactorial $n\$$ is defined as $n\$ = \underbrace{n!^{n!^{\cdot^{\cdot^{\cdot^{n!}}}}}}_{n!}$. What is the units digit of $4\$$?''
\textit{Ground truth}. 6. \textit{S3 prediction}. EXTRACTION\_FAILED.

The model correctly identifies that $4! = 24$ and begins constructing the tower $24^{(24^{(24^{\cdots}})}$, but then falls into an infinite loop, generating ``$24\hat{\phantom{x}}(24\hat{\phantom{x}}(24\hat{\phantom{x}}(\ldots$'' hundreds of times. The resulting statistics are as follows.
\begin{itemize}[nosep]
  \item \textbf{Token length}. 8{,}192 (truncated at limit)
  \item \textbf{TokenProbe}. 0.969 (among the highest in the entire dataset)
  \item \textbf{Max positive run}. 7{,}010 tokens, meaning that for 7{,}010 consecutive tokens, every single token is above the mean logprob
  \item \textbf{Zero-crossings}. 3.0 per 100 tokens (vs.\ 24.0 for correct S3 samples)
  \item \textbf{Second-half TokenProbe}. 1.000, indicating that in the second half, the logprob of \textit{every} token exceeds the mean
\end{itemize}

This case illustrates the core pathology. The model has learned to generate a syntactically plausible mathematical expression (exponent towers) with high confidence, but no longer reasons about the mathematical content needed to determine units digits via modular arithmetic. The degeneration is therefore largely invisible to logprob-based quality metrics: by such measures, the response appears high quality.

\paragraph{Case Study 3. MATH \#52, Correct vs.\ Infinite Loop.}
\textit{Problem}. ``Simplify. $\frac{\sqrt{2.5^2-0.7^2}}{2.7-2.5}$.''
\textit{Ground truth}. 12.

\begin{itemize}[nosep]
  \item \textbf{S3} (8B + Phi4). Correct in 568 tokens. Uses difference-of-squares factorization in the think section. $2.5^2 - 0.7^2 = (2.5-0.7)(2.5+0.7) = 1.8 \times 3.2 = 5.76$, $\sqrt{5.76} = 2.4$, answer $= 2.4/0.2 = 12$. TokenProbe = 0.803, zero-crossings = 24.3/100tok, reflecting a healthy and diverse reasoning trace.
  \item \textbf{S2} (4B + Qwen). Incorrect, 8{,}192 tokens. Empty think block, then enters a loop where ``$6.25 - 0.25 = 6.25 - 0.25 = \ldots$'' repeats 286 times. The model correctly identifies $2.5^2 = 6.25$ and $0.7^2 = 0.49$ but then \textit{miscomputes} $6.25 - 0.49$ as ``$6.25 - 0.25$'' and loops endlessly on this error because it has no self-correction mechanism.
\end{itemize}

This comparison further suggests that think tags and structured reasoning preserve a compact self-correction pathway, whereas the setting without think tags makes an arithmetic error and then loops because it has lost the ability to self-correct.

\subsubsection{Why SFT Fails While RL Succeeds. Structural Analysis}
\label{asec:sft_why}

The failures above are not implementation deficiencies fixable by hyperparameter tuning. They stem from six structural limitations of the SFT distillation paradigm.

\begin{enumerate}[leftmargin=2em]

\item \textbf{Self-correction marker destruction.}
The logprob-based distillation filter removes over 90\% of self-correction markers (``Wait'', ``Actually'', ``Let me re-check'', etc.) from the training data. In the Phi4 dataset, the average number of self-correction markers per sample drops from 8.53 to 0.58 (6.8\% retention). In the Qwen dataset, from 33.27 to 3.19 (9.6\% retention). This occurs because self-correction tokens, which represent moments of genuine model uncertainty, tend to have negative \normlp{}, placing them in the same region as wasteful redundancy. The filter cannot distinguish between a valuable ``Wait, let me reconsider the geometry\ldots'' (course correction after discovering a misconception) and a wasteful ``Wait, let me recalculate\ldots wait, let me recalculate again\ldots'' (repeatedly dwelling on a confirmed conclusion).

\item \textbf{Logprob filter ambiguity.}
Both valuable exploration and wasteful redundancy present as low-\normlp{} tokens.
Valuable corrections tend to occur at moments of model ``uncertainty'' (negative \normlp{}), and valueless repetitions also occur at moments of model ``uncertainty.''
The logprob filter is blind along this critical dimension and can only apply indiscriminate deletion.
This reflects a structural limitation: no static threshold on \normlp{} can separate the two categories because they overlap in logprob space.

\item \textbf{Causal context destruction.}
When the filter removes tokens from the middle of a reasoning chain, the remaining tokens are concatenated into a sequence that was \textit{never generated by any model}, as illustrated below.
\begin{equation*}
\underbrace{A \to B \to C \to [\text{Wait, C is wrong}] \to D \to E}_{\text{Original}} \quad\Longrightarrow\quad \underbrace{A \to B \to D \to E}_{\text{Distilled}}
\end{equation*}
In the original chain, $D$ was generated conditioned on the correction of $C$. In the distilled chain, $D$ follows $B$ directly, a transition that never occurred in any natural reasoning process. The model is forced to learn $P(D \mid A, B)$ when the correct generation actually depends on $P(D \mid A, B, C, [\text{correction}])$, a much richer context.

\item \textbf{NTP loss lacks credit assignment.}
The next-token prediction loss~\cite{radford2018improving} $\mathcal{L}_{\text{NTP}} = -\frac{1}{T}\sum_{t=1}^{T} \log P_\theta(x_t \mid x_{<t})$ applies equal weight to every token. There is no mechanism to distinguish core reasoning steps from redundant filler. In contrast, RL's GRPO~\cite{shao2024deepseekmath} loss weights tokens by advantage $\hat{A}_t$, enabling statistical credit assignment across multiple rollouts. Patterns consistently appearing in successful trajectories are reinforced. Patterns consistently appearing in failed trajectories are suppressed.

\item \textbf{No feedback loop.}
SFT training operates on a fixed, offline dataset. Once the distillation filter makes its decisions, the model has no mechanism to ``contest'' incorrect deletions during training. RL operates online. The model generates rollouts, receives rewards, and updates. If skipping self-correction leads to wrong answers (negative reward) while performing self-correction leads to correct answers (positive reward), the model learns when correction is valuable through thousands of such trial-and-error cycles.

\item \textbf{Compounded exposure bias.}
Standard SFT already suffers from exposure bias~\cite{ranzato2015sequence,bengio2015scheduled}. Training uses teacher forcing (ground-truth prefix), while inference uses autoregressive generation (model's own prefix). Distillation worsens this. The training data contains artificially compressed token transitions (from the filter), which diverge even further from the model's natural generation patterns at test time. The result is that small deviations at inference time cascade into large errors, with no recovery mechanism.
\end{enumerate}

\paragraph{Why RL avoids these pitfalls.}
RL preserves complete causal context (soft penalty via reward signals, never hard deletion), enables statistical credit assignment (GRPO~\cite{shao2024deepseekmath} advantage estimation across rollouts), trains on the model's own distribution (no exposure bias), and allows new behaviors to emerge. DeepSeek-R1~\cite{guo2025deepseek} demonstrated that self-correction behaviors can emerge from scratch through pure RL training, without SFT data containing such behaviors. The distinction is therefore structural: \textbf{SFT deletes information and then imitates the compressed trace, whereas RL preserves the full trajectory and reinforces it differentially.}

\subsection{Outlook. SFT as RL Warm-Start}
\label{asec:sft_outlook}

Although pure SFT distillation is unviable as a standalone training method, these experiments offer two valuable insights for the RL path.

\paragraph{SFT as cold-start initialization.}
The SFT distillation data retains value not as the final training objective, but as an initialization for RL training. SFT can provide an initial bias toward compact reasoning style, nudging the model to prefer shorter, more structured traces, while RL's online exploration subsequently recovers reasoning quality through reward-signal correction of the errors and degeneration introduced by SFT. This follows the established pipeline of DeepSeek-R1~\cite{guo2025deepseek}, which uses SFT warm-start before GRPO~\cite{shao2024deepseekmath} training. The key caveat is that one must verify after SFT warm-start that $r(\text{TokenProbe}, \text{correctness})$ remains positive, because otherwise the \normlp{}-based reward shaping would operate on an inverted signal.

\paragraph{New perspectives on RL efficiency training.}
The SFT experiments clarify why token efficiency is difficult to obtain through post-hoc data compression alone. The value of Chain-of-Thought reasoning often lies in the apparently ``redundant'' exploration process that enables self-correction. This insight motivates the RL-based approach: rather than compressing existing reasoning traces offline, the model should learn \textit{naturally efficient} reasoning strategies through online exploration with efficiency-aware rewards. Our data shows that models answering correctly use fewer tokens (Pearson $r(\text{tokens}, \text{correct}) \approx -0.59$ to $-0.74$), suggesting that the objective is not to remove self-correction, but to make it more precise: correcting decisively when needed and advancing decisively otherwise.

\paragraph{Alignment when transferring teacher distributions.}
If future warm-start methods transfer token distributions between models with different vocabularies, token alignment becomes an additional design consideration. Probabilistic Token Alignment for Large Language Model Fusion~\cite{zeng2025probabilistic} formulates this alignment through optimal transport and provides a soft mapping for model fusion. Such distribution-aware alignment could inform future teacher--student transfer, although its ability to preserve self-correction in compressed reasoning traces would require separate evaluation.

\section{Exploring Token Compression Limits}
\label{asec: token_limits}

\begin{figure*}[h]
    \centering
    \includegraphics[width=0.98\textwidth]{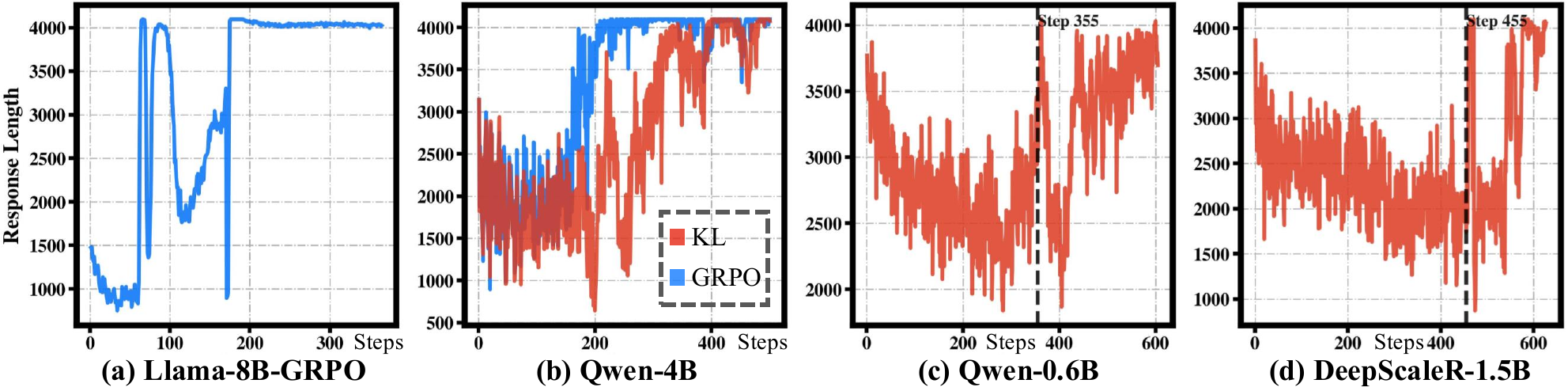}
    \vspace{-0.3cm}
    \caption{\textbf{Failure modes of length-controlled RL training.}
    \textbf{(A)} Llama-3.2-8B-Instruct under vanilla GRPO. Response length oscillates wildly and locks at the 4096-token cap after $\sim$200 steps.
    \textbf{(B)} Qwen3-4B-Instruct under both GRPO (blue) and our KL variant (red). Both runs saturate at the cap with degenerate, repetitive outputs.
    \textbf{(C, D)} Qwen3-0.6B-Base and DeepScaleR-1.5B trained with our method. Training is stable until the running compression scale crosses $\approx 1.6\times$ (dashed line, at step 355 for 0.6B and step 455 for 1.5B), after which length explodes, accuracy collapses to zero, and outputs degenerate into pure repetition.}
    \vspace{-0.5cm}
    \label{fig: failure}
\end{figure*}

\paragraph{Failure 1. Instruction-tuned models collapse under vanilla GRPO.}
Figure~\ref{fig: failure}(A) shows Llama-3.2-8B-Instruct trained with plain GRPO, without any compression-related modification. After an initial dip the response length oscillates violently and then saturates at the 4096-token cap, never recovering. We attribute this to two compounding effects of instruction tuning. First, SFT on instruction data sharply narrows the output entropy. The policy has already committed to a small set of high-probability templates, leaving little exploration headroom for the on-policy GRPO updates. Once a few reward gradients push the policy off these templates, the loss landscape becomes brittle and the model transitions abruptly between disparate output styles rather than improving smoothly. Second, instruction-tuned checkpoints carry strong stylistic priors, including preambles, headings, hedging phrases, and answer formatting, that act as implicit length regulators. Once the verifiable reward starts breaking those priors, the latent format constraint disappears and length grows without bound. Stable RLHF/RLVR pipelines on instruction models typically depend on heavy KL regularization to a frozen SFT reference, conservative learning-rate schedules, and reward models that explicitly penalize verbosity, none of which are present in vanilla GRPO. This is consistent with the common practice in the recent reasoning-RL literature of starting from base rather than instruction-tuned checkpoints.

\paragraph{Failure 2. Qwen3-4B-Instruct exhibits the same pathology with both GRPO and our KL method.}
Figure~\ref{fig: failure}(B) confirms that this failure is not Llama-specific. On Qwen3-4B-Instruct, both vanilla GRPO and our KL variant converge to the same terminal state. Length pinned at 4096 and outputs collapsed onto a narrow set of repeating $n$-grams. Manual inspection of late-training rollouts reveals that the model produces the same phrase or short paragraph dozens of times until the budget is exhausted. Once this regime is entered the verifier reward signal carries essentially no useful gradient information. The mechanism is a positive feedback loop characteristic of neural-text degeneration. The instruction-tuned policy starts in a low-entropy basin. The compression objective rewards short \emph{correct} answers but provides no penalty for repetitive padding when the answer is wrong. As accuracy degrades the model pads more, lowering reward variance and further attenuating gradients. Crucially, in this setting our KL anchor is the instruction-tuned policy itself, so anchoring offers no protection. It pulls the policy back toward a reference that is already prone to degeneration.

\paragraph{Failure 3. A hard compression limit at $\approx 1.6\times$.}
Figures~\ref{fig: failure}(C) and~\ref{fig: failure}(D) reveal a more fundamental boundary that appears even on base models trained with our full method. Qwen3-0.6B-Base (C) and DeepScaleR-1.5B (D) train stably for several hundred steps, with average response length decreasing as expected. However, the running \emph{compression scale}, defined as the ratio of the initial average CoT length to the current average length, keeps climbing, and at the moments marked by the dashed lines (step 355 for 0.6B and step 455 for 1.5B) it crosses approximately $1.6\times$. At that crossing both runs undergo the same transition. A sharp burst of variance, a brief partial recovery, and then a runaway increase in length until the cap is hit. Post-collapse rollouts are uniformly degenerate (zero accuracy, length saturated at the maximum, outputs purely repetitive), mirroring the instruction-model failures of panels A and B. We interpret $1.6\times$ as an empirical lower bound on the information density of the reasoning trace. Below the corresponding length, the chain-of-thought no longer carries enough intermediate computation to support the final answer, the verifier reward becomes effectively random, and the policy loses any signal that would keep it in the compressed regime. In our experiments, this boundary appears similar across the 0.6B and 1.5B checkpoints despite their different absolute reasoning budgets. This finding aligns with a growing body of work arguing that chain-of-thought length is not cosmetic but corresponds to actual sequential computation, and that for problems of a given difficulty there exists a minimum number of intermediate tokens below which an autoregressive solver cannot succeed. Once that floor is crossed, the model abandons reasoning altogether and emits the maximum-length repetitive output that minimizes the immediate negative reward.

\paragraph{Is the $\sim 1.6\times$ floor task-dependent?}
The $1.6\times$ value is observed on Qwen3-0.6B-Base and DeepScaleR-1.5B
trained on DeepScaleR (math-only) and evaluated on the four math
benchmarks of Tab.~\ref{tab:main_results}. We describe it as
\emph{backbone-agnostic} across these two checkpoints, not
\emph{task-agnostic}. Mechanistically the floor reflects the minimum
number of intermediate tokens an autoregressive solver requires for
problems drawn from a particular difficulty distribution, which is
necessarily task-dependent. The supplementary budget-gradient analysis
in \S\ref{asec: un_appendix_tables} is consistent with this view.
MATH-500, the easiest of the four math benchmarks for our checkpoints,
remains partially solvable down to $1$K tokens (Step-Flash-196B-11B
retains $48.6\%$ on MATH-500 at the $1$K cap. Tab.~\ref{tab:unlimited_1k}),
while AIME-2025 collapses to $0\%$ on every one of the five large models
once the cap drops below $4$K (Tab.~\ref{tab:unlimited_4k}). The
\emph{compression floor} is therefore lower on easier distributions and
higher on harder ones. Reporting one number tied to a single
distribution, DeepScaleR-style competition mathematics, is what we
mean by ``backbone-agnostic.'' We did not retrain at the frontier on
non-math tasks because the $1.6\times$ collapse is destructive
(Fig.~\ref{fig: failure}(C, D)) and replicating these failure runs per
task would be prohibitive. Characterising the floor as a function of
task difficulty distribution is left to future work.

\section{Limitations and Future Work}
\label{asec: limitations}

\subsection{Theoretical Certification}
\label{asec: theoretical_certification}

Our analysis identifies a strong and consistent empirical association between TokenProbe, reasoning accuracy, and token usage across model-level, quartile-level, and response-level measurements. The present work focuses on validating and leveraging this association as a practical inference-native signal for efficient reasoning. A natural direction for future work is to develop a deeper mechanistic account of why normalized token-level confidence aligns with reasoning quality and token efficiency. This question is also open for related log-probability-, entropy-, and confidence-based reasoning diagnostics, which are often motivated by empirical correlations with downstream quality indicators. We view our findings as an empirical basis for such future theoretical certification, while the proposed framework remains directly usable as a diagnostic and optimization tool.

\subsection{General Reasoning Efficiency}
\label{asec: general_reasoning_efficiency}

We also conducted a lightweight transfer diagnostic to examine whether the TokenProbe trend is tied only to mathematical reasoning. The diagnostic uses two domains. Math, consisting of AIME2025, AMC, MATH, and OlympiadBench, and General, consisting of MMLU, MMLU-Pro, LSAT, and GPQA. For each domain, we use 10 fixed samples per dataset, giving 40 samples per domain. We compare three Qwen3-4B variants. The base model, the KL-trained variant, and the RS-trained variant.

\begin{table}[h]
  \centering
  \caption{\textbf{TokenProbe on fixed 40-sample transfer diagnostics.}
  Each domain contains four datasets with 10 fixed samples per dataset. We report only TokenProbe.}
  \resizebox{0.4\linewidth}{!}{%
  \begin{tabular}{lccc}
  \toprule
  \textbf{Domain} & \textbf{Base} & \textbf{w/ KL} & \textbf{w/ RS} \\
  \midrule
  Math    & 0.782 & 0.887 & 0.846 \\
  General & 0.736 & 0.846 & 0.802 \\
  \bottomrule
  \end{tabular}
  }
  \label{tab:tokenprobe_transfer_appendix}
\end{table}

The same directional pattern appears in both domains. TokenProbe increases after applying either KL or RS. This provides trend-level evidence that the efficiency behavior is relatively task-agnostic rather than tailored only to math reasoning. We view this as an initial diagnostic rather than a complete generality claim, and future work will extend the study of general efficiency to broader settings such as code generation and agentic reasoning.

\paragraph{Efficient adaptation across tasks and modalities.}
Parameter-efficient adaptation offers further settings in which to examine whether TokenProbe remains informative after a change in training regime. MEPT~\cite{zeng2025mept} uses multiple prompt experts to adapt to diverse data distributions, while Visual Fourier Prompt Tuning~\cite{zeng2024visual} incorporates frequency-domain information into visual prompts for adapting vision Transformers. These approaches motivate future studies of combining efficient adaptation with efficient reasoning. For visual or multimodal settings, applying TokenProbe would require a generated reasoning trace with accessible token log-probabilities; the present experiments do not establish its validity for visual prompt tokens.

\providecommand{\Cref}[1]{\hyperref[#1]{\ref*{#1}}}
\providecommand{\cref}[1]{\hyperref[#1]{\ref*{#1}}}

\section{Reproducibility, Compute, and Responsible Use Details}
\label{asec:reproducibility_responsible_use}

\subsection{Open Access to Data and Code}
\label{asec:open_access_data_code}

The training data used for our efficient-GRPO runs is
DeepScaleR-Preview-Dataset, which is publicly available on Hugging Face at
\url{https://huggingface.co/datasets/agentica-org/DeepScaleR-Preview-Dataset}.
The dataset card lists the license as MIT. The appendix provides pseudocode for
the core implementation components: core-window selection in
\Cref{lst:core_window}, selective KL anchoring in \Cref{lst:kl_loss}, and reward
shaping in \Cref{lst:rs_loss}. 

\subsection{Statistical Significance}
\label{asec:statistical_significance}

To estimate run-to-run variability for the main efficient-GRPO conclusions, we
rerun two representative configurations three times and report the sample
standard deviation across runs, using denominator $n-1$. The reported values
measure variability across repeated training/evaluation runs under the same
benchmark protocol.
\Cref{tab:run_to_run_std} shows that the average-accuracy variance is small for
both a low-capacity KL setting and a mid-scale RS setting, while per-benchmark
variance is larger on AIME because it contains only 30 problems.

\begin{table}[h]
  \centering
  \caption{\textbf{Run-to-run standard deviation.}
  Each entry is the sample standard deviation across three repeated runs.
  We report standard deviation of accuracy in percentage points.}
  \label{tab:run_to_run_std}
  \resizebox{0.72\linewidth}{!}{%
  \begin{tabular}{lccccc}
  \toprule
  \textbf{Model} & \textbf{Average} & \textbf{AIME} & \textbf{MATH} & \textbf{AMC} & \textbf{Olympiad} \\
  \midrule
  Qwen-0.6B w/ KL & 0.90 & 1.96 & 1.70 & 1.83 & 0.95 \\
  Qwen-4B w/ RS   & 0.20 & 5.10 & 1.01 & 3.17 & 1.85 \\
  \bottomrule
  \end{tabular}
  }
\end{table}

\subsection{Compute Resources}
\label{asec:compute_resources}

All main training runs use one node with $8\times$ NVIDIA H100 80GB GPUs. The
training recipe is otherwise the same as in \Cref{asec: training_settings}.
\Cref{tab:compute_resources} reports the average wall-clock time for each
backbone family. These times cover the main efficient-GRPO training runs and do
not include exploratory failed runs, closed-API evaluation time, or offline
analysis scripts.

\begin{table}[h]
  \centering
  \caption{\textbf{Compute resources for main training runs.}
  All runs use a single node with $8\times$ NVIDIA H100 80GB GPUs.}
  \label{tab:compute_resources}
  \resizebox{0.72\linewidth}{!}{%
  \begin{tabular}{lll}
  \toprule
  \textbf{Backbone} & \textbf{Hardware} & \textbf{Average training time} \\
  \midrule
  Qwen-0.6B & $8\times$ H100 80GB & 12h 14m \\
  DeepScaleR-1.5B & $8\times$ H100 80GB & 9h 10m \\
  Qwen-4B & $8\times$ H100 80GB & 21h 56m \\
  Phi-Reasoning-4B & $8\times$ H100 80GB & 25h 13m \\
  Qwen-8B & $8\times$ H100 80GB & 38h 20m \\
  \bottomrule
  \end{tabular}
  }
\end{table}

The TokenProbe-specific operations add negligible training overhead. As reported
in \Cref{tab:tokenprobe_overhead}, the extra CPU-side computation accounts for
$0.009\%$--$0.210\%$ of a representative GRPO step across trace lengths from
1K to 16K tokens. At inference time, the overhead is zero because the trained
checkpoint is decoded with an unmodified generator.

\subsection{Broader Impacts}
\label{asec:broader_impacts}

The primary positive impact of TokenProbe-guided efficient reasoning is reduced
inference cost. By lowering generated-token usage while preserving reasoning
accuracy, the method can reduce serving cost and energy consumption and can make
reasoning-capable models more accessible in resource-constrained settings. This
is especially relevant for open sub-10B models, where lower inference cost can
make deployment feasible without relying on large closed APIs.

The same efficiency gains may also lower the cost of misuse. Cheaper reasoning
can make automated generation, persuasion, or decision-support pipelines easier
to scale, including undesirable uses. In addition, the method does not eliminate
standard risks of language models: a compressed reasoning trace may still be
incorrect, incomplete, or overconfident. We therefore view TokenProbe as an
efficiency and diagnostic method, not as a safety or factuality guarantee.
This work does not release a new high-risk foundation model; it studies a
training objective and diagnostic signal built on existing models and public
benchmarks.

\subsection{Licenses for Existing Assets}
\label{asec:asset_licenses}

We use existing models, datasets, and software frameworks, and credit the
corresponding creators through citations and model/dataset identifiers in
Tables~\Cref{tab:models_training}, \Cref{tab:models_opensource}, and
\Cref{tab:models_closedsource}.
\Cref{tab:asset_licenses} summarizes the licenses or terms listed by the
corresponding public model cards or provider documentation.

\begin{table}[p]
  \centering
  \rotatebox{90}{%
  \begin{minipage}{0.95\textheight}
  \centering
  \caption{\textbf{Licenses and terms for existing assets.}
  License entries follow the corresponding public model or dataset cards when
  available. Closed-API models are governed by provider API terms rather than
  open model licenses.}
  \label{tab:asset_licenses}
  \resizebox{\linewidth}{!}{%
  \begin{tabular}{p{3.2cm}p{10.5cm}p{3.7cm}p{5.2cm}}
  \toprule
  \textbf{Asset} & \textbf{Identifier / source} & \textbf{License / terms} & \textbf{Role} \\
  \midrule
  DeepScaleR-Preview-Dataset & \texttt{agentica-org/DeepScaleR-Preview-Dataset} & MIT & Training data \\
  Qwen family & \texttt{Qwen/Qwen3-*}, \texttt{Qwen/Qwen3.5-*}, and \texttt{Qwen/Qwen3-Next-*} & Apache-2.0 & Training and open baselines \\
  DeepScaleR-1.5B & \texttt{agentica-org/DeepScaleR-1.5B-Preview} & MIT & Training backbone \\
  Phi-4-mini-reasoning & \texttt{microsoft/Phi-4-mini-reasoning} & MIT & Training backbone \\
  DeepSeek-R1-Distill-Qwen & \texttt{deepseek-ai/DeepSeek-R1-Distill-Qwen-1.5B} & MIT & Open baseline \\
  GPT-OSS & \texttt{openai/gpt-oss-120b} & Apache-2.0 & Open baseline \\
  GLM family & \texttt{zai-org/GLM-4.5-Air}, \texttt{zai-org/GLM-4.7}, and \texttt{zai-org/GLM-5} & MIT & Open baselines \\
  Step-3.5-Flash & \texttt{stepfun-ai/Step-3.5-Flash} & Apache-2.0 & Open baseline \\
  DeepSeek V3/V4 family & \texttt{deepseek-ai/DeepSeek-V3.2}, \texttt{deepseek-ai/DeepSeek-V4-Pro}, and \texttt{deepseek-ai/DeepSeek-V4-Flash} & MIT & Open baselines \\
  Kimi-K2-Thinking & \texttt{moonshotai/Kimi-K2-Thinking} & Custom / other license & Open baseline \\
  Hunyuan3 & \texttt{tencent/Hy3-preview} & Custom / other license & Open baseline \\
  Closed-API models & GPT, Claude, Gemini, Grok, and Qwen API models & Provider API terms & Closed baselines \\
  veRL and vLLM & \texttt{verl} and \texttt{vllm} software frameworks & Apache-2.0 & Training and inference infrastructure \\
  \bottomrule
  \end{tabular}
  }
  \end{minipage}%
  }
\end{table}

\subsection{Declaration of LLM Usage}
\label{asec:llm_usage}

LLMs were used to assist with editing, including grammar, spelling, and word
choice, with data processing and filtering, and with understanding technical
concepts during the research and writing process. LLM-generated text  was
not used as an unverified source of experimental results. All quantitative
claims reported in the paper are based on the experiments and analyses described
in the main text and appendix.

\clearpage
\section{Supplementary Validation Experiments}
\label{asec:supplementary_validation_experiments}

\subsection{Within-problem analysis controlling for response-level factors}

We analyzed responses from the original Qwen3-4B thinking model on MATH and
OlympiadBench. Within each dataset, we standardized TokenProbe, log response
length, mean token log probability, and 4-gram repetition coverage. We then
demeaned correctness and these predictors within each problem and fit a
no-intercept regression. This compares responses to the same problem while
controlling for length, generation confidence, and repetition. TokenProbe
retained a positive coefficient on both datasets ($\beta_{\mathrm{TP}}=0.277$
on MATH and $0.148$ on OlympiadBench), indicating an incremental association
with correctness after these controls.

\subsection{Causal interventions on high-NormLP regions}

We used Qwen3-4B on MATH and OlympiadBench and compared high-NormLP, low-NormLP,
and random windows under the same token budget. For token deletion, we removed
the selected tokens from the reasoning trace, retained the original prompt and
the \verb|</think>| boundary, removed the original final answer, and
regenerated an answer with deterministic decoding. For attention-key masking,
we preserved the tokens but prevented the selected positions from serving as
attention keys for subsequent reasoning and answer tokens during prefill.

At a 20\% intervention budget, the reported correctness damage was:

\begin{table*}[h]
  \centering
  \small
  \caption{\textbf{Correctness damage under token deletion and attention-key masking.} Entries in each method column are ordered High-NormLP / Random / Low-NormLP.}
  \label{tab:supp_intervention_damage}
  \begin{tabular}{lcc}
  \toprule
  \textbf{Dataset} & \textbf{Token deletion} & \textbf{Attention-key masking} \\
  \midrule
  MATH & 2.6\% / 1.1\% / 0.8\% & 2.4\% / 0.8\% / 0.3\% \\
  OlympiadBench & 1.8\% / 0.7\% / 0.5\% & 1.5\% / 0.6\% / 0.4\% \\
  \bottomrule
  \end{tabular}
\end{table*}

High-NormLP deletion caused 1.5\% and 1.1\% more damage than random deletion
on MATH and OlympiadBench, respectively. For attention-key masking, the
corresponding differences from the tabulated results were 1.6\% and 0.9\%.
Both interventions showed the ordering High-NormLP $>$ Random $>$ Low-NormLP,
providing causal evidence that high-NormLP regions are more consequential to
downstream answers in these evaluations.

\clearpage
\section{Supplementary Closed-API Reasoning Results}
\label{asec:supplementary_closed_api}

\subsection{Closed-source models and API prices}

\begin{table*}[h]
  \centering
  \footnotesize
  \setlength{\tabcolsep}{3pt}
  \renewcommand{\arraystretch}{1.02}
  \caption{\textbf{Closed-source baseline models used in our experiments.}
  Version denotes the model version number. Links point to official announcements or model pages.
  Output prices are standard public list prices in USD per million tokens, as of September 24, 2026.}
  \label{tab:supp_closed_api_models}
  \begin{tabular}{p{2.7cm}p{4cm}p{1cm}p{1.5cm}>{\centering\arraybackslash}p{1.55cm}p{2.0cm}}
  \toprule
  \textbf{Model Name} & \textbf{API Identifier} & \textbf{Version} & \textbf{Mode} & \textbf{Output price} & \textbf{Official link} \\
  \midrule
  \multicolumn{6}{c}{\textbf{OpenAI}} \\
  \midrule
  GPT-4o & \texttt{gpt-4o} & 4o & Default & \$10.00 & \href{https://openai.com/index/introducing-gpt-4o-and-more-tools-to-chatgpt-free-users/}{Announcement} \\
  OpenAI o1 & \texttt{o1} & o1 & Reasoning & \$60.00 & \href{https://openai.com/index/learning-to-reason-with-llms/}{Announcement} \\
  OpenAI o3 & \texttt{o3} & o3 & Reasoning & \$8.00 & \href{https://openai.com/index/introducing-o3-and-o4-mini/}{Announcement} \\
  GPT-5.2 Pro & \texttt{gpt-5.2-pro} & 5.2 & Pro & \$168.00 & \href{https://openai.com/index/introducing-gpt-5-2/}{Announcement} \\
  GPT-5.4 Mini & \texttt{gpt-5.4-mini} & 5.4 & Mini & \$4.50 & \href{https://openai.com/index/introducing-gpt-5-4-mini-and-nano/}{Announcement} \\
  GPT-5.4 Pro & \texttt{gpt-5.4-pro} & 5.4 & Pro & \$180.00 & \href{https://openai.com/index/introducing-gpt-5-4/}{Announcement} \\
  GPT-5.5 & \texttt{gpt-5.5} & 5.5 & Standard & \$30.00 & \href{https://openai.com/index/introducing-gpt-5-5/}{Announcement} \\
  GPT-5.5 Pro & \texttt{gpt-5.5-pro} & 5.5 & Pro & \$180.00 & \href{https://openai.com/index/introducing-gpt-5-5/}{Announcement} \\
  GPT-5.6 Sol & \texttt{gpt-5.6-sol} & 5.6 & Sol & \$20.00 & \href{https://openai.com/index/gpt-5-6/}{Announcement} \\
  GPT-5.6 Terra & \texttt{gpt-5.6-terra} & 5.6 & Terra & \$12.00 & \href{https://openai.com/index/gpt-5-6/}{Announcement} \\
  GPT-5.6 Luna & \texttt{gpt-5.6-luna} & 5.6 & Luna & \$1.20 & \href{https://openai.com/index/gpt-5-6/}{Announcement} \\
  \midrule
  \multicolumn{6}{c}{\textbf{xAI}} \\
  \midrule
  Grok-4.1-Fast & \texttt{grok-4-1-fast-reasoning} & 4.1 & Fast & \$0.50 & \href{https://x.ai/news/grok-4-1-fast}{Announcement} \\
  Grok-4.3 & \texttt{grok-4.3} & 4.3 & Default & \$2.50 & \href{https://docs.x.ai/developers/models/grok-4.3}{Model page} \\
  Grok-4.20 & \texttt{grok-4.20-reasoning} & 4.20 & Reasoning & \$2.50 & \href{https://docs.x.ai/developers/models/grok-4.20}{Model page} \\
  Grok-4.20 & \texttt{grok-4.20-non-reasoning} & 4.20 & Non-reasoning & \$2.50 & \href{https://docs.x.ai/developers/models/grok-4.20}{Model page} \\
  Grok-4.20-0309 & \texttt{grok-4.20-0309-reasoning} & 4.20-0309 & Reasoning & \$2.50 & \href{https://data.x.ai/2026-04-07-grok-4-20-model-card.pdf}{Model card} \\
  Grok-4.20-0309 & \texttt{grok-4.20-0309-non-reasoning} & 4.20-0309 & Non-reasoning & \$2.50 & \href{https://data.x.ai/2026-04-07-grok-4-20-model-card.pdf}{Model card} \\
  \midrule
  \multicolumn{6}{c}{\textbf{Google}} \\
  \midrule
  Gemini-3-Flash & \texttt{gemini-3-flash-preview} & 3 & Flash & \$3.00 & \href{https://ai.google.dev/gemini-api/docs/gemini-3}{Model guide} \\
  Gemini-3.1-Pro & \texttt{gemini-3.1-pro-preview} & 3.1 & Pro & \$12.00 & \href{https://ai.google.dev/gemini-api/docs/gemini-3}{Model guide} \\
  Gemini-3.1-Flash-Lite & \texttt{gemini-3.1-flash-lite} & 3.1 & Flash-Lite & \$1.50 & \href{https://ai.google.dev/gemini-api/docs/models/gemini-3.1-flash-lite}{Model page} \\
  \midrule
  \multicolumn{6}{c}{\textbf{Anthropic}} \\
  \midrule
  Claude-4.5-Haiku & \texttt{claude-haiku-4-5} & 4.5 & Haiku & \$5.00 & \href{https://www.anthropic.com/news/claude-haiku-4-5}{Announcement} \\
  Claude-4.6-Sonnet & \texttt{claude-sonnet-4-6} & 4.6 & Sonnet & \$15.00 & \href{https://www.anthropic.com/news/claude-sonnet-4-6}{Announcement} \\
  Claude-4.6-Opus & \texttt{claude-opus-4-6} & 4.6 & Opus & \$25.00 & \href{https://www.anthropic.com/news/claude-opus-4-6}{Announcement} \\
  \midrule
  \multicolumn{6}{c}{\textbf{Alibaba Cloud}} \\
  \midrule
  Qwen-3.6-Plus & \texttt{qwen3.6-plus} & 3.6 & Plus & \$1.651 & \href{https://www.alibabacloud.com/help/en/model-studio/qwen3-6-plus}{Model page} \\
  \bottomrule
  \end{tabular}
\end{table*}

Output-token prices use standard public API rates and exclude input tokens, caching discounts, batch discounts, and tool fees. GPT-5.6 reasoning-effort variants share the same model price. Where prices vary with context length, the listed rates are for prompts below each provider's threshold. 

\subsection{Per-task results}

Table~\ref{tab:supp_closed_api_indomain} reports the four in-domain benchmarks. Table~\ref{tab:supp_closed_api_ood} reports the four out-of-domain benchmarks. 

\begin{table*}[t]
  \centering
  \footnotesize
  \setlength{\tabcolsep}{3pt}
  \renewcommand{\arraystretch}{1.03}
  \caption{\textbf{Per-task results on in-domain reasoning benchmarks.} \textbf{\#Tk.} is the average number of generated completion tokens per response, including billable reasoning tokens where present. \textbf{Acc.} is accuracy in percent.}
  \label{tab:supp_closed_api_indomain}
  \resizebox{\textwidth}{!}{%
  \begin{tabular}{lcccccccccc}
  \toprule
  \multirow{2}{*}{Model} & \multicolumn{2}{c}{\textbf{Average}} & \multicolumn{2}{c}{\textbf{AIME-2025}} & \multicolumn{2}{c}{\textbf{MATH-500}} & \multicolumn{2}{c}{\textbf{AMC}} & \multicolumn{2}{c}{\textbf{OlympiadBench}} \\
  \cmidrule(lr){2-3} \cmidrule(lr){4-5} \cmidrule(lr){6-7} \cmidrule(lr){8-9} \cmidrule(lr){10-11}
  & \#Tk. & Acc. & \#Tk. & Acc. & \#Tk. & Acc. & \#Tk. & Acc. & \#Tk. & Acc. \\
  \midrule
  \multicolumn{11}{c}{\textbf{Previously reported closed-API baselines}} \\
  \midrule
  GPT-5.4-Mini & 987 & 69.6 & 1{,}714 & 36.7 & 589 & 88.6 & 1{,}137 & 73.5 & 1{,}232 & 56.6 \\
  Claude-4.6-Sonnet & 1{,}221 & 61.3 & 1{,}907 & 20.0 & 801 & 86.0 & 1{,}413 & 51.8 & 1{,}478 & 46.1 \\
  Grok-4.1-Fast & 4{,}945 & 52.3 & 9{,}361 & 10.0 & 1{,}866 & 74.8 & 4{,}017 & 43.4 & 7{,}144 & 38.5 \\
  Gemini-3.1-Pro & 1{,}586 & 54.4 & 2{,}020 & 10.0 & 1{,}274 & 79.2 & 1{,}798 & 37.3 & 1{,}771 & 40.1 \\
  Gemini-3-Flash & 1{,}828 & 35.1 & 2{,}039 & 3.3 & 1{,}625 & 56.0 & 1{,}948 & 24.1 & 1{,}954 & 22.4 \\
  Claude-4.6-Opus & 1{,}130 & 70.0 & 1{,}852 & 30.0 & 689 & 93.2 & 1{,}375 & 65.1 & 1{,}395 & 55.1 \\
  Claude-4.5-Haiku & 1{,}339 & 72.0 & 1{,}803 & 23.3 & 1{,}061 & 89.8 & 1{,}533 & 66.3 & 1{,}500 & 61.6 \\
  Gemini-3.1-Flash-Lite & 778 & 80.4 & 1{,}036 & 36.7 & 614 & 94.6 & 893 & 81.9 & 875 & 71.6 \\
  \midrule
  \multicolumn{11}{c}{\textbf{OpenAI evaluations}} \\
  \midrule
  GPT-4o & 619 & 49.9 & 741 & 0.0 & 507 & 69.8 & 732 & 42.2 & 682 & 38.4 \\
  GPT-5.2 Pro & 978 & 64.4 & 1{,}679 & 43.3 & 601 & 83.8 & 1{,}116 & 85.5 & 1{,}209 & 48.4 \\
  GPT-5.4 Pro & 991 & 63.8 & 1{,}707 & 43.3 & 624 & 83.4 & 1{,}100 & 83.1 & 1{,}222 & 47.6 \\
  GPT-5.5 & 801 & 67.9 & 1{,}419 & 70.0 & 427 & 85.8 & 796 & 90.4 & 1{,}051 & 51.7 \\
  GPT-5.5 Pro & 981 & 62.0 & 1{,}723 & 36.7 & 585 & 82.0 & 1{,}052 & 79.5 & 1{,}233 & 46.2 \\
  GPT-5.6 Luna & 761 & 59.9 & 1{,}359 & 66.7 & 402 & 77.4 & 735 & 88.0 & 1{,}004 & 43.1 \\
  GPT-5.6 Sol & 652 & 63.2 & 1{,}211 & 76.7 & 302 & 79.2 & 602 & 84.3 & 893 & 48.1 \\
  GPT-5.6 Sol (high) & 695 & 63.0 & 1{,}365 & 56.7 & 332 & 80.4 & 647 & 85.5 & 940 & 47.6 \\
  GPT-5.6 Sol (low) & 587 & 65.0 & 1{,}074 & 80.0 & 272 & 80.0 & 526 & 85.5 & 806 & 50.7 \\
  GPT-5.6 Sol (max) & 837 & 56.8 & 1{,}556 & 43.3 & 444 & 75.8 & 870 & 77.1 & 1{,}092 & 40.9 \\
  GPT-5.6 Sol (xhigh) & 747 & 60.0 & 1{,}445 & 56.7 & 365 & 78.2 & 710 & 84.3 & 1{,}004 & 43.7 \\
  GPT-5.6 Terra & 628 & 65.4 & 1{,}176 & 83.3 & 292 & 81.2 & 551 & 92.8 & 862 & 49.5 \\
  OpenAI o1 & 1{,}390 & 40.5 & 1{,}976 & 0.0 & 994 & 62.4 & 1{,}651 & 28.9 & 1{,}625 & 27.4 \\
  OpenAI o3 & 1{,}282 & 44.4 & 1{,}947 & 13.3 & 886 & 63.8 & 1{,}497 & 50.6 & 1{,}518 & 30.7 \\
  \midrule
  \multicolumn{11}{c}{\textbf{xAI evaluations}} \\
  \midrule
  Grok-4.20 (non-reasoning) & 1{,}480 & 38.9 & 2{,}008 & 10.0 & 1{,}075 & 58.6 & 1{,}746 & 37.3 & 1{,}723 & 25.8 \\
  Grok-4.20 (reasoning) & 4{,}244 & 72.9 & 8{,}450 & 96.4 & 2{,}208 & 76.6 & 4{,}967 & 98.8 & 5{,}615 & 65.3 \\
  Grok-4.20-0309 (non-reasoning) & 1{,}477 & 39.4 & 2{,}006 & 10.0 & 1{,}081 & 60.0 & 1{,}763 & 32.5 & 1{,}712 & 26.2 \\
  Grok-4.3 & 1{,}703 & 64.8 & 3{,}036 & 60.0 & 1{,}065 & 73.0 & 1{,}844 & 94.0 & 2{,}099 & 55.4 \\
  \bottomrule
  \end{tabular}}
\end{table*}

\begin{table*}[t]
  \centering
  \footnotesize
  \setlength{\tabcolsep}{3pt}
  \renewcommand{\arraystretch}{1.03}
  \caption{\textbf{Per-task results on out-of-domain generalization.} The four tasks are MMLU, MMLU-Pro, LSAT, and GPQA. \textbf{\#Tk.} is the average number of generated completion tokens per response, including billable reasoning tokens where present. \textbf{Acc.} is accuracy in percent.}
  \label{tab:supp_closed_api_ood}
  \resizebox{\textwidth}{!}{%
  \begin{tabular}{lcccccccccc}
  \toprule
  \multirow{2}{*}{Model} & \multicolumn{2}{c}{\textbf{Average}} & \multicolumn{2}{c}{\textbf{MMLU}} & \multicolumn{2}{c}{\textbf{MMLU-Pro}} & \multicolumn{2}{c}{\textbf{LSAT}} & \multicolumn{2}{c}{\textbf{GPQA}} \\
  \cmidrule(lr){2-3} \cmidrule(lr){4-5} \cmidrule(lr){6-7} \cmidrule(lr){8-9} \cmidrule(lr){10-11}
  & \#Tk. & Acc. & \#Tk. & Acc. & \#Tk. & Acc. & \#Tk. & Acc. & \#Tk. & Acc. \\
  \midrule
  \multicolumn{11}{c}{\textbf{Previously reported closed-API baselines}} \\
  \midrule
  Grok-4.1-Fast & 2{,}892 & 88.2 & 1{,}026 & 90.9 & 2{,}288 & 84.1 & 2{,}797 & 96.1 & 5{,}459 & 81.8 \\
  Claude-4.5-Haiku & 1{,}090 & 72.0 & 706 & 88.6 & 883 & 79.8 & 1{,}521 & 54.8 & 1{,}250 & 64.7 \\
  Gemini-3.1-Pro & 1{,}294 & 68.6 & 681 & 92.8 & 1{,}055 & 80.3 & 1{,}727 & 54.4 & 1{,}714 & 47.0 \\
  Claude-4.6-Sonnet & 1{,}290 & 57.1 & 645 & 88.8 & 965 & 73.0 & 1{,}797 & 35.7 & 1{,}754 & 30.8 \\
  Gemini-3-Flash & 1{,}651 & 29.1 & 1{,}123 & 64.9 & 1{,}593 & 33.9 & 1{,}915 & 12.6 & 1{,}973 & 5.1 \\
  \midrule
  \multicolumn{11}{c}{\textbf{OpenAI evaluations}} \\
  \midrule
  GPT-4o & 443 & 71.9 & 345 & 86.8 & 474 & 70.2 & 563 & 38.3 & 646 & 43.9 \\
  GPT-5.2 Pro & 495 & 84.7 & 215 & 92.8 & 437 & 82.0 & 1{,}393 & 82.2 & 1{,}162 & 60.6 \\
  GPT-5.4 Pro & 667 & 85.7 & 378 & 93.1 & 670 & 83.0 & 1{,}371 & 86.5 & 1{,}307 & 61.1 \\
  GPT-5.5 & 373 & 88.7 & 208 & 94.4 & 363 & 85.1 & 709 & 96.1 & 865 & 69.2 \\
  GPT-5.5 Pro & 675 & 83.9 & 429 & 92.4 & 685 & 80.7 & 1{,}115 & 87.0 & 1{,}347 & 53.0 \\
  GPT-5.6 Luna & \textemdash & \textemdash & 169 & 24.3 & 341 & 13.6 & 726 & 70.9 & \textemdash & \textemdash \\
  GPT-5.6 Sol & \textemdash & \textemdash & 138 & 22.1 & 255 & 25.6 & 474 & 84.8 & \textemdash & \textemdash \\
  GPT-5.6 Sol (high) & \textemdash & \textemdash & 162 & 17.7 & 317 & 24.2 & 526 & 86.1 & \textemdash & \textemdash \\
  GPT-5.6 Sol (low) & \textemdash & \textemdash & 103 & 28.3 & 201 & 31.0 & 418 & 88.3 & \textemdash & \textemdash \\
  GPT-5.6 Sol (max) & \textemdash & \textemdash & 284 & 16.0 & 495 & 22.0 & 654 & 82.2 & \textemdash & \textemdash \\
  GPT-5.6 Sol (xhigh) & \textemdash & \textemdash & 188 & 16.0 & 371 & 22.2 & 556 & 82.6 & \textemdash & \textemdash \\
  GPT-5.6 Terra & \textemdash & \textemdash & 94 & 26.1 & 202 & 26.7 & 377 & 51.3 & \textemdash & \textemdash \\
  OpenAI o1 & 989 & 61.0 & 553 & 84.2 & 1{,}056 & 60.2 & 1{,}942 & 0.0 & 1{,}748 & 18.7 \\
  OpenAI o3 & 774 & 68.5 & 425 & 88.9 & 757 & 64.4 & 1{,}722 & 31.7 & 1{,}518 & 28.8 \\
  \midrule
  \multicolumn{11}{c}{\textbf{xAI evaluations}} \\
  \midrule
  Grok-4.20 (non-reasoning) & 1{,}229 & 67.3 & 929 & 84.3 & 1{,}252 & 67.4 & 1{,}901 & 23.9 & 1{,}850 & 30.8 \\
  Grok-4.20 (reasoning) & 2{,}526 & 90.5 & 1{,}393 & 92.8 & 2{,}707 & 86.8 & 3{,}794 & 96.1 & 5{,}951 & 91.2 \\
  Grok-4.20-0309 (non-reasoning) & 1{,}235 & 66.4 & 935 & 83.9 & 1{,}258 & 65.9 & 1{,}900 & 23.5 & 1{,}862 & 30.3 \\
  Grok-4.20-0309 (reasoning) & 2{,}501 & 91.1 & 1{,}384 & 93.3 & 2{,}705 & 87.3 & 3{,}767 & 96.1 & 5{,}744 & 93.2 \\
  Grok-4.3 & 898 & 89.7 & 629 & 92.9 & 850 & 86.1 & 1{,}810 & 95.2 & 1{,}443 & 84.8 \\
  \bottomrule
  \end{tabular}}
\end{table*}

\clearpage
\section{Appendix of Tables}
\label{asec: appendix_tables}

\subsection{More Results on Reasoning Benchmarks}
\label{asec: more_main}

\begin{table*}[h]
  \centering
  \caption{\textbf{Main results on four in-domain reasoning benchmarks.}
  \textbf{\#Tk.}\ is the average number of generated tokens per response
  (lower is better). \textbf{Acc.}\ is averaged accuracy in \% (higher is better).}
  \resizebox{\textwidth}{!}{%
  \begin{tabular}{l>{\cellcolor[gray]{0.9}}c>{\cellcolor[gray]{0.9}}ccccccccc}
  \toprule
  \multirow{2}{*}{Model} & \multicolumn{2}{c}{\textbf{Average}} & \multicolumn{2}{c}{\textbf{AIME}} & \multicolumn{2}{c}{\textbf{MATH}} & \multicolumn{2}{c}{\textbf{AMC}} & \multicolumn{2}{c}{\textbf{Olympid}} \\
  \cmidrule(lr){2-3} \cmidrule(lr){4-5} \cmidrule(lr){6-7} \cmidrule(lr){8-9} \cmidrule(lr){10-11}
  & \#Tk. & Acc. & \#Tk. & Acc. & \#Tk. & Acc. & \#Tk. & Acc. & \#Tk. & Acc. \\
  \midrule
  \multicolumn{11}{c}{\textbf{Closed-API Reasoning Models}}\\
  \midrule
  Claude-4.6-Opus        & 1{,}130 & 70.0 & 1{,}852 & 30.0 &   689 & 93.2 & 1{,}375 & 65.1 & 1{,}395 & 55.1 \\
  Claude-4.5-Haiku       & 1{,}339 & 72.0 & 1{,}803 & 23.3 & 1{,}061 & 89.8 & 1{,}533 & 66.3 & 1{,}500 & 61.6 \\
  Gemini-3.1-Flash-Lite  &     778 & 80.4 & 1{,}036 & 36.7 &     614 & 94.6 &     893 & 81.9 &     875 & 71.6 \\
  \midrule
  \multicolumn{11}{c}{\textbf{Open Source Reasoning Models}}\\
  \midrule
  GLM-744B-40B           & 1{,}742 & 31.1 & 2{,}048 &  0.0 & 1{,}487 & 52.0 & 1{,}900 & 16.9 & 1{,}898 & 18.8 \\
  DeepSeek-1.6T-49B      & 1{,}305 & 58.8 & 1{,}884 & 23.3 &     920 & 82.0 & 1{,}482 & 49.4 & 1{,}543 & 44.3 \\
  DeepSeek-284B-13B      & 1{,}308 & 57.3 & 1{,}892 & 16.7 &     906 & 79.2 & 1{,}510 & 50.6 & 1{,}554 & 43.7 \\
  \bottomrule
  \end{tabular}
  }
  \vspace{-0.5cm}
  \label{tab:main_results_apex}
\end{table*}

\subsection{Per-task Results on Unlimited Token Budget}
\label{asec: un_appendix_tables}

\begin{table*}[h]
  \centering
  \caption{\textbf{Token budget gradient results at 32K.}}
  \resizebox{\textwidth}{!}{%
  \begin{tabular}{lcccccccccc}
  \toprule
  \multirow{2}{*}{Model} & \multicolumn{2}{c}{\textbf{Average}} & \multicolumn{2}{c}{\textbf{AIME}} & \multicolumn{2}{c}{\textbf{MATH}} & \multicolumn{2}{c}{\textbf{AMC}} & \multicolumn{2}{c}{\textbf{OlympiadBench}} \\
  \cmidrule(lr){2-3} \cmidrule(lr){4-5} \cmidrule(lr){6-7} \cmidrule(lr){8-9} \cmidrule(lr){10-11}
  & \#Tk. & Acc. & \#Tk. & Acc. & \#Tk. & Acc. & \#Tk. & Acc. & \#Tk. & Acc. \\
  \midrule
  Qwen-3.6-Plus       & 15{,}207 & 80.0 & 25{,}812 & 56.7 & 8812 & 96.2 & 18{,}520 & 90.4 & 19{,}065 & 67.7 \\
  Gemini-3-Flash      & 4844     & 91.4 & 8332     & 93.3 & 2830 & 98.6 & 5049     & 96.4 & 6156     & 85.3 \\
  Qwen3-Next-80B-3B   & 8472     & 88.0 & 16{,}183 & 83.3 & 4185 & 97.8 & 9193     & 92.8 & 11{,}216 & 80.4 \\
  GLM-Air-106B-12B    & 11{,}278 & 86.1 & 20{,}859 & 83.3 & 6517 & 97.8 & 13{,}257 & 92.8 & 14{,}136 & 76.7 \\
  Step-Flash-196B-11B & 10{,}147 & 80.2 & 23{,}018 & 56.7 & 4124 & 97.0 & 10{,}868 & 88.0 & 13{,}949 & 67.9 \\
  Hunyuan3-295B-21B   & 9552     & 81.5 & 22{,}038 & 66.7 & 4238 & 97.4 & 10{,}685 & 86.7 & 12{,}794 & 69.8 \\
  \bottomrule
  \end{tabular}
  }
  \label{tab:unlimited_32k}
\end{table*}

\begin{table*}[h]
  \centering
  \caption{\textbf{Token budget gradient results at 16K.}}
  \resizebox{\textwidth}{!}{%
  \begin{tabular}{lcccccccccc}
  \toprule
  \multirow{2}{*}{Model} & \multicolumn{2}{c}{\textbf{Average}} & \multicolumn{2}{c}{\textbf{AIME}} & \multicolumn{2}{c}{\textbf{MATH}} & \multicolumn{2}{c}{\textbf{AMC}} & \multicolumn{2}{c}{\textbf{OlympiadBench}} \\
  \cmidrule(lr){2-3} \cmidrule(lr){4-5} \cmidrule(lr){6-7} \cmidrule(lr){8-9} \cmidrule(lr){10-11}
  & \#Tk. & Acc. & \#Tk. & Acc. & \#Tk. & Acc. & \#Tk. & Acc. & \#Tk. & Acc. \\
  \midrule
  Qwen-3.6-Plus       & 10{,}678 & 55.9 & 15{,}472 & 10.0 & 7432 & 80.4 & 13{,}440 & 42.2 & 12{,}530 & 41.5 \\
  Gemini-3-Flash      & 4802     & 90.4 & 8332     & 93.3 & 2830 & 98.6 & 4989     & 95.2 & 6083     & 83.6 \\
  Qwen3-Next-80B-3B   & 7045     & 77.3 & 12{,}568 & 56.7 & 3988 & 95.2 & 8334     & 83.1 & 8905     & 64.3 \\
  GLM-Air-106B-12B    & 8831     & 71.7 & 14{,}191 & 40.0 & 5970 & 90.4 & 10{,}837 & 71.1 & 10{,}465 & 59.3 \\
  Step-Flash-196B-11B & 6692     & 71.7 & 13{,}740 & 33.3 & 3338 & 92.0 & 7752     & 72.3 & 8734     & 58.2 \\
  Hunyuan3-295B-21B   & 6771     & 70.7 & 13{,}509 & 30.0 & 3728 & 92.0 & 8343     & 71.1 & 8532     & 56.7 \\
  \bottomrule
  \end{tabular}
  }
  \label{tab:unlimited_16k}
\end{table*}

\begin{table*}[h]
  \centering
  \caption{\textbf{Token budget gradient results at 8K.}}
  \resizebox{\textwidth}{!}{%
  \begin{tabular}{lcccccccccc}
  \toprule
  \multirow{2}{*}{Model} & \multicolumn{2}{c}{\textbf{Average}} & \multicolumn{2}{c}{\textbf{AIME}} & \multicolumn{2}{c}{\textbf{MATH}} & \multicolumn{2}{c}{\textbf{AMC}} & \multicolumn{2}{c}{\textbf{OlympiadBench}} \\
  \cmidrule(lr){2-3} \cmidrule(lr){4-5} \cmidrule(lr){6-7} \cmidrule(lr){8-9} \cmidrule(lr){10-11}
  & \#Tk. & Acc. & \#Tk. & Acc. & \#Tk. & Acc. & \#Tk. & Acc. & \#Tk. & Acc. \\
  \midrule
  Qwen-3.6-Plus       & 6492 & 38.5 & 7919 & 0.0  & 5196 & 64.8 & 7500 & 19.3 & 7265 & 23.1 \\
    Gemini-3-Flash      & 4106 & 73.8 & 6515 & 53.3 & 2673 & 92.4 & 4286 & 78.3 & 5039 & 60.3 \\
  Qwen3-Next-80B-3B   & 4991 & 60.2 & 7722 & 16.7 & 3326 & 83.0 & 6111 & 56.6 & 5965 & 45.8 \\
  GLM-Air-106B-12B    & 6033 & 49.6 & 7988 & 10.0 & 4711 & 72.8 & 7038 & 33.7 & 6801 & 36.1 \\
  Step-Flash-196B-11B & 4245 & 63.2 & 7484 & 16.7 & 2527 & 85.2 & 5022 & 57.8 & 5279 & 49.6 \\
  Hunyuan3-295B-21B   & 4406 & 60.6 & 7465 & 20.0 & 2877 & 83.4 & 5494 & 50.6 & 5269 & 46.7 \\
  \bottomrule
  \end{tabular}
  }
  \label{tab:unlimited_8k}
\end{table*}

\begin{table*}[h]
  \centering
  \caption{\textbf{Token budget gradient results at 4K.}}
  \resizebox{\textwidth}{!}{%
  \begin{tabular}{lcccccccccc}
  \toprule
  \multirow{2}{*}{Model} & \multicolumn{2}{c}{\textbf{Average}} & \multicolumn{2}{c}{\textbf{AIME}} & \multicolumn{2}{c}{\textbf{MATH}} & \multicolumn{2}{c}{\textbf{AMC}} & \multicolumn{2}{c}{\textbf{OlympiadBench}} \\
  \cmidrule(lr){2-3} \cmidrule(lr){4-5} \cmidrule(lr){6-7} \cmidrule(lr){8-9} \cmidrule(lr){10-11}
  & \#Tk. & Acc. & \#Tk. & Acc. & \#Tk. & Acc. & \#Tk. & Acc. & \#Tk. & Acc. \\
  \midrule
  Qwen-3.6-Plus       & 3727 & 21.0 & 3959 & 0.0  & 3361 & 43.2 & 3986 & 7.2  & 3957 & 7.3  \\
  Gemini-3-Flash      & 2860 & 58.6 & 3885 & 20.0 & 2217 & 82.2 & 3038 & 54.2 & 3269 & 43.4 \\
  Qwen3-Next-80B-3B   & 3101 & 42.8 & 4096 & 0.0  & 2391 & 69.2 & 3590 & 25.3 & 3522 & 27.3 \\
  GLM-Air-106B-12B    & 3606 & 28.3 & 4096 & 0.0  & 3179 & 49.0 & 3875 & 16.9 & 3867 & 15.7 \\
  Step-Flash-196B-11B & 2619 & 53.4 & 3962 & 10.0 & 1784 & 76.2 & 3004 & 45.8 & 3131 & 39.4 \\
  Hunyuan3-295B-21B   & 2698 & 46.3 & 3988 & 10.0 & 1994 & 70.8 & 3219 & 30.1 & 3098 & 31.7 \\
  \bottomrule
  \end{tabular}
  }
  \label{tab:unlimited_4k}
\end{table*}

\begin{table*}[h]
  \centering
  \caption{\textbf{Token budget gradient results at 1K.}}
  \resizebox{\textwidth}{!}{%
  \begin{tabular}{lcccccccccc}
  \toprule
  \multirow{2}{*}{Model} & \multicolumn{2}{c}{\textbf{Average}} & \multicolumn{2}{c}{\textbf{AIME}} & \multicolumn{2}{c}{\textbf{MATH}} & \multicolumn{2}{c}{\textbf{AMC}} & \multicolumn{2}{c}{\textbf{OlympiadBench}} \\
  \cmidrule(lr){2-3} \cmidrule(lr){4-5} \cmidrule(lr){6-7} \cmidrule(lr){8-9} \cmidrule(lr){10-11}
  & \#Tk. & Acc. & \#Tk. & Acc. & \#Tk. & Acc. & \#Tk. & Acc. & \#Tk. & Acc. \\
  \midrule
  Qwen-3.6-Plus       & 1016 & 0.1  & 990  & 0.0  & 1021 & 0.2  & 1024 & 0.0  & 1012 & 0.0  \\
  Gemini-3-Flash      & 1008 & 6.9  & 1024 & 0.0  & 986  & 16.4 & 1022 & 1.2  & 1023 & 0.9  \\
  Qwen3-Next-80B-3B   & 1007 & 7.6  & 1024 & 0.0  & 987  & 16.0 & 1021 & 3.6  & 1020 & 2.2  \\
  GLM-Air-106B-12B    & 1020 & 2.4  & 1024 & 0.0  & 1015 & 5.8  & 1024 & 0.0  & 1024 & 0.3  \\
  Step-Flash-196B-11B & 912  & 25.8 & 1024 & 0.0  & 794  & 48.6 & 982  & 13.3 & 985  & 11.6 \\
  Hunyuan3-295B-21B   & 905  & 22.4 & 1024 & 0.0  & 832  & 43.4 & 962  & 12.0 & 946  & 9.2  \\
  \bottomrule
  \end{tabular}
  }
  \label{tab:unlimited_1k}
\end{table*}

\subsection{Per-task Results on Out-of-domain Generalization}
\label{asec:unlimited_ood_tables}

\begin{table*}[h]
  \centering
  \caption{\textbf{Per-task results on out-of-domain generalization.}
  Models are evaluated on four general-domain tasks (MMLU, MMLU-Pro, LSAT, GPQA) under a 2{,}048-token rollout cap.
  \textbf{\#Tk.}\ is the average number of generated tokens per response (lower is better).
  \textbf{Acc.}\ is accuracy in \% (higher is better).}
  \resizebox{\textwidth}{!}{%
  \begin{tabular}{lcccccccccc}
  \toprule
  \multirow{2}{*}{Model} & \multicolumn{2}{c}{\textbf{Average}} & \multicolumn{2}{c}{\textbf{MMLU}} & \multicolumn{2}{c}{\textbf{MMLU-Pro}} & \multicolumn{2}{c}{\textbf{LSAT}} & \multicolumn{2}{c}{\textbf{GPQA}} \\
  \cmidrule(lr){2-3} \cmidrule(lr){4-5} \cmidrule(lr){6-7} \cmidrule(lr){8-9} \cmidrule(lr){10-11}
  & \#Tk. & Acc. & \#Tk. & Acc. & \#Tk. & Acc. & \#Tk. & Acc. & \#Tk. & Acc. \\
  \midrule
  \multicolumn{11}{c}{\textbf{Closed-API Reasoning Models}} \\
  \midrule
  Grok-4.1-Fast       & 2892 & 88.2 & 1026 & 90.9 & 2288 & 84.1 & 2797 & 96.1 & 5459 & 81.8 \\
  Claude-4.5-Haiku    & 1090 & 72.0 & 706  & 88.6 & 883  & 79.8 & 1521 & 54.8 & 1250 & 64.7 \\
  Gemini-3.1-Pro      & 1294 & 68.6 & 681  & 92.8 & 1055 & 80.3 & 1727 & 54.4 & 1714 & 47.0 \\
  Claude-4.6-Sonnet   & 1290 & 57.1 & 645  & 88.8 & 965  & 73.0 & 1797 & 35.7 & 1754 & 30.8 \\
  Gemini-3-Flash      & 1651 & 29.1 & 1123 & 64.9 & 1593 & 33.9 & 1915 & 12.6 & 1973 & 5.1  \\
  \midrule
  \multicolumn{11}{c}{\textbf{Open-Source Reasoning Models}} \\
  \midrule
  DeepSeek-671B-37B   & 1556 & 43.6 & 918  & 82.1 & 1442 & 54.5 & 1938 & 19.6 & 1926 & 18.2 \\
  GLM-358B-32B        & 1575 & 39.8 & 962  & 75.6 & 1460 & 50.3 & 1936 & 16.1 & 1942 & 17.2 \\
  Qwen3-Next-80B-3B   & 1843 & 24.5 & 1509 & 62.3 & 1802 & 32.8 & 2031 & 0.9  & 2032 & 2.0  \\
  GPT-117B-5B         & 980  & 15.3 & 289  & 21.6 & 597  & 17.7 & 1671 & 8.6  & 1363 & 13.2 \\
  \midrule
  \multicolumn{11}{c}{\textbf{Ours}} \\
  \midrule
  Qwen-4B~\cite{yang2025qwen3} & 4401 & 59.6 & 1775 & 80.3 & 3466 & 63.6 & 5907 & 61.7 & 6457 & 32.8 \\
  \hspace{0.4cm} -   \hspace{0.1cm}w/ KL & 935  & 60.6 & 446 & 77.9 & 721 & 64.6 & 1574 & 50.4 & 1000 & 49.5 \\
  \hspace{0.4cm} -   \hspace{0.1cm}w/ RS & 1179 & 57.7 & 664 & 78.1 & 933 & 63.5 & 1796 & 43.9 & 1323 & 45.5 \\
  \bottomrule
  \end{tabular}
  }
  \label{tab:ood_generalization}
\end{table*}

\clearpage
\input{checklist.tex}

\end{document}

%% file: case/wait1.tex
\begingroup
\hbadness=10000
\hfuzz=10pt
\begin{figure*}[htbp]
\centering
{\noindent

{\scriptsize
\setlength{\fboxrule}{.5pt}\fcolorbox{black}{green!10}{\parbox{0.98\textwidth}{\strut \textbf{Model:} Qwen-4B \\ \textbf{Ground Truth:} 49 \\ \textbf{Correct:} Yes \\ \textbf{Positive Ratio:} 0.7402 \\ \textbf{Prompt:} The operation @ is defined as \$\textbackslash{}frac\{m\}\{n\}@\textbackslash{}frac\{p\}\{q\} = (m)(p)\textbackslash{}left(\textbackslash{}frac\{q\}\{n\}\textbackslash{}right)\$ for simplified fractions \$\textbackslash{}frac\{p\}\{q\}\$.  What is the simplified value of \$\textbackslash{}frac\{7\}\{30\}@\textbackslash{}frac\{10\}\{21\}\$? Let's think step by step and output the final answer within \textbackslash{}boxed\{\}.}}

{\scriptsize
\setlength{\fboxrule}{.5pt}\fcolorbox{black}{gray!10}{\parbox{0.98\textwidth}{\raggedright\setlength{\parindent}{0pt}\def\\{\ifhmode\unskip\fi\par\noindent}\strut 
{\tiny\color{gray}1}\,%
\setlength{\fboxsep}{0pt}\fcolorbox{gray!10}{gray!10}{\strut
    \mycolorbox[text=\textbf{Response:}]%%
}
\setlength{\fboxsep}{0pt}\fcolorbox{gray!10}{gray!10}{\strut
    \mycolorbox[text=\strut{\textless{}think\textgreater{}}, color=green!17.634716]%
}
\\
{\tiny\color{gray}2}\,%
\setlength{\fboxsep}{0pt}\fcolorbox{gray!10}{gray!10}{\strut
    \mycolorbox[text=\strut{Okay}, color=green!17.632373]%
    \mycolorbox[text=\strut{,}, color=green!17.636577]%
}
\setlength{\fboxsep}{0pt}\fcolorbox{gray!10}{gray!10}{\strut
    \mycolorbox[text=\strut{so}, color=red!28.769062]%
}
\setlength{\fboxsep}{0pt}\fcolorbox{gray!10}{gray!10}{\strut
    \mycolorbox[text=\strut{I}, color=green!17.605463]%
}
\setlength{\fboxsep}{0pt}\fcolorbox{gray!10}{gray!10}{\strut
    \mycolorbox[text=\strut{need}, color=green!17.457303]%
}
\setlength{\fboxsep}{0pt}\fcolorbox{gray!10}{gray!10}{\strut
    \mycolorbox[text=\strut{to}, color=green!17.636594]%
}
\setlength{\fboxsep}{0pt}\fcolorbox{gray!10}{gray!10}{\strut
    \mycolorbox[text=\strut{figure}, color=green!11.517471]%
}
\setlength{\fboxsep}{0pt}\fcolorbox{gray!10}{gray!10}{\strut
    \mycolorbox[text=\strut{out}, color=green!17.636594]%
}
\setlength{\fboxsep}{0pt}\fcolorbox{gray!10}{gray!10}{\strut
    \mycolorbox[text=\strut{what}]%
}
\setlength{\fboxsep}{0pt}\fcolorbox{gray!10}{gray!10}{\strut
    \mycolorbox[text=\strut{the}, color=green!15.904788]%
}
\setlength{\fboxsep}{0pt}\fcolorbox{gray!10}{gray!10}{\strut
    \mycolorbox[text=\strut{operation}, color=green!17.621206]%
}
\setlength{\fboxsep}{0pt}\fcolorbox{gray!10}{gray!10}{\strut
    \mycolorbox[text=\strut{@}, color=green!17.633441]%
}
\setlength{\fboxsep}{0pt}\fcolorbox{gray!10}{gray!10}{\strut
    \mycolorbox[text=\strut{does}, color=green!17.595931]%
}
\setlength{\fboxsep}{0pt}\fcolorbox{gray!10}{gray!10}{\strut
    \mycolorbox[text=\strut{here}, color=green!17.605213]%
    \mycolorbox[text=\strut{.}, color=green!17.636560]%
}
\setlength{\fboxsep}{0pt}\fcolorbox{gray!10}{gray!10}{\strut
    \mycolorbox[text=\strut{The}, color=green!17.527485]%
}
\setlength{\fboxsep}{0pt}\fcolorbox{gray!10}{gray!10}{\strut
    \mycolorbox[text=\strut{problem}, color=green!17.636586]%
}
\setlength{\fboxsep}{0pt}\fcolorbox{gray!10}{gray!10}{\strut
    \mycolorbox[text=\strut{says}, color=green!17.633278]%
}
\setlength{\fboxsep}{0pt}\fcolorbox{gray!10}{gray!10}{\strut
    \mycolorbox[text=\strut{that}, color=green!17.629650]%
}
\setlength{\fboxsep}{0pt}\fcolorbox{gray!10}{gray!10}{\strut
    \mycolorbox[text=\strut{for}, color=red!42.300623]%
}
\setlength{\fboxsep}{0pt}\fcolorbox{gray!10}{gray!10}{\strut
    \mycolorbox[text=\strut{simplified}, color=green!12.901218]%
}
\setlength{\fboxsep}{0pt}\fcolorbox{gray!10}{gray!10}{\strut
    \mycolorbox[text=\strut{fractions}, color=green!17.636594]%
}
\setlength{\fboxsep}{0pt}\fcolorbox{gray!10}{gray!10}{\strut
    \mycolorbox[text=\strut{p}, color=green!14.488387]%
    \mycolorbox[text=\strut{/q}, color=green!17.636586]%
}
\setlength{\fboxsep}{0pt}\fcolorbox{gray!10}{gray!10}{\strut
    \mycolorbox[text=\strut{and}, color=green!17.406321]%
}
\setlength{\fboxsep}{0pt}\fcolorbox{gray!10}{gray!10}{\strut
    \mycolorbox[text=\strut{m}, color=green!17.636577]%
    \mycolorbox[text=\strut{/n}, color=green!17.636594]%
    \mycolorbox[text=\strut{,}, color=green!17.636594]%
}
\setlength{\fboxsep}{0pt}\fcolorbox{gray!10}{gray!10}{\strut
    \mycolorbox[text=\strut{the}, color=green!17.636594]%
}
\setlength{\fboxsep}{0pt}\fcolorbox{gray!10}{gray!10}{\strut
    \mycolorbox[text=\strut{operation}, color=green!17.636594]%
}
\setlength{\fboxsep}{0pt}\fcolorbox{gray!10}{gray!10}{\strut
    \mycolorbox[text=\strut{@}, color=red!5.626196]%
}
\setlength{\fboxsep}{0pt}\fcolorbox{gray!10}{gray!10}{\strut
    \mycolorbox[text=\strut{is}, color=green!17.636594]%
}
\setlength{\fboxsep}{0pt}\fcolorbox{gray!10}{gray!10}{\strut
    \mycolorbox[text=\strut{defined}, color=green!17.636594]%
}
\setlength{\fboxsep}{0pt}\fcolorbox{gray!10}{gray!10}{\strut
    \mycolorbox[text=\strut{as}, color=green!17.636594]%
}
\setlength{\fboxsep}{0pt}\fcolorbox{gray!10}{gray!10}{\strut
    \mycolorbox[text=\strut{(}, color=green!17.596603]%
    \mycolorbox[text=\strut{m}, color=green!17.636594]%
    \mycolorbox[text=\strut{/n}, color=green!16.028171]%
    \mycolorbox[text=\strut{)}, color=green!17.636594]%
}
\setlength{\fboxsep}{0pt}\fcolorbox{gray!10}{gray!10}{\strut
    \mycolorbox[text=\strut{@}, color=green!17.511546]%
}
\setlength{\fboxsep}{0pt}\fcolorbox{gray!10}{gray!10}{\strut
    \mycolorbox[text=\strut{(}, color=green!17.636594]%
    \mycolorbox[text=\strut{p}, color=green!17.636594]%
    \mycolorbox[text=\strut{/q}, color=green!17.636594]%
    \mycolorbox[text=\strut{)}, color=green!17.636594]%
}
\setlength{\fboxsep}{0pt}\fcolorbox{gray!10}{gray!10}{\strut
    \mycolorbox[text=\strut{=}, color=green!16.304369]%
}
\setlength{\fboxsep}{0pt}\fcolorbox{gray!10}{gray!10}{\strut
    \mycolorbox[text=\strut{(}, color=green!16.600557]%
    \mycolorbox[text=\strut{m}, color=green!17.636594]%
    \mycolorbox[text=\strut{)(}, color=green!17.636594]%
    \mycolorbox[text=\strut{p}, color=green!17.636594]%
    \mycolorbox[text=\strut{)(}, color=green!17.636508]%
    \mycolorbox[text=\strut{q}, color=green!17.636594]%
    \mycolorbox[text=\strut{/n}, color=green!17.636586]%
    \mycolorbox[text=\strut{).}, color=green!17.636594]%
}
\setlength{\fboxsep}{0pt}\fcolorbox{gray!10}{gray!10}{\strut
    \setlength{\fboxsep}{1pt}\fbox{\mycolorbox[text=\strut{Wait}, color=green!16.613731]}%
    \mycolorbox[text=\strut{,}, color=green!17.636594]%
}
\setlength{\fboxsep}{0pt}\fcolorbox{gray!10}{gray!10}{\strut
    \mycolorbox[text=\strut{let}, color=green!17.389094]%
}
\setlength{\fboxsep}{0pt}\fcolorbox{gray!10}{gray!10}{\strut
    \mycolorbox[text=\strut{me}, color=green!17.636594]%
}
\setlength{\fboxsep}{0pt}\fcolorbox{gray!10}{gray!10}{\strut
    \mycolorbox[text=\strut{check}, color=red!29.913870]%
}
\setlength{\fboxsep}{0pt}\fcolorbox{gray!10}{gray!10}{\strut
    \mycolorbox[text=\strut{that}, color=green!12.503641]%
}
\setlength{\fboxsep}{0pt}\fcolorbox{gray!10}{gray!10}{\strut
    \mycolorbox[text=\strut{again}, color=green!17.635492]%
    \mycolorbox[text=\strut{.}, color=green!17.631460]%
}
\setlength{\fboxsep}{0pt}\fcolorbox{gray!10}{gray!10}{\strut
    \mycolorbox[text=\strut{It}, color=red!57.754686]%
}
\setlength{\fboxsep}{0pt}\fcolorbox{gray!10}{gray!10}{\strut
    \mycolorbox[text=\strut{says}, color=green!17.627531]%
    \mycolorbox[text=\strut{:}, color=red!13.595980]%
}
\setlength{\fboxsep}{0pt}\fcolorbox{gray!10}{gray!10}{\strut
    \mycolorbox[text=\strut{"}, color=red!3.517854]%
    \mycolorbox[text=\strut{The}, color=green!15.427332]%
}
\setlength{\fboxsep}{0pt}\fcolorbox{gray!10}{gray!10}{\strut
    \mycolorbox[text=\strut{operation}, color=green!17.636594]%
}
\setlength{\fboxsep}{0pt}\fcolorbox{gray!10}{gray!10}{\strut
    \mycolorbox[text=\strut{@}, color=green!17.636594]%
}
\setlength{\fboxsep}{0pt}\fcolorbox{gray!10}{gray!10}{\strut
    \mycolorbox[text=\strut{is}, color=green!17.636594]%
}
\setlength{\fboxsep}{0pt}\fcolorbox{gray!10}{gray!10}{\strut
    \mycolorbox[text=\strut{defined}, color=green!17.636594]%
}
\setlength{\fboxsep}{0pt}\fcolorbox{gray!10}{gray!10}{\strut
    \mycolorbox[text=\strut{as}, color=green!17.636594]%
}
\setlength{\fboxsep}{0pt}\fcolorbox{gray!10}{gray!10}{\strut
    \mycolorbox[text=\strut{m}, color=green!9.287618]%
    \mycolorbox[text=\strut{/n}, color=green!17.636551]%
}
\setlength{\fboxsep}{0pt}\fcolorbox{gray!10}{gray!10}{\strut
    \mycolorbox[text=\strut{@}, color=green!17.636577]%
}
\setlength{\fboxsep}{0pt}\fcolorbox{gray!10}{gray!10}{\strut
    \mycolorbox[text=\strut{p}, color=green!17.636594]%
    \mycolorbox[text=\strut{/q}, color=green!17.636594]%
}
\setlength{\fboxsep}{0pt}\fcolorbox{gray!10}{gray!10}{\strut
    \mycolorbox[text=\strut{=}, color=green!17.636594]%
}
\setlength{\fboxsep}{0pt}\fcolorbox{gray!10}{gray!10}{\strut
    \mycolorbox[text=\strut{(}, color=green!17.636594]%
    \mycolorbox[text=\strut{m}, color=green!17.636594]%
    \mycolorbox[text=\strut{)(}, color=green!17.636594]%
    \mycolorbox[text=\strut{p}, color=green!17.636594]%
    \mycolorbox[text=\strut{)(}, color=green!17.636594]%
    \mycolorbox[text=\strut{q}, color=green!17.636569]%
    \mycolorbox[text=\strut{/n}, color=green!17.636586]%
    \mycolorbox[text=\strut{)}, color=green!17.500203]%
}
\setlength{\fboxsep}{0pt}\fcolorbox{gray!10}{gray!10}{\strut
    \mycolorbox[text=\strut{for}, color=green!17.636577]%
}
\setlength{\fboxsep}{0pt}\fcolorbox{gray!10}{gray!10}{\strut
    \mycolorbox[text=\strut{simplified}, color=green!17.636594]%
}
\setlength{\fboxsep}{0pt}\fcolorbox{gray!10}{gray!10}{\strut
    \mycolorbox[text=\strut{fractions}, color=green!17.636594]%
}
\setlength{\fboxsep}{0pt}\fcolorbox{gray!10}{gray!10}{\strut
    \mycolorbox[text=\strut{p}, color=green!17.636577]%
    \mycolorbox[text=\strut{/q}, color=green!17.636594]%
    \mycolorbox[text=\strut{."}, color=green!17.451849]%
}
\setlength{\fboxsep}{0pt}\fcolorbox{gray!10}{gray!10}{\strut
    \mycolorbox[text=\strut{Hmm}, color=green!13.966181]%
    \mycolorbox[text=\strut{,}, color=green!14.794839]%
}
\setlength{\fboxsep}{0pt}\fcolorbox{gray!10}{gray!10}{\strut
    \mycolorbox[text=\strut{maybe}]%
}
\setlength{\fboxsep}{0pt}\fcolorbox{gray!10}{gray!10}{\strut
    \mycolorbox[text=\strut{I}, color=green!17.610668]%
}
\setlength{\fboxsep}{0pt}\fcolorbox{gray!10}{gray!10}{\strut
    \mycolorbox[text=\strut{need}, color=green!5.607870]%
}
\setlength{\fboxsep}{0pt}\fcolorbox{gray!10}{gray!10}{\strut
    \mycolorbox[text=\strut{to}, color=green!17.636594]%
}
\setlength{\fboxsep}{0pt}\fcolorbox{gray!10}{gray!10}{\strut
    \mycolorbox[text=\strut{parse}, color=green!9.297385]%
}
\setlength{\fboxsep}{0pt}\fcolorbox{gray!10}{gray!10}{\strut
    \mycolorbox[text=\strut{that}, color=red!6.662666]%
}
\setlength{\fboxsep}{0pt}\fcolorbox{gray!10}{gray!10}{\strut
    \mycolorbox[text=\strut{more}, color=red!5.404330]%
}
\setlength{\fboxsep}{0pt}\fcolorbox{gray!10}{gray!10}{\strut
    \mycolorbox[text=\strut{carefully}, color=green!17.635388]%
    \mycolorbox[text=\strut{.}]%
}
\\
\\
{\tiny\color{gray}3}\,%
\setlength{\fboxsep}{0pt}\fcolorbox{gray!10}{gray!10}{\strut
    \setlength{\fboxsep}{1pt}\fbox{\mycolorbox[text=\strut{Wait}, color=green!8.554914]}%
    \mycolorbox[text=\strut{,}, color=green!17.636577]%
}
\setlength{\fboxsep}{0pt}\fcolorbox{gray!10}{gray!10}{\strut
    \mycolorbox[text=\strut{actually}, color=red!42.162882]%
    \mycolorbox[text=\strut{,}, color=green!17.340766]%
}
\setlength{\fboxsep}{0pt}\fcolorbox{gray!10}{gray!10}{\strut
    \mycolorbox[text=\strut{the}, color=green!9.038353]%
}
\setlength{\fboxsep}{0pt}\fcolorbox{gray!10}{gray!10}{\strut
    \mycolorbox[text=\strut{problem}, color=red!29.007966]%
}
\setlength{\fboxsep}{0pt}\fcolorbox{gray!10}{gray!10}{\strut
    \mycolorbox[text=\strut{says}, color=green!13.379440]%
    \mycolorbox[text=\strut{:}, color=green!13.625766]%
}
\setlength{\fboxsep}{0pt}\fcolorbox{gray!10}{gray!10}{\strut
    \mycolorbox[text=\strut{"}, color=green!14.560559]%
    \mycolorbox[text=\strut{The}, color=green!16.284266]%
}
\setlength{\fboxsep}{0pt}\fcolorbox{gray!10}{gray!10}{\strut
    \mycolorbox[text=\strut{operation}, color=green!17.636594]%
}
\setlength{\fboxsep}{0pt}\fcolorbox{gray!10}{gray!10}{\strut
    \mycolorbox[text=\strut{@}, color=green!17.636594]%
}
\setlength{\fboxsep}{0pt}\fcolorbox{gray!10}{gray!10}{\strut
    \mycolorbox[text=\strut{is}, color=green!17.636594]%
}
\setlength{\fboxsep}{0pt}\fcolorbox{gray!10}{gray!10}{\strut
    \mycolorbox[text=\strut{defined}, color=green!17.636594]%
}
\setlength{\fboxsep}{0pt}\fcolorbox{gray!10}{gray!10}{\strut
    \mycolorbox[text=\strut{as}, color=green!17.636594]%
}
\setlength{\fboxsep}{0pt}\fcolorbox{gray!10}{gray!10}{\strut
    \mycolorbox[text=\strut{(}, color=red!31.877117]%
    \mycolorbox[text=\strut{m}, color=green!17.636534]%
    \mycolorbox[text=\strut{/n}, color=green!17.636577]%
    \mycolorbox[text=\strut{)}, color=green!17.636594]%
}
\setlength{\fboxsep}{0pt}\fcolorbox{gray!10}{gray!10}{\strut
    \mycolorbox[text=\strut{@}]%
}
\setlength{\fboxsep}{0pt}\fcolorbox{gray!10}{gray!10}{\strut
    \mycolorbox[text=\strut{(}, color=green!17.636594]%
    \mycolorbox[text=\strut{p}, color=green!17.636594]%
    \mycolorbox[text=\strut{/q}, color=green!17.636594]%
    \mycolorbox[text=\strut{)}, color=green!17.636594]%
}
\setlength{\fboxsep}{0pt}\fcolorbox{gray!10}{gray!10}{\strut
    \mycolorbox[text=\strut{=}, color=green!17.636594]%
}
\setlength{\fboxsep}{0pt}\fcolorbox{gray!10}{gray!10}{\strut
    \mycolorbox[text=\strut{(}, color=green!17.636026]%
    \mycolorbox[text=\strut{m}, color=green!17.636594]%
    \mycolorbox[text=\strut{)(}, color=green!17.636594]%
    \mycolorbox[text=\strut{p}, color=green!17.636594]%
    \mycolorbox[text=\strut{)(}, color=green!17.636508]%
    \mycolorbox[text=\strut{q}, color=green!17.636551]%
    \mycolorbox[text=\strut{/n}, color=green!17.636534]%
    \mycolorbox[text=\strut{)}, color=green!17.566322]%
}
\setlength{\fboxsep}{0pt}\fcolorbox{gray!10}{gray!10}{\strut
    \mycolorbox[text=\strut{for}, color=green!17.636594]%
}
\setlength{\fboxsep}{0pt}\fcolorbox{gray!10}{gray!10}{\strut
    \mycolorbox[text=\strut{simplified}, color=green!17.636594]%
}
\setlength{\fboxsep}{0pt}\fcolorbox{gray!10}{gray!10}{\strut
    \mycolorbox[text=\strut{fractions}, color=green!17.636594]%
}
\setlength{\fboxsep}{0pt}\fcolorbox{gray!10}{gray!10}{\strut
    \mycolorbox[text=\strut{p}, color=green!17.486917]%
    \mycolorbox[text=\strut{/q}, color=green!17.636594]%
    \mycolorbox[text=\strut{."}, color=green!17.570279]%
}
\setlength{\fboxsep}{0pt}\fcolorbox{gray!10}{gray!10}{\strut
    \setlength{\fboxsep}{1pt}\fbox{\mycolorbox[text=\strut{Wait}, color=green!12.228076]}%
    \mycolorbox[text=\strut{,}, color=green!17.636594]%
}
\setlength{\fboxsep}{0pt}\fcolorbox{gray!10}{gray!10}{\strut
    \mycolorbox[text=\strut{maybe}, color=red!33.283583]%
}
\setlength{\fboxsep}{0pt}\fcolorbox{gray!10}{gray!10}{\strut
    \mycolorbox[text=\strut{there}, color=red!51.899116]%
    \mycolorbox[text=\strut{'s}, color=green!17.254558]%
}
\setlength{\fboxsep}{0pt}\fcolorbox{gray!10}{gray!10}{\strut
    \mycolorbox[text=\strut{a}, color=green!17.185610]%
}
\setlength{\fboxsep}{0pt}\fcolorbox{gray!10}{gray!10}{\strut
    \mycolorbox[text=\strut{typo}, color=green!17.510986]%
}
\setlength{\fboxsep}{0pt}\fcolorbox{gray!10}{gray!10}{\strut
    \mycolorbox[text=\strut{in}, color=red!64.759334]%
}
\setlength{\fboxsep}{0pt}\fcolorbox{gray!10}{gray!10}{\strut
    \mycolorbox[text=\strut{my}]%
}
\setlength{\fboxsep}{0pt}\fcolorbox{gray!10}{gray!10}{\strut
    \mycolorbox[text=\strut{understanding}, color=red!4.965692]%
    \mycolorbox[text=\strut{?}, color=red!16.751945]%
}
\setlength{\fboxsep}{0pt}\fcolorbox{gray!10}{gray!10}{\strut
    \mycolorbox[text=\strut{Let}, color=green!15.209543]%
}
\setlength{\fboxsep}{0pt}\fcolorbox{gray!10}{gray!10}{\strut
    \mycolorbox[text=\strut{me}, color=green!17.636594]%
}
\setlength{\fboxsep}{0pt}\fcolorbox{gray!10}{gray!10}{\strut
    \mycolorbox[text=\strut{check}, color=red!15.023298]%
}
\setlength{\fboxsep}{0pt}\fcolorbox{gray!10}{gray!10}{\strut
    \mycolorbox[text=\strut{again}, color=red!11.455345]%
    \mycolorbox[text=\strut{.}, color=red!28.955129]%
}
\setlength{\fboxsep}{0pt}\fcolorbox{gray!10}{gray!10}{\strut
    \mycolorbox[text=\strut{The}, color=green!15.202783]%
}
\setlength{\fboxsep}{0pt}\fcolorbox{gray!10}{gray!10}{\strut
    \mycolorbox[text=\strut{original}, color=green!12.809325]%
}
\setlength{\fboxsep}{0pt}\fcolorbox{gray!10}{gray!10}{\strut
    \mycolorbox[text=\strut{problem}, color=green!15.690902]%
}
\setlength{\fboxsep}{0pt}\fcolorbox{gray!10}{gray!10}{\strut
    \mycolorbox[text=\strut{says}, color=green!16.162187]%
    \mycolorbox[text=\strut{:}, color=red!67.385919]%
}
\setlength{\fboxsep}{0pt}\fcolorbox{gray!10}{gray!10}{\strut
    \mycolorbox[text=\strut{"}, color=green!17.000623]%
    \mycolorbox[text=\strut{The}, color=green!17.547638]%
}
\setlength{\fboxsep}{0pt}\fcolorbox{gray!10}{gray!10}{\strut
    \mycolorbox[text=\strut{operation}, color=green!17.636594]%
}
\setlength{\fboxsep}{0pt}\fcolorbox{gray!10}{gray!10}{\strut
    \mycolorbox[text=\strut{@}, color=green!17.636594]%
}
\setlength{\fboxsep}{0pt}\fcolorbox{gray!10}{gray!10}{\strut
    \mycolorbox[text=\strut{is}, color=green!17.636594]%
}
\setlength{\fboxsep}{0pt}\fcolorbox{gray!10}{gray!10}{\strut
    \mycolorbox[text=\strut{defined}, color=green!17.636594]%
}
\setlength{\fboxsep}{0pt}\fcolorbox{gray!10}{gray!10}{\strut
    \mycolorbox[text=\strut{as}, color=green!17.636594]%
}
\setlength{\fboxsep}{0pt}\fcolorbox{gray!10}{gray!10}{\strut
    \mycolorbox[text=\strut{m}, color=red!14.610710]%
    \mycolorbox[text=\strut{/n}, color=green!17.636569]%
}
\setlength{\fboxsep}{0pt}\fcolorbox{gray!10}{gray!10}{\strut
    \mycolorbox[text=\strut{@}, color=green!17.636258]%
}
\setlength{\fboxsep}{0pt}\fcolorbox{gray!10}{gray!10}{\strut
    \mycolorbox[text=\strut{p}, color=green!17.636594]%
    \mycolorbox[text=\strut{/q}, color=green!17.636594]%
}
\setlength{\fboxsep}{0pt}\fcolorbox{gray!10}{gray!10}{\strut
    \mycolorbox[text=\strut{=}, color=green!17.636594]%
}
\setlength{\fboxsep}{0pt}\fcolorbox{gray!10}{gray!10}{\strut
    \mycolorbox[text=\strut{(}, color=green!17.636327]%
    \mycolorbox[text=\strut{m}, color=green!17.636594]%
    \mycolorbox[text=\strut{)(}, color=green!17.636594]%
    \mycolorbox[text=\strut{p}, color=green!17.636594]%
    \mycolorbox[text=\strut{)(}, color=green!17.636569]%
    \mycolorbox[text=\strut{q}, color=green!17.636586]%
    \mycolorbox[text=\strut{/n}, color=green!17.636569]%
    \mycolorbox[text=\strut{)}, color=green!17.636577]%
}
\setlength{\fboxsep}{0pt}\fcolorbox{gray!10}{gray!10}{\strut
    \mycolorbox[text=\strut{for}, color=green!17.636594]%
}
\setlength{\fboxsep}{0pt}\fcolorbox{gray!10}{gray!10}{\strut
    \mycolorbox[text=\strut{simplified}, color=green!17.636594]%
}
\setlength{\fboxsep}{0pt}\fcolorbox{gray!10}{gray!10}{\strut
    \mycolorbox[text=\strut{fractions}, color=green!17.636594]%
}
\setlength{\fboxsep}{0pt}\fcolorbox{gray!10}{gray!10}{\strut
    \mycolorbox[text=\strut{p}, color=green!17.551932]%
    \mycolorbox[text=\strut{/q}, color=green!17.636594]%
    \mycolorbox[text=\strut{."}, color=green!16.524148]%
}
\setlength{\fboxsep}{0pt}\fcolorbox{gray!10}{gray!10}{\strut
    \setlength{\fboxsep}{1pt}\fbox{\mycolorbox[text=\strut{Wait}, color=red!41.726674]}%
    \mycolorbox[text=\strut{,}, color=green!17.636594]%
}
\setlength{\fboxsep}{0pt}\fcolorbox{gray!10}{gray!10}{\strut
    \mycolorbox[text=\strut{maybe}, color=red!22.754717]%
}
\setlength{\fboxsep}{0pt}\fcolorbox{gray!10}{gray!10}{\strut
    \mycolorbox[text=\strut{the}, color=green!5.842565]%
}
\setlength{\fboxsep}{0pt}\fcolorbox{gray!10}{gray!10}{\strut
    \mycolorbox[text=\strut{operation}, color=red!67.539731]%
}
\setlength{\fboxsep}{0pt}\fcolorbox{gray!10}{gray!10}{\strut
    \mycolorbox[text=\strut{is}, color=green!17.489462]%
}
\setlength{\fboxsep}{0pt}\fcolorbox{gray!10}{gray!10}{\strut
    \mycolorbox[text=\strut{defined}, color=red!29.633086]%
}
\setlength{\fboxsep}{0pt}\fcolorbox{gray!10}{gray!10}{\strut
    \mycolorbox[text=\strut{between}, color=red!64.189031]%
}
\setlength{\fboxsep}{0pt}\fcolorbox{gray!10}{gray!10}{\strut
    \mycolorbox[text=\strut{two}, color=green!17.494992]%
}
\setlength{\fboxsep}{0pt}\fcolorbox{gray!10}{gray!10}{\strut
    \mycolorbox[text=\strut{fractions}, color=green!17.406321]%
    \mycolorbox[text=\strut{,}, color=green!16.796700]%
}
\setlength{\fboxsep}{0pt}\fcolorbox{gray!10}{gray!10}{\strut
    \mycolorbox[text=\strut{m}]%
    \mycolorbox[text=\strut{/n}, color=green!17.636508]%
}
\setlength{\fboxsep}{0pt}\fcolorbox{gray!10}{gray!10}{\strut
    \mycolorbox[text=\strut{and}, color=green!17.636327]%
}
\setlength{\fboxsep}{0pt}\fcolorbox{gray!10}{gray!10}{\strut
    \mycolorbox[text=\strut{p}, color=green!17.636594]%
    \mycolorbox[text=\strut{/q}, color=green!17.636594]%
    \mycolorbox[text=\strut{,}, color=green!17.417252]%
}
\setlength{\fboxsep}{0pt}\fcolorbox{gray!10}{gray!10}{\strut
    \mycolorbox[text=\strut{and}, color=green!14.676586]%
}
\setlength{\fboxsep}{0pt}\fcolorbox{gray!10}{gray!10}{\strut
    \mycolorbox[text=\strut{the}, color=green!15.990776]%
}
\setlength{\fboxsep}{0pt}\fcolorbox{gray!10}{gray!10}{\strut
    \mycolorbox[text=\strut{result}, color=green!17.631072]%
}
\setlength{\fboxsep}{0pt}\fcolorbox{gray!10}{gray!10}{\strut
    \mycolorbox[text=\strut{is}, color=green!17.636431]%
}
\setlength{\fboxsep}{0pt}\fcolorbox{gray!10}{gray!10}{\strut
    \mycolorbox[text=\strut{m}, color=green!17.148198]%
    \mycolorbox[text=\strut{*p}, color=red!7.630906]%
    \mycolorbox[text=\strut{*(}, color=green!17.617527]%
    \mycolorbox[text=\strut{q}, color=green!17.636594]%
    \mycolorbox[text=\strut{/n}, color=green!17.636508]%
    \mycolorbox[text=\strut{).}, color=green!9.646544]%
}
\setlength{\fboxsep}{0pt}\fcolorbox{gray!10}{gray!10}{\strut
    \mycolorbox[text=\strut{So}, color=red!47.366567]%
    \mycolorbox[text=\strut{,}, color=green!9.849584]%
}
\setlength{\fboxsep}{0pt}\fcolorbox{gray!10}{gray!10}{\strut
    \mycolorbox[text=\strut{given}, color=red!57.044547]%
}
\setlength{\fboxsep}{0pt}\fcolorbox{gray!10}{gray!10}{\strut
    \mycolorbox[text=\strut{two}, color=red!40.841803]%
}
\setlength{\fboxsep}{0pt}\fcolorbox{gray!10}{gray!10}{\strut
    \mycolorbox[text=\strut{fractions}, color=green!13.972717]%
    \mycolorbox[text=\strut{,}, color=green!17.535954]%
}
\setlength{\fboxsep}{0pt}\fcolorbox{gray!10}{gray!10}{\strut
    \mycolorbox[text=\strut{you}, color=red!12.835079]%
}
\setlength{\fboxsep}{0pt}\fcolorbox{gray!10}{gray!10}{\strut
    \mycolorbox[text=\strut{take}, color=red!22.680195]%
}
\setlength{\fboxsep}{0pt}\fcolorbox{gray!10}{gray!10}{\strut
    \mycolorbox[text=\strut{the}, color=green!17.531797]%
}
\setlength{\fboxsep}{0pt}\fcolorbox{gray!10}{gray!10}{\strut
    \mycolorbox[text=\strut{numerator}, color=green!3.394486]%
}
\setlength{\fboxsep}{0pt}\fcolorbox{gray!10}{gray!10}{\strut
    \mycolorbox[text=\strut{of}, color=green!17.636594]%
}
\setlength{\fboxsep}{0pt}\fcolorbox{gray!10}{gray!10}{\strut
    \mycolorbox[text=\strut{the}, color=green!17.636594]%
}
\setlength{\fboxsep}{0pt}\fcolorbox{gray!10}{gray!10}{\strut
    \mycolorbox[text=\strut{first}, color=green!17.636594]%
    \mycolorbox[text=\strut{,}, color=red!16.820144]%
}
\setlength{\fboxsep}{0pt}\fcolorbox{gray!10}{gray!10}{\strut
    \mycolorbox[text=\strut{multiply}, color=green!16.894460]%
}
\setlength{\fboxsep}{0pt}\fcolorbox{gray!10}{gray!10}{\strut
    \mycolorbox[text=\strut{by}, color=green!17.570675]%
}
\setlength{\fboxsep}{0pt}\fcolorbox{gray!10}{gray!10}{\strut
    \mycolorbox[text=\strut{the}, color=green!17.631184]%
}
\setlength{\fboxsep}{0pt}\fcolorbox{gray!10}{gray!10}{\strut
    \mycolorbox[text=\strut{numerator}, color=green!17.636594]%
}
\setlength{\fboxsep}{0pt}\fcolorbox{gray!10}{gray!10}{\strut
    \mycolorbox[text=\strut{of}, color=green!17.636594]%
}
\setlength{\fboxsep}{0pt}\fcolorbox{gray!10}{gray!10}{\strut
    \mycolorbox[text=\strut{the}, color=green!17.636594]%
}
\setlength{\fboxsep}{0pt}\fcolorbox{gray!10}{gray!10}{\strut
    \mycolorbox[text=\strut{second}, color=green!17.636594]%
    \mycolorbox[text=\strut{,}, color=green!17.636594]%
}
\setlength{\fboxsep}{0pt}\fcolorbox{gray!10}{gray!10}{\strut
    \mycolorbox[text=\strut{then}, color=green!10.841168]%
}
\setlength{\fboxsep}{0pt}\fcolorbox{gray!10}{gray!10}{\strut
    \mycolorbox[text=\strut{multiply}, color=green!17.636138]%
}
\setlength{\fboxsep}{0pt}\fcolorbox{gray!10}{gray!10}{\strut
    \mycolorbox[text=\strut{by}, color=green!17.636594]%
}
\setlength{\fboxsep}{0pt}\fcolorbox{gray!10}{gray!10}{\strut
    \mycolorbox[text=\strut{the}, color=red!67.386175]%
}
\setlength{\fboxsep}{0pt}\fcolorbox{gray!10}{gray!10}{\strut
    \mycolorbox[text=\strut{denominator}, color=green!17.558700]%
}
\setlength{\fboxsep}{0pt}\fcolorbox{gray!10}{gray!10}{\strut
    \mycolorbox[text=\strut{of}, color=green!17.636594]%
}
\setlength{\fboxsep}{0pt}\fcolorbox{gray!10}{gray!10}{\strut
    \mycolorbox[text=\strut{the}, color=green!17.636594]%
}
\setlength{\fboxsep}{0pt}\fcolorbox{gray!10}{gray!10}{\strut
    \mycolorbox[text=\strut{first}, color=green!17.621887]%
}
\setlength{\fboxsep}{0pt}\fcolorbox{gray!10}{gray!10}{\strut
    \mycolorbox[text=\strut{divided}, color=green!17.337207]%
}
\setlength{\fboxsep}{0pt}\fcolorbox{gray!10}{gray!10}{\strut
    \mycolorbox[text=\strut{by}, color=green!17.636594]%
}
\setlength{\fboxsep}{0pt}\fcolorbox{gray!10}{gray!10}{\strut
    \mycolorbox[text=\strut{the}, color=green!17.633648]%
}
\setlength{\fboxsep}{0pt}\fcolorbox{gray!10}{gray!10}{\strut
    \mycolorbox[text=\strut{denominator}, color=green!17.611728]%
}
\setlength{\fboxsep}{0pt}\fcolorbox{gray!10}{gray!10}{\strut
    \mycolorbox[text=\strut{of}, color=green!17.636465]%
}
\setlength{\fboxsep}{0pt}\fcolorbox{gray!10}{gray!10}{\strut
    \mycolorbox[text=\strut{the}, color=green!17.636431]%
}
\setlength{\fboxsep}{0pt}\fcolorbox{gray!10}{gray!10}{\strut
    \mycolorbox[text=\strut{second}, color=green!17.585235]%
    \mycolorbox[text=\strut{?}, color=green!16.600427]%
}
\setlength{\fboxsep}{0pt}\fcolorbox{gray!10}{gray!10}{\strut
    \setlength{\fboxsep}{1pt}\fbox{\mycolorbox[text=\strut{Wait}]}%
    \mycolorbox[text=\strut{,}, color=green!17.570658]%
}
\setlength{\fboxsep}{0pt}\fcolorbox{gray!10}{gray!10}{\strut
    \mycolorbox[text=\strut{no}, color=green!10.096214]%
    \mycolorbox[text=\strut{,}, color=red!30.455467]%
}
\setlength{\fboxsep}{0pt}\fcolorbox{gray!10}{gray!10}{\strut
    \mycolorbox[text=\strut{the}, color=red!47.668593]%
}
\setlength{\fboxsep}{0pt}\fcolorbox{gray!10}{gray!10}{\strut
    \mycolorbox[text=\strut{definition}, color=red!58.108516]%
}
\setlength{\fboxsep}{0pt}\fcolorbox{gray!10}{gray!10}{\strut
    \mycolorbox[text=\strut{is}, color=red!10.863311]%
}
\setlength{\fboxsep}{0pt}\fcolorbox{gray!10}{gray!10}{\strut
    \mycolorbox[text=\strut{(}, color=red!38.173874]%
    \mycolorbox[text=\strut{m}, color=green!17.636594]%
    \mycolorbox[text=\strut{)(}, color=red!40.903455]%
    \mycolorbox[text=\strut{p}, color=green!17.636594]%
    \mycolorbox[text=\strut{)(}, color=green!17.636491]%
    \mycolorbox[text=\strut{q}, color=green!17.636586]%
    \mycolorbox[text=\strut{/n}, color=green!17.636026]%
    \mycolorbox[text=\strut{).}, color=green!17.624903]%
}
\setlength{\fboxsep}{0pt}\fcolorbox{gray!10}{gray!10}{\strut
    \mycolorbox[text=\strut{Let}, color=red!42.341074]%
}
\setlength{\fboxsep}{0pt}\fcolorbox{gray!10}{gray!10}{\strut
    \mycolorbox[text=\strut{me}, color=green!17.570675]%
}
\setlength{\fboxsep}{0pt}\fcolorbox{gray!10}{gray!10}{\strut
    \mycolorbox[text=\strut{write}, color=red!12.676579]%
}
\setlength{\fboxsep}{0pt}\fcolorbox{gray!10}{gray!10}{\strut
    \mycolorbox[text=\strut{that}, color=red!25.307104]%
}
\setlength{\fboxsep}{0pt}\fcolorbox{gray!10}{gray!10}{\strut
    \mycolorbox[text=\strut{down}, color=red!7.984962]%
    \mycolorbox[text=\strut{:}, color=red!31.537744]%
}
\\
\\
{\tiny\color{gray}4}\,%
\setlength{\fboxsep}{0pt}\fcolorbox{gray!10}{gray!10}{\strut
    \mycolorbox[text=\strut{If}, color=red!20.772744]%
}
\setlength{\fboxsep}{0pt}\fcolorbox{gray!10}{gray!10}{\strut
    \mycolorbox[text=\strut{we}, color=red!7.104852]%
}
\setlength{\fboxsep}{0pt}\fcolorbox{gray!10}{gray!10}{\strut
    \mycolorbox[text=\strut{have}, color=green!17.636319]%
}
\setlength{\fboxsep}{0pt}\fcolorbox{gray!10}{gray!10}{\strut
    \mycolorbox[text=\strut{two}, color=green!15.403044]%
}
\setlength{\fboxsep}{0pt}\fcolorbox{gray!10}{gray!10}{\strut
    \mycolorbox[text=\strut{fractions}, color=green!17.148492]%
    \mycolorbox[text=\strut{,}, color=green!14.545942]%
}
\setlength{\fboxsep}{0pt}\fcolorbox{gray!10}{gray!10}{\strut
    \mycolorbox[text=\strut{say}, color=red!37.897069]%
}
\setlength{\fboxsep}{0pt}\fcolorbox{gray!10}{gray!10}{\strut
    \mycolorbox[text=\strut{a}, color=red!49.951102]%
    \mycolorbox[text=\strut{/b}, color=green!17.634785]%
}
\setlength{\fboxsep}{0pt}\fcolorbox{gray!10}{gray!10}{\strut
    \mycolorbox[text=\strut{and}, color=green!4.134808]%
}
\setlength{\fboxsep}{0pt}\fcolorbox{gray!10}{gray!10}{\strut
    \mycolorbox[text=\strut{c}, color=green!17.636594]%
    \mycolorbox[text=\strut{/d}, color=green!17.636594]%
    \mycolorbox[text=\strut{,}, color=green!17.621155]%
}
\setlength{\fboxsep}{0pt}\fcolorbox{gray!10}{gray!10}{\strut
    \mycolorbox[text=\strut{then}, color=green!17.636319]%
}
\setlength{\fboxsep}{0pt}\fcolorbox{gray!10}{gray!10}{\strut
    \mycolorbox[text=\strut{a}, color=green!8.230681]%
    \mycolorbox[text=\strut{/b}, color=green!17.636586]%
}
\setlength{\fboxsep}{0pt}\fcolorbox{gray!10}{gray!10}{\strut
    \mycolorbox[text=\strut{@}, color=green!17.636586]%
}
\setlength{\fboxsep}{0pt}\fcolorbox{gray!10}{gray!10}{\strut
    \mycolorbox[text=\strut{c}, color=green!17.636594]%
    \mycolorbox[text=\strut{/d}, color=green!17.636594]%
}
\setlength{\fboxsep}{0pt}\fcolorbox{gray!10}{gray!10}{\strut
    \mycolorbox[text=\strut{=}, color=green!16.915301]%
}
\setlength{\fboxsep}{0pt}\fcolorbox{gray!10}{gray!10}{\strut
    \mycolorbox[text=\strut{a}]%
}
\setlength{\fboxsep}{0pt}\fcolorbox{gray!10}{gray!10}{\strut
    \mycolorbox[text=\strut{*}, color=green!12.906120]%
}
\setlength{\fboxsep}{0pt}\fcolorbox{gray!10}{gray!10}{\strut
    \mycolorbox[text=\strut{c}, color=green!17.636594]%
}
\setlength{\fboxsep}{0pt}\fcolorbox{gray!10}{gray!10}{\strut
    \mycolorbox[text=\strut{*}, color=green!17.636594]%
}
\setlength{\fboxsep}{0pt}\fcolorbox{gray!10}{gray!10}{\strut
    \mycolorbox[text=\strut{(}, color=green!17.636534]%
    \mycolorbox[text=\strut{d}, color=green!17.636594]%
}
\setlength{\fboxsep}{0pt}\fcolorbox{gray!10}{gray!10}{\strut
    \mycolorbox[text=\strut{/}, color=green!4.134798]%
}
\setlength{\fboxsep}{0pt}\fcolorbox{gray!10}{gray!10}{\strut
    \mycolorbox[text=\strut{b}, color=green!17.636594]%
    \mycolorbox[text=\strut{).}]%
}
\setlength{\fboxsep}{0pt}\fcolorbox{gray!10}{gray!10}{\strut
    \setlength{\fboxsep}{1pt}\fbox{\mycolorbox[text=\strut{Wait}, color=red!28.857534]}%
    \mycolorbox[text=\strut{,}, color=green!17.635268]%
}
\setlength{\fboxsep}{0pt}\fcolorbox{gray!10}{gray!10}{\strut
    \mycolorbox[text=\strut{is}, color=red!20.830410]%
}
\setlength{\fboxsep}{0pt}\fcolorbox{gray!10}{gray!10}{\strut
    \mycolorbox[text=\strut{that}, color=green!17.636586]%
}
\setlength{\fboxsep}{0pt}\fcolorbox{gray!10}{gray!10}{\strut
    \mycolorbox[text=\strut{right}, color=red!39.859620]%
    \mycolorbox[text=\strut{?}, color=green!17.256668]%
}
\setlength{\fboxsep}{0pt}\fcolorbox{gray!10}{gray!10}{\strut
    \mycolorbox[text=\strut{Because}, color=red!16.316753]%
}
\setlength{\fboxsep}{0pt}\fcolorbox{gray!10}{gray!10}{\strut
    \mycolorbox[text=\strut{according}, color=red!26.734149]%
}
\setlength{\fboxsep}{0pt}\fcolorbox{gray!10}{gray!10}{\strut
    \mycolorbox[text=\strut{to}, color=green!17.636594]%
}
\setlength{\fboxsep}{0pt}\fcolorbox{gray!10}{gray!10}{\strut
    \mycolorbox[text=\strut{the}, color=green!17.636586]%
}
\setlength{\fboxsep}{0pt}\fcolorbox{gray!10}{gray!10}{\strut
    \mycolorbox[text=\strut{problem}, color=green!15.096260]%
}
\setlength{\fboxsep}{0pt}\fcolorbox{gray!10}{gray!10}{\strut
    \mycolorbox[text=\strut{statement}, color=red!12.855558]%
    \mycolorbox[text=\strut{,}, color=red!10.950910]%
}
\setlength{\fboxsep}{0pt}\fcolorbox{gray!10}{gray!10}{\strut
    \mycolorbox[text=\strut{it}, color=red!25.019406]%
    \mycolorbox[text=\strut{'s}, color=green!16.193097]%
}
\setlength{\fboxsep}{0pt}\fcolorbox{gray!10}{gray!10}{\strut
    \mycolorbox[text=\strut{m}, color=red!34.960370]%
    \mycolorbox[text=\strut{/n}, color=green!17.570201]%
}
\setlength{\fboxsep}{0pt}\fcolorbox{gray!10}{gray!10}{\strut
    \mycolorbox[text=\strut{@}, color=green!17.636594]%
}
\setlength{\fboxsep}{0pt}\fcolorbox{gray!10}{gray!10}{\strut
    \mycolorbox[text=\strut{p}, color=green!17.636594]%
    \mycolorbox[text=\strut{/q}, color=green!17.636594]%
}
\setlength{\fboxsep}{0pt}\fcolorbox{gray!10}{gray!10}{\strut
    \mycolorbox[text=\strut{=}, color=green!17.605196]%
}
\setlength{\fboxsep}{0pt}\fcolorbox{gray!10}{gray!10}{\strut
    \mycolorbox[text=\strut{(}, color=red!40.828280]%
    \mycolorbox[text=\strut{m}, color=green!17.636594]%
    \mycolorbox[text=\strut{)(}, color=green!17.636569]%
    \mycolorbox[text=\strut{p}, color=green!17.636594]%
    \mycolorbox[text=\strut{)(}, color=green!17.636594]%
    \mycolorbox[text=\strut{q}, color=green!17.636594]%
    \mycolorbox[text=\strut{/n}, color=green!17.633312]%
    \mycolorbox[text=\strut{).}, color=green!17.636586]%
}
\setlength{\fboxsep}{0pt}\fcolorbox{gray!10}{gray!10}{\strut
    \mycolorbox[text=\strut{So}, color=green!9.279283]%
}
\setlength{\fboxsep}{0pt}\fcolorbox{gray!10}{gray!10}{\strut
    \mycolorbox[text=\strut{m}, color=red!54.816467]%
}
\setlength{\fboxsep}{0pt}\fcolorbox{gray!10}{gray!10}{\strut
    \mycolorbox[text=\strut{is}, color=green!17.429683]%
}
\setlength{\fboxsep}{0pt}\fcolorbox{gray!10}{gray!10}{\strut
    \mycolorbox[text=\strut{the}, color=green!13.952894]%
}
\setlength{\fboxsep}{0pt}\fcolorbox{gray!10}{gray!10}{\strut
    \mycolorbox[text=\strut{numerator}, color=green!17.636026]%
}
\setlength{\fboxsep}{0pt}\fcolorbox{gray!10}{gray!10}{\strut
    \mycolorbox[text=\strut{of}, color=green!17.636586]%
}
\setlength{\fboxsep}{0pt}\fcolorbox{gray!10}{gray!10}{\strut
    \mycolorbox[text=\strut{the}, color=green!17.636586]%
}
\setlength{\fboxsep}{0pt}\fcolorbox{gray!10}{gray!10}{\strut
    \mycolorbox[text=\strut{first}, color=green!17.636594]%
}
\setlength{\fboxsep}{0pt}\fcolorbox{gray!10}{gray!10}{\strut
    \mycolorbox[text=\strut{fraction}, color=green!17.635647]%
    \mycolorbox[text=\strut{,}, color=green!17.636465]%
}
\setlength{\fboxsep}{0pt}\fcolorbox{gray!10}{gray!10}{\strut
    \mycolorbox[text=\strut{p}, color=red!34.846969]%
}
\setlength{\fboxsep}{0pt}\fcolorbox{gray!10}{gray!10}{\strut
    \mycolorbox[text=\strut{is}, color=green!17.636594]%
}
\setlength{\fboxsep}{0pt}\fcolorbox{gray!10}{gray!10}{\strut
    \mycolorbox[text=\strut{the}, color=green!17.621887]%
}
\setlength{\fboxsep}{0pt}\fcolorbox{gray!10}{gray!10}{\strut
    \mycolorbox[text=\strut{numerator}, color=green!17.636594]%
}
\setlength{\fboxsep}{0pt}\fcolorbox{gray!10}{gray!10}{\strut
    \mycolorbox[text=\strut{of}, color=green!17.636594]%
}
\setlength{\fboxsep}{0pt}\fcolorbox{gray!10}{gray!10}{\strut
    \mycolorbox[text=\strut{the}, color=green!17.636594]%
}
\setlength{\fboxsep}{0pt}\fcolorbox{gray!10}{gray!10}{\strut
    \mycolorbox[text=\strut{second}, color=green!17.636594]%
}
\setlength{\fboxsep}{0pt}\fcolorbox{gray!10}{gray!10}{\strut
    \mycolorbox[text=\strut{fraction}, color=red!5.362822]%
    \mycolorbox[text=\strut{,}, color=green!17.616976]%
}
\setlength{\fboxsep}{0pt}\fcolorbox{gray!10}{gray!10}{\strut
    \mycolorbox[text=\strut{and}, color=green!11.484764]%
}
\setlength{\fboxsep}{0pt}\fcolorbox{gray!10}{gray!10}{\strut
    \mycolorbox[text=\strut{q}, color=green!9.694810]%
}
\setlength{\fboxsep}{0pt}\fcolorbox{gray!10}{gray!10}{\strut
    \mycolorbox[text=\strut{is}, color=red!16.681197]%
}
\setlength{\fboxsep}{0pt}\fcolorbox{gray!10}{gray!10}{\strut
    \mycolorbox[text=\strut{the}, color=green!17.636491]%
}
\setlength{\fboxsep}{0pt}\fcolorbox{gray!10}{gray!10}{\strut
    \mycolorbox[text=\strut{denominator}, color=green!17.636594]%
}
\setlength{\fboxsep}{0pt}\fcolorbox{gray!10}{gray!10}{\strut
    \mycolorbox[text=\strut{of}, color=green!17.636594]%
}
\setlength{\fboxsep}{0pt}\fcolorbox{gray!10}{gray!10}{\strut
    \mycolorbox[text=\strut{the}, color=green!17.636594]%
}
\setlength{\fboxsep}{0pt}\fcolorbox{gray!10}{gray!10}{\strut
    \mycolorbox[text=\strut{second}, color=green!17.636594]%
}
\setlength{\fboxsep}{0pt}\fcolorbox{gray!10}{gray!10}{\strut
    \mycolorbox[text=\strut{fraction}, color=green!17.636586]%
    \mycolorbox[text=\strut{,}, color=green!14.741601]%
}
\setlength{\fboxsep}{0pt}\fcolorbox{gray!10}{gray!10}{\strut
    \mycolorbox[text=\strut{and}, color=green!16.726653]%
}
\setlength{\fboxsep}{0pt}\fcolorbox{gray!10}{gray!10}{\strut
    \mycolorbox[text=\strut{n}, color=green!17.633286]%
}
\setlength{\fboxsep}{0pt}\fcolorbox{gray!10}{gray!10}{\strut
    \mycolorbox[text=\strut{is}, color=green!17.636594]%
}
\setlength{\fboxsep}{0pt}\fcolorbox{gray!10}{gray!10}{\strut
    \mycolorbox[text=\strut{the}, color=green!17.636569]%
}
\setlength{\fboxsep}{0pt}\fcolorbox{gray!10}{gray!10}{\strut
    \mycolorbox[text=\strut{denominator}, color=green!17.636594]%
}
\setlength{\fboxsep}{0pt}\fcolorbox{gray!10}{gray!10}{\strut
    \mycolorbox[text=\strut{of}, color=green!17.636594]%
}
\setlength{\fboxsep}{0pt}\fcolorbox{gray!10}{gray!10}{\strut
    \mycolorbox[text=\strut{the}, color=green!17.636594]%
}
\setlength{\fboxsep}{0pt}\fcolorbox{gray!10}{gray!10}{\strut
    \mycolorbox[text=\strut{first}, color=green!17.636594]%
}
\setlength{\fboxsep}{0pt}\fcolorbox{gray!10}{gray!10}{\strut
    \mycolorbox[text=\strut{fraction}, color=green!17.635897]%
    \mycolorbox[text=\strut{.}, color=green!9.625194]%
}
\setlength{\fboxsep}{0pt}\fcolorbox{gray!10}{gray!10}{\strut
    \mycolorbox[text=\strut{So}, color=red!18.706231]%
}
\setlength{\fboxsep}{0pt}\fcolorbox{gray!10}{gray!10}{\strut
    \mycolorbox[text=\strut{the}, color=red!37.239110]%
}
\setlength{\fboxsep}{0pt}\fcolorbox{gray!10}{gray!10}{\strut
    \mycolorbox[text=\strut{operation}, color=green!5.665409]%
}
\setlength{\fboxsep}{0pt}\fcolorbox{gray!10}{gray!10}{\strut
    \mycolorbox[text=\strut{is}, color=green!12.162299]%
}
\setlength{\fboxsep}{0pt}\fcolorbox{gray!10}{gray!10}{\strut
    \mycolorbox[text=\strut{(}, color=red!52.175133]%
    \mycolorbox[text=\strut{m}, color=green!15.283593]%
}
\setlength{\fboxsep}{0pt}\fcolorbox{gray!10}{gray!10}{\strut
    \mycolorbox[text=\strut{*}, color=red!28.303398]%
}
\setlength{\fboxsep}{0pt}\fcolorbox{gray!10}{gray!10}{\strut
    \mycolorbox[text=\strut{p}, color=green!17.636594]%
}
\setlength{\fboxsep}{0pt}\fcolorbox{gray!10}{gray!10}{\strut
    \mycolorbox[text=\strut{*}, color=green!17.585261]%
}
\setlength{\fboxsep}{0pt}\fcolorbox{gray!10}{gray!10}{\strut
    \mycolorbox[text=\strut{q}, color=green!9.702662]%
    \mycolorbox[text=\strut{)}, color=green!9.696592]%
}
\setlength{\fboxsep}{0pt}\fcolorbox{gray!10}{gray!10}{\strut
    \mycolorbox[text=\strut{/}, color=green!14.795052]%
}
\setlength{\fboxsep}{0pt}\fcolorbox{gray!10}{gray!10}{\strut
    \mycolorbox[text=\strut{n}, color=green!17.636594]%
    \mycolorbox[text=\strut{?}, color=green!14.065426]%
}
\setlength{\fboxsep}{0pt}\fcolorbox{gray!10}{gray!10}{\strut
    \setlength{\fboxsep}{1pt}\fbox{\mycolorbox[text=\strut{Wait}, color=green!16.358868]}%
    \mycolorbox[text=\strut{,}, color=green!17.398792]%
}
\setlength{\fboxsep}{0pt}\fcolorbox{gray!10}{gray!10}{\strut
    \mycolorbox[text=\strut{no}, color=green!12.410752]%
    \mycolorbox[text=\strut{.}, color=red!42.873597]%
}
\setlength{\fboxsep}{0pt}\fcolorbox{gray!10}{gray!10}{\strut
    \setlength{\fboxsep}{1pt}\fbox{\mycolorbox[text=\strut{Wait}, color=green!10.377048]}%
    \mycolorbox[text=\strut{,}, color=green!14.773379]%
}
\setlength{\fboxsep}{0pt}\fcolorbox{gray!10}{gray!10}{\strut
    \mycolorbox[text=\strut{the}, color=red!11.185475]%
}
\setlength{\fboxsep}{0pt}\fcolorbox{gray!10}{gray!10}{\strut
    \mycolorbox[text=\strut{problem}, color=red!20.964346]%
}
\setlength{\fboxsep}{0pt}\fcolorbox{gray!10}{gray!10}{\strut
    \mycolorbox[text=\strut{says}, color=green!17.487970]%
}
\setlength{\fboxsep}{0pt}\fcolorbox{gray!10}{gray!10}{\strut
    \mycolorbox[text=\strut{it}, color=red!64.759690]%
    \mycolorbox[text=\strut{'s}, color=green!15.668919]%
}
\setlength{\fboxsep}{0pt}\fcolorbox{gray!10}{gray!10}{\strut
    \mycolorbox[text=\strut{(}]%
    \mycolorbox[text=\strut{m}, color=green!17.636594]%
    \mycolorbox[text=\strut{)(}, color=green!17.636267]%
    \mycolorbox[text=\strut{p}, color=green!17.636594]%
    \mycolorbox[text=\strut{)(}, color=green!17.636388]%
    \mycolorbox[text=\strut{q}, color=green!17.636586]%
    \mycolorbox[text=\strut{/n}, color=green!17.625135]%
    \mycolorbox[text=\strut{).}, color=green!16.600531]%
}
\setlength{\fboxsep}{0pt}\fcolorbox{gray!10}{gray!10}{\strut
    \mycolorbox[text=\strut{So}, color=green!16.090536]%
}
\setlength{\fboxsep}{0pt}\fcolorbox{gray!10}{gray!10}{\strut
    \mycolorbox[text=\strut{that}, color=green!11.005369]%
}
\setlength{\fboxsep}{0pt}\fcolorbox{gray!10}{gray!10}{\strut
    \mycolorbox[text=\strut{would}]%
}
\setlength{\fboxsep}{0pt}\fcolorbox{gray!10}{gray!10}{\strut
    \mycolorbox[text=\strut{be}, color=green!17.628522]%
}
\setlength{\fboxsep}{0pt}\fcolorbox{gray!10}{gray!10}{\strut
    \mycolorbox[text=\strut{m}, color=green!16.284642]%
    \mycolorbox[text=\strut{*p}, color=red!11.592077]%
    \mycolorbox[text=\strut{*(}, color=green!6.145944]%
    \mycolorbox[text=\strut{q}, color=green!17.636594]%
    \mycolorbox[text=\strut{/n}, color=green!17.601447]%
    \mycolorbox[text=\strut{).}, color=green!12.664270]%
}
\setlength{\fboxsep}{0pt}\fcolorbox{gray!10}{gray!10}{\strut
    \mycolorbox[text=\strut{Which}, color=red!46.707646]%
}
\setlength{\fboxsep}{0pt}\fcolorbox{gray!10}{gray!10}{\strut
    \mycolorbox[text=\strut{is}, color=green!13.407211]%
}
\setlength{\fboxsep}{0pt}\fcolorbox{gray!10}{gray!10}{\strut
    \mycolorbox[text=\strut{(}, color=red!32.412841]%
    \mycolorbox[text=\strut{m}, color=green!17.635044]%
    \mycolorbox[text=\strut{*p}, color=green!13.840326]%
    \mycolorbox[text=\strut{*q}, color=green!17.636594]%
    \mycolorbox[text=\strut{)/}, color=green!17.636491]%
    \mycolorbox[text=\strut{n}, color=green!17.636594]%
    \mycolorbox[text=\strut{.}, color=green!14.860421]%
}
\setlength{\fboxsep}{0pt}\fcolorbox{gray!10}{gray!10}{\strut
    \mycolorbox[text=\strut{So}, color=green!13.196873]%
    \mycolorbox[text=\strut{,}, color=red!53.865024]%
}
\setlength{\fboxsep}{0pt}\fcolorbox{gray!10}{gray!10}{\strut
    \mycolorbox[text=\strut{for}, color=red!76.332742]%
}
\setlength{\fboxsep}{0pt}\fcolorbox{gray!10}{gray!10}{\strut
    \mycolorbox[text=\strut{example}, color=green!17.325577]%
    \mycolorbox[text=\strut{,}, color=green!17.636551]%
}
\setlength{\fboxsep}{0pt}\fcolorbox{gray!10}{gray!10}{\strut
    \mycolorbox[text=\strut{if}, color=green!17.390450]%
}
\setlength{\fboxsep}{0pt}\fcolorbox{gray!10}{gray!10}{\strut
    \mycolorbox[text=\strut{I}, color=red!65.320210]%
}
\setlength{\fboxsep}{0pt}\fcolorbox{gray!10}{gray!10}{\strut
    \mycolorbox[text=\strut{have}, color=green!17.569994]%
}
\setlength{\fboxsep}{0pt}\fcolorbox{gray!10}{gray!10}{\strut
    \mycolorbox[text=\strut{two}, color=red!24.139277]%
}
\setlength{\fboxsep}{0pt}\fcolorbox{gray!10}{gray!10}{\strut
    \mycolorbox[text=\strut{fractions}, color=green!17.633286]%
    \mycolorbox[text=\strut{,}, color=green!15.376547]%
}
\setlength{\fboxsep}{0pt}\fcolorbox{gray!10}{gray!10}{\strut
    \mycolorbox[text=\strut{say}, color=red!67.216020]%
    \mycolorbox[text=\strut{,}, color=red!47.530647]%
}
\setlength{\fboxsep}{0pt}\fcolorbox{gray!10}{gray!10}{\strut
    \mycolorbox[text=\strut{1}, color=red!28.057030]%
    \mycolorbox[text=\strut{/}, color=green!17.636594]%
    \mycolorbox[text=\strut{2}, color=green!17.636594]%
}
\setlength{\fboxsep}{0pt}\fcolorbox{gray!10}{gray!10}{\strut
    \mycolorbox[text=\strut{@}, color=green!13.972726]%
}
\setlength{\fboxsep}{0pt}\fcolorbox{gray!10}{gray!10}{\strut
    \mycolorbox[text=\strut{3}, color=green!17.636594]%
    \mycolorbox[text=\strut{/}, color=green!17.636594]%
    \mycolorbox[text=\strut{4}, color=green!17.636594]%
    \mycolorbox[text=\strut{,}, color=green!17.633975]%
}
\setlength{\fboxsep}{0pt}\fcolorbox{gray!10}{gray!10}{\strut
    \mycolorbox[text=\strut{then}, color=green!9.524926]%
}
\setlength{\fboxsep}{0pt}\fcolorbox{gray!10}{gray!10}{\strut
    \mycolorbox[text=\strut{it}, color=red!49.898073]%
}
\setlength{\fboxsep}{0pt}\fcolorbox{gray!10}{gray!10}{\strut
    \mycolorbox[text=\strut{would}, color=green!17.225310]%
}
\setlength{\fboxsep}{0pt}\fcolorbox{gray!10}{gray!10}{\strut
    \mycolorbox[text=\strut{be}, color=green!17.636543]%
}
\setlength{\fboxsep}{0pt}\fcolorbox{gray!10}{gray!10}{\strut
    \mycolorbox[text=\strut{1}, color=green!17.636594]%
    \mycolorbox[text=\strut{*}, color=green!9.702662]%
    \mycolorbox[text=\strut{3}, color=green!17.636594]%
    \mycolorbox[text=\strut{*(}, color=green!17.636594]%
    \mycolorbox[text=\strut{4}, color=green!17.636594]%
    \mycolorbox[text=\strut{/}, color=green!17.635130]%
    \mycolorbox[text=\strut{2}, color=green!17.636594]%
    \mycolorbox[text=\strut{)}, color=green!17.624765]%
}
\setlength{\fboxsep}{0pt}\fcolorbox{gray!10}{gray!10}{\strut
    \mycolorbox[text=\strut{=}, color=green!17.635388]%
}
\setlength{\fboxsep}{0pt}\fcolorbox{gray!10}{gray!10}{\strut
    \mycolorbox[text=\strut{1}, color=green!17.612348]%
    \mycolorbox[text=\strut{*}, color=green!17.627617]%
    \mycolorbox[text=\strut{3}, color=green!17.636594]%
    \mycolorbox[text=\strut{*}, color=green!17.636517]%
    \mycolorbox[text=\strut{2}, color=green!17.636594]%
}
\setlength{\fboxsep}{0pt}\fcolorbox{gray!10}{gray!10}{\strut
    \mycolorbox[text=\strut{=}, color=green!17.496985]%
}
\setlength{\fboxsep}{0pt}\fcolorbox{gray!10}{gray!10}{\strut
    \mycolorbox[text=\strut{6}, color=green!17.636594]%
    \mycolorbox[text=\strut{.}, color=red!4.588845]%
}
\setlength{\fboxsep}{0pt}\fcolorbox{gray!10}{gray!10}{\strut
    \mycolorbox[text=\strut{Is}, color=red!18.419652]%
}
\setlength{\fboxsep}{0pt}\fcolorbox{gray!10}{gray!10}{\strut
    \mycolorbox[text=\strut{that}, color=green!17.636517]%
}
\setlength{\fboxsep}{0pt}\fcolorbox{gray!10}{gray!10}{\strut
    \mycolorbox[text=\strut{right}, color=red!41.788530]%
    \mycolorbox[text=\strut{?}, color=red!40.828280]%
}
\\
\\
{\tiny\color{gray}5}\,%
\setlength{\fboxsep}{0pt}\fcolorbox{gray!10}{gray!10}{\strut
    \mycolorbox[text=\strut{But}, color=green!11.797703]%
}
\setlength{\fboxsep}{0pt}\fcolorbox{gray!10}{gray!10}{\strut
    \mycolorbox[text=\strut{let}, color=red!11.248387]%
}
\setlength{\fboxsep}{0pt}\fcolorbox{gray!10}{gray!10}{\strut
    \mycolorbox[text=\strut{me}, color=green!17.009447]%
}
\setlength{\fboxsep}{0pt}\fcolorbox{gray!10}{gray!10}{\strut
    \mycolorbox[text=\strut{check}, color=red!5.204822]%
}
\setlength{\fboxsep}{0pt}\fcolorbox{gray!10}{gray!10}{\strut
    \mycolorbox[text=\strut{if}, color=red!38.362014]%
}
\setlength{\fboxsep}{0pt}\fcolorbox{gray!10}{gray!10}{\strut
    \mycolorbox[text=\strut{I}, color=red!72.615853]%
}
\setlength{\fboxsep}{0pt}\fcolorbox{gray!10}{gray!10}{\strut
    \mycolorbox[text=\strut{got}, color=red!70.029592]%
}
\setlength{\fboxsep}{0pt}\fcolorbox{gray!10}{gray!10}{\strut
    \mycolorbox[text=\strut{the}, color=green!16.984221]%
}
\setlength{\fboxsep}{0pt}\fcolorbox{gray!10}{gray!10}{\strut
    \mycolorbox[text=\strut{definition}, color=green!15.019919]%
}
\setlength{\fboxsep}{0pt}\fcolorbox{gray!10}{gray!10}{\strut
    \mycolorbox[text=\strut{correctly}, color=red!16.067608]%
    \mycolorbox[text=\strut{.}, color=green!17.636396]%
}
\setlength{\fboxsep}{0pt}\fcolorbox{gray!10}{gray!10}{\strut
    \mycolorbox[text=\strut{The}, color=green!16.501381]%
}
\setlength{\fboxsep}{0pt}\fcolorbox{gray!10}{gray!10}{\strut
    \mycolorbox[text=\strut{problem}, color=green!16.491668]%
}
\setlength{\fboxsep}{0pt}\fcolorbox{gray!10}{gray!10}{\strut
    \mycolorbox[text=\strut{says}, color=green!15.877891]%
    \mycolorbox[text=\strut{:}]%
}
\setlength{\fboxsep}{0pt}\fcolorbox{gray!10}{gray!10}{\strut
    \mycolorbox[text=\strut{"}, color=green!12.414362]%
    \mycolorbox[text=\strut{The}, color=green!16.797420]%
}
\setlength{\fboxsep}{0pt}\fcolorbox{gray!10}{gray!10}{\strut
    \mycolorbox[text=\strut{operation}, color=green!17.636594]%
}
\setlength{\fboxsep}{0pt}\fcolorbox{gray!10}{gray!10}{\strut
    \mycolorbox[text=\strut{@}, color=green!17.636594]%
}
\setlength{\fboxsep}{0pt}\fcolorbox{gray!10}{gray!10}{\strut
    \mycolorbox[text=\strut{is}, color=green!17.636594]%
}
\setlength{\fboxsep}{0pt}\fcolorbox{gray!10}{gray!10}{\strut
    \mycolorbox[text=\strut{defined}, color=green!17.636594]%
}
\setlength{\fboxsep}{0pt}\fcolorbox{gray!10}{gray!10}{\strut
    \mycolorbox[text=\strut{as}, color=green!17.636594]%
}
\setlength{\fboxsep}{0pt}\fcolorbox{gray!10}{gray!10}{\strut
    \mycolorbox[text=\strut{m}, color=green!15.811885]%
    \mycolorbox[text=\strut{/n}, color=green!17.636594]%
}
\setlength{\fboxsep}{0pt}\fcolorbox{gray!10}{gray!10}{\strut
    \mycolorbox[text=\strut{@}, color=green!17.636594]%
}
\setlength{\fboxsep}{0pt}\fcolorbox{gray!10}{gray!10}{\strut
    \mycolorbox[text=\strut{p}, color=green!17.636594]%
    \mycolorbox[text=\strut{/q}, color=green!17.636594]%
}
\setlength{\fboxsep}{0pt}\fcolorbox{gray!10}{gray!10}{\strut
    \mycolorbox[text=\strut{=}, color=green!17.636594]%
}
\setlength{\fboxsep}{0pt}\fcolorbox{gray!10}{gray!10}{\strut
    \mycolorbox[text=\strut{(}, color=green!17.636551]%
    \mycolorbox[text=\strut{m}, color=green!17.636594]%
    \mycolorbox[text=\strut{)(}, color=green!17.636594]%
    \mycolorbox[text=\strut{p}, color=green!17.636594]%
    \mycolorbox[text=\strut{)(}, color=green!17.636586]%
    \mycolorbox[text=\strut{q}, color=green!17.636594]%
    \mycolorbox[text=\strut{/n}, color=green!17.636517]%
    \mycolorbox[text=\strut{)}, color=green!17.603145]%
}
\setlength{\fboxsep}{0pt}\fcolorbox{gray!10}{gray!10}{\strut
    \mycolorbox[text=\strut{for}, color=green!17.636569]%
}
\setlength{\fboxsep}{0pt}\fcolorbox{gray!10}{gray!10}{\strut
    \mycolorbox[text=\strut{simplified}, color=green!17.636594]%
}
\setlength{\fboxsep}{0pt}\fcolorbox{gray!10}{gray!10}{\strut
    \mycolorbox[text=\strut{fractions}, color=green!17.636594]%
}
\setlength{\fboxsep}{0pt}\fcolorbox{gray!10}{gray!10}{\strut
    \mycolorbox[text=\strut{p}, color=green!17.636327]%
    \mycolorbox[text=\strut{/q}, color=green!17.636594]%
    \mycolorbox[text=\strut{."}, color=green!17.354278]%
}
\setlength{\fboxsep}{0pt}\fcolorbox{gray!10}{gray!10}{\strut
    \setlength{\fboxsep}{1pt}\fbox{\mycolorbox[text=\strut{Wait}, color=red!6.059691]}%
    \mycolorbox[text=\strut{,}, color=green!17.636594]%
}
\setlength{\fboxsep}{0pt}\fcolorbox{gray!10}{gray!10}{\strut
    \mycolorbox[text=\strut{maybe}, color=red!23.703706]%
}
\setlength{\fboxsep}{0pt}\fcolorbox{gray!10}{gray!10}{\strut
    \mycolorbox[text=\strut{the}, color=red!20.037883]%
}
\setlength{\fboxsep}{0pt}\fcolorbox{gray!10}{gray!10}{\strut
    \mycolorbox[text=\strut{second}, color=red!17.625397]%
}
\setlength{\fboxsep}{0pt}\fcolorbox{gray!10}{gray!10}{\strut
    \mycolorbox[text=\strut{fraction}, color=green!17.636353]%
}
\setlength{\fboxsep}{0pt}\fcolorbox{gray!10}{gray!10}{\strut
    \mycolorbox[text=\strut{is}, color=green!17.187332]%
}
\setlength{\fboxsep}{0pt}\fcolorbox{gray!10}{gray!10}{\strut
    \mycolorbox[text=\strut{simplified}, color=red!29.056593]%
    \mycolorbox[text=\strut{?}, color=green!15.855730]%
}
\setlength{\fboxsep}{0pt}\fcolorbox{gray!10}{gray!10}{\strut
    \mycolorbox[text=\strut{So}, color=red!27.337954]%
}
\setlength{\fboxsep}{0pt}\fcolorbox{gray!10}{gray!10}{\strut
    \mycolorbox[text=\strut{maybe}, color=red!12.974928]%
}
\setlength{\fboxsep}{0pt}\fcolorbox{gray!10}{gray!10}{\strut
    \mycolorbox[text=\strut{p}, color=red!30.587068]%
    \mycolorbox[text=\strut{/q}, color=green!17.635862]%
}
\setlength{\fboxsep}{0pt}\fcolorbox{gray!10}{gray!10}{\strut
    \mycolorbox[text=\strut{is}, color=green!17.635793]%
}
\setlength{\fboxsep}{0pt}\fcolorbox{gray!10}{gray!10}{\strut
    \mycolorbox[text=\strut{simplified}, color=green!16.055859]%
    \mycolorbox[text=\strut{,}, color=green!10.996619]%
}
\setlength{\fboxsep}{0pt}\fcolorbox{gray!10}{gray!10}{\strut
    \mycolorbox[text=\strut{but}, color=green!17.597965]%
}
\setlength{\fboxsep}{0pt}\fcolorbox{gray!10}{gray!10}{\strut
    \mycolorbox[text=\strut{the}, color=red!35.148916]%
}
\setlength{\fboxsep}{0pt}\fcolorbox{gray!10}{gray!10}{\strut
    \mycolorbox[text=\strut{first}, color=green!17.636414]%
}
\setlength{\fboxsep}{0pt}\fcolorbox{gray!10}{gray!10}{\strut
    \mycolorbox[text=\strut{fraction}, color=green!9.702202]%
}
\setlength{\fboxsep}{0pt}\fcolorbox{gray!10}{gray!10}{\strut
    \mycolorbox[text=\strut{m}, color=green!11.430010]%
    \mycolorbox[text=\strut{/n}, color=green!17.636594]%
}
\setlength{\fboxsep}{0pt}\fcolorbox{gray!10}{gray!10}{\strut
    \mycolorbox[text=\strut{might}, color=red!9.849107]%
}
\setlength{\fboxsep}{0pt}\fcolorbox{gray!10}{gray!10}{\strut
    \mycolorbox[text=\strut{not}, color=green!17.627505]%
}
\setlength{\fboxsep}{0pt}\fcolorbox{gray!10}{gray!10}{\strut
    \mycolorbox[text=\strut{be}, color=green!15.916652]%
    \mycolorbox[text=\strut{?}, color=green!17.003978]%
}
\setlength{\fboxsep}{0pt}\fcolorbox{gray!10}{gray!10}{\strut
    \mycolorbox[text=\strut{Or}, color=red!78.888874]%
}
\setlength{\fboxsep}{0pt}\fcolorbox{gray!10}{gray!10}{\strut
    \mycolorbox[text=\strut{is}, color=red!17.946671]%
}
\setlength{\fboxsep}{0pt}\fcolorbox{gray!10}{gray!10}{\strut
    \mycolorbox[text=\strut{m}, color=red!72.164276]%
    \mycolorbox[text=\strut{/n}, color=green!17.636594]%
}
\setlength{\fboxsep}{0pt}\fcolorbox{gray!10}{gray!10}{\strut
    \mycolorbox[text=\strut{also}, color=red!13.352260]%
}
\setlength{\fboxsep}{0pt}\fcolorbox{gray!10}{gray!10}{\strut
    \mycolorbox[text=\strut{simplified}, color=green!17.225734]%
    \mycolorbox[text=\strut{?}, color=green!17.148501]%
}
\setlength{\fboxsep}{0pt}\fcolorbox{gray!10}{gray!10}{\strut
    \mycolorbox[text=\strut{The}]%
}
\setlength{\fboxsep}{0pt}\fcolorbox{gray!10}{gray!10}{\strut
    \mycolorbox[text=\strut{problem}, color=green!7.296653]%
}
\setlength{\fboxsep}{0pt}\fcolorbox{gray!10}{gray!10}{\strut
    \mycolorbox[text=\strut{says}, color=green!17.604024]%
}
\setlength{\fboxsep}{0pt}\fcolorbox{gray!10}{gray!10}{\strut
    \mycolorbox[text=\strut{"}, color=green!17.570925]%
    \mycolorbox[text=\strut{for}, color=green!17.621878]%
}
\setlength{\fboxsep}{0pt}\fcolorbox{gray!10}{gray!10}{\strut
    \mycolorbox[text=\strut{simplified}, color=green!17.636594]%
}
\setlength{\fboxsep}{0pt}\fcolorbox{gray!10}{gray!10}{\strut
    \mycolorbox[text=\strut{fractions}, color=green!17.636594]%
}
\setlength{\fboxsep}{0pt}\fcolorbox{gray!10}{gray!10}{\strut
    \mycolorbox[text=\strut{p}, color=green!17.636551]%
    \mycolorbox[text=\strut{/q}, color=green!17.636594]%
    \mycolorbox[text=\strut{".}, color=red!43.041839]%
}
\setlength{\fboxsep}{0pt}\fcolorbox{gray!10}{gray!10}{\strut
    \mycolorbox[text=\strut{So}, color=red!69.581672]%
}
\setlength{\fboxsep}{0pt}\fcolorbox{gray!10}{gray!10}{\strut
    \mycolorbox[text=\strut{maybe}, color=green!15.295806]%
}
\setlength{\fboxsep}{0pt}\fcolorbox{gray!10}{gray!10}{\strut
    \mycolorbox[text=\strut{p}, color=red!28.038230]%
    \mycolorbox[text=\strut{/q}, color=green!17.636586]%
}
\setlength{\fboxsep}{0pt}\fcolorbox{gray!10}{gray!10}{\strut
    \mycolorbox[text=\strut{is}, color=green!17.634682]%
}
\setlength{\fboxsep}{0pt}\fcolorbox{gray!10}{gray!10}{\strut
    \mycolorbox[text=\strut{simplified}, color=green!17.617113]%
    \mycolorbox[text=\strut{,}, color=green!17.567115]%
}
\setlength{\fboxsep}{0pt}\fcolorbox{gray!10}{gray!10}{\strut
    \mycolorbox[text=\strut{but}, color=green!17.617251]%
}
\setlength{\fboxsep}{0pt}\fcolorbox{gray!10}{gray!10}{\strut
    \mycolorbox[text=\strut{m}, color=green!17.185896]%
    \mycolorbox[text=\strut{/n}, color=green!17.636594]%
}
\setlength{\fboxsep}{0pt}\fcolorbox{gray!10}{gray!10}{\strut
    \mycolorbox[text=\strut{might}, color=red!47.843654]%
}
\setlength{\fboxsep}{0pt}\fcolorbox{gray!10}{gray!10}{\strut
    \mycolorbox[text=\strut{not}, color=green!17.252699]%
}
\setlength{\fboxsep}{0pt}\fcolorbox{gray!10}{gray!10}{\strut
    \mycolorbox[text=\strut{be}, color=green!13.839315]%
    \mycolorbox[text=\strut{?}, color=green!17.355289]%
}
\setlength{\fboxsep}{0pt}\fcolorbox{gray!10}{gray!10}{\strut
    \mycolorbox[text=\strut{Hmm}, color=red!67.717997]%
    \mycolorbox[text=\strut{.}, color=red!52.324441]%
}
\setlength{\fboxsep}{0pt}\fcolorbox{gray!10}{gray!10}{\strut
    \mycolorbox[text=\strut{But}, color=red!17.073744]%
}
\setlength{\fboxsep}{0pt}\fcolorbox{gray!10}{gray!10}{\strut
    \mycolorbox[text=\strut{in}, color=red!52.258063]%
}
\setlength{\fboxsep}{0pt}\fcolorbox{gray!10}{gray!10}{\strut
    \mycolorbox[text=\strut{the}, color=green!13.827568]%
}
\setlength{\fboxsep}{0pt}\fcolorbox{gray!10}{gray!10}{\strut
    \mycolorbox[text=\strut{problem}, color=green!15.549844]%
    \mycolorbox[text=\strut{,}, color=green!14.479806]%
}
\setlength{\fboxsep}{0pt}\fcolorbox{gray!10}{gray!10}{\strut
    \mycolorbox[text=\strut{they}, color=red!58.157916]%
}
\setlength{\fboxsep}{0pt}\fcolorbox{gray!10}{gray!10}{\strut
    \mycolorbox[text=\strut{are}, color=red!3.939552]%
}
\setlength{\fboxsep}{0pt}\fcolorbox{gray!10}{gray!10}{\strut
    \mycolorbox[text=\strut{asking}, color=red!35.196202]%
}
\setlength{\fboxsep}{0pt}\fcolorbox{gray!10}{gray!10}{\strut
    \mycolorbox[text=\strut{for}, color=green!7.020752]%
}
\setlength{\fboxsep}{0pt}\fcolorbox{gray!10}{gray!10}{\strut
    \mycolorbox[text=\strut{the}]%
}
\setlength{\fboxsep}{0pt}\fcolorbox{gray!10}{gray!10}{\strut
    \mycolorbox[text=\strut{value}]%
}
\setlength{\fboxsep}{0pt}\fcolorbox{gray!10}{gray!10}{\strut
    \mycolorbox[text=\strut{of}, color=green!17.636586]%
}
\setlength{\fboxsep}{0pt}\fcolorbox{gray!10}{gray!10}{\strut
    \mycolorbox[text=\strut{7}, color=green!17.636594]%
    \mycolorbox[text=\strut{/}, color=green!17.636594]%
    \mycolorbox[text=\strut{3}, color=green!17.636594]%
    \mycolorbox[text=\strut{0}, color=green!17.636594]%
}
\setlength{\fboxsep}{0pt}\fcolorbox{gray!10}{gray!10}{\strut
    \mycolorbox[text=\strut{@}, color=green!17.636594]%
}
\setlength{\fboxsep}{0pt}\fcolorbox{gray!10}{gray!10}{\strut
    \mycolorbox[text=\strut{1}, color=green!17.636594]%
    \mycolorbox[text=\strut{0}, color=green!17.636594]%
    \mycolorbox[text=\strut{/}, color=green!17.636594]%
    \mycolorbox[text=\strut{2}, color=green!17.636594]%
    \mycolorbox[text=\strut{1}, color=green!17.636594]%
    \mycolorbox[text=\strut{.}, color=green!15.891117]%
}
\setlength{\fboxsep}{0pt}\fcolorbox{gray!10}{gray!10}{\strut
    \mycolorbox[text=\strut{So}, color=red!16.224451]%
}
\setlength{\fboxsep}{0pt}\fcolorbox{gray!10}{gray!10}{\strut
    \mycolorbox[text=\strut{both}, color=red!19.679012]%
}
\setlength{\fboxsep}{0pt}\fcolorbox{gray!10}{gray!10}{\strut
    \mycolorbox[text=\strut{fractions}, color=green!6.680994]%
}
\setlength{\fboxsep}{0pt}\fcolorbox{gray!10}{gray!10}{\strut
    \mycolorbox[text=\strut{are}, color=green!15.905226]%
}
\setlength{\fboxsep}{0pt}\fcolorbox{gray!10}{gray!10}{\strut
    \mycolorbox[text=\strut{given}, color=green!16.470627]%
}
\setlength{\fboxsep}{0pt}\fcolorbox{gray!10}{gray!10}{\strut
    \mycolorbox[text=\strut{as}, color=red!8.375093]%
}
\setlength{\fboxsep}{0pt}\fcolorbox{gray!10}{gray!10}{\strut
    \mycolorbox[text=\strut{simplified}, color=green!15.914716]%
}
\setlength{\fboxsep}{0pt}\fcolorbox{gray!10}{gray!10}{\strut
    \mycolorbox[text=\strut{fractions}, color=red!23.531044]%
    \mycolorbox[text=\strut{?}, color=red!7.189080]%
}
\setlength{\fboxsep}{0pt}\fcolorbox{gray!10}{gray!10}{\strut
    \mycolorbox[text=\strut{Let}, color=green!16.059096]%
}
\setlength{\fboxsep}{0pt}\fcolorbox{gray!10}{gray!10}{\strut
    \mycolorbox[text=\strut{me}, color=green!16.722837]%
}
\setlength{\fboxsep}{0pt}\fcolorbox{gray!10}{gray!10}{\strut
    \mycolorbox[text=\strut{check}, color=green!17.633579]%
}
\setlength{\fboxsep}{0pt}\fcolorbox{gray!10}{gray!10}{\strut
    \mycolorbox[text=\strut{if}, color=red!15.947098]%
}
\setlength{\fboxsep}{0pt}\fcolorbox{gray!10}{gray!10}{\strut
    \mycolorbox[text=\strut{7}, color=green!17.636534]%
    \mycolorbox[text=\strut{/}, color=green!17.636577]%
    \mycolorbox[text=\strut{3}, color=green!17.636594]%
    \mycolorbox[text=\strut{0}, color=green!17.636594]%
}
\setlength{\fboxsep}{0pt}\fcolorbox{gray!10}{gray!10}{\strut
    \mycolorbox[text=\strut{and}, color=green!11.513034]%
}
\setlength{\fboxsep}{0pt}\fcolorbox{gray!10}{gray!10}{\strut
    \mycolorbox[text=\strut{1}, color=green!17.636594]%
    \mycolorbox[text=\strut{0}, color=green!17.636594]%
    \mycolorbox[text=\strut{/}, color=green!17.636594]%
    \mycolorbox[text=\strut{2}, color=green!17.636594]%
    \mycolorbox[text=\strut{1}, color=green!17.636594]%
}
\setlength{\fboxsep}{0pt}\fcolorbox{gray!10}{gray!10}{\strut
    \mycolorbox[text=\strut{are}, color=green!17.570141]%
}
\setlength{\fboxsep}{0pt}\fcolorbox{gray!10}{gray!10}{\strut
    \mycolorbox[text=\strut{simplified}, color=green!17.249300]%
    \mycolorbox[text=\strut{.}, color=red!34.856223]%
}
\\
\\
{\tiny\color{gray}6}\,%
\setlength{\fboxsep}{0pt}\fcolorbox{gray!10}{gray!10}{\strut
    \mycolorbox[text=\strut{7}, color=red!25.847631]%
}
\setlength{\fboxsep}{0pt}\fcolorbox{gray!10}{gray!10}{\strut
    \mycolorbox[text=\strut{and}, color=green!6.857505]%
}
\setlength{\fboxsep}{0pt}\fcolorbox{gray!10}{gray!10}{\strut
    \mycolorbox[text=\strut{3}, color=green!17.636594]%
    \mycolorbox[text=\strut{0}, color=green!17.636594]%
}
\setlength{\fboxsep}{0pt}\fcolorbox{gray!10}{gray!10}{\strut
    \mycolorbox[text=\strut{have}, color=red!36.056076]%
}
\setlength{\fboxsep}{0pt}\fcolorbox{gray!10}{gray!10}{\strut
    \mycolorbox[text=\strut{a}, color=red!46.655530]%
}
\setlength{\fboxsep}{0pt}\fcolorbox{gray!10}{gray!10}{\strut
    \mycolorbox[text=\strut{common}, color=red!23.760877]%
}
\setlength{\fboxsep}{0pt}\fcolorbox{gray!10}{gray!10}{\strut
    \mycolorbox[text=\strut{factor}, color=green!15.430546]%
    \mycolorbox[text=\strut{?}, color=green!4.134767]%
}
\setlength{\fboxsep}{0pt}\fcolorbox{gray!10}{gray!10}{\strut
    \mycolorbox[text=\strut{7}, color=green!17.636594]%
}
\setlength{\fboxsep}{0pt}\fcolorbox{gray!10}{gray!10}{\strut
    \mycolorbox[text=\strut{is}, color=green!17.612340]%
}
\setlength{\fboxsep}{0pt}\fcolorbox{gray!10}{gray!10}{\strut
    \mycolorbox[text=\strut{prime}, color=green!17.570667]%
    \mycolorbox[text=\strut{,}]%
}
\setlength{\fboxsep}{0pt}\fcolorbox{gray!10}{gray!10}{\strut
    \mycolorbox[text=\strut{7}, color=red!57.568165]%
}
\setlength{\fboxsep}{0pt}\fcolorbox{gray!10}{gray!10}{\strut
    \mycolorbox[text=\strut{doesn}, color=green!11.426589]%
    \mycolorbox[text=\strut{'t}, color=green!17.636594]%
}
\setlength{\fboxsep}{0pt}\fcolorbox{gray!10}{gray!10}{\strut
    \mycolorbox[text=\strut{divide}, color=green!17.636560]%
}
\setlength{\fboxsep}{0pt}\fcolorbox{gray!10}{gray!10}{\strut
    \mycolorbox[text=\strut{3}, color=green!17.636594]%
    \mycolorbox[text=\strut{0}, color=green!17.636594]%
    \mycolorbox[text=\strut{,}, color=green!9.701836]%
}
\setlength{\fboxsep}{0pt}\fcolorbox{gray!10}{gray!10}{\strut
    \mycolorbox[text=\strut{so}, color=green!17.636534]%
}
\setlength{\fboxsep}{0pt}\fcolorbox{gray!10}{gray!10}{\strut
    \mycolorbox[text=\strut{7}, color=green!17.636594]%
    \mycolorbox[text=\strut{/}, color=green!17.636594]%
    \mycolorbox[text=\strut{3}, color=green!17.636594]%
    \mycolorbox[text=\strut{0}, color=green!17.636594]%
}
\setlength{\fboxsep}{0pt}\fcolorbox{gray!10}{gray!10}{\strut
    \mycolorbox[text=\strut{is}, color=green!17.636586]%
}
\setlength{\fboxsep}{0pt}\fcolorbox{gray!10}{gray!10}{\strut
    \mycolorbox[text=\strut{simplified}, color=green!17.601326]%
    \mycolorbox[text=\strut{.}, color=green!17.636594]%
}
\setlength{\fboxsep}{0pt}\fcolorbox{gray!10}{gray!10}{\strut
    \mycolorbox[text=\strut{1}, color=green!17.636594]%
    \mycolorbox[text=\strut{0}, color=green!17.636594]%
    \mycolorbox[text=\strut{/}, color=red!5.321234]%
    \mycolorbox[text=\strut{2}, color=green!17.636594]%
    \mycolorbox[text=\strut{1}, color=green!17.636594]%
    \mycolorbox[text=\strut{:}, color=green!17.251825]%
}
\setlength{\fboxsep}{0pt}\fcolorbox{gray!10}{gray!10}{\strut
    \mycolorbox[text=\strut{1}, color=green!17.636465]%
    \mycolorbox[text=\strut{0}, color=green!17.636594]%
}
\setlength{\fboxsep}{0pt}\fcolorbox{gray!10}{gray!10}{\strut
    \mycolorbox[text=\strut{and}]%
}
\setlength{\fboxsep}{0pt}\fcolorbox{gray!10}{gray!10}{\strut
    \mycolorbox[text=\strut{2}, color=green!17.636594]%
    \mycolorbox[text=\strut{1}, color=green!17.636594]%
    \mycolorbox[text=\strut{,}, color=red!36.210636]%
}
\setlength{\fboxsep}{0pt}\fcolorbox{gray!10}{gray!10}{\strut
    \mycolorbox[text=\strut{1}, color=green!15.922759]%
    \mycolorbox[text=\strut{0}, color=green!17.636594]%
}
\setlength{\fboxsep}{0pt}\fcolorbox{gray!10}{gray!10}{\strut
    \mycolorbox[text=\strut{factors}, color=red!16.716523]%
}
\setlength{\fboxsep}{0pt}\fcolorbox{gray!10}{gray!10}{\strut
    \mycolorbox[text=\strut{are}, color=green!17.390191]%
}
\setlength{\fboxsep}{0pt}\fcolorbox{gray!10}{gray!10}{\strut
    \mycolorbox[text=\strut{2}, color=green!17.636146]%
}
\setlength{\fboxsep}{0pt}\fcolorbox{gray!10}{gray!10}{\strut
    \mycolorbox[text=\strut{and}, color=green!16.595040]%
}
\setlength{\fboxsep}{0pt}\fcolorbox{gray!10}{gray!10}{\strut
    \mycolorbox[text=\strut{5}, color=green!17.636594]%
    \mycolorbox[text=\strut{,}, color=red!22.807113]%
}
\setlength{\fboxsep}{0pt}\fcolorbox{gray!10}{gray!10}{\strut
    \mycolorbox[text=\strut{2}, color=green!17.636594]%
    \mycolorbox[text=\strut{1}, color=green!17.636594]%
}
\setlength{\fboxsep}{0pt}\fcolorbox{gray!10}{gray!10}{\strut
    \mycolorbox[text=\strut{is}, color=green!17.406321]%
}
\setlength{\fboxsep}{0pt}\fcolorbox{gray!10}{gray!10}{\strut
    \mycolorbox[text=\strut{3}, color=green!17.636594]%
}
\setlength{\fboxsep}{0pt}\fcolorbox{gray!10}{gray!10}{\strut
    \mycolorbox[text=\strut{and}, color=green!17.009438]%
}
\setlength{\fboxsep}{0pt}\fcolorbox{gray!10}{gray!10}{\strut
    \mycolorbox[text=\strut{7}, color=green!17.636594]%
    \mycolorbox[text=\strut{,}, color=red!40.863790]%
}
\setlength{\fboxsep}{0pt}\fcolorbox{gray!10}{gray!10}{\strut
    \mycolorbox[text=\strut{so}, color=green!11.516396]%
}
\setlength{\fboxsep}{0pt}\fcolorbox{gray!10}{gray!10}{\strut
    \mycolorbox[text=\strut{no}, color=green!16.710390]%
}
\setlength{\fboxsep}{0pt}\fcolorbox{gray!10}{gray!10}{\strut
    \mycolorbox[text=\strut{common}, color=green!17.635621]%
}
\setlength{\fboxsep}{0pt}\fcolorbox{gray!10}{gray!10}{\strut
    \mycolorbox[text=\strut{factors}, color=green!17.636138]%
    \mycolorbox[text=\strut{.}, color=green!9.701049]%
}
\setlength{\fboxsep}{0pt}\fcolorbox{gray!10}{gray!10}{\strut
    \mycolorbox[text=\strut{So}, color=green!16.299577]%
}
\setlength{\fboxsep}{0pt}\fcolorbox{gray!10}{gray!10}{\strut
    \mycolorbox[text=\strut{both}, color=green!9.474154]%
}
\setlength{\fboxsep}{0pt}\fcolorbox{gray!10}{gray!10}{\strut
    \mycolorbox[text=\strut{are}, color=red!62.622405]%
}
\setlength{\fboxsep}{0pt}\fcolorbox{gray!10}{gray!10}{\strut
    \mycolorbox[text=\strut{simplified}, color=green!17.627427]%
    \mycolorbox[text=\strut{.}, color=red!16.847090]%
}
\setlength{\fboxsep}{0pt}\fcolorbox{gray!10}{gray!10}{\strut
    \mycolorbox[text=\strut{So}, color=red!16.750696]%
}
\setlength{\fboxsep}{0pt}\fcolorbox{gray!10}{gray!10}{\strut
    \mycolorbox[text=\strut{maybe}, color=red!28.327843]%
}
\setlength{\fboxsep}{0pt}\fcolorbox{gray!10}{gray!10}{\strut
    \mycolorbox[text=\strut{the}, color=green!12.154666]%
}
\setlength{\fboxsep}{0pt}\fcolorbox{gray!10}{gray!10}{\strut
    \mycolorbox[text=\strut{operation}, color=red!29.400221]%
}
\setlength{\fboxsep}{0pt}\fcolorbox{gray!10}{gray!10}{\strut
    \mycolorbox[text=\strut{is}, color=green!16.110869]%
}
\setlength{\fboxsep}{0pt}\fcolorbox{gray!10}{gray!10}{\strut
    \mycolorbox[text=\strut{defined}, color=green!17.166901]%
}
\setlength{\fboxsep}{0pt}\fcolorbox{gray!10}{gray!10}{\strut
    \mycolorbox[text=\strut{for}, color=green!14.699975]%
}
\setlength{\fboxsep}{0pt}\fcolorbox{gray!10}{gray!10}{\strut
    \mycolorbox[text=\strut{simplified}]%
}
\setlength{\fboxsep}{0pt}\fcolorbox{gray!10}{gray!10}{\strut
    \mycolorbox[text=\strut{fractions}, color=green!17.636026]%
    \mycolorbox[text=\strut{?}, color=red!61.341094]%
}
\setlength{\fboxsep}{0pt}\fcolorbox{gray!10}{gray!10}{\strut
    \mycolorbox[text=\strut{Or}, color=red!64.790356]%
}
\setlength{\fboxsep}{0pt}\fcolorbox{gray!10}{gray!10}{\strut
    \mycolorbox[text=\strut{maybe}, color=red!4.757281]%
}
\setlength{\fboxsep}{0pt}\fcolorbox{gray!10}{gray!10}{\strut
    \mycolorbox[text=\strut{just}, color=red!63.325395]%
}
\setlength{\fboxsep}{0pt}\fcolorbox{gray!10}{gray!10}{\strut
    \mycolorbox[text=\strut{that}, color=red!52.151884]%
}
\setlength{\fboxsep}{0pt}\fcolorbox{gray!10}{gray!10}{\strut
    \mycolorbox[text=\strut{p}, color=red!10.895186]%
    \mycolorbox[text=\strut{/q}, color=green!17.636594]%
}
\setlength{\fboxsep}{0pt}\fcolorbox{gray!10}{gray!10}{\strut
    \mycolorbox[text=\strut{is}, color=green!17.635759]%
}
\setlength{\fboxsep}{0pt}\fcolorbox{gray!10}{gray!10}{\strut
    \mycolorbox[text=\strut{simplified}, color=green!17.636517]%
    \mycolorbox[text=\strut{?}, color=red!4.908750]%
}
\setlength{\fboxsep}{0pt}\fcolorbox{gray!10}{gray!10}{\strut
    \mycolorbox[text=\strut{The}, color=red!10.623262]%
}
\setlength{\fboxsep}{0pt}\fcolorbox{gray!10}{gray!10}{\strut
    \mycolorbox[text=\strut{wording}, color=red!34.944926]%
}
\setlength{\fboxsep}{0pt}\fcolorbox{gray!10}{gray!10}{\strut
    \mycolorbox[text=\strut{is}, color=green!11.510600]%
}
\setlength{\fboxsep}{0pt}\fcolorbox{gray!10}{gray!10}{\strut
    \mycolorbox[text=\strut{a}, color=green!11.678142]%
}
\setlength{\fboxsep}{0pt}\fcolorbox{gray!10}{gray!10}{\strut
    \mycolorbox[text=\strut{bit}, color=green!17.585261]%
}
\setlength{\fboxsep}{0pt}\fcolorbox{gray!10}{gray!10}{\strut
    \mycolorbox[text=\strut{confusing}, color=red!13.816119]%
    \mycolorbox[text=\strut{.}, color=red!5.844884]%
}
\setlength{\fboxsep}{0pt}\fcolorbox{gray!10}{gray!10}{\strut
    \mycolorbox[text=\strut{Let}, color=red!62.335556]%
}
\setlength{\fboxsep}{0pt}\fcolorbox{gray!10}{gray!10}{\strut
    \mycolorbox[text=\strut{me}, color=green!17.578416]%
}
\setlength{\fboxsep}{0pt}\fcolorbox{gray!10}{gray!10}{\strut
    \mycolorbox[text=\strut{check}, color=red!9.379235]%
}
\setlength{\fboxsep}{0pt}\fcolorbox{gray!10}{gray!10}{\strut
    \mycolorbox[text=\strut{again}, color=green!9.098112]%
    \mycolorbox[text=\strut{:}, color=red!33.848242]%
}
\\
\\
{\tiny\color{gray}7}\,%
\setlength{\fboxsep}{0pt}\fcolorbox{gray!10}{gray!10}{\strut
    \mycolorbox[text=\strut{"The}, color=green!14.842174]%
}
\setlength{\fboxsep}{0pt}\fcolorbox{gray!10}{gray!10}{\strut
    \mycolorbox[text=\strut{operation}, color=green!17.636594]%
}
\setlength{\fboxsep}{0pt}\fcolorbox{gray!10}{gray!10}{\strut
    \mycolorbox[text=\strut{@}, color=green!17.636594]%
}
\setlength{\fboxsep}{0pt}\fcolorbox{gray!10}{gray!10}{\strut
    \mycolorbox[text=\strut{is}, color=green!17.636594]%
}
\setlength{\fboxsep}{0pt}\fcolorbox{gray!10}{gray!10}{\strut
    \mycolorbox[text=\strut{defined}, color=green!17.636594]%
}
\setlength{\fboxsep}{0pt}\fcolorbox{gray!10}{gray!10}{\strut
    \mycolorbox[text=\strut{as}, color=green!17.636594]%
}
\setlength{\fboxsep}{0pt}\fcolorbox{gray!10}{gray!10}{\strut
    \mycolorbox[text=\strut{m}, color=green!17.391063]%
    \mycolorbox[text=\strut{/n}, color=green!17.636594]%
}
\setlength{\fboxsep}{0pt}\fcolorbox{gray!10}{gray!10}{\strut
    \mycolorbox[text=\strut{@}, color=green!17.636594]%
}
\setlength{\fboxsep}{0pt}\fcolorbox{gray!10}{gray!10}{\strut
    \mycolorbox[text=\strut{p}, color=green!17.636594]%
    \mycolorbox[text=\strut{/q}, color=green!17.636594]%
}
\setlength{\fboxsep}{0pt}\fcolorbox{gray!10}{gray!10}{\strut
    \mycolorbox[text=\strut{=}, color=green!17.636594]%
}
\setlength{\fboxsep}{0pt}\fcolorbox{gray!10}{gray!10}{\strut
    \mycolorbox[text=\strut{(}, color=green!17.636586]%
    \mycolorbox[text=\strut{m}, color=green!17.636594]%
    \mycolorbox[text=\strut{)(}, color=green!17.636594]%
    \mycolorbox[text=\strut{p}, color=green!17.636594]%
    \mycolorbox[text=\strut{)(}, color=green!17.636594]%
    \mycolorbox[text=\strut{q}, color=green!17.636594]%
    \mycolorbox[text=\strut{/n}, color=green!17.636586]%
    \mycolorbox[text=\strut{)}, color=green!17.636594]%
}
\setlength{\fboxsep}{0pt}\fcolorbox{gray!10}{gray!10}{\strut
    \mycolorbox[text=\strut{for}, color=green!17.636594]%
}
\setlength{\fboxsep}{0pt}\fcolorbox{gray!10}{gray!10}{\strut
    \mycolorbox[text=\strut{simplified}, color=green!17.636594]%
}
\setlength{\fboxsep}{0pt}\fcolorbox{gray!10}{gray!10}{\strut
    \mycolorbox[text=\strut{fractions}, color=green!17.636594]%
}
\setlength{\fboxsep}{0pt}\fcolorbox{gray!10}{gray!10}{\strut
    \mycolorbox[text=\strut{p}, color=green!17.636594]%
    \mycolorbox[text=\strut{/q}, color=green!17.636594]%
    \mycolorbox[text=\strut{."}, color=red!16.750560]%
}
\\
\\
{\tiny\color{gray}8}\,%
\setlength{\fboxsep}{0pt}\fcolorbox{gray!10}{gray!10}{\strut
    \mycolorbox[text=\strut{So}]%
}
\setlength{\fboxsep}{0pt}\fcolorbox{gray!10}{gray!10}{\strut
    \mycolorbox[text=\strut{maybe}, color=red!6.367917]%
}
\setlength{\fboxsep}{0pt}\fcolorbox{gray!10}{gray!10}{\strut
    \mycolorbox[text=\strut{the}, color=green!3.160818]%
}
\setlength{\fboxsep}{0pt}\fcolorbox{gray!10}{gray!10}{\strut
    \mycolorbox[text=\strut{operation}, color=red!38.190474]%
}
\setlength{\fboxsep}{0pt}\fcolorbox{gray!10}{gray!10}{\strut
    \mycolorbox[text=\strut{is}, color=green!17.489307]%
}
\setlength{\fboxsep}{0pt}\fcolorbox{gray!10}{gray!10}{\strut
    \mycolorbox[text=\strut{defined}, color=green!17.548785]%
}
\setlength{\fboxsep}{0pt}\fcolorbox{gray!10}{gray!10}{\strut
    \mycolorbox[text=\strut{for}, color=red!18.894489]%
}
\setlength{\fboxsep}{0pt}\fcolorbox{gray!10}{gray!10}{\strut
    \mycolorbox[text=\strut{simplified}, color=red!54.817038]%
}
\setlength{\fboxsep}{0pt}\fcolorbox{gray!10}{gray!10}{\strut
    \mycolorbox[text=\strut{fractions}, color=green!17.605316]%
}
\setlength{\fboxsep}{0pt}\fcolorbox{gray!10}{gray!10}{\strut
    \mycolorbox[text=\strut{p}, color=green!17.126954]%
    \mycolorbox[text=\strut{/q}, color=green!17.636594]%
    \mycolorbox[text=\strut{,}, color=red!24.244229]%
}
\setlength{\fboxsep}{0pt}\fcolorbox{gray!10}{gray!10}{\strut
    \mycolorbox[text=\strut{but}, color=green!12.210552]%
}
\setlength{\fboxsep}{0pt}\fcolorbox{gray!10}{gray!10}{\strut
    \mycolorbox[text=\strut{m}, color=green!14.152975]%
    \mycolorbox[text=\strut{/n}, color=green!17.636594]%
}
\setlength{\fboxsep}{0pt}\fcolorbox{gray!10}{gray!10}{\strut
    \mycolorbox[text=\strut{could}, color=red!33.959124]%
}
\setlength{\fboxsep}{0pt}\fcolorbox{gray!10}{gray!10}{\strut
    \mycolorbox[text=\strut{be}, color=green!17.636543]%
}
\setlength{\fboxsep}{0pt}\fcolorbox{gray!10}{gray!10}{\strut
    \mycolorbox[text=\strut{any}, color=green!12.830789]%
}
\setlength{\fboxsep}{0pt}\fcolorbox{gray!10}{gray!10}{\strut
    \mycolorbox[text=\strut{fraction}, color=green!17.542990]%
    \mycolorbox[text=\strut{?}, color=green!17.591932]%
}
\setlength{\fboxsep}{0pt}\fcolorbox{gray!10}{gray!10}{\strut
    \mycolorbox[text=\strut{Or}]%
}
\setlength{\fboxsep}{0pt}\fcolorbox{gray!10}{gray!10}{\strut
    \mycolorbox[text=\strut{maybe}, color=red!24.076381]%
}
\setlength{\fboxsep}{0pt}\fcolorbox{gray!10}{gray!10}{\strut
    \mycolorbox[text=\strut{both}, color=green!15.775136]%
}
\setlength{\fboxsep}{0pt}\fcolorbox{gray!10}{gray!10}{\strut
    \mycolorbox[text=\strut{are}, color=red!47.102216]%
}
\setlength{\fboxsep}{0pt}\fcolorbox{gray!10}{gray!10}{\strut
    \mycolorbox[text=\strut{simplified}, color=green!17.610788]%
    \mycolorbox[text=\strut{?}, color=green!17.157286]%
}
\setlength{\fboxsep}{0pt}\fcolorbox{gray!10}{gray!10}{\strut
    \mycolorbox[text=\strut{The}, color=red!9.315757]%
}
\setlength{\fboxsep}{0pt}\fcolorbox{gray!10}{gray!10}{\strut
    \mycolorbox[text=\strut{wording}, color=green!6.719537]%
}
\setlength{\fboxsep}{0pt}\fcolorbox{gray!10}{gray!10}{\strut
    \mycolorbox[text=\strut{is}, color=green!15.408337]%
}
\setlength{\fboxsep}{0pt}\fcolorbox{gray!10}{gray!10}{\strut
    \mycolorbox[text=\strut{a}, color=red!22.647272]%
}
\setlength{\fboxsep}{0pt}\fcolorbox{gray!10}{gray!10}{\strut
    \mycolorbox[text=\strut{bit}, color=green!17.585252]%
}
\setlength{\fboxsep}{0pt}\fcolorbox{gray!10}{gray!10}{\strut
    \mycolorbox[text=\strut{unclear}, color=red!16.728005]%
    \mycolorbox[text=\strut{.}, color=red!22.722477]%
}
\setlength{\fboxsep}{0pt}\fcolorbox{gray!10}{gray!10}{\strut
    \mycolorbox[text=\strut{But}, color=red!17.999315]%
}
\setlength{\fboxsep}{0pt}\fcolorbox{gray!10}{gray!10}{\strut
    \mycolorbox[text=\strut{since}, color=red!28.216198]%
}
\setlength{\fboxsep}{0pt}\fcolorbox{gray!10}{gray!10}{\strut
    \mycolorbox[text=\strut{in}, color=red!42.342559]%
}
\setlength{\fboxsep}{0pt}\fcolorbox{gray!10}{gray!10}{\strut
    \mycolorbox[text=\strut{the}, color=green!17.565382]%
}
\setlength{\fboxsep}{0pt}\fcolorbox{gray!10}{gray!10}{\strut
    \mycolorbox[text=\strut{problem}, color=green!17.083189]%
    \mycolorbox[text=\strut{,}, color=red!23.038193]%
}
\setlength{\fboxsep}{0pt}\fcolorbox{gray!10}{gray!10}{\strut
    \mycolorbox[text=\strut{they}, color=red!20.211754]%
}
\setlength{\fboxsep}{0pt}\fcolorbox{gray!10}{gray!10}{\strut
    \mycolorbox[text=\strut{are}, color=green!9.554308]%
}
\setlength{\fboxsep}{0pt}\fcolorbox{gray!10}{gray!10}{\strut
    \mycolorbox[text=\strut{giving}, color=green!13.676471]%
}
\setlength{\fboxsep}{0pt}\fcolorbox{gray!10}{gray!10}{\strut
    \mycolorbox[text=\strut{us}, color=red!25.994796]%
}
\setlength{\fboxsep}{0pt}\fcolorbox{gray!10}{gray!10}{\strut
    \mycolorbox[text=\strut{two}, color=red!12.875075]%
}
\setlength{\fboxsep}{0pt}\fcolorbox{gray!10}{gray!10}{\strut
    \mycolorbox[text=\strut{simplified}, color=red!10.864454]%
}
\setlength{\fboxsep}{0pt}\fcolorbox{gray!10}{gray!10}{\strut
    \mycolorbox[text=\strut{fractions}, color=green!17.636594]%
    \mycolorbox[text=\strut{,}, color=green!17.519042]%
}
\setlength{\fboxsep}{0pt}\fcolorbox{gray!10}{gray!10}{\strut
    \mycolorbox[text=\strut{maybe}, color=red!24.757380]%
}
\setlength{\fboxsep}{0pt}\fcolorbox{gray!10}{gray!10}{\strut
    \mycolorbox[text=\strut{we}, color=green!11.183239]%
}
\setlength{\fboxsep}{0pt}\fcolorbox{gray!10}{gray!10}{\strut
    \mycolorbox[text=\strut{don}, color=red!57.782994]%
    \mycolorbox[text=\strut{'t}, color=green!17.636586]%
}
\setlength{\fboxsep}{0pt}\fcolorbox{gray!10}{gray!10}{\strut
    \mycolorbox[text=\strut{have}, color=red!16.676471]%
}
\setlength{\fboxsep}{0pt}\fcolorbox{gray!10}{gray!10}{\strut
    \mycolorbox[text=\strut{to}, color=green!17.636586]%
}
\setlength{\fboxsep}{0pt}\fcolorbox{gray!10}{gray!10}{\strut
    \mycolorbox[text=\strut{worry}, color=green!17.636103]%
}
\setlength{\fboxsep}{0pt}\fcolorbox{gray!10}{gray!10}{\strut
    \mycolorbox[text=\strut{about}, color=green!17.636474]%
}
\setlength{\fboxsep}{0pt}\fcolorbox{gray!10}{gray!10}{\strut
    \mycolorbox[text=\strut{that}, color=red!34.887267]%
    \mycolorbox[text=\strut{.}, color=green!4.336063]%
}
\setlength{\fboxsep}{0pt}\fcolorbox{gray!10}{gray!10}{\strut
    \mycolorbox[text=\strut{So}, color=red!7.027882]%
}
\setlength{\fboxsep}{0pt}\fcolorbox{gray!10}{gray!10}{\strut
    \mycolorbox[text=\strut{regardless}, color=red!60.200027]%
    \mycolorbox[text=\strut{,}, color=green!17.605463]%
}
\setlength{\fboxsep}{0pt}\fcolorbox{gray!10}{gray!10}{\strut
    \mycolorbox[text=\strut{we}, color=red!36.776595]%
}
\setlength{\fboxsep}{0pt}\fcolorbox{gray!10}{gray!10}{\strut
    \mycolorbox[text=\strut{can}, color=green!15.618865]%
}
\setlength{\fboxsep}{0pt}\fcolorbox{gray!10}{gray!10}{\strut
    \mycolorbox[text=\strut{proceed}, color=red!60.408610]%
}
\setlength{\fboxsep}{0pt}\fcolorbox{gray!10}{gray!10}{\strut
    \mycolorbox[text=\strut{with}, color=green!9.674335]%
}
\setlength{\fboxsep}{0pt}\fcolorbox{gray!10}{gray!10}{\strut
    \mycolorbox[text=\strut{the}, color=green!16.951610]%
}
\setlength{\fboxsep}{0pt}\fcolorbox{gray!10}{gray!10}{\strut
    \mycolorbox[text=\strut{definition}, color=red!17.309908]%
}
\setlength{\fboxsep}{0pt}\fcolorbox{gray!10}{gray!10}{\strut
    \mycolorbox[text=\strut{given}, color=red!41.152161]%
    \mycolorbox[text=\strut{.}, color=green!16.497078]%
}
\\
\\
{\tiny\color{gray}9}\,%
\setlength{\fboxsep}{0pt}\fcolorbox{gray!10}{gray!10}{\strut
    \mycolorbox[text=\strut{So}, color=green!15.420323]%
}
\setlength{\fboxsep}{0pt}\fcolorbox{gray!10}{gray!10}{\strut
    \mycolorbox[text=\strut{given}, color=red!68.176980]%
}
\setlength{\fboxsep}{0pt}\fcolorbox{gray!10}{gray!10}{\strut
    \mycolorbox[text=\strut{that}, color=green!17.466304]%
    \mycolorbox[text=\strut{,}, color=green!17.573011]%
}
\setlength{\fboxsep}{0pt}\fcolorbox{gray!10}{gray!10}{\strut
    \mycolorbox[text=\strut{we}, color=red!69.886493]%
}
\setlength{\fboxsep}{0pt}\fcolorbox{gray!10}{gray!10}{\strut
    \mycolorbox[text=\strut{need}]%
}
\setlength{\fboxsep}{0pt}\fcolorbox{gray!10}{gray!10}{\strut
    \mycolorbox[text=\strut{to}, color=green!17.636594]%
}
\setlength{\fboxsep}{0pt}\fcolorbox{gray!10}{gray!10}{\strut
    \mycolorbox[text=\strut{compute}, color=green!17.256521]%
}
\setlength{\fboxsep}{0pt}\fcolorbox{gray!10}{gray!10}{\strut
    \mycolorbox[text=\strut{7}, color=green!17.636594]%
    \mycolorbox[text=\strut{/}, color=green!17.636594]%
    \mycolorbox[text=\strut{3}, color=green!17.636594]%
    \mycolorbox[text=\strut{0}, color=green!17.636594]%
}
\setlength{\fboxsep}{0pt}\fcolorbox{gray!10}{gray!10}{\strut
    \mycolorbox[text=\strut{@}, color=green!17.636594]%
}
\setlength{\fboxsep}{0pt}\fcolorbox{gray!10}{gray!10}{\strut
    \mycolorbox[text=\strut{1}, color=green!17.636594]%
    \mycolorbox[text=\strut{0}, color=green!17.636594]%
    \mycolorbox[text=\strut{/}, color=green!17.636594]%
    \mycolorbox[text=\strut{2}, color=green!17.636594]%
    \mycolorbox[text=\strut{1}, color=green!17.636594]%
    \mycolorbox[text=\strut{.}, color=green!7.302467]%
}
\setlength{\fboxsep}{0pt}\fcolorbox{gray!10}{gray!10}{\strut
    \mycolorbox[text=\strut{According}, color=red!39.780234]%
}
\setlength{\fboxsep}{0pt}\fcolorbox{gray!10}{gray!10}{\strut
    \mycolorbox[text=\strut{to}, color=green!17.636594]%
}
\setlength{\fboxsep}{0pt}\fcolorbox{gray!10}{gray!10}{\strut
    \mycolorbox[text=\strut{the}, color=green!17.636569]%
}
\setlength{\fboxsep}{0pt}\fcolorbox{gray!10}{gray!10}{\strut
    \mycolorbox[text=\strut{definition}, color=green!16.263549]%
    \mycolorbox[text=\strut{,}, color=green!17.570529]%
}
\setlength{\fboxsep}{0pt}\fcolorbox{gray!10}{gray!10}{\strut
    \mycolorbox[text=\strut{this}, color=red!74.413457]%
}
\setlength{\fboxsep}{0pt}\fcolorbox{gray!10}{gray!10}{\strut
    \mycolorbox[text=\strut{is}, color=red!52.289191]%
}
\setlength{\fboxsep}{0pt}\fcolorbox{gray!10}{gray!10}{\strut
    \mycolorbox[text=\strut{m}]%
    \mycolorbox[text=\strut{/n}, color=green!12.068298]%
}
\setlength{\fboxsep}{0pt}\fcolorbox{gray!10}{gray!10}{\strut
    \mycolorbox[text=\strut{@}, color=green!17.632097]%
}
\setlength{\fboxsep}{0pt}\fcolorbox{gray!10}{gray!10}{\strut
    \mycolorbox[text=\strut{p}, color=green!17.636594]%
    \mycolorbox[text=\strut{/q}, color=green!17.636594]%
}
\setlength{\fboxsep}{0pt}\fcolorbox{gray!10}{gray!10}{\strut
    \mycolorbox[text=\strut{where}, color=green!3.159064]%
}
\setlength{\fboxsep}{0pt}\fcolorbox{gray!10}{gray!10}{\strut
    \mycolorbox[text=\strut{m}, color=green!17.636594]%
}
\setlength{\fboxsep}{0pt}\fcolorbox{gray!10}{gray!10}{\strut
    \mycolorbox[text=\strut{=}, color=red!81.087259]%
}
\setlength{\fboxsep}{0pt}\fcolorbox{gray!10}{gray!10}{\strut
    \mycolorbox[text=\strut{7}, color=green!17.636594]%
    \mycolorbox[text=\strut{,}, color=green!17.636594]%
}
\setlength{\fboxsep}{0pt}\fcolorbox{gray!10}{gray!10}{\strut
    \mycolorbox[text=\strut{n}, color=green!17.636594]%
}
\setlength{\fboxsep}{0pt}\fcolorbox{gray!10}{gray!10}{\strut
    \mycolorbox[text=\strut{=}, color=green!17.635044]%
}
\setlength{\fboxsep}{0pt}\fcolorbox{gray!10}{gray!10}{\strut
    \mycolorbox[text=\strut{3}, color=green!17.636594]%
    \mycolorbox[text=\strut{0}, color=green!17.636594]%
    \mycolorbox[text=\strut{,}, color=green!17.636594]%
}
\setlength{\fboxsep}{0pt}\fcolorbox{gray!10}{gray!10}{\strut
    \mycolorbox[text=\strut{p}, color=green!17.636551]%
}
\setlength{\fboxsep}{0pt}\fcolorbox{gray!10}{gray!10}{\strut
    \mycolorbox[text=\strut{=}, color=green!17.636586]%
}
\setlength{\fboxsep}{0pt}\fcolorbox{gray!10}{gray!10}{\strut
    \mycolorbox[text=\strut{1}, color=green!17.636594]%
    \mycolorbox[text=\strut{0}, color=green!17.636594]%
    \mycolorbox[text=\strut{,}, color=green!17.636594]%
}
\setlength{\fboxsep}{0pt}\fcolorbox{gray!10}{gray!10}{\strut
    \mycolorbox[text=\strut{q}, color=green!17.621887]%
}
\setlength{\fboxsep}{0pt}\fcolorbox{gray!10}{gray!10}{\strut
    \mycolorbox[text=\strut{=}, color=green!17.636594]%
}
\setlength{\fboxsep}{0pt}\fcolorbox{gray!10}{gray!10}{\strut
    \mycolorbox[text=\strut{2}, color=green!17.636594]%
    \mycolorbox[text=\strut{1}, color=green!17.636594]%
    \mycolorbox[text=\strut{.}, color=green!15.430617]%
}
\setlength{\fboxsep}{0pt}\fcolorbox{gray!10}{gray!10}{\strut
    \mycolorbox[text=\strut{Therefore}, color=red!8.324497]%
    \mycolorbox[text=\strut{,}, color=green!17.605902]%
}
\setlength{\fboxsep}{0pt}\fcolorbox{gray!10}{gray!10}{\strut
    \mycolorbox[text=\strut{the}, color=red!30.211083]%
}
\setlength{\fboxsep}{0pt}\fcolorbox{gray!10}{gray!10}{\strut
    \mycolorbox[text=\strut{result}, color=red!40.887374]%
}
\setlength{\fboxsep}{0pt}\fcolorbox{gray!10}{gray!10}{\strut
    \mycolorbox[text=\strut{is}]%
}
\setlength{\fboxsep}{0pt}\fcolorbox{gray!10}{gray!10}{\strut
    \mycolorbox[text=\strut{m}, color=green!9.676118]%
    \mycolorbox[text=\strut{*p}, color=red!10.857613]%
    \mycolorbox[text=\strut{*(}, color=green!17.636026]%
    \mycolorbox[text=\strut{q}, color=green!17.636594]%
    \mycolorbox[text=\strut{/n}, color=green!17.627677]%
    \mycolorbox[text=\strut{)}]%
}
\setlength{\fboxsep}{0pt}\fcolorbox{gray!10}{gray!10}{\strut
    \mycolorbox[text=\strut{=}, color=green!17.457277]%
}
\setlength{\fboxsep}{0pt}\fcolorbox{gray!10}{gray!10}{\strut
    \mycolorbox[text=\strut{7}, color=green!17.636594]%
}
\setlength{\fboxsep}{0pt}\fcolorbox{gray!10}{gray!10}{\strut
    \mycolorbox[text=\strut{*}, color=red!28.769059]%
}
\setlength{\fboxsep}{0pt}\fcolorbox{gray!10}{gray!10}{\strut
    \mycolorbox[text=\strut{1}, color=green!17.636594]%
    \mycolorbox[text=\strut{0}, color=green!17.636594]%
}
\setlength{\fboxsep}{0pt}\fcolorbox{gray!10}{gray!10}{\strut
    \mycolorbox[text=\strut{*}, color=green!17.636594]%
}
\setlength{\fboxsep}{0pt}\fcolorbox{gray!10}{gray!10}{\strut
    \mycolorbox[text=\strut{(}, color=green!17.636594]%
    \mycolorbox[text=\strut{2}, color=green!17.636594]%
    \mycolorbox[text=\strut{1}, color=green!17.636594]%
}
\setlength{\fboxsep}{0pt}\fcolorbox{gray!10}{gray!10}{\strut
    \mycolorbox[text=\strut{/}, color=green!17.256668]%
}
\setlength{\fboxsep}{0pt}\fcolorbox{gray!10}{gray!10}{\strut
    \mycolorbox[text=\strut{3}, color=green!17.636594]%
    \mycolorbox[text=\strut{0}, color=green!17.636594]%
    \mycolorbox[text=\strut{).}, color=green!13.972717]%
}
\setlength{\fboxsep}{0pt}\fcolorbox{gray!10}{gray!10}{\strut
    \mycolorbox[text=\strut{Let}, color=green!16.403503]%
}
\setlength{\fboxsep}{0pt}\fcolorbox{gray!10}{gray!10}{\strut
    \mycolorbox[text=\strut{me}, color=green!16.461809]%
}
\setlength{\fboxsep}{0pt}\fcolorbox{gray!10}{gray!10}{\strut
    \mycolorbox[text=\strut{compute}, color=green!16.303574]%
}
\setlength{\fboxsep}{0pt}\fcolorbox{gray!10}{gray!10}{\strut
    \mycolorbox[text=\strut{that}, color=green!15.430634]%
}
\setlength{\fboxsep}{0pt}\fcolorbox{gray!10}{gray!10}{\strut
    \mycolorbox[text=\strut{step}, color=red!28.847106]%
}
\setlength{\fboxsep}{0pt}\fcolorbox{gray!10}{gray!10}{\strut
    \mycolorbox[text=\strut{by}, color=green!17.636577]%
}
\setlength{\fboxsep}{0pt}\fcolorbox{gray!10}{gray!10}{\strut
    \mycolorbox[text=\strut{step}, color=green!17.636594]%
    \mycolorbox[text=\strut{.}, color=green!17.617708]%
}
\\
\\
{\tiny\color{gray}10}\,%
\setlength{\fboxsep}{0pt}\fcolorbox{gray!10}{gray!10}{\strut
    \mycolorbox[text=\strut{First}, color=green!17.636491]%
    \mycolorbox[text=\strut{,}, color=green!16.594987]%
}
\setlength{\fboxsep}{0pt}\fcolorbox{gray!10}{gray!10}{\strut
    \mycolorbox[text=\strut{compute}, color=green!11.248735]%
}
\setlength{\fboxsep}{0pt}\fcolorbox{gray!10}{gray!10}{\strut
    \mycolorbox[text=\strut{7}, color=red!10.857570]%
}
\setlength{\fboxsep}{0pt}\fcolorbox{gray!10}{gray!10}{\strut
    \mycolorbox[text=\strut{*}, color=green!15.876331]%
}
\setlength{\fboxsep}{0pt}\fcolorbox{gray!10}{gray!10}{\strut
    \mycolorbox[text=\strut{1}, color=green!17.636594]%
    \mycolorbox[text=\strut{0}, color=green!17.636594]%
    \mycolorbox[text=\strut{.}, color=red!14.452041]%
}
\setlength{\fboxsep}{0pt}\fcolorbox{gray!10}{gray!10}{\strut
    \mycolorbox[text=\strut{That}, color=green!16.828814]%
    \mycolorbox[text=\strut{'s}]%
}
\setlength{\fboxsep}{0pt}\fcolorbox{gray!10}{gray!10}{\strut
    \mycolorbox[text=\strut{7}, color=green!17.636594]%
    \mycolorbox[text=\strut{0}, color=green!17.636560]%
    \mycolorbox[text=\strut{.}, color=green!15.430634]%
}
\setlength{\fboxsep}{0pt}\fcolorbox{gray!10}{gray!10}{\strut
    \mycolorbox[text=\strut{Then}, color=green!17.636388]%
}
\setlength{\fboxsep}{0pt}\fcolorbox{gray!10}{gray!10}{\strut
    \mycolorbox[text=\strut{compute}, color=green!9.585794]%
}
\setlength{\fboxsep}{0pt}\fcolorbox{gray!10}{gray!10}{\strut
    \mycolorbox[text=\strut{2}, color=green!17.636594]%
    \mycolorbox[text=\strut{1}, color=green!17.636594]%
}
\setlength{\fboxsep}{0pt}\fcolorbox{gray!10}{gray!10}{\strut
    \mycolorbox[text=\strut{divided}, color=green!12.791388]%
}
\setlength{\fboxsep}{0pt}\fcolorbox{gray!10}{gray!10}{\strut
    \mycolorbox[text=\strut{by}, color=green!17.636594]%
}
\setlength{\fboxsep}{0pt}\fcolorbox{gray!10}{gray!10}{\strut
    \mycolorbox[text=\strut{3}, color=green!17.636594]%
    \mycolorbox[text=\strut{0}, color=green!17.636594]%
    \mycolorbox[text=\strut{.}, color=green!17.256659]%
}
\setlength{\fboxsep}{0pt}\fcolorbox{gray!10}{gray!10}{\strut
    \mycolorbox[text=\strut{2}, color=green!17.636594]%
    \mycolorbox[text=\strut{1}, color=green!17.636594]%
    \mycolorbox[text=\strut{/}, color=green!17.477487]%
    \mycolorbox[text=\strut{3}, color=green!17.636594]%
    \mycolorbox[text=\strut{0}, color=green!17.636594]%
}
\setlength{\fboxsep}{0pt}\fcolorbox{gray!10}{gray!10}{\strut
    \mycolorbox[text=\strut{simpl}, color=red!22.847800]%
    \mycolorbox[text=\strut{ifies}, color=green!17.636594]%
}
\setlength{\fboxsep}{0pt}\fcolorbox{gray!10}{gray!10}{\strut
    \mycolorbox[text=\strut{to}, color=green!17.570451]%
}
\setlength{\fboxsep}{0pt}\fcolorbox{gray!10}{gray!10}{\strut
    \mycolorbox[text=\strut{7}, color=green!17.636594]%
    \mycolorbox[text=\strut{/}, color=green!17.636594]%
    \mycolorbox[text=\strut{1}, color=green!17.636594]%
    \mycolorbox[text=\strut{0}, color=green!17.636594]%
    \mycolorbox[text=\strut{.}, color=green!14.790469]%
}
\setlength{\fboxsep}{0pt}\fcolorbox{gray!10}{gray!10}{\strut
    \mycolorbox[text=\strut{So}]%
}
\setlength{\fboxsep}{0pt}\fcolorbox{gray!10}{gray!10}{\strut
    \mycolorbox[text=\strut{then}, color=green!3.780672]%
    \mycolorbox[text=\strut{,}, color=red!32.178771]%
}
\setlength{\fboxsep}{0pt}\fcolorbox{gray!10}{gray!10}{\strut
    \mycolorbox[text=\strut{7}, color=green!17.636594]%
    \mycolorbox[text=\strut{0}, color=green!17.636560]%
}
\setlength{\fboxsep}{0pt}\fcolorbox{gray!10}{gray!10}{\strut
    \mycolorbox[text=\strut{*}, color=red!11.111171]%
}
\setlength{\fboxsep}{0pt}\fcolorbox{gray!10}{gray!10}{\strut
    \mycolorbox[text=\strut{(}, color=red!5.321179]%
    \mycolorbox[text=\strut{7}, color=green!17.633312]%
    \mycolorbox[text=\strut{/}, color=green!17.636491]%
    \mycolorbox[text=\strut{1}, color=green!17.636594]%
    \mycolorbox[text=\strut{0}, color=green!17.636594]%
    \mycolorbox[text=\strut{).}, color=green!13.972699]%
}
\setlength{\fboxsep}{0pt}\fcolorbox{gray!10}{gray!10}{\strut
    \mycolorbox[text=\strut{Let}, color=green!16.109662]%
}
\setlength{\fboxsep}{0pt}\fcolorbox{gray!10}{gray!10}{\strut
    \mycolorbox[text=\strut{me}, color=green!4.134798]%
}
\setlength{\fboxsep}{0pt}\fcolorbox{gray!10}{gray!10}{\strut
    \mycolorbox[text=\strut{compute}, color=red!12.314033]%
}
\setlength{\fboxsep}{0pt}\fcolorbox{gray!10}{gray!10}{\strut
    \mycolorbox[text=\strut{that}, color=green!15.411332]%
    \mycolorbox[text=\strut{.}, color=green!13.463671]%
}
\setlength{\fboxsep}{0pt}\fcolorbox{gray!10}{gray!10}{\strut
    \mycolorbox[text=\strut{7}, color=green!17.636594]%
    \mycolorbox[text=\strut{0}, color=green!17.636594]%
}
\setlength{\fboxsep}{0pt}\fcolorbox{gray!10}{gray!10}{\strut
    \mycolorbox[text=\strut{divided}, color=green!9.461097]%
}
\setlength{\fboxsep}{0pt}\fcolorbox{gray!10}{gray!10}{\strut
    \mycolorbox[text=\strut{by}, color=green!17.636594]%
}
\setlength{\fboxsep}{0pt}\fcolorbox{gray!10}{gray!10}{\strut
    \mycolorbox[text=\strut{1}, color=green!17.636594]%
    \mycolorbox[text=\strut{0}, color=green!17.636594]%
}
\setlength{\fboxsep}{0pt}\fcolorbox{gray!10}{gray!10}{\strut
    \mycolorbox[text=\strut{is}, color=green!17.636594]%
}
\setlength{\fboxsep}{0pt}\fcolorbox{gray!10}{gray!10}{\strut
    \mycolorbox[text=\strut{7}, color=green!17.636594]%
    \mycolorbox[text=\strut{,}, color=green!17.496968]%
}
\setlength{\fboxsep}{0pt}\fcolorbox{gray!10}{gray!10}{\strut
    \mycolorbox[text=\strut{times}, color=red!45.735570]%
}
\setlength{\fboxsep}{0pt}\fcolorbox{gray!10}{gray!10}{\strut
    \mycolorbox[text=\strut{7}, color=green!17.636594]%
}
\setlength{\fboxsep}{0pt}\fcolorbox{gray!10}{gray!10}{\strut
    \mycolorbox[text=\strut{is}, color=green!17.636586]%
}
\setlength{\fboxsep}{0pt}\fcolorbox{gray!10}{gray!10}{\strut
    \mycolorbox[text=\strut{4}, color=green!17.636594]%
    \mycolorbox[text=\strut{9}, color=green!17.636594]%
    \mycolorbox[text=\strut{.}, color=green!17.634587]%
}
\setlength{\fboxsep}{0pt}\fcolorbox{gray!10}{gray!10}{\strut
    \mycolorbox[text=\strut{So}, color=red!7.033933]%
}
\setlength{\fboxsep}{0pt}\fcolorbox{gray!10}{gray!10}{\strut
    \mycolorbox[text=\strut{the}, color=green!6.439138]%
}
\setlength{\fboxsep}{0pt}\fcolorbox{gray!10}{gray!10}{\strut
    \mycolorbox[text=\strut{result}, color=green!17.414981]%
}
\setlength{\fboxsep}{0pt}\fcolorbox{gray!10}{gray!10}{\strut
    \mycolorbox[text=\strut{is}, color=green!16.292742]%
}
\setlength{\fboxsep}{0pt}\fcolorbox{gray!10}{gray!10}{\strut
    \mycolorbox[text=\strut{4}, color=green!17.636594]%
    \mycolorbox[text=\strut{9}, color=green!17.636594]%
    \mycolorbox[text=\strut{?}, color=red!16.503212]%
}
\setlength{\fboxsep}{0pt}\fcolorbox{gray!10}{gray!10}{\strut
    \setlength{\fboxsep}{1pt}\fbox{\mycolorbox[text=\strut{Wait}]}%
    \mycolorbox[text=\strut{,}, color=green!17.636026]%
}
\setlength{\fboxsep}{0pt}\fcolorbox{gray!10}{gray!10}{\strut
    \mycolorbox[text=\strut{so}, color=red!77.434606]%
}
\setlength{\fboxsep}{0pt}\fcolorbox{gray!10}{gray!10}{\strut
    \mycolorbox[text=\strut{is}, color=red!32.098740]%
}
\setlength{\fboxsep}{0pt}\fcolorbox{gray!10}{gray!10}{\strut
    \mycolorbox[text=\strut{that}, color=red!19.400186]%
}
\setlength{\fboxsep}{0pt}\fcolorbox{gray!10}{gray!10}{\strut
    \mycolorbox[text=\strut{the}, color=green!17.273389]%
}
\setlength{\fboxsep}{0pt}\fcolorbox{gray!10}{gray!10}{\strut
    \mycolorbox[text=\strut{answer}, color=green!15.252494]%
    \mycolorbox[text=\strut{?}, color=green!7.312668]%
}
\setlength{\fboxsep}{0pt}\fcolorbox{gray!10}{gray!10}{\strut
    \mycolorbox[text=\strut{Let}, color=red!48.633650]%
}
\setlength{\fboxsep}{0pt}\fcolorbox{gray!10}{gray!10}{\strut
    \mycolorbox[text=\strut{me}, color=green!17.636594]%
}
\setlength{\fboxsep}{0pt}\fcolorbox{gray!10}{gray!10}{\strut
    \mycolorbox[text=\strut{check}, color=green!9.562399]%
}
\setlength{\fboxsep}{0pt}\fcolorbox{gray!10}{gray!10}{\strut
    \mycolorbox[text=\strut{again}, color=green!6.239614]%
    \mycolorbox[text=\strut{.}, color=green!10.829596]%
}
\\
\\
{\tiny\color{gray}11}\,%
\setlength{\fboxsep}{0pt}\fcolorbox{gray!10}{gray!10}{\strut
    \mycolorbox[text=\strut{Alternatively}, color=red!15.558967]%
    \mycolorbox[text=\strut{,}, color=green!17.636594]%
}
\setlength{\fboxsep}{0pt}\fcolorbox{gray!10}{gray!10}{\strut
    \mycolorbox[text=\strut{maybe}, color=green!15.666705]%
}
\setlength{\fboxsep}{0pt}\fcolorbox{gray!10}{gray!10}{\strut
    \mycolorbox[text=\strut{I}, color=green!16.161156]%
}
\setlength{\fboxsep}{0pt}\fcolorbox{gray!10}{gray!10}{\strut
    \mycolorbox[text=\strut{should}, color=red!17.660928]%
}
\setlength{\fboxsep}{0pt}\fcolorbox{gray!10}{gray!10}{\strut
    \mycolorbox[text=\strut{compute}, color=red!31.466733]%
}
\setlength{\fboxsep}{0pt}\fcolorbox{gray!10}{gray!10}{\strut
    \mycolorbox[text=\strut{it}, color=red!68.098456]%
}
\setlength{\fboxsep}{0pt}\fcolorbox{gray!10}{gray!10}{\strut
    \mycolorbox[text=\strut{as}, color=green!11.929977]%
}
\setlength{\fboxsep}{0pt}\fcolorbox{gray!10}{gray!10}{\strut
    \mycolorbox[text=\strut{fractions}, color=green!15.319627]%
    \mycolorbox[text=\strut{.}, color=green!10.542933]%
}
\setlength{\fboxsep}{0pt}\fcolorbox{gray!10}{gray!10}{\strut
    \mycolorbox[text=\strut{Let}]%
}
\setlength{\fboxsep}{0pt}\fcolorbox{gray!10}{gray!10}{\strut
    \mycolorbox[text=\strut{me}, color=red!16.676255]%
}
\setlength{\fboxsep}{0pt}\fcolorbox{gray!10}{gray!10}{\strut
    \mycolorbox[text=\strut{do}, color=red!38.618355]%
}
\setlength{\fboxsep}{0pt}\fcolorbox{gray!10}{gray!10}{\strut
    \mycolorbox[text=\strut{that}, color=green!15.070705]%
    \mycolorbox[text=\strut{.}, color=red!48.310112]%
}
\setlength{\fboxsep}{0pt}\fcolorbox{gray!10}{gray!10}{\strut
    \mycolorbox[text=\strut{7}, color=green!17.340810]%
}
\setlength{\fboxsep}{0pt}\fcolorbox{gray!10}{gray!10}{\strut
    \mycolorbox[text=\strut{*}, color=green!10.371166]%
}
\setlength{\fboxsep}{0pt}\fcolorbox{gray!10}{gray!10}{\strut
    \mycolorbox[text=\strut{1}, color=green!17.636594]%
    \mycolorbox[text=\strut{0}, color=green!17.636594]%
}
\setlength{\fboxsep}{0pt}\fcolorbox{gray!10}{gray!10}{\strut
    \mycolorbox[text=\strut{*}, color=green!3.191158]%
}
\setlength{\fboxsep}{0pt}\fcolorbox{gray!10}{gray!10}{\strut
    \mycolorbox[text=\strut{(}, color=green!17.617708]%
    \mycolorbox[text=\strut{2}, color=green!17.636594]%
    \mycolorbox[text=\strut{1}, color=green!17.636594]%
}
\setlength{\fboxsep}{0pt}\fcolorbox{gray!10}{gray!10}{\strut
    \mycolorbox[text=\strut{/}, color=red!16.676400]%
}
\setlength{\fboxsep}{0pt}\fcolorbox{gray!10}{gray!10}{\strut
    \mycolorbox[text=\strut{3}, color=green!17.636594]%
    \mycolorbox[text=\strut{0}, color=green!17.636594]%
    \mycolorbox[text=\strut{).}, color=green!4.134555]%
}
\setlength{\fboxsep}{0pt}\fcolorbox{gray!10}{gray!10}{\strut
    \mycolorbox[text=\strut{So}, color=red!25.409785]%
}
\setlength{\fboxsep}{0pt}\fcolorbox{gray!10}{gray!10}{\strut
    \mycolorbox[text=\strut{that}, color=red!44.164175]%
    \mycolorbox[text=\strut{'s}, color=red!35.433577]%
}
\setlength{\fboxsep}{0pt}\fcolorbox{gray!10}{gray!10}{\strut
    \mycolorbox[text=\strut{(}, color=red!34.853475]%
    \mycolorbox[text=\strut{7}, color=green!17.636594]%
}
\setlength{\fboxsep}{0pt}\fcolorbox{gray!10}{gray!10}{\strut
    \mycolorbox[text=\strut{*}, color=green!14.794998]%
}
\setlength{\fboxsep}{0pt}\fcolorbox{gray!10}{gray!10}{\strut
    \mycolorbox[text=\strut{1}, color=green!17.636594]%
    \mycolorbox[text=\strut{0}, color=green!17.636594]%
}
\setlength{\fboxsep}{0pt}\fcolorbox{gray!10}{gray!10}{\strut
    \mycolorbox[text=\strut{*}, color=green!17.596612]%
}
\setlength{\fboxsep}{0pt}\fcolorbox{gray!10}{gray!10}{\strut
    \mycolorbox[text=\strut{2}, color=green!17.636594]%
    \mycolorbox[text=\strut{1}, color=green!17.636594]%
    \mycolorbox[text=\strut{)}, color=green!17.008875]%
}
\setlength{\fboxsep}{0pt}\fcolorbox{gray!10}{gray!10}{\strut
    \mycolorbox[text=\strut{/}, color=green!16.600548]%
}
\setlength{\fboxsep}{0pt}\fcolorbox{gray!10}{gray!10}{\strut
    \mycolorbox[text=\strut{3}, color=green!17.636594]%
    \mycolorbox[text=\strut{0}, color=green!17.636594]%
    \mycolorbox[text=\strut{.}, color=green!17.527882]%
}
\setlength{\fboxsep}{0pt}\fcolorbox{gray!10}{gray!10}{\strut
    \mycolorbox[text=\strut{Let}, color=green!16.715162]%
}
\setlength{\fboxsep}{0pt}\fcolorbox{gray!10}{gray!10}{\strut
    \mycolorbox[text=\strut{me}, color=red!13.737564]%
}
\setlength{\fboxsep}{0pt}\fcolorbox{gray!10}{gray!10}{\strut
    \mycolorbox[text=\strut{compute}, color=green!16.827164]%
}
\setlength{\fboxsep}{0pt}\fcolorbox{gray!10}{gray!10}{\strut
    \mycolorbox[text=\strut{numerator}, color=green!10.527874]%
}
\setlength{\fboxsep}{0pt}\fcolorbox{gray!10}{gray!10}{\strut
    \mycolorbox[text=\strut{and}, color=red!12.599912]%
}
\setlength{\fboxsep}{0pt}\fcolorbox{gray!10}{gray!10}{\strut
    \mycolorbox[text=\strut{denominator}, color=green!17.636474]%
    \mycolorbox[text=\strut{.}, color=red!32.678418]%
}
\\
\\
{\tiny\color{gray}12}\,%
\setlength{\fboxsep}{0pt}\fcolorbox{gray!10}{gray!10}{\strut
    \mycolorbox[text=\strut{N}, color=green!9.624292]%
    \mycolorbox[text=\strut{umerator}, color=green!17.636586]%
    \mycolorbox[text=\strut{:}, color=green!17.406321]%
}
\setlength{\fboxsep}{0pt}\fcolorbox{gray!10}{gray!10}{\strut
    \mycolorbox[text=\strut{7}, color=green!17.636594]%
}
\setlength{\fboxsep}{0pt}\fcolorbox{gray!10}{gray!10}{\strut
    \mycolorbox[text=\strut{*}, color=green!12.906111]%
}
\setlength{\fboxsep}{0pt}\fcolorbox{gray!10}{gray!10}{\strut
    \mycolorbox[text=\strut{1}, color=green!17.636594]%
    \mycolorbox[text=\strut{0}, color=green!17.636594]%
}
\setlength{\fboxsep}{0pt}\fcolorbox{gray!10}{gray!10}{\strut
    \mycolorbox[text=\strut{*}, color=red!22.051957]%
}
\setlength{\fboxsep}{0pt}\fcolorbox{gray!10}{gray!10}{\strut
    \mycolorbox[text=\strut{2}, color=green!17.636594]%
    \mycolorbox[text=\strut{1}, color=green!17.636594]%
    \mycolorbox[text=\strut{.}, color=green!17.008485]%
}
\setlength{\fboxsep}{0pt}\fcolorbox{gray!10}{gray!10}{\strut
    \mycolorbox[text=\strut{Let}, color=red!7.005545]%
}
\setlength{\fboxsep}{0pt}\fcolorbox{gray!10}{gray!10}{\strut
    \mycolorbox[text=\strut{me}, color=red!40.828285]%
}
\setlength{\fboxsep}{0pt}\fcolorbox{gray!10}{gray!10}{\strut
    \mycolorbox[text=\strut{compute}, color=green!16.056821]%
}
\setlength{\fboxsep}{0pt}\fcolorbox{gray!10}{gray!10}{\strut
    \mycolorbox[text=\strut{7}, color=green!17.636465]%
    \mycolorbox[text=\strut{*}, color=red!28.769059]%
    \mycolorbox[text=\strut{1}, color=green!17.635862]%
    \mycolorbox[text=\strut{0}, color=green!17.636594]%
}
\setlength{\fboxsep}{0pt}\fcolorbox{gray!10}{gray!10}{\strut
    \mycolorbox[text=\strut{first}, color=green!15.534523]%
    \mycolorbox[text=\strut{,}, color=green!9.536704]%
}
\setlength{\fboxsep}{0pt}\fcolorbox{gray!10}{gray!10}{\strut
    \mycolorbox[text=\strut{which}, color=red!10.858300]%
}
\setlength{\fboxsep}{0pt}\fcolorbox{gray!10}{gray!10}{\strut
    \mycolorbox[text=\strut{is}, color=green!17.636594]%
}
\setlength{\fboxsep}{0pt}\fcolorbox{gray!10}{gray!10}{\strut
    \mycolorbox[text=\strut{7}, color=green!17.636594]%
    \mycolorbox[text=\strut{0}, color=green!17.636594]%
    \mycolorbox[text=\strut{.}, color=red!28.769923]%
}
\setlength{\fboxsep}{0pt}\fcolorbox{gray!10}{gray!10}{\strut
    \mycolorbox[text=\strut{Then}, color=green!15.922461]%
}
\setlength{\fboxsep}{0pt}\fcolorbox{gray!10}{gray!10}{\strut
    \mycolorbox[text=\strut{7}, color=green!17.636594]%
    \mycolorbox[text=\strut{0}, color=green!17.636594]%
}
\setlength{\fboxsep}{0pt}\fcolorbox{gray!10}{gray!10}{\strut
    \mycolorbox[text=\strut{*}, color=red!40.828288]%
}
\setlength{\fboxsep}{0pt}\fcolorbox{gray!10}{gray!10}{\strut
    \mycolorbox[text=\strut{2}, color=green!17.636594]%
    \mycolorbox[text=\strut{1}, color=green!17.636594]%
    \mycolorbox[text=\strut{.}, color=green!17.636103]%
}
\setlength{\fboxsep}{0pt}\fcolorbox{gray!10}{gray!10}{\strut
    \mycolorbox[text=\strut{Let}, color=red!25.094840]%
    \mycolorbox[text=\strut{'s}, color=red!65.039887]%
}
\setlength{\fboxsep}{0pt}\fcolorbox{gray!10}{gray!10}{\strut
    \mycolorbox[text=\strut{do}, color=red!48.599683]%
}
\setlength{\fboxsep}{0pt}\fcolorbox{gray!10}{gray!10}{\strut
    \mycolorbox[text=\strut{7}, color=green!17.636594]%
    \mycolorbox[text=\strut{0}, color=green!17.636594]%
    \mycolorbox[text=\strut{*}, color=green!12.905397]%
    \mycolorbox[text=\strut{2}, color=green!17.636594]%
    \mycolorbox[text=\strut{0}, color=green!17.636327]%
    \mycolorbox[text=\strut{=}, color=red!61.503157]%
    \mycolorbox[text=\strut{1}, color=green!17.636594]%
    \mycolorbox[text=\strut{4}, color=green!17.636569]%
    \mycolorbox[text=\strut{0}, color=green!17.636594]%
    \mycolorbox[text=\strut{0}, color=green!17.636594]%
}
\setlength{\fboxsep}{0pt}\fcolorbox{gray!10}{gray!10}{\strut
    \mycolorbox[text=\strut{and}, color=red!28.781388]%
}
\setlength{\fboxsep}{0pt}\fcolorbox{gray!10}{gray!10}{\strut
    \mycolorbox[text=\strut{7}, color=green!17.636594]%
    \mycolorbox[text=\strut{0}, color=green!17.636594]%
    \mycolorbox[text=\strut{*}, color=green!17.636594]%
    \mycolorbox[text=\strut{1}, color=green!17.636594]%
    \mycolorbox[text=\strut{=}, color=green!17.636594]%
    \mycolorbox[text=\strut{7}, color=green!17.636594]%
    \mycolorbox[text=\strut{0}, color=green!17.636594]%
    \mycolorbox[text=\strut{,}, color=green!17.453057]%
}
\setlength{\fboxsep}{0pt}\fcolorbox{gray!10}{gray!10}{\strut
    \mycolorbox[text=\strut{so}, color=green!17.008537]%
}
\setlength{\fboxsep}{0pt}\fcolorbox{gray!10}{gray!10}{\strut
    \mycolorbox[text=\strut{total}, color=green!12.905704]%
}
\setlength{\fboxsep}{0pt}\fcolorbox{gray!10}{gray!10}{\strut
    \mycolorbox[text=\strut{is}, color=red!28.769597]%
}
\setlength{\fboxsep}{0pt}\fcolorbox{gray!10}{gray!10}{\strut
    \mycolorbox[text=\strut{1}, color=green!17.636594]%
    \mycolorbox[text=\strut{4}, color=green!17.636594]%
    \mycolorbox[text=\strut{7}, color=green!4.134798]%
    \mycolorbox[text=\strut{0}, color=green!17.636594]%
    \mycolorbox[text=\strut{.}, color=green!9.702662]%
}
\setlength{\fboxsep}{0pt}\fcolorbox{gray!10}{gray!10}{\strut
    \mycolorbox[text=\strut{So}, color=red!42.622372]%
}
\setlength{\fboxsep}{0pt}\fcolorbox{gray!10}{gray!10}{\strut
    \mycolorbox[text=\strut{numerator}, color=green!12.787286]%
}
\setlength{\fboxsep}{0pt}\fcolorbox{gray!10}{gray!10}{\strut
    \mycolorbox[text=\strut{is}, color=green!17.636594]%
}
\setlength{\fboxsep}{0pt}\fcolorbox{gray!10}{gray!10}{\strut
    \mycolorbox[text=\strut{1}, color=green!17.636594]%
    \mycolorbox[text=\strut{4}, color=green!17.636594]%
    \mycolorbox[text=\strut{7}, color=green!17.636594]%
    \mycolorbox[text=\strut{0}, color=green!17.636594]%
    \mycolorbox[text=\strut{.}, color=red!6.143450]%
}
\setlength{\fboxsep}{0pt}\fcolorbox{gray!10}{gray!10}{\strut
    \mycolorbox[text=\strut{Den}, color=green!17.521301]%
    \mycolorbox[text=\strut{ominator}, color=green!17.636594]%
}
\setlength{\fboxsep}{0pt}\fcolorbox{gray!10}{gray!10}{\strut
    \mycolorbox[text=\strut{is}, color=green!17.636594]%
}
\setlength{\fboxsep}{0pt}\fcolorbox{gray!10}{gray!10}{\strut
    \mycolorbox[text=\strut{3}, color=green!17.636594]%
    \mycolorbox[text=\strut{0}, color=green!17.636594]%
    \mycolorbox[text=\strut{.}, color=green!15.430625]%
}
\setlength{\fboxsep}{0pt}\fcolorbox{gray!10}{gray!10}{\strut
    \mycolorbox[text=\strut{So}, color=red!18.490819]%
}
\setlength{\fboxsep}{0pt}\fcolorbox{gray!10}{gray!10}{\strut
    \mycolorbox[text=\strut{1}, color=green!17.636594]%
    \mycolorbox[text=\strut{4}, color=green!17.636594]%
    \mycolorbox[text=\strut{7}, color=green!17.636594]%
    \mycolorbox[text=\strut{0}, color=green!17.636594]%
}
\setlength{\fboxsep}{0pt}\fcolorbox{gray!10}{gray!10}{\strut
    \mycolorbox[text=\strut{divided}, color=red!22.889346]%
}
\setlength{\fboxsep}{0pt}\fcolorbox{gray!10}{gray!10}{\strut
    \mycolorbox[text=\strut{by}, color=green!17.636594]%
}
\setlength{\fboxsep}{0pt}\fcolorbox{gray!10}{gray!10}{\strut
    \mycolorbox[text=\strut{3}, color=green!17.636594]%
    \mycolorbox[text=\strut{0}, color=green!17.636594]%
    \mycolorbox[text=\strut{.}, color=green!13.938729]%
}
\setlength{\fboxsep}{0pt}\fcolorbox{gray!10}{gray!10}{\strut
    \mycolorbox[text=\strut{Let}, color=green!11.430479]%
    \mycolorbox[text=\strut{'s}, color=red!67.385849]%
}
\setlength{\fboxsep}{0pt}\fcolorbox{gray!10}{gray!10}{\strut
    \mycolorbox[text=\strut{simplify}, color=red!53.903471]%
}
\setlength{\fboxsep}{0pt}\fcolorbox{gray!10}{gray!10}{\strut
    \mycolorbox[text=\strut{that}, color=green!12.804137]%
    \mycolorbox[text=\strut{.}, color=red!52.382867]%
}
\\
\\
{\tiny\color{gray}13}\,%
\setlength{\fboxsep}{0pt}\fcolorbox{gray!10}{gray!10}{\strut
    \mycolorbox[text=\strut{Div}, color=green!16.939685]%
    \mycolorbox[text=\strut{ide}, color=green!17.631175]%
}
\setlength{\fboxsep}{0pt}\fcolorbox{gray!10}{gray!10}{\strut
    \mycolorbox[text=\strut{numerator}, color=green!17.629297]%
}
\setlength{\fboxsep}{0pt}\fcolorbox{gray!10}{gray!10}{\strut
    \mycolorbox[text=\strut{and}, color=green!17.636594]%
}
\setlength{\fboxsep}{0pt}\fcolorbox{gray!10}{gray!10}{\strut
    \mycolorbox[text=\strut{denominator}, color=green!17.636594]%
}
\setlength{\fboxsep}{0pt}\fcolorbox{gray!10}{gray!10}{\strut
    \mycolorbox[text=\strut{by}, color=green!17.636594]%
}
\setlength{\fboxsep}{0pt}\fcolorbox{gray!10}{gray!10}{\strut
    \mycolorbox[text=\strut{1}, color=green!17.636594]%
    \mycolorbox[text=\strut{0}, color=green!17.636594]%
    \mycolorbox[text=\strut{:}, color=green!17.399215]%
}
\setlength{\fboxsep}{0pt}\fcolorbox{gray!10}{gray!10}{\strut
    \mycolorbox[text=\strut{1}, color=green!17.636594]%
    \mycolorbox[text=\strut{4}, color=green!17.636594]%
    \mycolorbox[text=\strut{7}, color=green!17.636594]%
}
\setlength{\fboxsep}{0pt}\fcolorbox{gray!10}{gray!10}{\strut
    \mycolorbox[text=\strut{/}, color=green!15.997605]%
}
\setlength{\fboxsep}{0pt}\fcolorbox{gray!10}{gray!10}{\strut
    \mycolorbox[text=\strut{3}, color=green!17.636594]%
    \mycolorbox[text=\strut{.}, color=green!17.625135]%
}
\setlength{\fboxsep}{0pt}\fcolorbox{gray!10}{gray!10}{\strut
    \mycolorbox[text=\strut{1}, color=green!16.600557]%
    \mycolorbox[text=\strut{4}, color=green!17.636594]%
    \mycolorbox[text=\strut{7}, color=green!17.636594]%
}
\setlength{\fboxsep}{0pt}\fcolorbox{gray!10}{gray!10}{\strut
    \mycolorbox[text=\strut{divided}, color=green!17.629978]%
}
\setlength{\fboxsep}{0pt}\fcolorbox{gray!10}{gray!10}{\strut
    \mycolorbox[text=\strut{by}, color=green!17.636594]%
}
\setlength{\fboxsep}{0pt}\fcolorbox{gray!10}{gray!10}{\strut
    \mycolorbox[text=\strut{3}, color=green!17.636594]%
}
\setlength{\fboxsep}{0pt}\fcolorbox{gray!10}{gray!10}{\strut
    \mycolorbox[text=\strut{is}, color=green!15.284854]%
}
\setlength{\fboxsep}{0pt}\fcolorbox{gray!10}{gray!10}{\strut
    \mycolorbox[text=\strut{4}, color=green!17.636594]%
    \mycolorbox[text=\strut{9}, color=green!17.636594]%
    \mycolorbox[text=\strut{.}, color=green!17.621784]%
}
\setlength{\fboxsep}{0pt}\fcolorbox{gray!10}{gray!10}{\strut
    \mycolorbox[text=\strut{So}, color=green!16.190860]%
}
\setlength{\fboxsep}{0pt}\fcolorbox{gray!10}{gray!10}{\strut
    \mycolorbox[text=\strut{yes}, color=red!9.794077]%
    \mycolorbox[text=\strut{,}, color=green!17.636594]%
}
\setlength{\fboxsep}{0pt}\fcolorbox{gray!10}{gray!10}{\strut
    \mycolorbox[text=\strut{same}, color=red!68.736518]%
}
\setlength{\fboxsep}{0pt}\fcolorbox{gray!10}{gray!10}{\strut
    \mycolorbox[text=\strut{answer}, color=red!67.814334]%
    \mycolorbox[text=\strut{,}, color=red!59.497357]%
}
\setlength{\fboxsep}{0pt}\fcolorbox{gray!10}{gray!10}{\strut
    \mycolorbox[text=\strut{4}, color=green!17.636594]%
    \mycolorbox[text=\strut{9}, color=green!17.636594]%
    \mycolorbox[text=\strut{.}, color=green!17.009143]%
}
\setlength{\fboxsep}{0pt}\fcolorbox{gray!10}{gray!10}{\strut
    \mycolorbox[text=\strut{So}, color=red!10.914878]%
}
\setlength{\fboxsep}{0pt}\fcolorbox{gray!10}{gray!10}{\strut
    \mycolorbox[text=\strut{the}, color=red!21.894038]%
}
\setlength{\fboxsep}{0pt}\fcolorbox{gray!10}{gray!10}{\strut
    \mycolorbox[text=\strut{result}, color=red!6.387811]%
}
\setlength{\fboxsep}{0pt}\fcolorbox{gray!10}{gray!10}{\strut
    \mycolorbox[text=\strut{is}, color=green!17.256590]%
}
\setlength{\fboxsep}{0pt}\fcolorbox{gray!10}{gray!10}{\strut
    \mycolorbox[text=\strut{4}, color=green!17.636594]%
    \mycolorbox[text=\strut{9}, color=green!17.636594]%
    \mycolorbox[text=\strut{.}, color=green!13.519100]%
}
\setlength{\fboxsep}{0pt}\fcolorbox{gray!10}{gray!10}{\strut
    \mycolorbox[text=\strut{But}, color=red!17.258531]%
}
\setlength{\fboxsep}{0pt}\fcolorbox{gray!10}{gray!10}{\strut
    \setlength{\fboxsep}{1pt}\fbox{\mycolorbox[text=\strut{wait}]}%
    \mycolorbox[text=\strut{,}, color=green!17.636577]%
}
\setlength{\fboxsep}{0pt}\fcolorbox{gray!10}{gray!10}{\strut
    \mycolorbox[text=\strut{4}, color=green!17.636594]%
    \mycolorbox[text=\strut{9}, color=green!17.636594]%
}
\setlength{\fboxsep}{0pt}\fcolorbox{gray!10}{gray!10}{\strut
    \mycolorbox[text=\strut{is}, color=green!17.636491]%
}
\setlength{\fboxsep}{0pt}\fcolorbox{gray!10}{gray!10}{\strut
    \mycolorbox[text=\strut{an}, color=green!17.600869]%
}
\setlength{\fboxsep}{0pt}\fcolorbox{gray!10}{gray!10}{\strut
    \mycolorbox[text=\strut{integer}, color=green!17.636594]%
    \mycolorbox[text=\strut{.}, color=red!16.711904]%
}
\setlength{\fboxsep}{0pt}\fcolorbox{gray!10}{gray!10}{\strut
    \mycolorbox[text=\strut{But}, color=red!41.691343]%
}
\setlength{\fboxsep}{0pt}\fcolorbox{gray!10}{gray!10}{\strut
    \mycolorbox[text=\strut{the}, color=green!6.483255]%
}
\setlength{\fboxsep}{0pt}\fcolorbox{gray!10}{gray!10}{\strut
    \mycolorbox[text=\strut{operation}, color=red!48.612749]%
}
\setlength{\fboxsep}{0pt}\fcolorbox{gray!10}{gray!10}{\strut
    \mycolorbox[text=\strut{is}, color=green!12.905044]%
}
\setlength{\fboxsep}{0pt}\fcolorbox{gray!10}{gray!10}{\strut
    \mycolorbox[text=\strut{defined}, color=green!16.540581]%
}
\setlength{\fboxsep}{0pt}\fcolorbox{gray!10}{gray!10}{\strut
    \mycolorbox[text=\strut{for}, color=red!24.811430]%
}
\setlength{\fboxsep}{0pt}\fcolorbox{gray!10}{gray!10}{\strut
    \mycolorbox[text=\strut{fractions}, color=green!17.617544]%
    \mycolorbox[text=\strut{,}, color=red!59.223604]%
}
\setlength{\fboxsep}{0pt}\fcolorbox{gray!10}{gray!10}{\strut
    \mycolorbox[text=\strut{but}, color=green!9.673284]%
}
\setlength{\fboxsep}{0pt}\fcolorbox{gray!10}{gray!10}{\strut
    \mycolorbox[text=\strut{maybe}, color=red!57.934389]%
}
\setlength{\fboxsep}{0pt}\fcolorbox{gray!10}{gray!10}{\strut
    \mycolorbox[text=\strut{the}, color=green!13.917218]%
}
\setlength{\fboxsep}{0pt}\fcolorbox{gray!10}{gray!10}{\strut
    \mycolorbox[text=\strut{result}, color=green!17.522871]%
}
\setlength{\fboxsep}{0pt}\fcolorbox{gray!10}{gray!10}{\strut
    \mycolorbox[text=\strut{can}, color=red!22.882015]%
}
\setlength{\fboxsep}{0pt}\fcolorbox{gray!10}{gray!10}{\strut
    \mycolorbox[text=\strut{be}, color=green!17.636586]%
}
\setlength{\fboxsep}{0pt}\fcolorbox{gray!10}{gray!10}{\strut
    \mycolorbox[text=\strut{an}, color=green!17.349414]%
}
\setlength{\fboxsep}{0pt}\fcolorbox{gray!10}{gray!10}{\strut
    \mycolorbox[text=\strut{integer}, color=green!17.636594]%
    \mycolorbox[text=\strut{.}, color=red!25.456832]%
}
\setlength{\fboxsep}{0pt}\fcolorbox{gray!10}{gray!10}{\strut
    \mycolorbox[text=\strut{So}, color=red!40.347114]%
}
\setlength{\fboxsep}{0pt}\fcolorbox{gray!10}{gray!10}{\strut
    \mycolorbox[text=\strut{is}, color=red!38.392007]%
}
\setlength{\fboxsep}{0pt}\fcolorbox{gray!10}{gray!10}{\strut
    \mycolorbox[text=\strut{4}, color=green!17.636594]%
    \mycolorbox[text=\strut{9}, color=green!17.636594]%
}
\setlength{\fboxsep}{0pt}\fcolorbox{gray!10}{gray!10}{\strut
    \mycolorbox[text=\strut{the}, color=green!17.339557]%
}
\setlength{\fboxsep}{0pt}\fcolorbox{gray!10}{gray!10}{\strut
    \mycolorbox[text=\strut{simplified}, color=green!4.428873]%
}
\setlength{\fboxsep}{0pt}\fcolorbox{gray!10}{gray!10}{\strut
    \mycolorbox[text=\strut{value}, color=green!17.574330]%
    \mycolorbox[text=\strut{?}, color=red!5.465839]%
}
\setlength{\fboxsep}{0pt}\fcolorbox{gray!10}{gray!10}{\strut
    \setlength{\fboxsep}{1pt}\fbox{\mycolorbox[text=\strut{Wait}, color=red!70.001888]}%
    \mycolorbox[text=\strut{,}, color=green!17.636560]%
}
\setlength{\fboxsep}{0pt}\fcolorbox{gray!10}{gray!10}{\strut
    \mycolorbox[text=\strut{but}, color=red!30.462695]%
}
\setlength{\fboxsep}{0pt}\fcolorbox{gray!10}{gray!10}{\strut
    \mycolorbox[text=\strut{the}, color=red!32.239962]%
}
\setlength{\fboxsep}{0pt}\fcolorbox{gray!10}{gray!10}{\strut
    \mycolorbox[text=\strut{problem}]%
}
\setlength{\fboxsep}{0pt}\fcolorbox{gray!10}{gray!10}{\strut
    \mycolorbox[text=\strut{says}, color=green!17.631641]%
}
\setlength{\fboxsep}{0pt}\fcolorbox{gray!10}{gray!10}{\strut
    \mycolorbox[text=\strut{"}, color=green!17.591716]%
    \mycolorbox[text=\strut{s}, color=red!16.798045]%
    \mycolorbox[text=\strut{implified}, color=green!17.627479]%
}
\setlength{\fboxsep}{0pt}\fcolorbox{gray!10}{gray!10}{\strut
    \mycolorbox[text=\strut{value}, color=green!17.636250]%
}
\setlength{\fboxsep}{0pt}\fcolorbox{gray!10}{gray!10}{\strut
    \mycolorbox[text=\strut{of}, color=red!54.937794]%
}
\setlength{\fboxsep}{0pt}\fcolorbox{gray!10}{gray!10}{\strut
    \mycolorbox[text=\strut{7}, color=green!17.636594]%
    \mycolorbox[text=\strut{/}, color=green!17.636594]%
    \mycolorbox[text=\strut{3}, color=green!17.636594]%
    \mycolorbox[text=\strut{0}, color=green!17.636594]%
}
\setlength{\fboxsep}{0pt}\fcolorbox{gray!10}{gray!10}{\strut
    \mycolorbox[text=\strut{@}, color=green!17.636293]%
}
\setlength{\fboxsep}{0pt}\fcolorbox{gray!10}{gray!10}{\strut
    \mycolorbox[text=\strut{1}, color=green!17.636594]%
    \mycolorbox[text=\strut{0}, color=green!17.636594]%
    \mycolorbox[text=\strut{/}, color=green!17.636594]%
    \mycolorbox[text=\strut{2}, color=green!17.636594]%
    \mycolorbox[text=\strut{1}, color=green!17.636594]%
    \mycolorbox[text=\strut{".}, color=green!13.660128]%
}
\setlength{\fboxsep}{0pt}\fcolorbox{gray!10}{gray!10}{\strut
    \mycolorbox[text=\strut{So}, color=red!5.931790]%
}
\setlength{\fboxsep}{0pt}\fcolorbox{gray!10}{gray!10}{\strut
    \mycolorbox[text=\strut{maybe}, color=red!28.396830]%
}
\setlength{\fboxsep}{0pt}\fcolorbox{gray!10}{gray!10}{\strut
    \mycolorbox[text=\strut{4}, color=green!17.636594]%
    \mycolorbox[text=\strut{9}, color=green!17.636594]%
}
\setlength{\fboxsep}{0pt}\fcolorbox{gray!10}{gray!10}{\strut
    \mycolorbox[text=\strut{is}, color=green!17.635440]%
}
\setlength{\fboxsep}{0pt}\fcolorbox{gray!10}{gray!10}{\strut
    \mycolorbox[text=\strut{already}, color=red!16.831819]%
}
\setlength{\fboxsep}{0pt}\fcolorbox{gray!10}{gray!10}{\strut
    \mycolorbox[text=\strut{simplified}, color=green!17.598930]%
    \mycolorbox[text=\strut{?}, color=green!8.064230]%
}
\setlength{\fboxsep}{0pt}\fcolorbox{gray!10}{gray!10}{\strut
    \mycolorbox[text=\strut{But}, color=red!59.611989]%
}
\setlength{\fboxsep}{0pt}\fcolorbox{gray!10}{gray!10}{\strut
    \mycolorbox[text=\strut{4}, color=green!17.636594]%
    \mycolorbox[text=\strut{9}, color=green!17.636594]%
}
\setlength{\fboxsep}{0pt}\fcolorbox{gray!10}{gray!10}{\strut
    \mycolorbox[text=\strut{is}, color=green!17.636422]%
}
\setlength{\fboxsep}{0pt}\fcolorbox{gray!10}{gray!10}{\strut
    \mycolorbox[text=\strut{7}, color=red!28.769062]%
}
\setlength{\fboxsep}{0pt}\fcolorbox{gray!10}{gray!10}{\strut
    \mycolorbox[text=\strut{squared}, color=green!15.784428]%
    \mycolorbox[text=\strut{,}, color=red!10.860624]%
}
\setlength{\fboxsep}{0pt}\fcolorbox{gray!10}{gray!10}{\strut
    \mycolorbox[text=\strut{so}]%
}
\setlength{\fboxsep}{0pt}\fcolorbox{gray!10}{gray!10}{\strut
    \mycolorbox[text=\strut{it}, color=green!10.383078]%
    \mycolorbox[text=\strut{'s}, color=green!17.020523]%
}
\setlength{\fboxsep}{0pt}\fcolorbox{gray!10}{gray!10}{\strut
    \mycolorbox[text=\strut{an}, color=green!16.981100]%
}
\setlength{\fboxsep}{0pt}\fcolorbox{gray!10}{gray!10}{\strut
    \mycolorbox[text=\strut{integer}, color=green!17.636594]%
    \mycolorbox[text=\strut{.}, color=red!17.839374]%
}
\setlength{\fboxsep}{0pt}\fcolorbox{gray!10}{gray!10}{\strut
    \mycolorbox[text=\strut{So}]%
}
\setlength{\fboxsep}{0pt}\fcolorbox{gray!10}{gray!10}{\strut
    \mycolorbox[text=\strut{the}, color=red!55.614395]%
}
\setlength{\fboxsep}{0pt}\fcolorbox{gray!10}{gray!10}{\strut
    \mycolorbox[text=\strut{answer}, color=red!10.881890]%
}
\setlength{\fboxsep}{0pt}\fcolorbox{gray!10}{gray!10}{\strut
    \mycolorbox[text=\strut{is}, color=green!15.247352]%
}
\setlength{\fboxsep}{0pt}\fcolorbox{gray!10}{gray!10}{\strut
    \mycolorbox[text=\strut{4}, color=green!17.636594]%
    \mycolorbox[text=\strut{9}, color=green!17.636594]%
    \mycolorbox[text=\strut{.}, color=red!39.776058]%
}
\setlength{\fboxsep}{0pt}\fcolorbox{gray!10}{gray!10}{\strut
    \mycolorbox[text=\strut{But}, color=red!31.186310]%
}
\setlength{\fboxsep}{0pt}\fcolorbox{gray!10}{gray!10}{\strut
    \mycolorbox[text=\strut{let}, color=green!13.292376]%
}
\setlength{\fboxsep}{0pt}\fcolorbox{gray!10}{gray!10}{\strut
    \mycolorbox[text=\strut{me}, color=green!17.636594]%
}
\setlength{\fboxsep}{0pt}\fcolorbox{gray!10}{gray!10}{\strut
    \mycolorbox[text=\strut{check}, color=green!12.805188]%
}
\setlength{\fboxsep}{0pt}\fcolorbox{gray!10}{gray!10}{\strut
    \mycolorbox[text=\strut{again}, color=red!17.612458]%
}
\setlength{\fboxsep}{0pt}\fcolorbox{gray!10}{gray!10}{\strut
    \mycolorbox[text=\strut{if}, color=red!21.010144]%
}
\setlength{\fboxsep}{0pt}\fcolorbox{gray!10}{gray!10}{\strut
    \mycolorbox[text=\strut{I}, color=green!13.733637]%
}
\setlength{\fboxsep}{0pt}\fcolorbox{gray!10}{gray!10}{\strut
    \mycolorbox[text=\strut{interpreted}, color=red!66.006729]%
}
\setlength{\fboxsep}{0pt}\fcolorbox{gray!10}{gray!10}{\strut
    \mycolorbox[text=\strut{the}, color=green!17.636594]%
}
\setlength{\fboxsep}{0pt}\fcolorbox{gray!10}{gray!10}{\strut
    \mycolorbox[text=\strut{operation}, color=green!11.499888]%
}
\setlength{\fboxsep}{0pt}\fcolorbox{gray!10}{gray!10}{\strut
    \mycolorbox[text=\strut{correctly}, color=green!17.632950]%
    \mycolorbox[text=\strut{.}, color=green!17.379785]%
}
\\
\\
{\tiny\color{gray}14}\,%
\setlength{\fboxsep}{0pt}\fcolorbox{gray!10}{gray!10}{\strut
    \setlength{\fboxsep}{1pt}\fbox{\mycolorbox[text=\strut{Wait}]}%
    \mycolorbox[text=\strut{,}, color=green!17.443520]%
}
\setlength{\fboxsep}{0pt}\fcolorbox{gray!10}{gray!10}{\strut
    \mycolorbox[text=\strut{let}, color=red!64.224190]%
}
\setlength{\fboxsep}{0pt}\fcolorbox{gray!10}{gray!10}{\strut
    \mycolorbox[text=\strut{me}, color=green!16.600548]%
}
\setlength{\fboxsep}{0pt}\fcolorbox{gray!10}{gray!10}{\strut
    \mycolorbox[text=\strut{check}, color=red!22.204343]%
}
\setlength{\fboxsep}{0pt}\fcolorbox{gray!10}{gray!10}{\strut
    \mycolorbox[text=\strut{the}, color=red!10.402941]%
}
\setlength{\fboxsep}{0pt}\fcolorbox{gray!10}{gray!10}{\strut
    \mycolorbox[text=\strut{definition}, color=red!43.682402]%
}
\setlength{\fboxsep}{0pt}\fcolorbox{gray!10}{gray!10}{\strut
    \mycolorbox[text=\strut{again}]%
    \mycolorbox[text=\strut{.}, color=green!15.865619]%
}
\setlength{\fboxsep}{0pt}\fcolorbox{gray!10}{gray!10}{\strut
    \mycolorbox[text=\strut{The}, color=green!10.190491]%
}
\setlength{\fboxsep}{0pt}\fcolorbox{gray!10}{gray!10}{\strut
    \mycolorbox[text=\strut{operation}, color=red!16.824423]%
}
\setlength{\fboxsep}{0pt}\fcolorbox{gray!10}{gray!10}{\strut
    \mycolorbox[text=\strut{is}, color=green!9.702024]%
}
\setlength{\fboxsep}{0pt}\fcolorbox{gray!10}{gray!10}{\strut
    \mycolorbox[text=\strut{defined}, color=red!35.390049]%
}
\setlength{\fboxsep}{0pt}\fcolorbox{gray!10}{gray!10}{\strut
    \mycolorbox[text=\strut{as}, color=green!17.633295]%
}
\setlength{\fboxsep}{0pt}\fcolorbox{gray!10}{gray!10}{\strut
    \mycolorbox[text=\strut{m}, color=green!9.687388]%
    \mycolorbox[text=\strut{/n}, color=green!17.636526]%
}
\setlength{\fboxsep}{0pt}\fcolorbox{gray!10}{gray!10}{\strut
    \mycolorbox[text=\strut{@}, color=green!17.636586]%
}
\setlength{\fboxsep}{0pt}\fcolorbox{gray!10}{gray!10}{\strut
    \mycolorbox[text=\strut{p}, color=green!17.636594]%
    \mycolorbox[text=\strut{/q}, color=green!17.636594]%
}
\setlength{\fboxsep}{0pt}\fcolorbox{gray!10}{gray!10}{\strut
    \mycolorbox[text=\strut{=}, color=green!17.494992]%
}
\setlength{\fboxsep}{0pt}\fcolorbox{gray!10}{gray!10}{\strut
    \mycolorbox[text=\strut{(}, color=red!34.847021]%
    \mycolorbox[text=\strut{m}, color=green!17.636594]%
    \mycolorbox[text=\strut{)(}, color=green!17.633114]%
    \mycolorbox[text=\strut{p}, color=green!17.636594]%
    \mycolorbox[text=\strut{)(}, color=green!17.636586]%
    \mycolorbox[text=\strut{q}, color=green!17.636586]%
    \mycolorbox[text=\strut{/n}, color=green!17.605446]%
    \mycolorbox[text=\strut{).}, color=green!17.636594]%
}
\setlength{\fboxsep}{0pt}\fcolorbox{gray!10}{gray!10}{\strut
    \mycolorbox[text=\strut{So}, color=green!10.182553]%
}
\setlength{\fboxsep}{0pt}\fcolorbox{gray!10}{gray!10}{\strut
    \mycolorbox[text=\strut{that}, color=red!40.007174]%
    \mycolorbox[text=\strut{'s}, color=red!60.841087]%
}
\setlength{\fboxsep}{0pt}\fcolorbox{gray!10}{gray!10}{\strut
    \mycolorbox[text=\strut{m}, color=green!16.360909]%
    \mycolorbox[text=\strut{*p}, color=red!23.771996]%
    \mycolorbox[text=\strut{*(}, color=green!14.792313]%
    \mycolorbox[text=\strut{q}, color=green!17.636586]%
    \mycolorbox[text=\strut{/n}, color=green!17.610306]%
    \mycolorbox[text=\strut{).}, color=green!17.617553]%
}
\setlength{\fboxsep}{0pt}\fcolorbox{gray!10}{gray!10}{\strut
    \mycolorbox[text=\strut{So}, color=red!4.526581]%
}
\setlength{\fboxsep}{0pt}\fcolorbox{gray!10}{gray!10}{\strut
    \mycolorbox[text=\strut{that}, color=red!74.233341]%
}
\setlength{\fboxsep}{0pt}\fcolorbox{gray!10}{gray!10}{\strut
    \mycolorbox[text=\strut{is}, color=red!32.137651]%
}
\setlength{\fboxsep}{0pt}\fcolorbox{gray!10}{gray!10}{\strut
    \mycolorbox[text=\strut{m}, color=red!21.225816]%
    \mycolorbox[text=\strut{*p}, color=green!16.546084]%
    \mycolorbox[text=\strut{*q}, color=green!17.381667]%
}
\setlength{\fboxsep}{0pt}\fcolorbox{gray!10}{gray!10}{\strut
    \mycolorbox[text=\strut{/}, color=red!31.877191]%
}
\setlength{\fboxsep}{0pt}\fcolorbox{gray!10}{gray!10}{\strut
    \mycolorbox[text=\strut{n}, color=green!17.009447]%
    \mycolorbox[text=\strut{.}, color=green!16.151655]%
}
\setlength{\fboxsep}{0pt}\fcolorbox{gray!10}{gray!10}{\strut
    \mycolorbox[text=\strut{So}, color=red!11.410527]%
}
\setlength{\fboxsep}{0pt}\fcolorbox{gray!10}{gray!10}{\strut
    \mycolorbox[text=\strut{in}, color=red!31.967929]%
}
\setlength{\fboxsep}{0pt}\fcolorbox{gray!10}{gray!10}{\strut
    \mycolorbox[text=\strut{this}, color=green!3.354518]%
}
\setlength{\fboxsep}{0pt}\fcolorbox{gray!10}{gray!10}{\strut
    \mycolorbox[text=\strut{case}, color=green!17.633226]%
    \mycolorbox[text=\strut{,}, color=green!17.599861]%
}
\setlength{\fboxsep}{0pt}\fcolorbox{gray!10}{gray!10}{\strut
    \mycolorbox[text=\strut{m}, color=red!6.423859]%
    \mycolorbox[text=\strut{=}, color=red!29.746129]%
    \mycolorbox[text=\strut{7}, color=green!17.636594]%
    \mycolorbox[text=\strut{,}, color=green!17.636577]%
}
\setlength{\fboxsep}{0pt}\fcolorbox{gray!10}{gray!10}{\strut
    \mycolorbox[text=\strut{p}, color=green!17.617708]%
    \mycolorbox[text=\strut{=}, color=green!17.636594]%
    \mycolorbox[text=\strut{1}, color=green!17.636594]%
    \mycolorbox[text=\strut{0}, color=green!17.636594]%
    \mycolorbox[text=\strut{,}, color=green!17.636594]%
}
\setlength{\fboxsep}{0pt}\fcolorbox{gray!10}{gray!10}{\strut
    \mycolorbox[text=\strut{q}, color=green!17.636594]%
    \mycolorbox[text=\strut{=}, color=green!17.636594]%
    \mycolorbox[text=\strut{2}, color=green!17.636594]%
    \mycolorbox[text=\strut{1}, color=green!17.636594]%
    \mycolorbox[text=\strut{,}, color=green!17.636594]%
}
\setlength{\fboxsep}{0pt}\fcolorbox{gray!10}{gray!10}{\strut
    \mycolorbox[text=\strut{n}, color=green!17.635862]%
    \mycolorbox[text=\strut{=}, color=green!17.636594]%
    \mycolorbox[text=\strut{3}, color=green!17.636594]%
    \mycolorbox[text=\strut{0}, color=green!17.636594]%
    \mycolorbox[text=\strut{.}, color=green!16.924280]%
}
\setlength{\fboxsep}{0pt}\fcolorbox{gray!10}{gray!10}{\strut
    \mycolorbox[text=\strut{So}, color=red!16.944630]%
}
\setlength{\fboxsep}{0pt}\fcolorbox{gray!10}{gray!10}{\strut
    \mycolorbox[text=\strut{7}, color=green!17.636594]%
    \mycolorbox[text=\strut{*}, color=green!15.430599]%
    \mycolorbox[text=\strut{1}, color=green!17.636594]%
    \mycolorbox[text=\strut{0}, color=green!17.636594]%
    \mycolorbox[text=\strut{*}, color=green!17.636594]%
    \mycolorbox[text=\strut{2}, color=green!17.636594]%
    \mycolorbox[text=\strut{1}, color=green!17.636594]%
}
\setlength{\fboxsep}{0pt}\fcolorbox{gray!10}{gray!10}{\strut
    \mycolorbox[text=\strut{/}, color=green!15.536195]%
}
\setlength{\fboxsep}{0pt}\fcolorbox{gray!10}{gray!10}{\strut
    \mycolorbox[text=\strut{3}, color=green!17.636594]%
    \mycolorbox[text=\strut{0}, color=green!17.636594]%
    \mycolorbox[text=\strut{.}, color=green!17.063405]%
}
\setlength{\fboxsep}{0pt}\fcolorbox{gray!10}{gray!10}{\strut
    \mycolorbox[text=\strut{Which}, color=green!12.993253]%
}
\setlength{\fboxsep}{0pt}\fcolorbox{gray!10}{gray!10}{\strut
    \mycolorbox[text=\strut{is}, color=green!16.983311]%
}
\setlength{\fboxsep}{0pt}\fcolorbox{gray!10}{gray!10}{\strut
    \mycolorbox[text=\strut{indeed}, color=red!64.437539]%
}
\setlength{\fboxsep}{0pt}\fcolorbox{gray!10}{gray!10}{\strut
    \mycolorbox[text=\strut{1}, color=green!15.921165]%
    \mycolorbox[text=\strut{4}, color=green!17.636594]%
    \mycolorbox[text=\strut{7}, color=green!17.636594]%
    \mycolorbox[text=\strut{0}, color=green!17.636594]%
}
\setlength{\fboxsep}{0pt}\fcolorbox{gray!10}{gray!10}{\strut
    \mycolorbox[text=\strut{/}, color=green!15.430467]%
}
\setlength{\fboxsep}{0pt}\fcolorbox{gray!10}{gray!10}{\strut
    \mycolorbox[text=\strut{3}, color=green!17.636594]%
    \mycolorbox[text=\strut{0}, color=green!17.636594]%
}
\setlength{\fboxsep}{0pt}\fcolorbox{gray!10}{gray!10}{\strut
    \mycolorbox[text=\strut{=}, color=green!15.786060]%
}
\setlength{\fboxsep}{0pt}\fcolorbox{gray!10}{gray!10}{\strut
    \mycolorbox[text=\strut{4}, color=green!17.636594]%
    \mycolorbox[text=\strut{9}, color=green!17.636594]%
    \mycolorbox[text=\strut{.}, color=green!17.309598]%
}
\setlength{\fboxsep}{0pt}\fcolorbox{gray!10}{gray!10}{\strut
    \mycolorbox[text=\strut{So}, color=green!14.670028]%
}
\setlength{\fboxsep}{0pt}\fcolorbox{gray!10}{gray!10}{\strut
    \mycolorbox[text=\strut{that}, color=green!11.807875]%
    \mycolorbox[text=\strut{'s}, color=red!41.536305]%
}
\setlength{\fboxsep}{0pt}\fcolorbox{gray!10}{gray!10}{\strut
    \mycolorbox[text=\strut{correct}, color=green!16.701132]%
    \mycolorbox[text=\strut{.}, color=red!22.680247]%
}
\\
\\
{\tiny\color{gray}15}\,%
\setlength{\fboxsep}{0pt}\fcolorbox{gray!10}{gray!10}{\strut
    \mycolorbox[text=\strut{Alternatively}, color=red!41.973566]%
    \mycolorbox[text=\strut{,}, color=green!17.636594]%
}
\setlength{\fboxsep}{0pt}\fcolorbox{gray!10}{gray!10}{\strut
    \mycolorbox[text=\strut{maybe}, color=green!7.368001]%
}
\setlength{\fboxsep}{0pt}\fcolorbox{gray!10}{gray!10}{\strut
    \mycolorbox[text=\strut{there}, color=red!42.661545]%
}
\setlength{\fboxsep}{0pt}\fcolorbox{gray!10}{gray!10}{\strut
    \mycolorbox[text=\strut{is}, color=red!52.475953]%
}
\setlength{\fboxsep}{0pt}\fcolorbox{gray!10}{gray!10}{\strut
    \mycolorbox[text=\strut{a}, color=green!6.471055]%
}
\setlength{\fboxsep}{0pt}\fcolorbox{gray!10}{gray!10}{\strut
    \mycolorbox[text=\strut{different}, color=red!19.879861]%
}
\setlength{\fboxsep}{0pt}\fcolorbox{gray!10}{gray!10}{\strut
    \mycolorbox[text=\strut{interpretation}, color=red!5.382781]%
    \mycolorbox[text=\strut{?}, color=red!27.587167]%
}
\setlength{\fboxsep}{0pt}\fcolorbox{gray!10}{gray!10}{\strut
    \mycolorbox[text=\strut{For}, color=red!44.253779]%
}
\setlength{\fboxsep}{0pt}\fcolorbox{gray!10}{gray!10}{\strut
    \mycolorbox[text=\strut{example}, color=green!15.430634]%
    \mycolorbox[text=\strut{,}, color=green!17.636594]%
}
\setlength{\fboxsep}{0pt}\fcolorbox{gray!10}{gray!10}{\strut
    \mycolorbox[text=\strut{maybe}, color=red!19.864738]%
}
\setlength{\fboxsep}{0pt}\fcolorbox{gray!10}{gray!10}{\strut
    \mycolorbox[text=\strut{the}, color=green!13.574091]%
}
\setlength{\fboxsep}{0pt}\fcolorbox{gray!10}{gray!10}{\strut
    \mycolorbox[text=\strut{operation}, color=green!16.641563]%
}
\setlength{\fboxsep}{0pt}\fcolorbox{gray!10}{gray!10}{\strut
    \mycolorbox[text=\strut{is}, color=green!17.425030]%
}
\setlength{\fboxsep}{0pt}\fcolorbox{gray!10}{gray!10}{\strut
    \mycolorbox[text=\strut{(}, color=red!45.459813]%
    \mycolorbox[text=\strut{m}, color=green!17.635776]%
    \mycolorbox[text=\strut{/n}, color=green!17.245902]%
    \mycolorbox[text=\strut{)}, color=green!17.596112]%
}
\setlength{\fboxsep}{0pt}\fcolorbox{gray!10}{gray!10}{\strut
    \mycolorbox[text=\strut{@}, color=green!17.556467]%
}
\setlength{\fboxsep}{0pt}\fcolorbox{gray!10}{gray!10}{\strut
    \mycolorbox[text=\strut{(}, color=green!17.636569]%
    \mycolorbox[text=\strut{p}, color=green!17.636594]%
    \mycolorbox[text=\strut{/q}, color=green!17.636594]%
    \mycolorbox[text=\strut{)}, color=green!17.636594]%
}
\setlength{\fboxsep}{0pt}\fcolorbox{gray!10}{gray!10}{\strut
    \mycolorbox[text=\strut{=}, color=green!16.367652]%
}
\setlength{\fboxsep}{0pt}\fcolorbox{gray!10}{gray!10}{\strut
    \mycolorbox[text=\strut{m}, color=red!6.751551]%
}
\setlength{\fboxsep}{0pt}\fcolorbox{gray!10}{gray!10}{\strut
    \mycolorbox[text=\strut{*}, color=red!16.729461]%
}
\setlength{\fboxsep}{0pt}\fcolorbox{gray!10}{gray!10}{\strut
    \mycolorbox[text=\strut{p}, color=green!17.340697]%
}
\setlength{\fboxsep}{0pt}\fcolorbox{gray!10}{gray!10}{\strut
    \mycolorbox[text=\strut{*}, color=green!17.605420]%
}
\setlength{\fboxsep}{0pt}\fcolorbox{gray!10}{gray!10}{\strut
    \mycolorbox[text=\strut{(}, color=green!17.633303]%
    \mycolorbox[text=\strut{q}, color=green!17.635862]%
}
\setlength{\fboxsep}{0pt}\fcolorbox{gray!10}{gray!10}{\strut
    \mycolorbox[text=\strut{/}, color=red!5.321679]%
}
\setlength{\fboxsep}{0pt}\fcolorbox{gray!10}{gray!10}{\strut
    \mycolorbox[text=\strut{n}, color=green!17.629650]%
    \mycolorbox[text=\strut{).}, color=green!16.774792]%
}
\setlength{\fboxsep}{0pt}\fcolorbox{gray!10}{gray!10}{\strut
    \mycolorbox[text=\strut{Which}, color=red!16.456946]%
}
\setlength{\fboxsep}{0pt}\fcolorbox{gray!10}{gray!10}{\strut
    \mycolorbox[text=\strut{is}, color=green!17.080227]%
}
\setlength{\fboxsep}{0pt}\fcolorbox{gray!10}{gray!10}{\strut
    \mycolorbox[text=\strut{the}, color=red!23.844838]%
}
\setlength{\fboxsep}{0pt}\fcolorbox{gray!10}{gray!10}{\strut
    \mycolorbox[text=\strut{same}, color=green!17.636379]%
}
\setlength{\fboxsep}{0pt}\fcolorbox{gray!10}{gray!10}{\strut
    \mycolorbox[text=\strut{thing}, color=red!57.583085]%
    \mycolorbox[text=\strut{.}, color=green!12.956047]%
}
\setlength{\fboxsep}{0pt}\fcolorbox{gray!10}{gray!10}{\strut
    \mycolorbox[text=\strut{So}, color=green!16.271826]%
}
\setlength{\fboxsep}{0pt}\fcolorbox{gray!10}{gray!10}{\strut
    \mycolorbox[text=\strut{that}, color=red!38.171968]%
    \mycolorbox[text=\strut{'s}, color=red!32.313677]%
}
\setlength{\fboxsep}{0pt}\fcolorbox{gray!10}{gray!10}{\strut
    \mycolorbox[text=\strut{7}, color=green!17.635793]%
}
\setlength{\fboxsep}{0pt}\fcolorbox{gray!10}{gray!10}{\strut
    \mycolorbox[text=\strut{*}, color=red!40.828491]%
}
\setlength{\fboxsep}{0pt}\fcolorbox{gray!10}{gray!10}{\strut
    \mycolorbox[text=\strut{1}, color=green!17.636594]%
    \mycolorbox[text=\strut{0}, color=green!17.636594]%
}
\setlength{\fboxsep}{0pt}\fcolorbox{gray!10}{gray!10}{\strut
    \mycolorbox[text=\strut{*}, color=green!17.636431]%
}
\setlength{\fboxsep}{0pt}\fcolorbox{gray!10}{gray!10}{\strut
    \mycolorbox[text=\strut{(}, color=green!17.596612]%
    \mycolorbox[text=\strut{2}, color=green!17.636586]%
    \mycolorbox[text=\strut{1}, color=green!17.636594]%
}
\setlength{\fboxsep}{0pt}\fcolorbox{gray!10}{gray!10}{\strut
    \mycolorbox[text=\strut{/}, color=green!16.304299]%
}
\setlength{\fboxsep}{0pt}\fcolorbox{gray!10}{gray!10}{\strut
    \mycolorbox[text=\strut{3}, color=green!17.636594]%
    \mycolorbox[text=\strut{0}, color=green!17.636594]%
    \mycolorbox[text=\strut{)}, color=red!62.630366]%
}
\setlength{\fboxsep}{0pt}\fcolorbox{gray!10}{gray!10}{\strut
    \mycolorbox[text=\strut{=}, color=green!16.003251]%
}
\setlength{\fboxsep}{0pt}\fcolorbox{gray!10}{gray!10}{\strut
    \mycolorbox[text=\strut{7}, color=green!16.125096]%
    \mycolorbox[text=\strut{*}, color=red!54.404983]%
    \mycolorbox[text=\strut{1}, color=green!17.636594]%
    \mycolorbox[text=\strut{0}, color=green!17.636594]%
    \mycolorbox[text=\strut{*(}, color=red!5.323287]%
    \mycolorbox[text=\strut{7}, color=green!17.633097]%
    \mycolorbox[text=\strut{/}, color=green!17.636491]%
    \mycolorbox[text=\strut{1}, color=green!17.636594]%
    \mycolorbox[text=\strut{0}, color=green!17.636594]%
    \mycolorbox[text=\strut{)}, color=green!16.596302]%
}
\setlength{\fboxsep}{0pt}\fcolorbox{gray!10}{gray!10}{\strut
    \mycolorbox[text=\strut{=}, color=green!17.615743]%
}
\setlength{\fboxsep}{0pt}\fcolorbox{gray!10}{gray!10}{\strut
    \mycolorbox[text=\strut{7}, color=green!17.340827]%
    \mycolorbox[text=\strut{*}, color=green!17.617260]%
    \mycolorbox[text=\strut{7}, color=green!17.625144]%
}
\setlength{\fboxsep}{0pt}\fcolorbox{gray!10}{gray!10}{\strut
    \mycolorbox[text=\strut{=}, color=green!4.134656]%
}
\setlength{\fboxsep}{0pt}\fcolorbox{gray!10}{gray!10}{\strut
    \mycolorbox[text=\strut{4}, color=green!17.636594]%
    \mycolorbox[text=\strut{9}, color=green!17.636594]%
    \mycolorbox[text=\strut{.}, color=green!17.572330]%
}
\setlength{\fboxsep}{0pt}\fcolorbox{gray!10}{gray!10}{\strut
    \mycolorbox[text=\strut{So}, color=red!35.276982]%
}
\setlength{\fboxsep}{0pt}\fcolorbox{gray!10}{gray!10}{\strut
    \mycolorbox[text=\strut{that}, color=red!12.448974]%
    \mycolorbox[text=\strut{'s}, color=green!7.580357]%
}
\setlength{\fboxsep}{0pt}\fcolorbox{gray!10}{gray!10}{\strut
    \mycolorbox[text=\strut{the}, color=red!51.889210]%
}
\setlength{\fboxsep}{0pt}\fcolorbox{gray!10}{gray!10}{\strut
    \mycolorbox[text=\strut{same}, color=green!17.631503]%
    \mycolorbox[text=\strut{.}, color=red!53.535638]%
}
\setlength{\fboxsep}{0pt}\fcolorbox{gray!10}{gray!10}{\strut
    \mycolorbox[text=\strut{So}, color=green!6.438133]%
}
\setlength{\fboxsep}{0pt}\fcolorbox{gray!10}{gray!10}{\strut
    \mycolorbox[text=\strut{that}, color=red!17.261938]%
}
\setlength{\fboxsep}{0pt}\fcolorbox{gray!10}{gray!10}{\strut
    \mycolorbox[text=\strut{seems}, color=red!22.011934]%
}
\setlength{\fboxsep}{0pt}\fcolorbox{gray!10}{gray!10}{\strut
    \mycolorbox[text=\strut{to}, color=red!54.899972]%
}
\setlength{\fboxsep}{0pt}\fcolorbox{gray!10}{gray!10}{\strut
    \mycolorbox[text=\strut{check}, color=green!4.734082]%
}
\setlength{\fboxsep}{0pt}\fcolorbox{gray!10}{gray!10}{\strut
    \mycolorbox[text=\strut{out}, color=green!17.627565]%
    \mycolorbox[text=\strut{.}]%
}
\\
\\
{\tiny\color{gray}16}\,%
\setlength{\fboxsep}{0pt}\fcolorbox{gray!10}{gray!10}{\strut
    \mycolorbox[text=\strut{Alternatively}, color=red!28.961451]%
    \mycolorbox[text=\strut{,}, color=green!17.636594]%
}
\setlength{\fboxsep}{0pt}\fcolorbox{gray!10}{gray!10}{\strut
    \mycolorbox[text=\strut{maybe}, color=red!9.444025]%
}
\setlength{\fboxsep}{0pt}\fcolorbox{gray!10}{gray!10}{\strut
    \mycolorbox[text=\strut{the}, color=red!37.410474]%
}
\setlength{\fboxsep}{0pt}\fcolorbox{gray!10}{gray!10}{\strut
    \mycolorbox[text=\strut{operation}, color=red!12.322015]%
}
\setlength{\fboxsep}{0pt}\fcolorbox{gray!10}{gray!10}{\strut
    \mycolorbox[text=\strut{is}, color=green!16.257009]%
}
\setlength{\fboxsep}{0pt}\fcolorbox{gray!10}{gray!10}{\strut
    \mycolorbox[text=\strut{(}, color=red!58.817928]%
    \mycolorbox[text=\strut{m}, color=green!17.627212]%
    \mycolorbox[text=\strut{/n}, color=green!17.076823]%
    \mycolorbox[text=\strut{)}, color=green!17.325222]%
}
\setlength{\fboxsep}{0pt}\fcolorbox{gray!10}{gray!10}{\strut
    \mycolorbox[text=\strut{@}, color=green!16.992319]%
}
\setlength{\fboxsep}{0pt}\fcolorbox{gray!10}{gray!10}{\strut
    \mycolorbox[text=\strut{(}, color=green!17.636465]%
    \mycolorbox[text=\strut{p}, color=green!17.636594]%
    \mycolorbox[text=\strut{/q}, color=green!17.636594]%
    \mycolorbox[text=\strut{)}, color=green!17.636586]%
}
\setlength{\fboxsep}{0pt}\fcolorbox{gray!10}{gray!10}{\strut
    \mycolorbox[text=\strut{=}, color=green!17.443106]%
}
\setlength{\fboxsep}{0pt}\fcolorbox{gray!10}{gray!10}{\strut
    \mycolorbox[text=\strut{(}, color=red!47.186873]%
    \mycolorbox[text=\strut{m}, color=green!17.632657]%
}
\setlength{\fboxsep}{0pt}\fcolorbox{gray!10}{gray!10}{\strut
    \mycolorbox[text=\strut{*}, color=green!7.135762]%
}
\setlength{\fboxsep}{0pt}\fcolorbox{gray!10}{gray!10}{\strut
    \mycolorbox[text=\strut{p}, color=green!17.636491]%
    \mycolorbox[text=\strut{)}, color=green!16.571775]%
}
\setlength{\fboxsep}{0pt}\fcolorbox{gray!10}{gray!10}{\strut
    \mycolorbox[text=\strut{/}, color=green!17.547724]%
}
\setlength{\fboxsep}{0pt}\fcolorbox{gray!10}{gray!10}{\strut
    \mycolorbox[text=\strut{(}, color=green!17.636594]%
    \mycolorbox[text=\strut{n}, color=green!17.636586]%
}
\setlength{\fboxsep}{0pt}\fcolorbox{gray!10}{gray!10}{\strut
    \mycolorbox[text=\strut{/}, color=red!29.500721]%
}
\setlength{\fboxsep}{0pt}\fcolorbox{gray!10}{gray!10}{\strut
    \mycolorbox[text=\strut{q}, color=green!17.636594]%
    \mycolorbox[text=\strut{)?}, color=red!21.513063]%
}
\setlength{\fboxsep}{0pt}\fcolorbox{gray!10}{gray!10}{\strut
    \mycolorbox[text=\strut{But}, color=red!29.822675]%
}
\setlength{\fboxsep}{0pt}\fcolorbox{gray!10}{gray!10}{\strut
    \mycolorbox[text=\strut{that}, color=green!17.288586]%
}
\setlength{\fboxsep}{0pt}\fcolorbox{gray!10}{gray!10}{\strut
    \mycolorbox[text=\strut{would}, color=green!17.617346]%
}
\setlength{\fboxsep}{0pt}\fcolorbox{gray!10}{gray!10}{\strut
    \mycolorbox[text=\strut{be}, color=green!17.612090]%
}
\setlength{\fboxsep}{0pt}\fcolorbox{gray!10}{gray!10}{\strut
    \mycolorbox[text=\strut{different}, color=green!11.249823]%
    \mycolorbox[text=\strut{.}, color=green!15.922312]%
}
\setlength{\fboxsep}{0pt}\fcolorbox{gray!10}{gray!10}{\strut
    \setlength{\fboxsep}{1pt}\fbox{\mycolorbox[text=\strut{Wait}, color=red!39.771071]}%
    \mycolorbox[text=\strut{,}, color=green!17.623361]%
}
\setlength{\fboxsep}{0pt}\fcolorbox{gray!10}{gray!10}{\strut
    \mycolorbox[text=\strut{but}, color=red!3.503821]%
}
\setlength{\fboxsep}{0pt}\fcolorbox{gray!10}{gray!10}{\strut
    \mycolorbox[text=\strut{according}, color=red!37.443386]%
}
\setlength{\fboxsep}{0pt}\fcolorbox{gray!10}{gray!10}{\strut
    \mycolorbox[text=\strut{to}, color=green!17.636594]%
}
\setlength{\fboxsep}{0pt}\fcolorbox{gray!10}{gray!10}{\strut
    \mycolorbox[text=\strut{the}, color=green!17.635724]%
}
\setlength{\fboxsep}{0pt}\fcolorbox{gray!10}{gray!10}{\strut
    \mycolorbox[text=\strut{problem}, color=green!16.945756]%
}
\setlength{\fboxsep}{0pt}\fcolorbox{gray!10}{gray!10}{\strut
    \mycolorbox[text=\strut{statement}, color=green!9.439963]%
    \mycolorbox[text=\strut{,}, color=green!16.924019]%
}
\setlength{\fboxsep}{0pt}\fcolorbox{gray!10}{gray!10}{\strut
    \mycolorbox[text=\strut{the}, color=red!76.069935]%
}
\setlength{\fboxsep}{0pt}\fcolorbox{gray!10}{gray!10}{\strut
    \mycolorbox[text=\strut{operation}, color=red!22.895903]%
}
\setlength{\fboxsep}{0pt}\fcolorbox{gray!10}{gray!10}{\strut
    \mycolorbox[text=\strut{is}, color=green!17.633622]%
}
\setlength{\fboxsep}{0pt}\fcolorbox{gray!10}{gray!10}{\strut
    \mycolorbox[text=\strut{defined}, color=red!22.051669]%
}
\setlength{\fboxsep}{0pt}\fcolorbox{gray!10}{gray!10}{\strut
    \mycolorbox[text=\strut{as}, color=green!17.636586]%
}
\setlength{\fboxsep}{0pt}\fcolorbox{gray!10}{gray!10}{\strut
    \mycolorbox[text=\strut{m}, color=red!46.688335]%
    \mycolorbox[text=\strut{/n}, color=green!17.624877]%
}
\setlength{\fboxsep}{0pt}\fcolorbox{gray!10}{gray!10}{\strut
    \mycolorbox[text=\strut{@}, color=green!17.636508]%
}
\setlength{\fboxsep}{0pt}\fcolorbox{gray!10}{gray!10}{\strut
    \mycolorbox[text=\strut{p}, color=green!17.636586]%
    \mycolorbox[text=\strut{/q}, color=green!17.636594]%
}
\setlength{\fboxsep}{0pt}\fcolorbox{gray!10}{gray!10}{\strut
    \mycolorbox[text=\strut{=}, color=green!17.492758]%
}
\setlength{\fboxsep}{0pt}\fcolorbox{gray!10}{gray!10}{\strut
    \mycolorbox[text=\strut{(}, color=green!17.009369]%
    \mycolorbox[text=\strut{m}, color=green!17.636594]%
    \mycolorbox[text=\strut{)(}, color=green!17.636224]%
    \mycolorbox[text=\strut{p}, color=green!17.636594]%
    \mycolorbox[text=\strut{)(}, color=green!17.636577]%
    \mycolorbox[text=\strut{q}, color=green!17.636586]%
    \mycolorbox[text=\strut{/n}, color=green!17.632381]%
    \mycolorbox[text=\strut{).}, color=green!17.605463]%
}
\setlength{\fboxsep}{0pt}\fcolorbox{gray!10}{gray!10}{\strut
    \mycolorbox[text=\strut{So}, color=green!9.627691]%
}
\setlength{\fboxsep}{0pt}\fcolorbox{gray!10}{gray!10}{\strut
    \mycolorbox[text=\strut{that}]%
}
\setlength{\fboxsep}{0pt}\fcolorbox{gray!10}{gray!10}{\strut
    \mycolorbox[text=\strut{is}, color=red!29.085533]%
}
\setlength{\fboxsep}{0pt}\fcolorbox{gray!10}{gray!10}{\strut
    \mycolorbox[text=\strut{m}, color=green!13.680336]%
    \mycolorbox[text=\strut{*p}, color=red!3.221324]%
    \mycolorbox[text=\strut{*(}, color=green!10.836401]%
    \mycolorbox[text=\strut{q}, color=green!17.636569]%
    \mycolorbox[text=\strut{/n}, color=green!17.608582]%
    \mycolorbox[text=\strut{).}, color=green!15.422577]%
}
\setlength{\fboxsep}{0pt}\fcolorbox{gray!10}{gray!10}{\strut
    \mycolorbox[text=\strut{Which}, color=red!62.862146]%
}
\setlength{\fboxsep}{0pt}\fcolorbox{gray!10}{gray!10}{\strut
    \mycolorbox[text=\strut{is}, color=green!15.464158]%
}
\setlength{\fboxsep}{0pt}\fcolorbox{gray!10}{gray!10}{\strut
    \mycolorbox[text=\strut{the}, color=red!64.626659]%
}
\setlength{\fboxsep}{0pt}\fcolorbox{gray!10}{gray!10}{\strut
    \mycolorbox[text=\strut{same}, color=green!17.612770]%
}

}}}}}
\vspace{-.1in}
\caption{Token-level visualization for Case 2: Qwen-4B (Part 1/3)}
\label{fig:case_2_1}
\vspace{-.1in}
\end{figure*}

\endgroup

%% file: case/wait2.tex
\begin{figure*}[htbp]
\centering
{\noindent

{\scriptsize
\setlength{\fboxrule}{.5pt}\fcolorbox{black}{green!10}{\parbox{1\textwidth}{\textbf{Model:} Qwen-4B \\ \textit{(Continued from Part 1/3 of Case 2)}}}

{\scriptsize
\setlength{\fboxrule}{.5pt}\fcolorbox{black}{gray!10}{\parbox{\textwidth}{
{\tiny\color{gray}17}\,%
\setlength{\fboxsep}{0pt}\fcolorbox{gray!10}{gray!10}{\strut
    \mycolorbox[text=\textbf{Response (cont.):}]%%
}
\setlength{\fboxsep}{0pt}\fcolorbox{gray!10}{gray!10}{\strut
    \mycolorbox[text=\strut{as}, color=green!17.617708]%
}
\setlength{\fboxsep}{0pt}\fcolorbox{gray!10}{gray!10}{\strut
    \mycolorbox[text=\strut{m}, color=red!22.901229]%
    \mycolorbox[text=\strut{*p}, color=green!17.459348]%
    \mycolorbox[text=\strut{*q}, color=green!17.633295]%
}
\setlength{\fboxsep}{0pt}\fcolorbox{gray!10}{gray!10}{\strut
    \mycolorbox[text=\strut{/}, color=green!11.502635]%
}
\setlength{\fboxsep}{0pt}\fcolorbox{gray!10}{gray!10}{\strut
    \mycolorbox[text=\strut{n}, color=green!15.922777]%
    \mycolorbox[text=\strut{.}, color=red!6.662276]%
}
\setlength{\fboxsep}{0pt}\fcolorbox{gray!10}{gray!10}{\strut
    \mycolorbox[text=\strut{So}, color=green!10.009956]%
}
\setlength{\fboxsep}{0pt}\fcolorbox{gray!10}{gray!10}{\strut
    \mycolorbox[text=\strut{that}, color=red!10.547636]%
}
\setlength{\fboxsep}{0pt}\fcolorbox{gray!10}{gray!10}{\strut
    \mycolorbox[text=\strut{is}, color=red!49.432909]%
}
\setlength{\fboxsep}{0pt}\fcolorbox{gray!10}{gray!10}{\strut
    \mycolorbox[text=\strut{correct}, color=red!51.008624]%
    \mycolorbox[text=\strut{.}, color=green!7.312620]%
}
\\
\\
{\tiny\color{gray}18}\,%
\setlength{\fboxsep}{0pt}\fcolorbox{gray!10}{gray!10}{\strut
    \mycolorbox[text=\strut{Alternatively}, color=red!7.426986]%
    \mycolorbox[text=\strut{,}, color=green!17.636594]%
}
\setlength{\fboxsep}{0pt}\fcolorbox{gray!10}{gray!10}{\strut
    \mycolorbox[text=\strut{maybe}, color=red!21.557987]%
}
\setlength{\fboxsep}{0pt}\fcolorbox{gray!10}{gray!10}{\strut
    \mycolorbox[text=\strut{I}, color=red!70.188732]%
}
\setlength{\fboxsep}{0pt}\fcolorbox{gray!10}{gray!10}{\strut
    \mycolorbox[text=\strut{mis}, color=red!41.453909]%
    \mycolorbox[text=\strut{read}, color=green!17.344706]%
}
\setlength{\fboxsep}{0pt}\fcolorbox{gray!10}{gray!10}{\strut
    \mycolorbox[text=\strut{the}, color=green!17.406442]%
}
\setlength{\fboxsep}{0pt}\fcolorbox{gray!10}{gray!10}{\strut
    \mycolorbox[text=\strut{definition}, color=red!66.128462]%
    \mycolorbox[text=\strut{.}, color=red!10.912035]%
}
\setlength{\fboxsep}{0pt}\fcolorbox{gray!10}{gray!10}{\strut
    \mycolorbox[text=\strut{Let}, color=green!11.643945]%
}
\setlength{\fboxsep}{0pt}\fcolorbox{gray!10}{gray!10}{\strut
    \mycolorbox[text=\strut{me}, color=green!17.636388]%
}
\setlength{\fboxsep}{0pt}\fcolorbox{gray!10}{gray!10}{\strut
    \mycolorbox[text=\strut{check}, color=green!16.850336]%
}
\setlength{\fboxsep}{0pt}\fcolorbox{gray!10}{gray!10}{\strut
    \mycolorbox[text=\strut{again}]%
    \mycolorbox[text=\strut{.}, color=red!76.223362]%
}
\\
\\
{\tiny\color{gray}19}\,%
\setlength{\fboxsep}{0pt}\fcolorbox{gray!10}{gray!10}{\strut
    \mycolorbox[text=\strut{Original}, color=green!13.578922]%
}
\setlength{\fboxsep}{0pt}\fcolorbox{gray!10}{gray!10}{\strut
    \mycolorbox[text=\strut{problem}, color=green!16.801972]%
    \mycolorbox[text=\strut{:}, color=red!9.315414]%
}
\setlength{\fboxsep}{0pt}\fcolorbox{gray!10}{gray!10}{\strut
    \mycolorbox[text=\strut{"}, color=green!16.304054]%
    \mycolorbox[text=\strut{The}, color=green!17.636388]%
}
\setlength{\fboxsep}{0pt}\fcolorbox{gray!10}{gray!10}{\strut
    \mycolorbox[text=\strut{operation}, color=green!17.636594]%
}
\setlength{\fboxsep}{0pt}\fcolorbox{gray!10}{gray!10}{\strut
    \mycolorbox[text=\strut{@}, color=green!17.636594]%
}
\setlength{\fboxsep}{0pt}\fcolorbox{gray!10}{gray!10}{\strut
    \mycolorbox[text=\strut{is}, color=green!17.636594]%
}
\setlength{\fboxsep}{0pt}\fcolorbox{gray!10}{gray!10}{\strut
    \mycolorbox[text=\strut{defined}, color=green!17.636594]%
}
\setlength{\fboxsep}{0pt}\fcolorbox{gray!10}{gray!10}{\strut
    \mycolorbox[text=\strut{as}, color=green!17.636594]%
}
\setlength{\fboxsep}{0pt}\fcolorbox{gray!10}{gray!10}{\strut
    \mycolorbox[text=\strut{m}, color=green!16.002393]%
    \mycolorbox[text=\strut{/n}, color=green!17.636594]%
}
\setlength{\fboxsep}{0pt}\fcolorbox{gray!10}{gray!10}{\strut
    \mycolorbox[text=\strut{@}, color=green!17.636586]%
}
\setlength{\fboxsep}{0pt}\fcolorbox{gray!10}{gray!10}{\strut
    \mycolorbox[text=\strut{p}, color=green!17.636594]%
    \mycolorbox[text=\strut{/q}, color=green!17.636594]%
}
\setlength{\fboxsep}{0pt}\fcolorbox{gray!10}{gray!10}{\strut
    \mycolorbox[text=\strut{=}, color=green!17.636594]%
}
\setlength{\fboxsep}{0pt}\fcolorbox{gray!10}{gray!10}{\strut
    \mycolorbox[text=\strut{(}, color=green!17.636577]%
    \mycolorbox[text=\strut{m}, color=green!17.636594]%
    \mycolorbox[text=\strut{)(}, color=green!17.636594]%
    \mycolorbox[text=\strut{p}, color=green!17.636594]%
    \mycolorbox[text=\strut{)(}, color=green!17.636569]%
    \mycolorbox[text=\strut{q}, color=green!17.636586]%
    \mycolorbox[text=\strut{/n}, color=green!17.636534]%
    \mycolorbox[text=\strut{)}, color=green!17.636439]%
}
\setlength{\fboxsep}{0pt}\fcolorbox{gray!10}{gray!10}{\strut
    \mycolorbox[text=\strut{for}, color=green!17.636594]%
}
\setlength{\fboxsep}{0pt}\fcolorbox{gray!10}{gray!10}{\strut
    \mycolorbox[text=\strut{simplified}, color=green!17.636594]%
}
\setlength{\fboxsep}{0pt}\fcolorbox{gray!10}{gray!10}{\strut
    \mycolorbox[text=\strut{fractions}, color=green!17.636594]%
}
\setlength{\fboxsep}{0pt}\fcolorbox{gray!10}{gray!10}{\strut
    \mycolorbox[text=\strut{p}, color=green!17.636560]%
    \mycolorbox[text=\strut{/q}, color=green!17.636594]%
    \mycolorbox[text=\strut{."}, color=green!3.860062]%
}
\\
\\
{\tiny\color{gray}20}\,%
\setlength{\fboxsep}{0pt}\fcolorbox{gray!10}{gray!10}{\strut
    \mycolorbox[text=\strut{So}]%
    \mycolorbox[text=\strut{,}, color=red!60.703913]%
}
\setlength{\fboxsep}{0pt}\fcolorbox{gray!10}{gray!10}{\strut
    \mycolorbox[text=\strut{the}, color=red!34.340006]%
}
\setlength{\fboxsep}{0pt}\fcolorbox{gray!10}{gray!10}{\strut
    \mycolorbox[text=\strut{operation}, color=red!13.285816]%
}
\setlength{\fboxsep}{0pt}\fcolorbox{gray!10}{gray!10}{\strut
    \mycolorbox[text=\strut{is}, color=green!9.826871]%
}
\setlength{\fboxsep}{0pt}\fcolorbox{gray!10}{gray!10}{\strut
    \mycolorbox[text=\strut{m}, color=red!51.767052]%
    \mycolorbox[text=\strut{/n}, color=green!17.612960]%
}
\setlength{\fboxsep}{0pt}\fcolorbox{gray!10}{gray!10}{\strut
    \mycolorbox[text=\strut{@}, color=green!17.631124]%
}
\setlength{\fboxsep}{0pt}\fcolorbox{gray!10}{gray!10}{\strut
    \mycolorbox[text=\strut{p}, color=green!17.636594]%
    \mycolorbox[text=\strut{/q}, color=green!17.636594]%
}
\setlength{\fboxsep}{0pt}\fcolorbox{gray!10}{gray!10}{\strut
    \mycolorbox[text=\strut{equals}, color=green!3.101861]%
}
\setlength{\fboxsep}{0pt}\fcolorbox{gray!10}{gray!10}{\strut
    \mycolorbox[text=\strut{m}, color=green!15.892923]%
}
\setlength{\fboxsep}{0pt}\fcolorbox{gray!10}{gray!10}{\strut
    \mycolorbox[text=\strut{times}, color=red!29.258663]%
}
\setlength{\fboxsep}{0pt}\fcolorbox{gray!10}{gray!10}{\strut
    \mycolorbox[text=\strut{p}, color=green!17.636594]%
}
\setlength{\fboxsep}{0pt}\fcolorbox{gray!10}{gray!10}{\strut
    \mycolorbox[text=\strut{times}, color=green!17.636594]%
}
\setlength{\fboxsep}{0pt}\fcolorbox{gray!10}{gray!10}{\strut
    \mycolorbox[text=\strut{(}, color=green!15.922549]%
    \mycolorbox[text=\strut{q}, color=green!17.634604]%
}
\setlength{\fboxsep}{0pt}\fcolorbox{gray!10}{gray!10}{\strut
    \mycolorbox[text=\strut{over}, color=red!24.212272]%
}
\setlength{\fboxsep}{0pt}\fcolorbox{gray!10}{gray!10}{\strut
    \mycolorbox[text=\strut{n}, color=green!17.636594]%
    \mycolorbox[text=\strut{).}, color=green!17.635759]%
}
\setlength{\fboxsep}{0pt}\fcolorbox{gray!10}{gray!10}{\strut
    \mycolorbox[text=\strut{So}, color=green!14.987860]%
}
\setlength{\fboxsep}{0pt}\fcolorbox{gray!10}{gray!10}{\strut
    \mycolorbox[text=\strut{that}, color=red!38.644401]%
}
\setlength{\fboxsep}{0pt}\fcolorbox{gray!10}{gray!10}{\strut
    \mycolorbox[text=\strut{is}, color=red!10.628737]%
}
\setlength{\fboxsep}{0pt}\fcolorbox{gray!10}{gray!10}{\strut
    \mycolorbox[text=\strut{m}, color=green!7.319477]%
    \mycolorbox[text=\strut{*p}, color=green!8.515298]%
    \mycolorbox[text=\strut{*(}, color=green!4.134113]%
    \mycolorbox[text=\strut{q}, color=green!17.636586]%
    \mycolorbox[text=\strut{/n}, color=green!17.619311]%
    \mycolorbox[text=\strut{).}, color=green!17.592328]%
}
\setlength{\fboxsep}{0pt}\fcolorbox{gray!10}{gray!10}{\strut
    \mycolorbox[text=\strut{So}, color=red!10.218895]%
}
\setlength{\fboxsep}{0pt}\fcolorbox{gray!10}{gray!10}{\strut
    \mycolorbox[text=\strut{yes}, color=red!50.919806]%
    \mycolorbox[text=\strut{,}, color=green!8.572527]%
}
\setlength{\fboxsep}{0pt}\fcolorbox{gray!10}{gray!10}{\strut
    \mycolorbox[text=\strut{that}, color=red!14.081687]%
    \mycolorbox[text=\strut{'s}, color=red!36.974469]%
}
\setlength{\fboxsep}{0pt}\fcolorbox{gray!10}{gray!10}{\strut
    \mycolorbox[text=\strut{correct}, color=red!75.866191]%
    \mycolorbox[text=\strut{.}, color=red!22.680214]%
}
\\
\\
{\tiny\color{gray}21}\,%
\setlength{\fboxsep}{0pt}\fcolorbox{gray!10}{gray!10}{\strut
    \mycolorbox[text=\strut{So}, color=red!53.397616]%
}
\setlength{\fboxsep}{0pt}\fcolorbox{gray!10}{gray!10}{\strut
    \mycolorbox[text=\strut{in}, color=red!79.041449]%
}
\setlength{\fboxsep}{0pt}\fcolorbox{gray!10}{gray!10}{\strut
    \mycolorbox[text=\strut{this}, color=red!24.828887]%
}
\setlength{\fboxsep}{0pt}\fcolorbox{gray!10}{gray!10}{\strut
    \mycolorbox[text=\strut{case}, color=green!17.001689]%
    \mycolorbox[text=\strut{,}, color=green!17.636414]%
}
\setlength{\fboxsep}{0pt}\fcolorbox{gray!10}{gray!10}{\strut
    \mycolorbox[text=\strut{7}, color=green!17.636431]%
    \mycolorbox[text=\strut{/}, color=green!17.582459]%
    \mycolorbox[text=\strut{3}, color=green!17.636594]%
    \mycolorbox[text=\strut{0}, color=green!17.636594]%
}
\setlength{\fboxsep}{0pt}\fcolorbox{gray!10}{gray!10}{\strut
    \mycolorbox[text=\strut{@}, color=green!17.635190]%
}
\setlength{\fboxsep}{0pt}\fcolorbox{gray!10}{gray!10}{\strut
    \mycolorbox[text=\strut{1}, color=green!17.636594]%
    \mycolorbox[text=\strut{0}, color=green!17.636594]%
    \mycolorbox[text=\strut{/}, color=green!17.636594]%
    \mycolorbox[text=\strut{2}, color=green!17.636594]%
    \mycolorbox[text=\strut{1}, color=green!17.636594]%
}
\setlength{\fboxsep}{0pt}\fcolorbox{gray!10}{gray!10}{\strut
    \mycolorbox[text=\strut{is}]%
}
\setlength{\fboxsep}{0pt}\fcolorbox{gray!10}{gray!10}{\strut
    \mycolorbox[text=\strut{7}, color=green!17.636594]%
    \mycolorbox[text=\strut{*}, color=green!4.133278]%
    \mycolorbox[text=\strut{1}, color=green!17.636594]%
    \mycolorbox[text=\strut{0}, color=green!17.636594]%
    \mycolorbox[text=\strut{*(}, color=green!17.636388]%
    \mycolorbox[text=\strut{2}, color=green!17.636586]%
    \mycolorbox[text=\strut{1}, color=green!17.636594]%
    \mycolorbox[text=\strut{/}, color=green!17.583261]%
    \mycolorbox[text=\strut{3}, color=green!17.636594]%
    \mycolorbox[text=\strut{0}, color=green!17.636594]%
    \mycolorbox[text=\strut{).}, color=red!16.762470]%
}
\setlength{\fboxsep}{0pt}\fcolorbox{gray!10}{gray!10}{\strut
    \mycolorbox[text=\strut{Which}, color=green!3.618559]%
}
\setlength{\fboxsep}{0pt}\fcolorbox{gray!10}{gray!10}{\strut
    \mycolorbox[text=\strut{is}]%
}
\setlength{\fboxsep}{0pt}\fcolorbox{gray!10}{gray!10}{\strut
    \mycolorbox[text=\strut{7}, color=green!17.634251]%
    \mycolorbox[text=\strut{*}, color=red!16.688077]%
    \mycolorbox[text=\strut{1}, color=green!17.636594]%
    \mycolorbox[text=\strut{0}, color=green!17.636594]%
    \mycolorbox[text=\strut{*(}, color=red!5.509695]%
    \mycolorbox[text=\strut{7}, color=green!17.630426]%
    \mycolorbox[text=\strut{/}, color=green!17.636594]%
    \mycolorbox[text=\strut{1}, color=green!17.636594]%
    \mycolorbox[text=\strut{0}, color=green!17.636594]%
    \mycolorbox[text=\strut{)}, color=red!5.712426]%
}
\setlength{\fboxsep}{0pt}\fcolorbox{gray!10}{gray!10}{\strut
    \mycolorbox[text=\strut{=}]%
}
\setlength{\fboxsep}{0pt}\fcolorbox{gray!10}{gray!10}{\strut
    \mycolorbox[text=\strut{7}, color=green!15.430599]%
    \mycolorbox[text=\strut{*}, color=green!17.635940]%
    \mycolorbox[text=\strut{7}, color=green!17.635862]%
}
\setlength{\fboxsep}{0pt}\fcolorbox{gray!10}{gray!10}{\strut
    \mycolorbox[text=\strut{=}, color=green!7.307887]%
}
\setlength{\fboxsep}{0pt}\fcolorbox{gray!10}{gray!10}{\strut
    \mycolorbox[text=\strut{4}, color=green!17.636594]%
    \mycolorbox[text=\strut{9}, color=green!17.636594]%
    \mycolorbox[text=\strut{.}, color=green!12.791977]%
}
\setlength{\fboxsep}{0pt}\fcolorbox{gray!10}{gray!10}{\strut
    \mycolorbox[text=\strut{So}, color=red!29.752072]%
}
\setlength{\fboxsep}{0pt}\fcolorbox{gray!10}{gray!10}{\strut
    \mycolorbox[text=\strut{that}, color=red!19.096753]%
    \mycolorbox[text=\strut{'s}, color=red!3.292171]%
}
\setlength{\fboxsep}{0pt}\fcolorbox{gray!10}{gray!10}{\strut
    \mycolorbox[text=\strut{the}, color=red!42.197095]%
}
\setlength{\fboxsep}{0pt}\fcolorbox{gray!10}{gray!10}{\strut
    \mycolorbox[text=\strut{answer}, color=green!15.991354]%
    \mycolorbox[text=\strut{.}, color=red!40.832810]%
}
\setlength{\fboxsep}{0pt}\fcolorbox{gray!10}{gray!10}{\strut
    \mycolorbox[text=\strut{But}, color=red!67.464716]%
}
\setlength{\fboxsep}{0pt}\fcolorbox{gray!10}{gray!10}{\strut
    \mycolorbox[text=\strut{maybe}, color=red!65.360290]%
}
\setlength{\fboxsep}{0pt}\fcolorbox{gray!10}{gray!10}{\strut
    \mycolorbox[text=\strut{I}, color=red!28.432196]%
}
\setlength{\fboxsep}{0pt}\fcolorbox{gray!10}{gray!10}{\strut
    \mycolorbox[text=\strut{need}, color=red!5.835751]%
}
\setlength{\fboxsep}{0pt}\fcolorbox{gray!10}{gray!10}{\strut
    \mycolorbox[text=\strut{to}, color=green!17.636594]%
}
\setlength{\fboxsep}{0pt}\fcolorbox{gray!10}{gray!10}{\strut
    \mycolorbox[text=\strut{check}, color=red!14.869809]%
}
\setlength{\fboxsep}{0pt}\fcolorbox{gray!10}{gray!10}{\strut
    \mycolorbox[text=\strut{if}, color=green!15.003223]%
}
\setlength{\fboxsep}{0pt}\fcolorbox{gray!10}{gray!10}{\strut
    \mycolorbox[text=\strut{there}, color=red!50.026651]%
    \mycolorbox[text=\strut{'s}, color=red!35.309794]%
}
\setlength{\fboxsep}{0pt}\fcolorbox{gray!10}{gray!10}{\strut
    \mycolorbox[text=\strut{a}, color=red!36.243293]%
}
\setlength{\fboxsep}{0pt}\fcolorbox{gray!10}{gray!10}{\strut
    \mycolorbox[text=\strut{different}, color=red!60.663169]%
}
\setlength{\fboxsep}{0pt}\fcolorbox{gray!10}{gray!10}{\strut
    \mycolorbox[text=\strut{way}, color=green!8.195045]%
}
\setlength{\fboxsep}{0pt}\fcolorbox{gray!10}{gray!10}{\strut
    \mycolorbox[text=\strut{to}, color=green!3.356206]%
}
\setlength{\fboxsep}{0pt}\fcolorbox{gray!10}{gray!10}{\strut
    \mycolorbox[text=\strut{interpret}, color=green!16.260502]%
}
\setlength{\fboxsep}{0pt}\fcolorbox{gray!10}{gray!10}{\strut
    \mycolorbox[text=\strut{the}]%
}
\setlength{\fboxsep}{0pt}\fcolorbox{gray!10}{gray!10}{\strut
    \mycolorbox[text=\strut{operation}]%
    \mycolorbox[text=\strut{.}, color=red!51.137446]%
}
\setlength{\fboxsep}{0pt}\fcolorbox{gray!10}{gray!10}{\strut
    \mycolorbox[text=\strut{For}, color=red!7.773196]%
}
\setlength{\fboxsep}{0pt}\fcolorbox{gray!10}{gray!10}{\strut
    \mycolorbox[text=\strut{example}, color=green!15.430634]%
    \mycolorbox[text=\strut{,}, color=green!17.636594]%
}
\setlength{\fboxsep}{0pt}\fcolorbox{gray!10}{gray!10}{\strut
    \mycolorbox[text=\strut{maybe}, color=red!22.761608]%
}
\setlength{\fboxsep}{0pt}\fcolorbox{gray!10}{gray!10}{\strut
    \mycolorbox[text=\strut{the}, color=green!5.185339]%
}
\setlength{\fboxsep}{0pt}\fcolorbox{gray!10}{gray!10}{\strut
    \mycolorbox[text=\strut{operation}, color=green!13.556295]%
}
\setlength{\fboxsep}{0pt}\fcolorbox{gray!10}{gray!10}{\strut
    \mycolorbox[text=\strut{is}, color=green!17.572072]%
}
\setlength{\fboxsep}{0pt}\fcolorbox{gray!10}{gray!10}{\strut
    \mycolorbox[text=\strut{(}, color=red!29.733480]%
    \mycolorbox[text=\strut{m}, color=green!17.625015]%
    \mycolorbox[text=\strut{/n}, color=green!17.583002]%
    \mycolorbox[text=\strut{)}, color=green!17.143065]%
}
\setlength{\fboxsep}{0pt}\fcolorbox{gray!10}{gray!10}{\strut
    \mycolorbox[text=\strut{@}, color=green!16.939528]%
}
\setlength{\fboxsep}{0pt}\fcolorbox{gray!10}{gray!10}{\strut
    \mycolorbox[text=\strut{(}, color=green!17.635388]%
    \mycolorbox[text=\strut{p}, color=green!17.636594]%
    \mycolorbox[text=\strut{/q}, color=green!17.636594]%
    \mycolorbox[text=\strut{)}, color=green!17.636586]%
}
\setlength{\fboxsep}{0pt}\fcolorbox{gray!10}{gray!10}{\strut
    \mycolorbox[text=\strut{=}, color=green!15.620508]%
}
\setlength{\fboxsep}{0pt}\fcolorbox{gray!10}{gray!10}{\strut
    \mycolorbox[text=\strut{m}, color=red!26.849226]%
}
\setlength{\fboxsep}{0pt}\fcolorbox{gray!10}{gray!10}{\strut
    \mycolorbox[text=\strut{*}, color=green!9.250917]%
}
\setlength{\fboxsep}{0pt}\fcolorbox{gray!10}{gray!10}{\strut
    \mycolorbox[text=\strut{p}, color=green!7.291475]%
}
\setlength{\fboxsep}{0pt}\fcolorbox{gray!10}{gray!10}{\strut
    \mycolorbox[text=\strut{*}, color=green!11.516561]%
}
\setlength{\fboxsep}{0pt}\fcolorbox{gray!10}{gray!10}{\strut
    \mycolorbox[text=\strut{(}, color=green!15.429384]%
    \mycolorbox[text=\strut{q}, color=green!17.631149]%
}
\setlength{\fboxsep}{0pt}\fcolorbox{gray!10}{gray!10}{\strut
    \mycolorbox[text=\strut{/}, color=red!5.346278]%
}
\setlength{\fboxsep}{0pt}\fcolorbox{gray!10}{gray!10}{\strut
    \mycolorbox[text=\strut{n}, color=green!17.634001]%
    \mycolorbox[text=\strut{).}, color=green!8.551565]%
}
\setlength{\fboxsep}{0pt}\fcolorbox{gray!10}{gray!10}{\strut
    \mycolorbox[text=\strut{Which}, color=red!31.532255]%
}
\setlength{\fboxsep}{0pt}\fcolorbox{gray!10}{gray!10}{\strut
    \mycolorbox[text=\strut{is}, color=green!16.329240]%
}
\setlength{\fboxsep}{0pt}\fcolorbox{gray!10}{gray!10}{\strut
    \mycolorbox[text=\strut{the}, color=red!28.041858]%
}
\setlength{\fboxsep}{0pt}\fcolorbox{gray!10}{gray!10}{\strut
    \mycolorbox[text=\strut{same}, color=green!17.635061]%
}
\setlength{\fboxsep}{0pt}\fcolorbox{gray!10}{gray!10}{\strut
    \mycolorbox[text=\strut{as}, color=red!29.764842]%
}
\setlength{\fboxsep}{0pt}\fcolorbox{gray!10}{gray!10}{\strut
    \mycolorbox[text=\strut{above}, color=green!5.407824]%
    \mycolorbox[text=\strut{.}, color=red!40.986916]%
}
\\
\\
{\tiny\color{gray}22}\,%
\setlength{\fboxsep}{0pt}\fcolorbox{gray!10}{gray!10}{\strut
    \mycolorbox[text=\strut{Alternatively}, color=green!15.761219]%
    \mycolorbox[text=\strut{,}, color=green!17.636594]%
}
\setlength{\fboxsep}{0pt}\fcolorbox{gray!10}{gray!10}{\strut
    \mycolorbox[text=\strut{if}, color=red!67.220369]%
}
\setlength{\fboxsep}{0pt}\fcolorbox{gray!10}{gray!10}{\strut
    \mycolorbox[text=\strut{the}, color=red!34.796558]%
}
\setlength{\fboxsep}{0pt}\fcolorbox{gray!10}{gray!10}{\strut
    \mycolorbox[text=\strut{operation}, color=green!14.893824]%
}
\setlength{\fboxsep}{0pt}\fcolorbox{gray!10}{gray!10}{\strut
    \mycolorbox[text=\strut{was}]%
}
\setlength{\fboxsep}{0pt}\fcolorbox{gray!10}{gray!10}{\strut
    \mycolorbox[text=\strut{defined}, color=red!81.133011]%
}
\setlength{\fboxsep}{0pt}\fcolorbox{gray!10}{gray!10}{\strut
    \mycolorbox[text=\strut{as}, color=green!14.198802]%
}
\setlength{\fboxsep}{0pt}\fcolorbox{gray!10}{gray!10}{\strut
    \mycolorbox[text=\strut{(}, color=red!35.077357]%
    \mycolorbox[text=\strut{m}, color=green!17.634802]%
    \mycolorbox[text=\strut{/n}, color=green!16.536401]%
    \mycolorbox[text=\strut{)}, color=green!17.531745]%
}
\setlength{\fboxsep}{0pt}\fcolorbox{gray!10}{gray!10}{\strut
    \mycolorbox[text=\strut{@}, color=green!12.376262]%
}
\setlength{\fboxsep}{0pt}\fcolorbox{gray!10}{gray!10}{\strut
    \mycolorbox[text=\strut{(}, color=green!17.636146]%
    \mycolorbox[text=\strut{p}, color=green!17.636594]%
    \mycolorbox[text=\strut{/q}, color=green!17.636594]%
    \mycolorbox[text=\strut{)}, color=green!17.636594]%
}
\setlength{\fboxsep}{0pt}\fcolorbox{gray!10}{gray!10}{\strut
    \mycolorbox[text=\strut{=}, color=green!17.569908]%
}
\setlength{\fboxsep}{0pt}\fcolorbox{gray!10}{gray!10}{\strut
    \mycolorbox[text=\strut{(}, color=red!23.040084]%
    \mycolorbox[text=\strut{m}, color=green!17.632597]%
}
\setlength{\fboxsep}{0pt}\fcolorbox{gray!10}{gray!10}{\strut
    \mycolorbox[text=\strut{*}, color=green!16.128217]%
}
\setlength{\fboxsep}{0pt}\fcolorbox{gray!10}{gray!10}{\strut
    \mycolorbox[text=\strut{p}, color=green!17.636121]%
    \mycolorbox[text=\strut{)}, color=green!17.464397]%
}
\setlength{\fboxsep}{0pt}\fcolorbox{gray!10}{gray!10}{\strut
    \mycolorbox[text=\strut{/}, color=green!17.616476]%
}
\setlength{\fboxsep}{0pt}\fcolorbox{gray!10}{gray!10}{\strut
    \mycolorbox[text=\strut{(}, color=green!17.636594]%
    \mycolorbox[text=\strut{n}, color=green!17.636465]%
}
\setlength{\fboxsep}{0pt}\fcolorbox{gray!10}{gray!10}{\strut
    \mycolorbox[text=\strut{/}, color=green!15.780681]%
}
\setlength{\fboxsep}{0pt}\fcolorbox{gray!10}{gray!10}{\strut
    \mycolorbox[text=\strut{q}, color=green!17.636594]%
    \mycolorbox[text=\strut{),}, color=red!17.590758]%
}
\setlength{\fboxsep}{0pt}\fcolorbox{gray!10}{gray!10}{\strut
    \mycolorbox[text=\strut{which}, color=red!38.909506]%
}
\setlength{\fboxsep}{0pt}\fcolorbox{gray!10}{gray!10}{\strut
    \mycolorbox[text=\strut{would}, color=green!15.922742]%
}
\setlength{\fboxsep}{0pt}\fcolorbox{gray!10}{gray!10}{\strut
    \mycolorbox[text=\strut{be}, color=green!17.634992]%
}
\setlength{\fboxsep}{0pt}\fcolorbox{gray!10}{gray!10}{\strut
    \mycolorbox[text=\strut{m}, color=red!35.412357]%
    \mycolorbox[text=\strut{*p}, color=green!17.044137]%
    \mycolorbox[text=\strut{*q}, color=green!15.899381]%
}
\setlength{\fboxsep}{0pt}\fcolorbox{gray!10}{gray!10}{\strut
    \mycolorbox[text=\strut{/}]%
}
\setlength{\fboxsep}{0pt}\fcolorbox{gray!10}{gray!10}{\strut
    \mycolorbox[text=\strut{n}, color=green!16.600557]%
    \mycolorbox[text=\strut{.}, color=red!22.683088]%
}
\setlength{\fboxsep}{0pt}\fcolorbox{gray!10}{gray!10}{\strut
    \mycolorbox[text=\strut{Which}]%
}
\setlength{\fboxsep}{0pt}\fcolorbox{gray!10}{gray!10}{\strut
    \mycolorbox[text=\strut{is}, color=green!17.520379]%
}
\setlength{\fboxsep}{0pt}\fcolorbox{gray!10}{gray!10}{\strut
    \mycolorbox[text=\strut{the}, color=red!17.273392]%
}
\setlength{\fboxsep}{0pt}\fcolorbox{gray!10}{gray!10}{\strut
    \mycolorbox[text=\strut{same}, color=green!17.636586]%
}
\setlength{\fboxsep}{0pt}\fcolorbox{gray!10}{gray!10}{\strut
    \mycolorbox[text=\strut{thing}, color=red!44.166592]%
    \mycolorbox[text=\strut{.}, color=green!16.087369]%
}
\setlength{\fboxsep}{0pt}\fcolorbox{gray!10}{gray!10}{\strut
    \mycolorbox[text=\strut{So}, color=green!17.338676]%
}
\setlength{\fboxsep}{0pt}\fcolorbox{gray!10}{gray!10}{\strut
    \mycolorbox[text=\strut{same}, color=red!60.264358]%
}
\setlength{\fboxsep}{0pt}\fcolorbox{gray!10}{gray!10}{\strut
    \mycolorbox[text=\strut{result}, color=green!3.918714]%
    \mycolorbox[text=\strut{.}, color=red!10.858005]%
}
\\
\\
{\tiny\color{gray}23}\,%
\setlength{\fboxsep}{0pt}\fcolorbox{gray!10}{gray!10}{\strut
    \mycolorbox[text=\strut{Alternatively}, color=red!3.231706]%
    \mycolorbox[text=\strut{,}, color=green!17.636594]%
}
\setlength{\fboxsep}{0pt}\fcolorbox{gray!10}{gray!10}{\strut
    \mycolorbox[text=\strut{maybe}, color=red!13.870229]%
}
\setlength{\fboxsep}{0pt}\fcolorbox{gray!10}{gray!10}{\strut
    \mycolorbox[text=\strut{the}, color=red!12.014868]%
}
\setlength{\fboxsep}{0pt}\fcolorbox{gray!10}{gray!10}{\strut
    \mycolorbox[text=\strut{operation}, color=red!17.826367]%
}
\setlength{\fboxsep}{0pt}\fcolorbox{gray!10}{gray!10}{\strut
    \mycolorbox[text=\strut{is}, color=green!16.046111]%
}
\setlength{\fboxsep}{0pt}\fcolorbox{gray!10}{gray!10}{\strut
    \mycolorbox[text=\strut{(}, color=red!31.983187]%
    \mycolorbox[text=\strut{m}, color=green!17.621835]%
    \mycolorbox[text=\strut{/n}, color=green!17.361128]%
    \mycolorbox[text=\strut{)}, color=green!17.528330]%
}
\setlength{\fboxsep}{0pt}\fcolorbox{gray!10}{gray!10}{\strut
    \mycolorbox[text=\strut{@}, color=green!16.356033]%
}
\setlength{\fboxsep}{0pt}\fcolorbox{gray!10}{gray!10}{\strut
    \mycolorbox[text=\strut{(}, color=green!17.635655]%
    \mycolorbox[text=\strut{p}, color=green!17.636594]%
    \mycolorbox[text=\strut{/q}, color=green!17.636594]%
    \mycolorbox[text=\strut{)}, color=green!17.636577]%
}
\setlength{\fboxsep}{0pt}\fcolorbox{gray!10}{gray!10}{\strut
    \mycolorbox[text=\strut{=}, color=green!17.574305]%
}
\setlength{\fboxsep}{0pt}\fcolorbox{gray!10}{gray!10}{\strut
    \mycolorbox[text=\strut{m}, color=red!29.149196]%
}
\setlength{\fboxsep}{0pt}\fcolorbox{gray!10}{gray!10}{\strut
    \mycolorbox[text=\strut{*}, color=green!16.144408]%
}
\setlength{\fboxsep}{0pt}\fcolorbox{gray!10}{gray!10}{\strut
    \mycolorbox[text=\strut{p}, color=red!28.770484]%
}
\setlength{\fboxsep}{0pt}\fcolorbox{gray!10}{gray!10}{\strut
    \mycolorbox[text=\strut{*}, color=red!5.322661]%
}
\setlength{\fboxsep}{0pt}\fcolorbox{gray!10}{gray!10}{\strut
    \mycolorbox[text=\strut{(}, color=red!34.847079]%
    \mycolorbox[text=\strut{q}, color=green!17.634036]%
}
\setlength{\fboxsep}{0pt}\fcolorbox{gray!10}{gray!10}{\strut
    \mycolorbox[text=\strut{/}, color=green!7.259867]%
}
\setlength{\fboxsep}{0pt}\fcolorbox{gray!10}{gray!10}{\strut
    \mycolorbox[text=\strut{n}, color=green!17.635423]%
    \mycolorbox[text=\strut{).}, color=green!14.930047]%
}
\setlength{\fboxsep}{0pt}\fcolorbox{gray!10}{gray!10}{\strut
    \mycolorbox[text=\strut{Which}, color=red!20.495555]%
}
\setlength{\fboxsep}{0pt}\fcolorbox{gray!10}{gray!10}{\strut
    \mycolorbox[text=\strut{is}, color=green!5.913499]%
}
\setlength{\fboxsep}{0pt}\fcolorbox{gray!10}{gray!10}{\strut
    \mycolorbox[text=\strut{the}, color=red!38.059088]%
}
\setlength{\fboxsep}{0pt}\fcolorbox{gray!10}{gray!10}{\strut
    \mycolorbox[text=\strut{same}, color=green!17.633760]%
}
\setlength{\fboxsep}{0pt}\fcolorbox{gray!10}{gray!10}{\strut
    \mycolorbox[text=\strut{as}, color=red!27.188572]%
}
\setlength{\fboxsep}{0pt}\fcolorbox{gray!10}{gray!10}{\strut
    \mycolorbox[text=\strut{above}, color=red!35.337705]%
    \mycolorbox[text=\strut{.}, color=red!28.779562]%
}
\\
\\
{\tiny\color{gray}24}\,%
\setlength{\fboxsep}{0pt}\fcolorbox{gray!10}{gray!10}{\strut
    \mycolorbox[text=\strut{So}, color=red!29.284118]%
}
\setlength{\fboxsep}{0pt}\fcolorbox{gray!10}{gray!10}{\strut
    \mycolorbox[text=\strut{regardless}, color=red!43.162241]%
}
\setlength{\fboxsep}{0pt}\fcolorbox{gray!10}{gray!10}{\strut
    \mycolorbox[text=\strut{of}, color=red!34.848047]%
}
\setlength{\fboxsep}{0pt}\fcolorbox{gray!10}{gray!10}{\strut
    \mycolorbox[text=\strut{the}, color=red!63.172958]%
}
\setlength{\fboxsep}{0pt}\fcolorbox{gray!10}{gray!10}{\strut
    \mycolorbox[text=\strut{interpretation}, color=red!69.995226]%
    \mycolorbox[text=\strut{,}, color=green!17.634544]%
}
\setlength{\fboxsep}{0pt}\fcolorbox{gray!10}{gray!10}{\strut
    \mycolorbox[text=\strut{the}, color=red!20.225433]%
}
\setlength{\fboxsep}{0pt}\fcolorbox{gray!10}{gray!10}{\strut
    \mycolorbox[text=\strut{result}, color=red!52.934139]%
}
\setlength{\fboxsep}{0pt}\fcolorbox{gray!10}{gray!10}{\strut
    \mycolorbox[text=\strut{is}, color=green!16.979956]%
}
\setlength{\fboxsep}{0pt}\fcolorbox{gray!10}{gray!10}{\strut
    \mycolorbox[text=\strut{4}, color=green!16.830576]%
    \mycolorbox[text=\strut{9}, color=green!17.636594]%
    \mycolorbox[text=\strut{.}, color=red!5.625729]%
}
\setlength{\fboxsep}{0pt}\fcolorbox{gray!10}{gray!10}{\strut
    \mycolorbox[text=\strut{So}, color=red!61.950800]%
}
\setlength{\fboxsep}{0pt}\fcolorbox{gray!10}{gray!10}{\strut
    \mycolorbox[text=\strut{maybe}, color=red!50.264427]%
}
\setlength{\fboxsep}{0pt}\fcolorbox{gray!10}{gray!10}{\strut
    \mycolorbox[text=\strut{4}, color=green!17.636594]%
    \mycolorbox[text=\strut{9}, color=green!17.636594]%
}
\setlength{\fboxsep}{0pt}\fcolorbox{gray!10}{gray!10}{\strut
    \mycolorbox[text=\strut{is}, color=green!17.636594]%
}
\setlength{\fboxsep}{0pt}\fcolorbox{gray!10}{gray!10}{\strut
    \mycolorbox[text=\strut{the}, color=green!17.006873]%
}
\setlength{\fboxsep}{0pt}\fcolorbox{gray!10}{gray!10}{\strut
    \mycolorbox[text=\strut{answer}, color=green!11.368247]%
    \mycolorbox[text=\strut{.}, color=red!52.530230]%
}
\\
\\
{\tiny\color{gray}25}\,%
\setlength{\fboxsep}{0pt}\fcolorbox{gray!10}{gray!10}{\strut
    \mycolorbox[text=\strut{But}, color=green!11.112461]%
}
\setlength{\fboxsep}{0pt}\fcolorbox{gray!10}{gray!10}{\strut
    \mycolorbox[text=\strut{let}, color=red!7.947704]%
}
\setlength{\fboxsep}{0pt}\fcolorbox{gray!10}{gray!10}{\strut
    \mycolorbox[text=\strut{me}, color=green!17.636517]%
}
\setlength{\fboxsep}{0pt}\fcolorbox{gray!10}{gray!10}{\strut
    \mycolorbox[text=\strut{check}, color=green!3.121191]%
}
\setlength{\fboxsep}{0pt}\fcolorbox{gray!10}{gray!10}{\strut
    \mycolorbox[text=\strut{once}, color=red!67.475048]%
}
\setlength{\fboxsep}{0pt}\fcolorbox{gray!10}{gray!10}{\strut
    \mycolorbox[text=\strut{more}, color=red!40.828290]%
}
\setlength{\fboxsep}{0pt}\fcolorbox{gray!10}{gray!10}{\strut
    \mycolorbox[text=\strut{with}, color=red!36.863809]%
}
\setlength{\fboxsep}{0pt}\fcolorbox{gray!10}{gray!10}{\strut
    \mycolorbox[text=\strut{numbers}, color=red!66.711531]%
    \mycolorbox[text=\strut{.}, color=green!9.510738]%
}
\setlength{\fboxsep}{0pt}\fcolorbox{gray!10}{gray!10}{\strut
    \mycolorbox[text=\strut{Let}, color=green!7.685589]%
}
\setlength{\fboxsep}{0pt}\fcolorbox{gray!10}{gray!10}{\strut
    \mycolorbox[text=\strut{me}, color=red!19.661391]%
}
\setlength{\fboxsep}{0pt}\fcolorbox{gray!10}{gray!10}{\strut
    \mycolorbox[text=\strut{take}]%
}
\setlength{\fboxsep}{0pt}\fcolorbox{gray!10}{gray!10}{\strut
    \mycolorbox[text=\strut{another}, color=red!61.536435]%
}
\setlength{\fboxsep}{0pt}\fcolorbox{gray!10}{gray!10}{\strut
    \mycolorbox[text=\strut{example}, color=green!17.256123]%
    \mycolorbox[text=\strut{.}, color=green!9.282923]%
}
\setlength{\fboxsep}{0pt}\fcolorbox{gray!10}{gray!10}{\strut
    \mycolorbox[text=\strut{Suppose}]%
}
\setlength{\fboxsep}{0pt}\fcolorbox{gray!10}{gray!10}{\strut
    \mycolorbox[text=\strut{I}]%
}
\setlength{\fboxsep}{0pt}\fcolorbox{gray!10}{gray!10}{\strut
    \mycolorbox[text=\strut{have}, color=green!13.658254]%
}
\setlength{\fboxsep}{0pt}\fcolorbox{gray!10}{gray!10}{\strut
    \mycolorbox[text=\strut{1}, color=green!17.635655]%
    \mycolorbox[text=\strut{/}, color=green!17.636594]%
    \mycolorbox[text=\strut{2}, color=green!17.406321]%
}
\setlength{\fboxsep}{0pt}\fcolorbox{gray!10}{gray!10}{\strut
    \mycolorbox[text=\strut{@}, color=green!17.636594]%
}
\setlength{\fboxsep}{0pt}\fcolorbox{gray!10}{gray!10}{\strut
    \mycolorbox[text=\strut{3}, color=green!17.590751]%
    \mycolorbox[text=\strut{/}, color=green!17.636594]%
    \mycolorbox[text=\strut{4}, color=green!17.636594]%
    \mycolorbox[text=\strut{.}, color=green!17.565391]%
}
\setlength{\fboxsep}{0pt}\fcolorbox{gray!10}{gray!10}{\strut
    \mycolorbox[text=\strut{According}]%
}
\setlength{\fboxsep}{0pt}\fcolorbox{gray!10}{gray!10}{\strut
    \mycolorbox[text=\strut{to}, color=green!17.636594]%
}
\setlength{\fboxsep}{0pt}\fcolorbox{gray!10}{gray!10}{\strut
    \mycolorbox[text=\strut{the}, color=green!16.688299]%
}
\setlength{\fboxsep}{0pt}\fcolorbox{gray!10}{gray!10}{\strut
    \mycolorbox[text=\strut{definition}, color=red!16.725215]%
    \mycolorbox[text=\strut{,}, color=green!17.634966]%
}
\setlength{\fboxsep}{0pt}\fcolorbox{gray!10}{gray!10}{\strut
    \mycolorbox[text=\strut{that}, color=red!4.487843]%
}
\setlength{\fboxsep}{0pt}\fcolorbox{gray!10}{gray!10}{\strut
    \mycolorbox[text=\strut{would}, color=red!27.592390]%
}
\setlength{\fboxsep}{0pt}\fcolorbox{gray!10}{gray!10}{\strut
    \mycolorbox[text=\strut{be}, color=green!17.636569]%
}
\setlength{\fboxsep}{0pt}\fcolorbox{gray!10}{gray!10}{\strut
    \mycolorbox[text=\strut{1}, color=green!17.636594]%
    \mycolorbox[text=\strut{*}, color=green!4.134808]%
    \mycolorbox[text=\strut{3}, color=green!17.636594]%
    \mycolorbox[text=\strut{*(}, color=green!17.636594]%
    \mycolorbox[text=\strut{4}, color=green!17.636594]%
    \mycolorbox[text=\strut{/}, color=green!17.612340]%
    \mycolorbox[text=\strut{2}, color=green!17.636594]%
    \mycolorbox[text=\strut{)}, color=green!16.596302]%
}
\setlength{\fboxsep}{0pt}\fcolorbox{gray!10}{gray!10}{\strut
    \mycolorbox[text=\strut{=}, color=green!17.636026]%
}
\setlength{\fboxsep}{0pt}\fcolorbox{gray!10}{gray!10}{\strut
    \mycolorbox[text=\strut{1}, color=green!11.517957]%
    \mycolorbox[text=\strut{*}, color=green!17.634604]%
    \mycolorbox[text=\strut{3}, color=green!17.636594]%
    \mycolorbox[text=\strut{*}, color=green!17.636327]%
    \mycolorbox[text=\strut{2}, color=green!17.636594]%
}
\setlength{\fboxsep}{0pt}\fcolorbox{gray!10}{gray!10}{\strut
    \mycolorbox[text=\strut{=}, color=green!15.430634]%
}
\setlength{\fboxsep}{0pt}\fcolorbox{gray!10}{gray!10}{\strut
    \mycolorbox[text=\strut{6}, color=green!17.636594]%
    \mycolorbox[text=\strut{.}, color=green!17.632829]%
}
\setlength{\fboxsep}{0pt}\fcolorbox{gray!10}{gray!10}{\strut
    \mycolorbox[text=\strut{Alternatively}, color=red!20.985896]%
    \mycolorbox[text=\strut{,}, color=green!17.636465]%
}
\setlength{\fboxsep}{0pt}\fcolorbox{gray!10}{gray!10}{\strut
    \mycolorbox[text=\strut{if}, color=red!29.133999]%
}
\setlength{\fboxsep}{0pt}\fcolorbox{gray!10}{gray!10}{\strut
    \mycolorbox[text=\strut{I}, color=green!14.131557]%
}
\setlength{\fboxsep}{0pt}\fcolorbox{gray!10}{gray!10}{\strut
    \mycolorbox[text=\strut{compute}, color=red!18.328082]%
}
\setlength{\fboxsep}{0pt}\fcolorbox{gray!10}{gray!10}{\strut
    \mycolorbox[text=\strut{1}, color=green!17.636517]%
    \mycolorbox[text=\strut{*}, color=green!16.957724]%
    \mycolorbox[text=\strut{3}, color=green!17.636594]%
    \mycolorbox[text=\strut{*}, color=green!17.406321]%
    \mycolorbox[text=\strut{4}, color=green!17.625144]%
}
\setlength{\fboxsep}{0pt}\fcolorbox{gray!10}{gray!10}{\strut
    \mycolorbox[text=\strut{/}, color=green!9.450785]%
}
\setlength{\fboxsep}{0pt}\fcolorbox{gray!10}{gray!10}{\strut
    \mycolorbox[text=\strut{2}, color=green!17.636594]%
}
\setlength{\fboxsep}{0pt}\fcolorbox{gray!10}{gray!10}{\strut
    \mycolorbox[text=\strut{=}, color=green!17.148388]%
}
\setlength{\fboxsep}{0pt}\fcolorbox{gray!10}{gray!10}{\strut
    \mycolorbox[text=\strut{1}, color=green!17.636594]%
    \mycolorbox[text=\strut{2}, color=green!17.636586]%
}
\setlength{\fboxsep}{0pt}\fcolorbox{gray!10}{gray!10}{\strut
    \mycolorbox[text=\strut{/}, color=green!9.702662]%
}
\setlength{\fboxsep}{0pt}\fcolorbox{gray!10}{gray!10}{\strut
    \mycolorbox[text=\strut{2}, color=green!17.636594]%
}
\setlength{\fboxsep}{0pt}\fcolorbox{gray!10}{gray!10}{\strut
    \mycolorbox[text=\strut{=}, color=green!17.636594]%
}
\setlength{\fboxsep}{0pt}\fcolorbox{gray!10}{gray!10}{\strut
    \mycolorbox[text=\strut{6}, color=green!17.636594]%
    \mycolorbox[text=\strut{.}, color=green!17.340801]%
}
\setlength{\fboxsep}{0pt}\fcolorbox{gray!10}{gray!10}{\strut
    \mycolorbox[text=\strut{So}]%
}
\setlength{\fboxsep}{0pt}\fcolorbox{gray!10}{gray!10}{\strut
    \mycolorbox[text=\strut{that}, color=red!24.104910]%
    \mycolorbox[text=\strut{'s}, color=red!31.286311]%
}
\setlength{\fboxsep}{0pt}\fcolorbox{gray!10}{gray!10}{\strut
    \mycolorbox[text=\strut{correct}]%
    \mycolorbox[text=\strut{.}, color=red!10.857598]%
}
\setlength{\fboxsep}{0pt}\fcolorbox{gray!10}{gray!10}{\strut
    \mycolorbox[text=\strut{So}, color=red!44.946120]%
}
\setlength{\fboxsep}{0pt}\fcolorbox{gray!10}{gray!10}{\strut
    \mycolorbox[text=\strut{if}, color=red!47.736926]%
}
\setlength{\fboxsep}{0pt}\fcolorbox{gray!10}{gray!10}{\strut
    \mycolorbox[text=\strut{I}, color=green!13.593564]%
}
\setlength{\fboxsep}{0pt}\fcolorbox{gray!10}{gray!10}{\strut
    \mycolorbox[text=\strut{take}, color=red!65.645837]%
}
\setlength{\fboxsep}{0pt}\fcolorbox{gray!10}{gray!10}{\strut
    \mycolorbox[text=\strut{1}, color=green!17.491024]%
    \mycolorbox[text=\strut{/}, color=green!17.636594]%
    \mycolorbox[text=\strut{2}, color=green!17.636491]%
}
\setlength{\fboxsep}{0pt}\fcolorbox{gray!10}{gray!10}{\strut
    \mycolorbox[text=\strut{@}, color=green!17.624998]%
}
\setlength{\fboxsep}{0pt}\fcolorbox{gray!10}{gray!10}{\strut
    \mycolorbox[text=\strut{3}, color=green!17.636569]%
    \mycolorbox[text=\strut{/}, color=green!17.636594]%
    \mycolorbox[text=\strut{4}, color=green!17.636594]%
}
\setlength{\fboxsep}{0pt}\fcolorbox{gray!10}{gray!10}{\strut
    \mycolorbox[text=\strut{=}, color=red!17.633426]%
}
\setlength{\fboxsep}{0pt}\fcolorbox{gray!10}{gray!10}{\strut
    \mycolorbox[text=\strut{6}, color=green!17.636594]%
    \mycolorbox[text=\strut{.}, color=red!8.311791]%
}
\setlength{\fboxsep}{0pt}\fcolorbox{gray!10}{gray!10}{\strut
    \mycolorbox[text=\strut{If}, color=red!36.701580]%
}
\setlength{\fboxsep}{0pt}\fcolorbox{gray!10}{gray!10}{\strut
    \mycolorbox[text=\strut{I}, color=green!17.504162]%
}
\setlength{\fboxsep}{0pt}\fcolorbox{gray!10}{gray!10}{\strut
    \mycolorbox[text=\strut{compute}, color=red!75.986379]%
}
\setlength{\fboxsep}{0pt}\fcolorbox{gray!10}{gray!10}{\strut
    \mycolorbox[text=\strut{1}, color=green!9.697849]%
    \mycolorbox[text=\strut{/}, color=green!16.301663]%
    \mycolorbox[text=\strut{2}, color=green!17.636594]%
}
\setlength{\fboxsep}{0pt}\fcolorbox{gray!10}{gray!10}{\strut
    \mycolorbox[text=\strut{@}, color=green!17.135187]%
}
\setlength{\fboxsep}{0pt}\fcolorbox{gray!10}{gray!10}{\strut
    \mycolorbox[text=\strut{3}, color=green!17.636052]%
    \mycolorbox[text=\strut{/}, color=green!17.636594]%
    \mycolorbox[text=\strut{4}, color=green!17.636594]%
    \mycolorbox[text=\strut{,}]%
}
\setlength{\fboxsep}{0pt}\fcolorbox{gray!10}{gray!10}{\strut
    \mycolorbox[text=\strut{is}, color=red!43.047133]%
}
\setlength{\fboxsep}{0pt}\fcolorbox{gray!10}{gray!10}{\strut
    \mycolorbox[text=\strut{that}, color=red!39.704969]%
}
\setlength{\fboxsep}{0pt}\fcolorbox{gray!10}{gray!10}{\strut
    \mycolorbox[text=\strut{6}, color=green!17.633312]%
    \mycolorbox[text=\strut{?}, color=green!17.256486]%
}
\setlength{\fboxsep}{0pt}\fcolorbox{gray!10}{gray!10}{\strut
    \mycolorbox[text=\strut{Let}, color=red!53.783756]%
}
\setlength{\fboxsep}{0pt}\fcolorbox{gray!10}{gray!10}{\strut
    \mycolorbox[text=\strut{me}, color=green!5.839042]%
}
\setlength{\fboxsep}{0pt}\fcolorbox{gray!10}{gray!10}{\strut
    \mycolorbox[text=\strut{see}, color=red!49.862811]%
    \mycolorbox[text=\strut{.}, color=red!21.864772]%
}
\setlength{\fboxsep}{0pt}\fcolorbox{gray!10}{gray!10}{\strut
    \mycolorbox[text=\strut{If}, color=red!36.555669]%
}
\setlength{\fboxsep}{0pt}\fcolorbox{gray!10}{gray!10}{\strut
    \mycolorbox[text=\strut{I}, color=red!5.787649]%
}
\setlength{\fboxsep}{0pt}\fcolorbox{gray!10}{gray!10}{\strut
    \mycolorbox[text=\strut{think}, color=red!35.183720]%
}
\setlength{\fboxsep}{0pt}\fcolorbox{gray!10}{gray!10}{\strut
    \mycolorbox[text=\strut{of}, color=green!14.602050]%
}
\setlength{\fboxsep}{0pt}\fcolorbox{gray!10}{gray!10}{\strut
    \mycolorbox[text=\strut{the}, color=red!17.448917]%
}
\setlength{\fboxsep}{0pt}\fcolorbox{gray!10}{gray!10}{\strut
    \mycolorbox[text=\strut{operation}, color=green!17.504602]%
}
\setlength{\fboxsep}{0pt}\fcolorbox{gray!10}{gray!10}{\strut
    \mycolorbox[text=\strut{as}, color=green!15.055936]%
}
\setlength{\fboxsep}{0pt}\fcolorbox{gray!10}{gray!10}{\strut
    \mycolorbox[text=\strut{multiplying}, color=red!28.248974]%
}
\setlength{\fboxsep}{0pt}\fcolorbox{gray!10}{gray!10}{\strut
    \mycolorbox[text=\strut{the}, color=green!6.384316]%
}
\setlength{\fboxsep}{0pt}\fcolorbox{gray!10}{gray!10}{\strut
    \mycolorbox[text=\strut{numer}, color=green!17.550492]%
    \mycolorbox[text=\strut{ators}, color=green!17.636594]%
}
\setlength{\fboxsep}{0pt}\fcolorbox{gray!10}{gray!10}{\strut
    \mycolorbox[text=\strut{and}, color=green!16.686516]%
}
\setlength{\fboxsep}{0pt}\fcolorbox{gray!10}{gray!10}{\strut
    \mycolorbox[text=\strut{then}, color=green!13.388599]%
}
\setlength{\fboxsep}{0pt}\fcolorbox{gray!10}{gray!10}{\strut
    \mycolorbox[text=\strut{multiplying}, color=green!17.568779]%
}
\setlength{\fboxsep}{0pt}\fcolorbox{gray!10}{gray!10}{\strut
    \mycolorbox[text=\strut{by}, color=green!17.636319]%
}
\setlength{\fboxsep}{0pt}\fcolorbox{gray!10}{gray!10}{\strut
    \mycolorbox[text=\strut{the}, color=green!14.172898]%
}
\setlength{\fboxsep}{0pt}\fcolorbox{gray!10}{gray!10}{\strut
    \mycolorbox[text=\strut{denominator}, color=green!17.636293]%
}
\setlength{\fboxsep}{0pt}\fcolorbox{gray!10}{gray!10}{\strut
    \mycolorbox[text=\strut{of}, color=green!17.636594]%
}
\setlength{\fboxsep}{0pt}\fcolorbox{gray!10}{gray!10}{\strut
    \mycolorbox[text=\strut{the}, color=green!17.636594]%
}
\setlength{\fboxsep}{0pt}\fcolorbox{gray!10}{gray!10}{\strut
    \mycolorbox[text=\strut{second}, color=green!17.635854]%
}
\setlength{\fboxsep}{0pt}\fcolorbox{gray!10}{gray!10}{\strut
    \mycolorbox[text=\strut{fraction}, color=green!17.551768]%
}
\setlength{\fboxsep}{0pt}\fcolorbox{gray!10}{gray!10}{\strut
    \mycolorbox[text=\strut{divided}, color=green!4.133851]%
}
\setlength{\fboxsep}{0pt}\fcolorbox{gray!10}{gray!10}{\strut
    \mycolorbox[text=\strut{by}, color=green!17.636594]%
}
\setlength{\fboxsep}{0pt}\fcolorbox{gray!10}{gray!10}{\strut
    \mycolorbox[text=\strut{the}, color=green!17.256668]%
}
\setlength{\fboxsep}{0pt}\fcolorbox{gray!10}{gray!10}{\strut
    \mycolorbox[text=\strut{denominator}, color=green!17.636560]%
}
\setlength{\fboxsep}{0pt}\fcolorbox{gray!10}{gray!10}{\strut
    \mycolorbox[text=\strut{of}, color=green!17.636594]%
}
\setlength{\fboxsep}{0pt}\fcolorbox{gray!10}{gray!10}{\strut
    \mycolorbox[text=\strut{the}, color=green!17.636586]%
}
\setlength{\fboxsep}{0pt}\fcolorbox{gray!10}{gray!10}{\strut
    \mycolorbox[text=\strut{first}, color=green!17.636594]%
    \mycolorbox[text=\strut{?}, color=red!76.300519]%
}
\setlength{\fboxsep}{0pt}\fcolorbox{gray!10}{gray!10}{\strut
    \setlength{\fboxsep}{1pt}\fbox{\mycolorbox[text=\strut{Wait}]}%
    \mycolorbox[text=\strut{,}, color=green!17.578959]%
}
\setlength{\fboxsep}{0pt}\fcolorbox{gray!10}{gray!10}{\strut
    \mycolorbox[text=\strut{no}, color=red!36.726505]%
    \mycolorbox[text=\strut{.}, color=red!23.274146]%
}
\setlength{\fboxsep}{0pt}\fcolorbox{gray!10}{gray!10}{\strut
    \setlength{\fboxsep}{1pt}\fbox{\mycolorbox[text=\strut{Wait}]}%
    \mycolorbox[text=\strut{,}, color=green!17.355349]%
}
\setlength{\fboxsep}{0pt}\fcolorbox{gray!10}{gray!10}{\strut
    \mycolorbox[text=\strut{the}, color=red!36.767477]%
}
\setlength{\fboxsep}{0pt}\fcolorbox{gray!10}{gray!10}{\strut
    \mycolorbox[text=\strut{operation}, color=red!12.662053]%
}
\setlength{\fboxsep}{0pt}\fcolorbox{gray!10}{gray!10}{\strut
    \mycolorbox[text=\strut{is}, color=green!17.601610]%
}
\setlength{\fboxsep}{0pt}\fcolorbox{gray!10}{gray!10}{\strut
    \mycolorbox[text=\strut{m}, color=green!4.128276]%
    \mycolorbox[text=\strut{/n}, color=green!7.752991]%
}
\setlength{\fboxsep}{0pt}\fcolorbox{gray!10}{gray!10}{\strut
    \mycolorbox[text=\strut{@}, color=green!17.636474]%
}
\setlength{\fboxsep}{0pt}\fcolorbox{gray!10}{gray!10}{\strut
    \mycolorbox[text=\strut{p}, color=green!17.636594]%
    \mycolorbox[text=\strut{/q}, color=green!17.636594]%
}
\setlength{\fboxsep}{0pt}\fcolorbox{gray!10}{gray!10}{\strut
    \mycolorbox[text=\strut{=}, color=green!14.650386]%
}
\setlength{\fboxsep}{0pt}\fcolorbox{gray!10}{gray!10}{\strut
    \mycolorbox[text=\strut{m}, color=green!16.600200]%
    \mycolorbox[text=\strut{*p}, color=green!9.697952]%
    \mycolorbox[text=\strut{*(}, color=green!17.551932]%
    \mycolorbox[text=\strut{q}, color=green!17.636586]%
    \mycolorbox[text=\strut{/n}, color=green!17.617708]%
    \mycolorbox[text=\strut{).}, color=green!17.631684]%
}
\setlength{\fboxsep}{0pt}\fcolorbox{gray!10}{gray!10}{\strut
    \mycolorbox[text=\strut{So}, color=green!7.280731]%
}
\setlength{\fboxsep}{0pt}\fcolorbox{gray!10}{gray!10}{\strut
    \mycolorbox[text=\strut{that}, color=red!54.942435]%
}
\setlength{\fboxsep}{0pt}\fcolorbox{gray!10}{gray!10}{\strut
    \mycolorbox[text=\strut{is}, color=red!28.865739]%
}
\setlength{\fboxsep}{0pt}\fcolorbox{gray!10}{gray!10}{\strut
    \mycolorbox[text=\strut{m}, color=red!25.438930]%
    \mycolorbox[text=\strut{*p}, color=green!11.287397]%
    \mycolorbox[text=\strut{*q}, color=green!6.949470]%
}
\setlength{\fboxsep}{0pt}\fcolorbox{gray!10}{gray!10}{\strut
    \mycolorbox[text=\strut{/}, color=green!8.438020]%
}
\setlength{\fboxsep}{0pt}\fcolorbox{gray!10}{gray!10}{\strut
    \mycolorbox[text=\strut{n}, color=green!14.795052]%
    \mycolorbox[text=\strut{.}, color=green!4.005408]%
}
\setlength{\fboxsep}{0pt}\fcolorbox{gray!10}{gray!10}{\strut
    \mycolorbox[text=\strut{So}, color=green!6.397390]%
}
\setlength{\fboxsep}{0pt}\fcolorbox{gray!10}{gray!10}{\strut
    \mycolorbox[text=\strut{in}, color=red!36.906872]%
}
\setlength{\fboxsep}{0pt}\fcolorbox{gray!10}{gray!10}{\strut
    \mycolorbox[text=\strut{this}]%
}
\setlength{\fboxsep}{0pt}\fcolorbox{gray!10}{gray!10}{\strut
    \mycolorbox[text=\strut{case}, color=green!15.430608]%
    \mycolorbox[text=\strut{,}, color=green!17.132737]%
}
\setlength{\fboxsep}{0pt}\fcolorbox{gray!10}{gray!10}{\strut
    \mycolorbox[text=\strut{m}, color=red!42.975965]%
    \mycolorbox[text=\strut{=}, color=green!13.180959]%
    \mycolorbox[text=\strut{1}, color=green!17.636594]%
    \mycolorbox[text=\strut{,}, color=green!17.636250]%
}
\setlength{\fboxsep}{0pt}\fcolorbox{gray!10}{gray!10}{\strut
    \mycolorbox[text=\strut{p}, color=green!4.134808]%
    \mycolorbox[text=\strut{=}, color=green!17.636594]%
    \mycolorbox[text=\strut{3}, color=green!17.636594]%
    \mycolorbox[text=\strut{,}, color=green!17.636594]%
}
\setlength{\fboxsep}{0pt}\fcolorbox{gray!10}{gray!10}{\strut
    \mycolorbox[text=\strut{q}, color=green!17.636017]%
    \mycolorbox[text=\strut{=}, color=green!17.636594]%
    \mycolorbox[text=\strut{4}, color=green!17.636594]%
    \mycolorbox[text=\strut{,}, color=green!17.636594]%
}
\setlength{\fboxsep}{0pt}\fcolorbox{gray!10}{gray!10}{\strut
    \mycolorbox[text=\strut{n}, color=green!17.636560]%
    \mycolorbox[text=\strut{=}, color=green!17.636594]%
    \mycolorbox[text=\strut{2}, color=green!17.636594]%
    \mycolorbox[text=\strut{.}, color=green!14.682602]%
}
\setlength{\fboxsep}{0pt}\fcolorbox{gray!10}{gray!10}{\strut
    \mycolorbox[text=\strut{So}, color=green!15.325824]%
}
\setlength{\fboxsep}{0pt}\fcolorbox{gray!10}{gray!10}{\strut
    \mycolorbox[text=\strut{1}, color=green!17.636594]%
    \mycolorbox[text=\strut{*}, color=green!17.596612]%
    \mycolorbox[text=\strut{3}, color=green!17.636594]%
    \mycolorbox[text=\strut{*}, color=green!17.636594]%
    \mycolorbox[text=\strut{4}, color=green!17.636594]%
}
\setlength{\fboxsep}{0pt}\fcolorbox{gray!10}{gray!10}{\strut
    \mycolorbox[text=\strut{/}, color=green!17.403662]%
}
\setlength{\fboxsep}{0pt}\fcolorbox{gray!10}{gray!10}{\strut
    \mycolorbox[text=\strut{2}, color=green!17.636594]%
}
\setlength{\fboxsep}{0pt}\fcolorbox{gray!10}{gray!10}{\strut
    \mycolorbox[text=\strut{=}, color=green!17.569382]%
}
\setlength{\fboxsep}{0pt}\fcolorbox{gray!10}{gray!10}{\strut
    \mycolorbox[text=\strut{1}, color=green!17.636586]%
    \mycolorbox[text=\strut{2}, color=green!17.636594]%
}
\setlength{\fboxsep}{0pt}\fcolorbox{gray!10}{gray!10}{\strut
    \mycolorbox[text=\strut{/}, color=red!5.321179]%
}
\setlength{\fboxsep}{0pt}\fcolorbox{gray!10}{gray!10}{\strut
    \mycolorbox[text=\strut{2}, color=green!17.636594]%
}
\setlength{\fboxsep}{0pt}\fcolorbox{gray!10}{gray!10}{\strut
    \mycolorbox[text=\strut{=}, color=green!17.634604]%
}
\setlength{\fboxsep}{0pt}\fcolorbox{gray!10}{gray!10}{\strut
    \mycolorbox[text=\strut{6}, color=green!17.636594]%
    \mycolorbox[text=\strut{.}, color=green!15.786314]%
}
\setlength{\fboxsep}{0pt}\fcolorbox{gray!10}{gray!10}{\strut
    \mycolorbox[text=\strut{So}, color=red!6.557137]%
}
\setlength{\fboxsep}{0pt}\fcolorbox{gray!10}{gray!10}{\strut
    \mycolorbox[text=\strut{that}, color=green!4.769772]%
    \mycolorbox[text=\strut{'s}, color=red!22.239666]%
}
\setlength{\fboxsep}{0pt}\fcolorbox{gray!10}{gray!10}{\strut
    \mycolorbox[text=\strut{correct}, color=green!15.188881]%
    \mycolorbox[text=\strut{.}, color=red!52.234522]%
}
\setlength{\fboxsep}{0pt}\fcolorbox{gray!10}{gray!10}{\strut
    \mycolorbox[text=\strut{So}, color=green!11.447946]%
}
\setlength{\fboxsep}{0pt}\fcolorbox{gray!10}{gray!10}{\strut
    \mycolorbox[text=\strut{that}, color=red!24.574158]%
}
\setlength{\fboxsep}{0pt}\fcolorbox{gray!10}{gray!10}{\strut
    \mycolorbox[text=\strut{seems}, color=green!10.055735]%
}
\setlength{\fboxsep}{0pt}\fcolorbox{gray!10}{gray!10}{\strut
    \mycolorbox[text=\strut{to}, color=green!14.635872]%
}
\setlength{\fboxsep}{0pt}\fcolorbox{gray!10}{gray!10}{\strut
    \mycolorbox[text=\strut{be}, color=red!56.295656]%
}
\setlength{\fboxsep}{0pt}\fcolorbox{gray!10}{gray!10}{\strut
    \mycolorbox[text=\strut{the}, color=red!3.137113]%
}
\setlength{\fboxsep}{0pt}\fcolorbox{gray!10}{gray!10}{\strut
    \mycolorbox[text=\strut{case}, color=green!6.359383]%
    \mycolorbox[text=\strut{.}, color=green!12.906120]%
}
\\
\\
{\tiny\color{gray}26}\,%
\setlength{\fboxsep}{0pt}\fcolorbox{gray!10}{gray!10}{\strut
    \mycolorbox[text=\strut{Alternatively}, color=red!26.928739]%
    \mycolorbox[text=\strut{,}, color=green!17.636594]%
}
\setlength{\fboxsep}{0pt}\fcolorbox{gray!10}{gray!10}{\strut
    \mycolorbox[text=\strut{if}, color=red!11.325477]%
}
\setlength{\fboxsep}{0pt}\fcolorbox{gray!10}{gray!10}{\strut
    \mycolorbox[text=\strut{I}]%
}
\setlength{\fboxsep}{0pt}\fcolorbox{gray!10}{gray!10}{\strut
    \mycolorbox[text=\strut{take}, color=red!82.231728]%
}
\setlength{\fboxsep}{0pt}\fcolorbox{gray!10}{gray!10}{\strut
    \mycolorbox[text=\strut{another}, color=red!29.046266]%
}
\setlength{\fboxsep}{0pt}\fcolorbox{gray!10}{gray!10}{\strut
    \mycolorbox[text=\strut{example}, color=green!17.611418]%
    \mycolorbox[text=\strut{.}, color=red!27.699564]%
}
\setlength{\fboxsep}{0pt}\fcolorbox{gray!10}{gray!10}{\strut
    \mycolorbox[text=\strut{Let}, color=green!6.645531]%
    \mycolorbox[text=\strut{'s}, color=red!40.829761]%
}
\setlength{\fboxsep}{0pt}\fcolorbox{gray!10}{gray!10}{\strut
    \mycolorbox[text=\strut{say}, color=green!3.911549]%
}
\setlength{\fboxsep}{0pt}\fcolorbox{gray!10}{gray!10}{\strut
    \mycolorbox[text=\strut{2}, color=green!17.636086]%
    \mycolorbox[text=\strut{/}, color=green!17.636594]%
    \mycolorbox[text=\strut{3}, color=green!17.585209]%
}
\setlength{\fboxsep}{0pt}\fcolorbox{gray!10}{gray!10}{\strut
    \mycolorbox[text=\strut{@}, color=green!17.636594]%
}
\setlength{\fboxsep}{0pt}\fcolorbox{gray!10}{gray!10}{\strut
    \mycolorbox[text=\strut{4}, color=green!7.757227]%
    \mycolorbox[text=\strut{/}, color=green!17.636594]%
    \mycolorbox[text=\strut{5}, color=green!17.636327]%
    \mycolorbox[text=\strut{.}, color=green!17.629556]%
}
\setlength{\fboxsep}{0pt}\fcolorbox{gray!10}{gray!10}{\strut
    \mycolorbox[text=\strut{Then}, color=red!31.826417]%
}
\setlength{\fboxsep}{0pt}\fcolorbox{gray!10}{gray!10}{\strut
    \mycolorbox[text=\strut{according}]%
}
\setlength{\fboxsep}{0pt}\fcolorbox{gray!10}{gray!10}{\strut
    \mycolorbox[text=\strut{to}, color=green!17.636594]%
}
\setlength{\fboxsep}{0pt}\fcolorbox{gray!10}{gray!10}{\strut
    \mycolorbox[text=\strut{the}, color=green!16.484864]%
}
\setlength{\fboxsep}{0pt}\fcolorbox{gray!10}{gray!10}{\strut
    \mycolorbox[text=\strut{definition}, color=red!45.341243]%
    \mycolorbox[text=\strut{,}, color=green!17.493914]%
}
\setlength{\fboxsep}{0pt}\fcolorbox{gray!10}{gray!10}{\strut
    \mycolorbox[text=\strut{that}, color=red!32.400557]%
}
\setlength{\fboxsep}{0pt}\fcolorbox{gray!10}{gray!10}{\strut
    \mycolorbox[text=\strut{would}, color=red!28.166150]%
}
\setlength{\fboxsep}{0pt}\fcolorbox{gray!10}{gray!10}{\strut
    \mycolorbox[text=\strut{be}, color=green!17.636586]%
}
\setlength{\fboxsep}{0pt}\fcolorbox{gray!10}{gray!10}{\strut
    \mycolorbox[text=\strut{2}, color=green!17.636594]%
    \mycolorbox[text=\strut{*}, color=green!16.600557]%
    \mycolorbox[text=\strut{4}, color=green!17.636594]%
    \mycolorbox[text=\strut{*(}, color=green!17.636577]%
    \mycolorbox[text=\strut{5}, color=green!17.636594]%
    \mycolorbox[text=\strut{/}, color=green!17.340697]%
    \mycolorbox[text=\strut{3}, color=green!17.636594]%
    \mycolorbox[text=\strut{)}, color=green!17.625437]%
}
\setlength{\fboxsep}{0pt}\fcolorbox{gray!10}{gray!10}{\strut
    \mycolorbox[text=\strut{=}, color=green!17.636465]%
}
\setlength{\fboxsep}{0pt}\fcolorbox{gray!10}{gray!10}{\strut
    \mycolorbox[text=\strut{8}, color=green!17.551829]%
    \mycolorbox[text=\strut{*(}, color=green!17.338831]%
    \mycolorbox[text=\strut{5}, color=green!17.636594]%
    \mycolorbox[text=\strut{/}, color=green!17.636327]%
    \mycolorbox[text=\strut{3}, color=green!17.636594]%
    \mycolorbox[text=\strut{)}, color=green!17.256668]%
}
\setlength{\fboxsep}{0pt}\fcolorbox{gray!10}{gray!10}{\strut
    \mycolorbox[text=\strut{=}, color=green!17.636594]%
}
\setlength{\fboxsep}{0pt}\fcolorbox{gray!10}{gray!10}{\strut
    \mycolorbox[text=\strut{4}, color=green!17.636594]%
    \mycolorbox[text=\strut{0}, color=green!17.636594]%
    \mycolorbox[text=\strut{/}, color=green!17.636250]%
    \mycolorbox[text=\strut{3}, color=green!17.636594]%
    \mycolorbox[text=\strut{.}, color=red!23.245170]%
}
\setlength{\fboxsep}{0pt}\fcolorbox{gray!10}{gray!10}{\strut
    \mycolorbox[text=\strut{Which}, color=green!10.156589]%
}
\setlength{\fboxsep}{0pt}\fcolorbox{gray!10}{gray!10}{\strut
    \mycolorbox[text=\strut{is}, color=green!17.424270]%
}
\setlength{\fboxsep}{0pt}\fcolorbox{gray!10}{gray!10}{\strut
    \mycolorbox[text=\strut{approximately}, color=green!17.468185]%
}
\setlength{\fboxsep}{0pt}\fcolorbox{gray!10}{gray!10}{\strut
    \mycolorbox[text=\strut{1}, color=green!17.636594]%
    \mycolorbox[text=\strut{3}, color=green!17.636594]%
    \mycolorbox[text=\strut{.}, color=green!17.634682]%
    \mycolorbox[text=\strut{3}, color=green!17.636594]%
    \mycolorbox[text=\strut{3}, color=green!17.636586]%
    \mycolorbox[text=\strut{3}, color=green!17.189391]%
    \mycolorbox[text=\strut{...}, color=green!15.618267]%
}
\setlength{\fboxsep}{0pt}\fcolorbox{gray!10}{gray!10}{\strut
    \mycolorbox[text=\strut{So}, color=red!13.739562]%
}
\setlength{\fboxsep}{0pt}\fcolorbox{gray!10}{gray!10}{\strut
    \mycolorbox[text=\strut{that}, color=red!44.652039]%
}
\setlength{\fboxsep}{0pt}\fcolorbox{gray!10}{gray!10}{\strut
    \mycolorbox[text=\strut{seems}, color=red!61.929423]%
}
\setlength{\fboxsep}{0pt}\fcolorbox{gray!10}{gray!10}{\strut
    \mycolorbox[text=\strut{to}, color=red!24.963259]%
}
\setlength{\fboxsep}{0pt}\fcolorbox{gray!10}{gray!10}{\strut
    \mycolorbox[text=\strut{be}, color=green!13.296654]%
}
\setlength{\fboxsep}{0pt}\fcolorbox{gray!10}{gray!10}{\strut
    \mycolorbox[text=\strut{the}, color=green!4.695478]%
}
\setlength{\fboxsep}{0pt}\fcolorbox{gray!10}{gray!10}{\strut
    \mycolorbox[text=\strut{case}, color=red!15.226166]%
    \mycolorbox[text=\strut{.}, color=red!40.828383]%
}
\setlength{\fboxsep}{0pt}\fcolorbox{gray!10}{gray!10}{\strut
    \mycolorbox[text=\strut{So}, color=green!6.421323]%
}
\setlength{\fboxsep}{0pt}\fcolorbox{gray!10}{gray!10}{\strut
    \mycolorbox[text=\strut{if}, color=red!26.225505]%
}
\setlength{\fboxsep}{0pt}\fcolorbox{gray!10}{gray!10}{\strut
    \mycolorbox[text=\strut{I}, color=red!11.952365]%
}
\setlength{\fboxsep}{0pt}\fcolorbox{gray!10}{gray!10}{\strut
    \mycolorbox[text=\strut{follow}, color=red!55.198410]%
}
\setlength{\fboxsep}{0pt}\fcolorbox{gray!10}{gray!10}{\strut
    \mycolorbox[text=\strut{that}, color=red!14.616632]%
    \mycolorbox[text=\strut{,}, color=green!11.258879]%
}
\setlength{\fboxsep}{0pt}\fcolorbox{gray!10}{gray!10}{\strut
    \mycolorbox[text=\strut{then}, color=green!16.337773]%
}
\setlength{\fboxsep}{0pt}\fcolorbox{gray!10}{gray!10}{\strut
    \mycolorbox[text=\strut{in}, color=red!48.777822]%
}
\setlength{\fboxsep}{0pt}\fcolorbox{gray!10}{gray!10}{\strut
    \mycolorbox[text=\strut{the}, color=red!52.241877]%
}
\setlength{\fboxsep}{0pt}\fcolorbox{gray!10}{gray!10}{\strut
    \mycolorbox[text=\strut{original}, color=green!12.177863]%
}
\setlength{\fboxsep}{0pt}\fcolorbox{gray!10}{gray!10}{\strut
    \mycolorbox[text=\strut{problem}, color=green!17.570667]%
    \mycolorbox[text=\strut{,}, color=green!17.598310]%
}
\setlength{\fboxsep}{0pt}\fcolorbox{gray!10}{gray!10}{\strut
    \mycolorbox[text=\strut{the}, color=red!57.381778]%
}
\setlength{\fboxsep}{0pt}\fcolorbox{gray!10}{gray!10}{\strut
    \mycolorbox[text=\strut{answer}, color=green!15.895806]%
}
\setlength{\fboxsep}{0pt}\fcolorbox{gray!10}{gray!10}{\strut
    \mycolorbox[text=\strut{is}, color=green!17.560520]%
}
\setlength{\fboxsep}{0pt}\fcolorbox{gray!10}{gray!10}{\strut
    \mycolorbox[text=\strut{4}, color=green!17.636146]%
    \mycolorbox[text=\strut{9}, color=green!17.636594]%
    \mycolorbox[text=\strut{.}, color=red!42.317895]%
}
\setlength{\fboxsep}{0pt}\fcolorbox{gray!10}{gray!10}{\strut
    \mycolorbox[text=\strut{So}, color=red!41.077641]%
}
\setlength{\fboxsep}{0pt}\fcolorbox{gray!10}{gray!10}{\strut
    \mycolorbox[text=\strut{maybe}, color=red!67.540153]%
}
\setlength{\fboxsep}{0pt}\fcolorbox{gray!10}{gray!10}{\strut
    \mycolorbox[text=\strut{that}, color=red!6.100927]%
    \mycolorbox[text=\strut{'s}, color=red!28.769059]%
}
\setlength{\fboxsep}{0pt}\fcolorbox{gray!10}{gray!10}{\strut
    \mycolorbox[text=\strut{the}, color=red!62.940074]%
}
\setlength{\fboxsep}{0pt}\fcolorbox{gray!10}{gray!10}{\strut
    \mycolorbox[text=\strut{answer}, color=green!8.230672]%
    \mycolorbox[text=\strut{.}]%
}
\\
\\
{\tiny\color{gray}27}\,%
\setlength{\fboxsep}{0pt}\fcolorbox{gray!10}{gray!10}{\strut
    \mycolorbox[text=\strut{But}]%
}
\setlength{\fboxsep}{0pt}\fcolorbox{gray!10}{gray!10}{\strut
    \mycolorbox[text=\strut{let}, color=red!24.500508]%
}
\setlength{\fboxsep}{0pt}\fcolorbox{gray!10}{gray!10}{\strut
    \mycolorbox[text=\strut{me}, color=green!17.636491]%
}
\setlength{\fboxsep}{0pt}\fcolorbox{gray!10}{gray!10}{\strut
    \mycolorbox[text=\strut{check}, color=green!10.960408]%
}
\setlength{\fboxsep}{0pt}\fcolorbox{gray!10}{gray!10}{\strut
    \mycolorbox[text=\strut{if}, color=red!14.454734]%
}
\setlength{\fboxsep}{0pt}\fcolorbox{gray!10}{gray!10}{\strut
    \mycolorbox[text=\strut{there}, color=red!24.231189]%
}
\setlength{\fboxsep}{0pt}\fcolorbox{gray!10}{gray!10}{\strut
    \mycolorbox[text=\strut{is}, color=red!35.022880]%
}
\setlength{\fboxsep}{0pt}\fcolorbox{gray!10}{gray!10}{\strut
    \mycolorbox[text=\strut{a}, color=red!34.823047]%
}
\setlength{\fboxsep}{0pt}\fcolorbox{gray!10}{gray!10}{\strut
    \mycolorbox[text=\strut{different}, color=red!67.082860]%
}
\setlength{\fboxsep}{0pt}\fcolorbox{gray!10}{gray!10}{\strut
    \mycolorbox[text=\strut{interpretation}, color=red!47.419681]%
    \mycolorbox[text=\strut{.}, color=red!4.846489]%
}
\setlength{\fboxsep}{0pt}\fcolorbox{gray!10}{gray!10}{\strut
    \mycolorbox[text=\strut{For}, color=red!19.439897]%
}
\setlength{\fboxsep}{0pt}\fcolorbox{gray!10}{gray!10}{\strut
    \mycolorbox[text=\strut{example}, color=green!9.702662]%
    \mycolorbox[text=\strut{,}, color=green!17.636594]%
}
\setlength{\fboxsep}{0pt}\fcolorbox{gray!10}{gray!10}{\strut
    \mycolorbox[text=\strut{if}, color=red!61.705521]%
}
\setlength{\fboxsep}{0pt}\fcolorbox{gray!10}{gray!10}{\strut
    \mycolorbox[text=\strut{the}, color=green!14.525999]%
}
\setlength{\fboxsep}{0pt}\fcolorbox{gray!10}{gray!10}{\strut
    \mycolorbox[text=\strut{operation}, color=green!16.221498]%
}
\setlength{\fboxsep}{0pt}\fcolorbox{gray!10}{gray!10}{\strut
    \mycolorbox[text=\strut{was}, color=red!12.001128]%
}
\setlength{\fboxsep}{0pt}\fcolorbox{gray!10}{gray!10}{\strut
    \mycolorbox[text=\strut{defined}, color=red!46.635126]%
}
\setlength{\fboxsep}{0pt}\fcolorbox{gray!10}{gray!10}{\strut
    \mycolorbox[text=\strut{as}, color=green!17.269438]%
}
\setlength{\fboxsep}{0pt}\fcolorbox{gray!10}{gray!10}{\strut
    \mycolorbox[text=\strut{(}]%
    \mycolorbox[text=\strut{m}, color=green!17.635750]%
    \mycolorbox[text=\strut{/n}, color=green!17.611426]%
    \mycolorbox[text=\strut{)}, color=green!17.569537]%
}
\setlength{\fboxsep}{0pt}\fcolorbox{gray!10}{gray!10}{\strut
    \mycolorbox[text=\strut{@}, color=green!17.504378]%
}
\setlength{\fboxsep}{0pt}\fcolorbox{gray!10}{gray!10}{\strut
    \mycolorbox[text=\strut{(}, color=green!17.636146]%
    \mycolorbox[text=\strut{p}, color=green!17.636594]%
    \mycolorbox[text=\strut{/q}, color=green!17.636594]%
    \mycolorbox[text=\strut{)}, color=green!17.636594]%
}
\setlength{\fboxsep}{0pt}\fcolorbox{gray!10}{gray!10}{\strut
    \mycolorbox[text=\strut{=}, color=green!17.494915]%
}
\setlength{\fboxsep}{0pt}\fcolorbox{gray!10}{gray!10}{\strut
    \mycolorbox[text=\strut{(}, color=red!6.348581]%
    \mycolorbox[text=\strut{m}, color=green!17.632786]%
}
\setlength{\fboxsep}{0pt}\fcolorbox{gray!10}{gray!10}{\strut
    \mycolorbox[text=\strut{*}, color=green!15.691956]%
}
\setlength{\fboxsep}{0pt}\fcolorbox{gray!10}{gray!10}{\strut
    \mycolorbox[text=\strut{p}, color=green!17.636146]%
    \mycolorbox[text=\strut{)}, color=green!17.329482]%
}
\setlength{\fboxsep}{0pt}\fcolorbox{gray!10}{gray!10}{\strut
    \mycolorbox[text=\strut{/}, color=green!17.445341]%
}
\setlength{\fboxsep}{0pt}\fcolorbox{gray!10}{gray!10}{\strut
    \mycolorbox[text=\strut{(}, color=green!17.636586]%
    \mycolorbox[text=\strut{n}, color=green!17.634578]%
}
\setlength{\fboxsep}{0pt}\fcolorbox{gray!10}{gray!10}{\strut
    \mycolorbox[text=\strut{/}, color=green!16.122333]%
}
\setlength{\fboxsep}{0pt}\fcolorbox{gray!10}{gray!10}{\strut
    \mycolorbox[text=\strut{q}, color=green!17.636594]%
    \mycolorbox[text=\strut{).}, color=green!6.762565]%
}
\setlength{\fboxsep}{0pt}\fcolorbox{gray!10}{gray!10}{\strut
    \mycolorbox[text=\strut{But}, color=red!77.292574]%
}
\setlength{\fboxsep}{0pt}\fcolorbox{gray!10}{gray!10}{\strut
    \mycolorbox[text=\strut{that}, color=green!16.036100]%
}
\setlength{\fboxsep}{0pt}\fcolorbox{gray!10}{gray!10}{\strut
    \mycolorbox[text=\strut{would}, color=green!15.247951]%
}
\setlength{\fboxsep}{0pt}\fcolorbox{gray!10}{gray!10}{\strut
    \mycolorbox[text=\strut{be}, color=green!17.614097]%
}
\setlength{\fboxsep}{0pt}\fcolorbox{gray!10}{gray!10}{\strut
    \mycolorbox[text=\strut{m}, color=green!4.871004]%
    \mycolorbox[text=\strut{*p}, color=green!17.274660]%
    \mycolorbox[text=\strut{*q}, color=red!4.761495]%
}
\setlength{\fboxsep}{0pt}\fcolorbox{gray!10}{gray!10}{\strut
    \mycolorbox[text=\strut{/}, color=green!15.430608]%
}
\setlength{\fboxsep}{0pt}\fcolorbox{gray!10}{gray!10}{\strut
    \mycolorbox[text=\strut{n}, color=green!16.600557]%
    \mycolorbox[text=\strut{.}, color=red!28.789960]%
}
\setlength{\fboxsep}{0pt}\fcolorbox{gray!10}{gray!10}{\strut
    \mycolorbox[text=\strut{Which}, color=green!14.043211]%
}
\setlength{\fboxsep}{0pt}\fcolorbox{gray!10}{gray!10}{\strut
    \mycolorbox[text=\strut{is}, color=green!17.589751]%
}
\setlength{\fboxsep}{0pt}\fcolorbox{gray!10}{gray!10}{\strut
    \mycolorbox[text=\strut{same}, color=red!52.457148]%
}
\setlength{\fboxsep}{0pt}\fcolorbox{gray!10}{gray!10}{\strut
    \mycolorbox[text=\strut{as}, color=green!16.294217]%
}
\setlength{\fboxsep}{0pt}\fcolorbox{gray!10}{gray!10}{\strut
    \mycolorbox[text=\strut{above}, color=red!30.131604]%
    \mycolorbox[text=\strut{.}, color=green!11.427693]%
}
\setlength{\fboxsep}{0pt}\fcolorbox{gray!10}{gray!10}{\strut
    \mycolorbox[text=\strut{So}, color=green!17.317410]%
}
\setlength{\fboxsep}{0pt}\fcolorbox{gray!10}{gray!10}{\strut
    \mycolorbox[text=\strut{same}]%
}
\setlength{\fboxsep}{0pt}\fcolorbox{gray!10}{gray!10}{\strut
    \mycolorbox[text=\strut{result}, color=red!3.916675]%
    \mycolorbox[text=\strut{.}, color=red!28.769092]%
}
\\
\\
{\tiny\color{gray}28}\,%
\setlength{\fboxsep}{0pt}\fcolorbox{gray!10}{gray!10}{\strut
    \mycolorbox[text=\strut{Alternatively}, color=green!16.835621]%
    \mycolorbox[text=\strut{,}, color=green!17.636594]%
}
\setlength{\fboxsep}{0pt}\fcolorbox{gray!10}{gray!10}{\strut
    \mycolorbox[text=\strut{maybe}, color=red!32.005277]%
}
\setlength{\fboxsep}{0pt}\fcolorbox{gray!10}{gray!10}{\strut
    \mycolorbox[text=\strut{the}, color=green!8.300848]%
}
\setlength{\fboxsep}{0pt}\fcolorbox{gray!10}{gray!10}{\strut
    \mycolorbox[text=\strut{operation}, color=green!7.988256]%
}
\setlength{\fboxsep}{0pt}\fcolorbox{gray!10}{gray!10}{\strut
    \mycolorbox[text=\strut{is}, color=green!15.416616]%
}
\setlength{\fboxsep}{0pt}\fcolorbox{gray!10}{gray!10}{\strut
    \mycolorbox[text=\strut{defined}, color=red!72.268667]%
}
\setlength{\fboxsep}{0pt}\fcolorbox{gray!10}{gray!10}{\strut
    \mycolorbox[text=\strut{as}, color=green!15.397742]%
}
\setlength{\fboxsep}{0pt}\fcolorbox{gray!10}{gray!10}{\strut
    \mycolorbox[text=\strut{m}, color=red!52.403692]%
    \mycolorbox[text=\strut{/n}, color=green!17.369360]%
}
\setlength{\fboxsep}{0pt}\fcolorbox{gray!10}{gray!10}{\strut
    \mycolorbox[text=\strut{@}, color=green!14.710656]%
}
\setlength{\fboxsep}{0pt}\fcolorbox{gray!10}{gray!10}{\strut
    \mycolorbox[text=\strut{p}, color=green!17.636577]%
    \mycolorbox[text=\strut{/q}, color=green!17.636594]%
}
\setlength{\fboxsep}{0pt}\fcolorbox{gray!10}{gray!10}{\strut
    \mycolorbox[text=\strut{=}, color=green!17.430935]%
}
\setlength{\fboxsep}{0pt}\fcolorbox{gray!10}{gray!10}{\strut
    \mycolorbox[text=\strut{(}, color=red!22.728040]%
    \mycolorbox[text=\strut{m}, color=green!17.634251]%
}
\setlength{\fboxsep}{0pt}\fcolorbox{gray!10}{gray!10}{\strut
    \mycolorbox[text=\strut{*}, color=green!8.525232]%
}
\setlength{\fboxsep}{0pt}\fcolorbox{gray!10}{gray!10}{\strut
    \mycolorbox[text=\strut{p}, color=green!17.614933]%
    \mycolorbox[text=\strut{)}, color=green!17.089945]%
}
\setlength{\fboxsep}{0pt}\fcolorbox{gray!10}{gray!10}{\strut
    \mycolorbox[text=\strut{/}, color=green!12.052486]%
}
\setlength{\fboxsep}{0pt}\fcolorbox{gray!10}{gray!10}{\strut
    \mycolorbox[text=\strut{(}, color=green!17.630822]%
    \mycolorbox[text=\strut{n}, color=green!17.584416]%
}
\setlength{\fboxsep}{0pt}\fcolorbox{gray!10}{gray!10}{\strut
    \mycolorbox[text=\strut{/}, color=green!7.095048]%
}
\setlength{\fboxsep}{0pt}\fcolorbox{gray!10}{gray!10}{\strut
    \mycolorbox[text=\strut{q}, color=green!17.636560]%
    \mycolorbox[text=\strut{).}, color=green!12.433274]%
}
\setlength{\fboxsep}{0pt}\fcolorbox{gray!10}{gray!10}{\strut
    \mycolorbox[text=\strut{Which}, color=red!5.472564]%
}
\setlength{\fboxsep}{0pt}\fcolorbox{gray!10}{gray!10}{\strut
    \mycolorbox[text=\strut{is}, color=red!9.291400]%
}
\setlength{\fboxsep}{0pt}\fcolorbox{gray!10}{gray!10}{\strut
    \mycolorbox[text=\strut{same}, color=red!15.629754]%
}
\setlength{\fboxsep}{0pt}\fcolorbox{gray!10}{gray!10}{\strut
    \mycolorbox[text=\strut{as}, color=red!5.832556]%
}
\setlength{\fboxsep}{0pt}\fcolorbox{gray!10}{gray!10}{\strut
    \mycolorbox[text=\strut{m}, color=green!5.788879]%
    \mycolorbox[text=\strut{*p}, color=green!16.636874]%
    \mycolorbox[text=\strut{*q}, color=green!17.573348]%
}
\setlength{\fboxsep}{0pt}\fcolorbox{gray!10}{gray!10}{\strut
    \mycolorbox[text=\strut{/}, color=green!7.312601]%
}
\setlength{\fboxsep}{0pt}\fcolorbox{gray!10}{gray!10}{\strut
    \mycolorbox[text=\strut{n}, color=green!11.517957]%
    \mycolorbox[text=\strut{.}, color=green!8.248689]%
}
\setlength{\fboxsep}{0pt}\fcolorbox{gray!10}{gray!10}{\strut
    \mycolorbox[text=\strut{So}, color=red!6.733776]%
}
\setlength{\fboxsep}{0pt}\fcolorbox{gray!10}{gray!10}{\strut
    \mycolorbox[text=\strut{same}]%
}
\setlength{\fboxsep}{0pt}\fcolorbox{gray!10}{gray!10}{\strut
    \mycolorbox[text=\strut{result}, color=red!60.868637]%
    \mycolorbox[text=\strut{.}, color=red!28.786541]%
}
\\
\\
{\tiny\color{gray}29}\,%
\setlength{\fboxsep}{0pt}\fcolorbox{gray!10}{gray!10}{\strut
    \mycolorbox[text=\strut{Alternatively}, color=red!10.017019]%
    \mycolorbox[text=\strut{,}, color=green!17.636594]%
}
\setlength{\fboxsep}{0pt}\fcolorbox{gray!10}{gray!10}{\strut
    \mycolorbox[text=\strut{maybe}, color=red!9.356550]%
}
\setlength{\fboxsep}{0pt}\fcolorbox{gray!10}{gray!10}{\strut
    \mycolorbox[text=\strut{the}]%
}
\setlength{\fboxsep}{0pt}\fcolorbox{gray!10}{gray!10}{\strut
    \mycolorbox[text=\strut{problem}, color=red!65.287850]%
}
\setlength{\fboxsep}{0pt}\fcolorbox{gray!10}{gray!10}{\strut
    \mycolorbox[text=\strut{is}, color=red!32.414034]%
}
\setlength{\fboxsep}{0pt}\fcolorbox{gray!10}{gray!10}{\strut
    \mycolorbox[text=\strut{written}, color=red!3.624221]%
}
\setlength{\fboxsep}{0pt}\fcolorbox{gray!10}{gray!10}{\strut
    \mycolorbox[text=\strut{with}, color=red!72.284366]%
}
\setlength{\fboxsep}{0pt}\fcolorbox{gray!10}{gray!10}{\strut
    \mycolorbox[text=\strut{different}, color=red!66.568547]%
}
\setlength{\fboxsep}{0pt}\fcolorbox{gray!10}{gray!10}{\strut
    \mycolorbox[text=\strut{parentheses}, color=red!18.720670]%
    \mycolorbox[text=\strut{?}, color=red!5.522477]%
}
\setlength{\fboxsep}{0pt}\fcolorbox{gray!10}{gray!10}{\strut
    \mycolorbox[text=\strut{For}, color=red!41.409521]%
}
\setlength{\fboxsep}{0pt}\fcolorbox{gray!10}{gray!10}{\strut
    \mycolorbox[text=\strut{example}, color=green!16.600557]%
    \mycolorbox[text=\strut{,}, color=green!17.635862]%
}
\setlength{\fboxsep}{0pt}\fcolorbox{gray!10}{gray!10}{\strut
    \mycolorbox[text=\strut{maybe}, color=red!38.456734]%
}
\setlength{\fboxsep}{0pt}\fcolorbox{gray!10}{gray!10}{\strut
    \mycolorbox[text=\strut{it}, color=red!49.378286]%
    \mycolorbox[text=\strut{'s}, color=green!14.610567]%
}
\setlength{\fboxsep}{0pt}\fcolorbox{gray!10}{gray!10}{\strut
    \mycolorbox[text=\strut{m}, color=green!3.444549]%
    \mycolorbox[text=\strut{/n}, color=green!9.335365]%
}
\setlength{\fboxsep}{0pt}\fcolorbox{gray!10}{gray!10}{\strut
    \mycolorbox[text=\strut{@}, color=green!17.573399]%
}
\setlength{\fboxsep}{0pt}\fcolorbox{gray!10}{gray!10}{\strut
    \mycolorbox[text=\strut{p}, color=green!17.636491]%
    \mycolorbox[text=\strut{/q}, color=green!17.636594]%
}
\setlength{\fboxsep}{0pt}\fcolorbox{gray!10}{gray!10}{\strut
    \mycolorbox[text=\strut{=}, color=green!17.610427]%
}
\setlength{\fboxsep}{0pt}\fcolorbox{gray!10}{gray!10}{\strut
    \mycolorbox[text=\strut{m}, color=green!13.928014]%
}
\setlength{\fboxsep}{0pt}\fcolorbox{gray!10}{gray!10}{\strut
    \mycolorbox[text=\strut{*}, color=green!8.937027]%
}
\setlength{\fboxsep}{0pt}\fcolorbox{gray!10}{gray!10}{\strut
    \mycolorbox[text=\strut{(}, color=red!16.772996]%
    \mycolorbox[text=\strut{p}, color=green!17.636250]%
}
\setlength{\fboxsep}{0pt}\fcolorbox{gray!10}{gray!10}{\strut
    \mycolorbox[text=\strut{/}, color=red!9.957678]%
}
\setlength{\fboxsep}{0pt}\fcolorbox{gray!10}{gray!10}{\strut
    \mycolorbox[text=\strut{(}, color=green!15.128283]%
    \mycolorbox[text=\strut{n}, color=green!17.406321]%
}
\setlength{\fboxsep}{0pt}\fcolorbox{gray!10}{gray!10}{\strut
    \mycolorbox[text=\strut{/}, color=red!47.647531]%
}
\setlength{\fboxsep}{0pt}\fcolorbox{gray!10}{gray!10}{\strut
    \mycolorbox[text=\strut{q}, color=green!17.636594]%
    \mycolorbox[text=\strut{))?}, color=green!15.193948]%
}
\setlength{\fboxsep}{0pt}\fcolorbox{gray!10}{gray!10}{\strut
    \mycolorbox[text=\strut{But}, color=red!50.037724]%
}
\setlength{\fboxsep}{0pt}\fcolorbox{gray!10}{gray!10}{\strut
    \mycolorbox[text=\strut{that}, color=green!17.415465]%
}
\setlength{\fboxsep}{0pt}\fcolorbox{gray!10}{gray!10}{\strut
    \mycolorbox[text=\strut{would}, color=green!17.077343]%
}
\setlength{\fboxsep}{0pt}\fcolorbox{gray!10}{gray!10}{\strut
    \mycolorbox[text=\strut{be}, color=green!17.623938]%
}
\setlength{\fboxsep}{0pt}\fcolorbox{gray!10}{gray!10}{\strut
    \mycolorbox[text=\strut{m}, color=green!16.585327]%
}
\setlength{\fboxsep}{0pt}\fcolorbox{gray!10}{gray!10}{\strut
    \mycolorbox[text=\strut{*}, color=green!9.680050]%
}
\setlength{\fboxsep}{0pt}\fcolorbox{gray!10}{gray!10}{\strut
    \mycolorbox[text=\strut{p}, color=green!15.921822]%
}
\setlength{\fboxsep}{0pt}\fcolorbox{gray!10}{gray!10}{\strut
    \mycolorbox[text=\strut{*}, color=green!11.517232]%
}
\setlength{\fboxsep}{0pt}\fcolorbox{gray!10}{gray!10}{\strut
    \mycolorbox[text=\strut{q}, color=red!10.857578]%
}
\setlength{\fboxsep}{0pt}\fcolorbox{gray!10}{gray!10}{\strut
    \mycolorbox[text=\strut{/}, color=green!17.629650]%
}
\setlength{\fboxsep}{0pt}\fcolorbox{gray!10}{gray!10}{\strut
    \mycolorbox[text=\strut{n}, color=green!17.256659]%
    \mycolorbox[text=\strut{.}, color=green!4.104404]%
}
\setlength{\fboxsep}{0pt}\fcolorbox{gray!10}{gray!10}{\strut
    \mycolorbox[text=\strut{Which}]%
}
\setlength{\fboxsep}{0pt}\fcolorbox{gray!10}{gray!10}{\strut
    \mycolorbox[text=\strut{is}, color=green!15.420288]%
}
\setlength{\fboxsep}{0pt}\fcolorbox{gray!10}{gray!10}{\strut
    \mycolorbox[text=\strut{same}, color=green!7.544250]%
}
\setlength{\fboxsep}{0pt}\fcolorbox{gray!10}{gray!10}{\strut
    \mycolorbox[text=\strut{as}, color=red!8.850958]%
}
\setlength{\fboxsep}{0pt}\fcolorbox{gray!10}{gray!10}{\strut
    \mycolorbox[text=\strut{above}, color=red!23.089758]%
    \mycolorbox[text=\strut{.}, color=red!40.828300]%
}
\setlength{\fboxsep}{0pt}\fcolorbox{gray!10}{gray!10}{\strut
    \mycolorbox[text=\strut{So}, color=green!17.534945]%
}
\setlength{\fboxsep}{0pt}\fcolorbox{gray!10}{gray!10}{\strut
    \mycolorbox[text=\strut{same}, color=red!52.100798]%
}
\setlength{\fboxsep}{0pt}\fcolorbox{gray!10}{gray!10}{\strut
    \mycolorbox[text=\strut{result}, color=red!3.598215]%
    \mycolorbox[text=\strut{.}]%
}
\\
\\
{\tiny\color{gray}30}\,%
\setlength{\fboxsep}{0pt}\fcolorbox{gray!10}{gray!10}{\strut
    \mycolorbox[text=\strut{Therefore}, color=red!45.598998]%
    \mycolorbox[text=\strut{,}, color=green!17.634604]%
}
\setlength{\fboxsep}{0pt}\fcolorbox{gray!10}{gray!10}{\strut
    \mycolorbox[text=\strut{regardless}, color=green!10.544568]%
}
\setlength{\fboxsep}{0pt}\fcolorbox{gray!10}{gray!10}{\strut
    \mycolorbox[text=\strut{of}, color=green!17.445833]%
}
\setlength{\fboxsep}{0pt}\fcolorbox{gray!10}{gray!10}{\strut
    \mycolorbox[text=\strut{how}, color=red!4.404121]%
}
\setlength{\fboxsep}{0pt}\fcolorbox{gray!10}{gray!10}{\strut
    \mycolorbox[text=\strut{I}, color=green!13.564251]%
}
\setlength{\fboxsep}{0pt}\fcolorbox{gray!10}{gray!10}{\strut
    \mycolorbox[text=\strut{parse}, color=green!14.527928]%
}
\setlength{\fboxsep}{0pt}\fcolorbox{gray!10}{gray!10}{\strut
    \mycolorbox[text=\strut{the}, color=green!9.681918]%
}
\setlength{\fboxsep}{0pt}\fcolorbox{gray!10}{gray!10}{\strut
    \mycolorbox[text=\strut{operation}, color=red!34.734889]%
    \mycolorbox[text=\strut{,}, color=green!17.305406]%
}
\setlength{\fboxsep}{0pt}\fcolorbox{gray!10}{gray!10}{\strut
    \mycolorbox[text=\strut{it}, color=red!28.051339]%
}
\setlength{\fboxsep}{0pt}\fcolorbox{gray!10}{gray!10}{\strut
    \mycolorbox[text=\strut{seems}]%
}
\setlength{\fboxsep}{0pt}\fcolorbox{gray!10}{gray!10}{\strut
    \mycolorbox[text=\strut{that}, color=green!9.282734]%
}
\setlength{\fboxsep}{0pt}\fcolorbox{gray!10}{gray!10}{\strut
    \mycolorbox[text=\strut{the}, color=green!12.893141]%
}
\setlength{\fboxsep}{0pt}\fcolorbox{gray!10}{gray!10}{\strut
    \mycolorbox[text=\strut{answer}, color=red!47.137978]%
}
\setlength{\fboxsep}{0pt}\fcolorbox{gray!10}{gray!10}{\strut
    \mycolorbox[text=\strut{is}, color=green!17.576830]%
}
\setlength{\fboxsep}{0pt}\fcolorbox{gray!10}{gray!10}{\strut
    \mycolorbox[text=\strut{4}, color=green!17.634337]%
    \mycolorbox[text=\strut{9}, color=green!17.636594]%
    \mycolorbox[text=\strut{.}, color=red!34.848197]%
}
\setlength{\fboxsep}{0pt}\fcolorbox{gray!10}{gray!10}{\strut
    \mycolorbox[text=\strut{Therefore}, color=red!20.444685]%
    \mycolorbox[text=\strut{,}, color=green!17.636586]%
}
\setlength{\fboxsep}{0pt}\fcolorbox{gray!10}{gray!10}{\strut
    \mycolorbox[text=\strut{I}, color=red!30.119770]%
}
\setlength{\fboxsep}{0pt}\fcolorbox{gray!10}{gray!10}{\strut
    \mycolorbox[text=\strut{think}, color=green!17.610883]%
}
\setlength{\fboxsep}{0pt}\fcolorbox{gray!10}{gray!10}{\strut
    \mycolorbox[text=\strut{that}, color=red!25.767257]%
}
\setlength{\fboxsep}{0pt}\fcolorbox{gray!10}{gray!10}{\strut
    \mycolorbox[text=\strut{is}, color=red!7.008317]%
}
\setlength{\fboxsep}{0pt}\fcolorbox{gray!10}{gray!10}{\strut
    \mycolorbox[text=\strut{the}]%
}
\setlength{\fboxsep}{0pt}\fcolorbox{gray!10}{gray!10}{\strut
    \mycolorbox[text=\strut{correct}, color=green!3.913759]%
}
\setlength{\fboxsep}{0pt}\fcolorbox{gray!10}{gray!10}{\strut
    \mycolorbox[text=\strut{answer}, color=green!17.618957]%
    \mycolorbox[text=\strut{.}, color=red!67.385881]%
}
\setlength{\fboxsep}{0pt}\fcolorbox{gray!10}{gray!10}{\strut
    \mycolorbox[text=\strut{However}, color=red!65.070524]%
    \mycolorbox[text=\strut{,}, color=green!17.636594]%
}
\setlength{\fboxsep}{0pt}\fcolorbox{gray!10}{gray!10}{\strut
    \mycolorbox[text=\strut{let}, color=red!41.944581]%
}
\setlength{\fboxsep}{0pt}\fcolorbox{gray!10}{gray!10}{\strut
    \mycolorbox[text=\strut{me}, color=green!17.636560]%
}
\setlength{\fboxsep}{0pt}\fcolorbox{gray!10}{gray!10}{\strut
    \mycolorbox[text=\strut{check}, color=green!14.099350]%
}
\setlength{\fboxsep}{0pt}\fcolorbox{gray!10}{gray!10}{\strut
    \mycolorbox[text=\strut{once}, color=red!37.407073]%
}
\setlength{\fboxsep}{0pt}\fcolorbox{gray!10}{gray!10}{\strut
    \mycolorbox[text=\strut{again}, color=red!28.769102]%
}
\setlength{\fboxsep}{0pt}\fcolorbox{gray!10}{gray!10}{\strut
    \mycolorbox[text=\strut{with}, color=red!33.000191]%
}
\setlength{\fboxsep}{0pt}\fcolorbox{gray!10}{gray!10}{\strut
    \mycolorbox[text=\strut{the}, color=green!14.366459]%
}
\setlength{\fboxsep}{0pt}\fcolorbox{gray!10}{gray!10}{\strut
    \mycolorbox[text=\strut{original}, color=red!23.834667]%
}
\setlength{\fboxsep}{0pt}\fcolorbox{gray!10}{gray!10}{\strut
    \mycolorbox[text=\strut{problem}, color=red!16.460619]%
    \mycolorbox[text=\strut{.}, color=red!37.595676]%
}
\\
\\
{\tiny\color{gray}31}\,%
\setlength{\fboxsep}{0pt}\fcolorbox{gray!10}{gray!10}{\strut
    \mycolorbox[text=\strut{Original}, color=green!14.460170]%
}
\setlength{\fboxsep}{0pt}\fcolorbox{gray!10}{gray!10}{\strut
    \mycolorbox[text=\strut{problem}, color=green!17.183144]%
    \mycolorbox[text=\strut{:}, color=red!3.915467]%
}
\setlength{\fboxsep}{0pt}\fcolorbox{gray!10}{gray!10}{\strut
    \mycolorbox[text=\strut{The}, color=red!77.307522]%
}
\setlength{\fboxsep}{0pt}\fcolorbox{gray!10}{gray!10}{\strut
    \mycolorbox[text=\strut{operation}, color=green!17.636586]%
}
\setlength{\fboxsep}{0pt}\fcolorbox{gray!10}{gray!10}{\strut
    \mycolorbox[text=\strut{@}, color=green!16.600557]%
}
\setlength{\fboxsep}{0pt}\fcolorbox{gray!10}{gray!10}{\strut
    \mycolorbox[text=\strut{is}, color=green!17.636594]%
}
\setlength{\fboxsep}{0pt}\fcolorbox{gray!10}{gray!10}{\strut
    \mycolorbox[text=\strut{defined}, color=green!17.636594]%
}
\setlength{\fboxsep}{0pt}\fcolorbox{gray!10}{gray!10}{\strut
    \mycolorbox[text=\strut{as}, color=green!17.636267]%
}
\setlength{\fboxsep}{0pt}\fcolorbox{gray!10}{gray!10}{\strut
    \mycolorbox[text=\strut{m}, color=green!17.142260]%
    \mycolorbox[text=\strut{/n}, color=green!17.636594]%
}
\setlength{\fboxsep}{0pt}\fcolorbox{gray!10}{gray!10}{\strut
    \mycolorbox[text=\strut{@}, color=green!17.636586]%
}
\setlength{\fboxsep}{0pt}\fcolorbox{gray!10}{gray!10}{\strut
    \mycolorbox[text=\strut{p}, color=green!17.636594]%
    \mycolorbox[text=\strut{/q}, color=green!17.636594]%
}
\setlength{\fboxsep}{0pt}\fcolorbox{gray!10}{gray!10}{\strut
    \mycolorbox[text=\strut{=}, color=green!17.636465]%
}
\setlength{\fboxsep}{0pt}\fcolorbox{gray!10}{gray!10}{\strut
    \mycolorbox[text=\strut{(}, color=green!17.340827]%
    \mycolorbox[text=\strut{m}, color=green!17.636594]%
    \mycolorbox[text=\strut{)(}, color=green!17.636577]%
    \mycolorbox[text=\strut{p}, color=green!17.636594]%
    \mycolorbox[text=\strut{)(}, color=green!17.636594]%
    \mycolorbox[text=\strut{q}, color=green!17.636594]%
    \mycolorbox[text=\strut{/n}, color=green!17.636250]%
    \mycolorbox[text=\strut{)}, color=green!11.517783]%
}
\setlength{\fboxsep}{0pt}\fcolorbox{gray!10}{gray!10}{\strut
    \mycolorbox[text=\strut{for}, color=green!17.636577]%
}
\setlength{\fboxsep}{0pt}\fcolorbox{gray!10}{gray!10}{\strut
    \mycolorbox[text=\strut{simplified}, color=green!17.636594]%
}
\setlength{\fboxsep}{0pt}\fcolorbox{gray!10}{gray!10}{\strut
    \mycolorbox[text=\strut{fractions}, color=green!17.636594]%
}
\setlength{\fboxsep}{0pt}\fcolorbox{gray!10}{gray!10}{\strut
    \mycolorbox[text=\strut{p}, color=green!17.636594]%
    \mycolorbox[text=\strut{/q}, color=green!17.636594]%
    \mycolorbox[text=\strut{.}, color=green!12.906120]%
}
\setlength{\fboxsep}{0pt}\fcolorbox{gray!10}{gray!10}{\strut
    \mycolorbox[text=\strut{So}, color=red!23.319119]%
    \mycolorbox[text=\strut{,}, color=red!46.280705]%
}
\setlength{\fboxsep}{0pt}\fcolorbox{gray!10}{gray!10}{\strut
    \mycolorbox[text=\strut{in}, color=red!65.806989]%
}
\setlength{\fboxsep}{0pt}\fcolorbox{gray!10}{gray!10}{\strut
    \mycolorbox[text=\strut{the}, color=red!38.327142]%
}
\setlength{\fboxsep}{0pt}\fcolorbox{gray!10}{gray!10}{\strut
    \mycolorbox[text=\strut{problem}, color=green!10.918697]%
    \mycolorbox[text=\strut{,}, color=red!5.527693]%
}
\setlength{\fboxsep}{0pt}\fcolorbox{gray!10}{gray!10}{\strut
    \mycolorbox[text=\strut{they}, color=red!27.734911]%
}
\setlength{\fboxsep}{0pt}\fcolorbox{gray!10}{gray!10}{\strut
    \mycolorbox[text=\strut{specify}, color=red!63.286753]%
}
\setlength{\fboxsep}{0pt}\fcolorbox{gray!10}{gray!10}{\strut
    \mycolorbox[text=\strut{that}, color=green!12.829449]%
}
\setlength{\fboxsep}{0pt}\fcolorbox{gray!10}{gray!10}{\strut
    \mycolorbox[text=\strut{p}, color=green!14.658960]%
    \mycolorbox[text=\strut{/q}, color=green!17.636594]%
}
\setlength{\fboxsep}{0pt}\fcolorbox{gray!10}{gray!10}{\strut
    \mycolorbox[text=\strut{is}, color=green!17.550803]%
}
\setlength{\fboxsep}{0pt}\fcolorbox{gray!10}{gray!10}{\strut
    \mycolorbox[text=\strut{simplified}, color=green!16.600392]%
    \mycolorbox[text=\strut{.}, color=green!11.401557]%
}
\setlength{\fboxsep}{0pt}\fcolorbox{gray!10}{gray!10}{\strut
    \mycolorbox[text=\strut{So}, color=red!40.168449]%
}
\setlength{\fboxsep}{0pt}\fcolorbox{gray!10}{gray!10}{\strut
    \mycolorbox[text=\strut{maybe}, color=red!5.599061]%
}
\setlength{\fboxsep}{0pt}\fcolorbox{gray!10}{gray!10}{\strut
    \mycolorbox[text=\strut{m}, color=red!60.883105]%
    \mycolorbox[text=\strut{/n}, color=green!17.636594]%
}
\setlength{\fboxsep}{0pt}\fcolorbox{gray!10}{gray!10}{\strut
    \mycolorbox[text=\strut{is}, color=green!10.548841]%
}
\setlength{\fboxsep}{0pt}\fcolorbox{gray!10}{gray!10}{\strut
    \mycolorbox[text=\strut{not}, color=green!17.417709]%
}
\setlength{\fboxsep}{0pt}\fcolorbox{gray!10}{gray!10}{\strut
    \mycolorbox[text=\strut{necessarily}, color=green!9.605454]%
}
\setlength{\fboxsep}{0pt}\fcolorbox{gray!10}{gray!10}{\strut
    \mycolorbox[text=\strut{simplified}, color=green!17.636233]%
    \mycolorbox[text=\strut{?}, color=green!16.989232]%
}
\setlength{\fboxsep}{0pt}\fcolorbox{gray!10}{gray!10}{\strut
    \mycolorbox[text=\strut{But}, color=green!6.555308]%
}
\setlength{\fboxsep}{0pt}\fcolorbox{gray!10}{gray!10}{\strut
    \mycolorbox[text=\strut{in}, color=green!15.911938]%
}
\setlength{\fboxsep}{0pt}\fcolorbox{gray!10}{gray!10}{\strut
    \mycolorbox[text=\strut{the}, color=red!48.433101]%
}
\setlength{\fboxsep}{0pt}\fcolorbox{gray!10}{gray!10}{\strut
    \mycolorbox[text=\strut{problem}, color=red!9.774608]%
    \mycolorbox[text=\strut{,}, color=green!14.109751]%
}
\setlength{\fboxsep}{0pt}\fcolorbox{gray!10}{gray!10}{\strut
    \mycolorbox[text=\strut{we}, color=red!33.559131]%
}
\setlength{\fboxsep}{0pt}\fcolorbox{gray!10}{gray!10}{\strut
    \mycolorbox[text=\strut{are}, color=red!5.836846]%
}
\setlength{\fboxsep}{0pt}\fcolorbox{gray!10}{gray!10}{\strut
    \mycolorbox[text=\strut{given}, color=green!15.902247]%
}
\setlength{\fboxsep}{0pt}\fcolorbox{gray!10}{gray!10}{\strut
    \mycolorbox[text=\strut{7}, color=green!17.636594]%
    \mycolorbox[text=\strut{/}, color=green!17.636594]%
    \mycolorbox[text=\strut{3}, color=green!17.636594]%
    \mycolorbox[text=\strut{0}, color=green!17.636594]%
}
\setlength{\fboxsep}{0pt}\fcolorbox{gray!10}{gray!10}{\strut
    \mycolorbox[text=\strut{and}, color=red!67.385863]%
}
\setlength{\fboxsep}{0pt}\fcolorbox{gray!10}{gray!10}{\strut
    \mycolorbox[text=\strut{1}, color=green!17.636594]%
    \mycolorbox[text=\strut{0}, color=green!17.636594]%
    \mycolorbox[text=\strut{/}, color=green!17.636594]%
    \mycolorbox[text=\strut{2}, color=green!17.636594]%
    \mycolorbox[text=\strut{1}, color=green!17.636594]%
    \mycolorbox[text=\strut{,}, color=green!12.704152]%
}
\setlength{\fboxsep}{0pt}\fcolorbox{gray!10}{gray!10}{\strut
    \mycolorbox[text=\strut{which}, color=red!10.884278]%
}
\setlength{\fboxsep}{0pt}\fcolorbox{gray!10}{gray!10}{\strut
    \mycolorbox[text=\strut{are}, color=green!17.634811]%
}
\setlength{\fboxsep}{0pt}\fcolorbox{gray!10}{gray!10}{\strut
    \mycolorbox[text=\strut{simplified}, color=red!62.635461]%
    \mycolorbox[text=\strut{.}, color=green!10.700154]%
}
\setlength{\fboxsep}{0pt}\fcolorbox{gray!10}{gray!10}{\strut
    \mycolorbox[text=\strut{So}, color=red!18.056689]%
}
\setlength{\fboxsep}{0pt}\fcolorbox{gray!10}{gray!10}{\strut
    \mycolorbox[text=\strut{maybe}, color=red!6.177672]%
}
\setlength{\fboxsep}{0pt}\fcolorbox{gray!10}{gray!10}{\strut
    \mycolorbox[text=\strut{that}, color=red!51.107435]%
}
\setlength{\fboxsep}{0pt}\fcolorbox{gray!10}{gray!10}{\strut
    \mycolorbox[text=\strut{is}, color=red!53.351299]%
}
\setlength{\fboxsep}{0pt}\fcolorbox{gray!10}{gray!10}{\strut
    \mycolorbox[text=\strut{just}, color=red!44.084905]%
}
\setlength{\fboxsep}{0pt}\fcolorbox{gray!10}{gray!10}{\strut
    \mycolorbox[text=\strut{a}, color=red!43.883465]%
}
\setlength{\fboxsep}{0pt}\fcolorbox{gray!10}{gray!10}{\strut
    \mycolorbox[text=\strut{condition}, color=red!44.203894]%
}
\setlength{\fboxsep}{0pt}\fcolorbox{gray!10}{gray!10}{\strut
    \mycolorbox[text=\strut{for}, color=red!29.477764]%
}
\setlength{\fboxsep}{0pt}\fcolorbox{gray!10}{gray!10}{\strut
    \mycolorbox[text=\strut{the}, color=green!12.878170]%
}
\setlength{\fboxsep}{0pt}\fcolorbox{gray!10}{gray!10}{\strut
    \mycolorbox[text=\strut{definition}]%
    \mycolorbox[text=\strut{?}, color=red!41.236097]%
}
\setlength{\fboxsep}{0pt}\fcolorbox{gray!10}{gray!10}{\strut
    \mycolorbox[text=\strut{Maybe}, color=red!12.174749]%
}
\setlength{\fboxsep}{0pt}\fcolorbox{gray!10}{gray!10}{\strut
    \mycolorbox[text=\strut{the}, color=red!7.710956]%
}
\setlength{\fboxsep}{0pt}\fcolorbox{gray!10}{gray!10}{\strut
    \mycolorbox[text=\strut{operation}, color=red!5.413837]%
}
\setlength{\fboxsep}{0pt}\fcolorbox{gray!10}{gray!10}{\strut
    \mycolorbox[text=\strut{is}, color=green!17.497011]%
}
\setlength{\fboxsep}{0pt}\fcolorbox{gray!10}{gray!10}{\strut
    \mycolorbox[text=\strut{defined}, color=green!16.122245]%
}
\setlength{\fboxsep}{0pt}\fcolorbox{gray!10}{gray!10}{\strut
    \mycolorbox[text=\strut{for}, color=green!4.573019]%
}
\setlength{\fboxsep}{0pt}\fcolorbox{gray!10}{gray!10}{\strut
    \mycolorbox[text=\strut{simplified}, color=red!25.281134]%
}
\setlength{\fboxsep}{0pt}\fcolorbox{gray!10}{gray!10}{\strut
    \mycolorbox[text=\strut{fractions}, color=green!13.972440]%
}
\setlength{\fboxsep}{0pt}\fcolorbox{gray!10}{gray!10}{\strut
    \mycolorbox[text=\strut{p}, color=green!17.056570]%
    \mycolorbox[text=\strut{/q}, color=green!17.636594]%
    \mycolorbox[text=\strut{,}, color=green!14.228818]%
}
\setlength{\fboxsep}{0pt}\fcolorbox{gray!10}{gray!10}{\strut
    \mycolorbox[text=\strut{but}, color=green!16.595901]%
}
\setlength{\fboxsep}{0pt}\fcolorbox{gray!10}{gray!10}{\strut
    \mycolorbox[text=\strut{m}, color=green!15.941550]%
    \mycolorbox[text=\strut{/n}, color=green!17.636594]%
}
\setlength{\fboxsep}{0pt}\fcolorbox{gray!10}{gray!10}{\strut
    \mycolorbox[text=\strut{could}, color=red!42.132807]%
}
\setlength{\fboxsep}{0pt}\fcolorbox{gray!10}{gray!10}{\strut
    \mycolorbox[text=\strut{be}, color=green!17.635509]%
}
\setlength{\fboxsep}{0pt}\fcolorbox{gray!10}{gray!10}{\strut
    \mycolorbox[text=\strut{any}]%
}
\setlength{\fboxsep}{0pt}\fcolorbox{gray!10}{gray!10}{\strut
    \mycolorbox[text=\strut{fraction}, color=green!17.406934]%
    \mycolorbox[text=\strut{?}, color=red!6.920844]%
}
\setlength{\fboxsep}{0pt}\fcolorbox{gray!10}{gray!10}{\strut
    \mycolorbox[text=\strut{But}, color=red!12.934684]%
}
\setlength{\fboxsep}{0pt}\fcolorbox{gray!10}{gray!10}{\strut
    \mycolorbox[text=\strut{in}, color=red!6.333931]%
}
\setlength{\fboxsep}{0pt}\fcolorbox{gray!10}{gray!10}{\strut
    \mycolorbox[text=\strut{our}, color=red!46.268941]%
}
\setlength{\fboxsep}{0pt}\fcolorbox{gray!10}{gray!10}{\strut
    \mycolorbox[text=\strut{case}, color=green!11.400601]%
    \mycolorbox[text=\strut{,}, color=green!17.636534]%
}
\setlength{\fboxsep}{0pt}\fcolorbox{gray!10}{gray!10}{\strut
    \mycolorbox[text=\strut{both}, color=red!53.243017]%
}
\setlength{\fboxsep}{0pt}\fcolorbox{gray!10}{gray!10}{\strut
    \mycolorbox[text=\strut{are}]%
}
\setlength{\fboxsep}{0pt}\fcolorbox{gray!10}{gray!10}{\strut
    \mycolorbox[text=\strut{simplified}, color=green!17.633217]%
    \mycolorbox[text=\strut{.}, color=red!27.640421]%
}
\setlength{\fboxsep}{0pt}\fcolorbox{gray!10}{gray!10}{\strut
    \mycolorbox[text=\strut{So}, color=red!4.677800]%
}
\setlength{\fboxsep}{0pt}\fcolorbox{gray!10}{gray!10}{\strut
    \mycolorbox[text=\strut{maybe}, color=red!51.562703]%
}
\setlength{\fboxsep}{0pt}\fcolorbox{gray!10}{gray!10}{\strut
    \mycolorbox[text=\strut{the}, color=red!73.315300]%
}
\setlength{\fboxsep}{0pt}\fcolorbox{gray!10}{gray!10}{\strut
    \mycolorbox[text=\strut{operation}, color=red!55.428604]%
}
\setlength{\fboxsep}{0pt}\fcolorbox{gray!10}{gray!10}{\strut
    \mycolorbox[text=\strut{is}, color=green!14.639760]%
}
\setlength{\fboxsep}{0pt}\fcolorbox{gray!10}{gray!10}{\strut
    \mycolorbox[text=\strut{defined}, color=green!7.504360]%
}
\setlength{\fboxsep}{0pt}\fcolorbox{gray!10}{gray!10}{\strut
    \mycolorbox[text=\strut{as}, color=red!39.155670]%
}
\setlength{\fboxsep}{0pt}\fcolorbox{gray!10}{gray!10}{\strut
    \mycolorbox[text=\strut{m}, color=red!49.275992]%
    \mycolorbox[text=\strut{/n}, color=green!16.605656]%
}
\setlength{\fboxsep}{0pt}\fcolorbox{gray!10}{gray!10}{\strut
    \mycolorbox[text=\strut{@}, color=green!17.636034]%
}
\setlength{\fboxsep}{0pt}\fcolorbox{gray!10}{gray!10}{\strut
    \mycolorbox[text=\strut{p}, color=green!17.636129]%
    \mycolorbox[text=\strut{/q}, color=green!17.636594]%
}
\setlength{\fboxsep}{0pt}\fcolorbox{gray!10}{gray!10}{\strut
    \mycolorbox[text=\strut{=}, color=green!14.167019]%
}
\setlength{\fboxsep}{0pt}\fcolorbox{gray!10}{gray!10}{\strut
    \mycolorbox[text=\strut{m}, color=green!16.302841]%
    \mycolorbox[text=\strut{*p}, color=green!11.488385]%
    \mycolorbox[text=\strut{*(}, color=green!17.148483]%
    \mycolorbox[text=\strut{q}, color=green!17.636586]%
    \mycolorbox[text=\strut{/n}, color=green!17.570632]%
    \mycolorbox[text=\strut{).}, color=red!28.829586]%
}
\setlength{\fboxsep}{0pt}\fcolorbox{gray!10}{gray!10}{\strut
    \mycolorbox[text=\strut{So}, color=red!9.918566]%
    \mycolorbox[text=\strut{,}, color=red!69.411829]%
}
\setlength{\fboxsep}{0pt}\fcolorbox{gray!10}{gray!10}{\strut
    \mycolorbox[text=\strut{in}, color=red!66.919344]%
}
\setlength{\fboxsep}{0pt}\fcolorbox{gray!10}{gray!10}{\strut
    \mycolorbox[text=\strut{our}, color=red!50.011474]%
}
\setlength{\fboxsep}{0pt}\fcolorbox{gray!10}{gray!10}{\strut
    \mycolorbox[text=\strut{case}]%
    \mycolorbox[text=\strut{,}, color=green!17.615993]%
}
\setlength{\fboxsep}{0pt}\fcolorbox{gray!10}{gray!10}{\strut
    \mycolorbox[text=\strut{7}, color=green!17.634036]%
    \mycolorbox[text=\strut{/}, color=red!11.055183]%
    \mycolorbox[text=\strut{3}, color=green!17.636594]%
    \mycolorbox[text=\strut{0}, color=green!17.636594]%
}
\setlength{\fboxsep}{0pt}\fcolorbox{gray!10}{gray!10}{\strut
    \mycolorbox[text=\strut{@}, color=green!17.606531]%
}
\setlength{\fboxsep}{0pt}\fcolorbox{gray!10}{gray!10}{\strut
    \mycolorbox[text=\strut{1}, color=green!17.636594]%
    \mycolorbox[text=\strut{0}, color=green!17.636594]%
    \mycolorbox[text=\strut{/}, color=green!17.636594]%
    \mycolorbox[text=\strut{2}, color=green!17.636594]%
    \mycolorbox[text=\strut{1}, color=green!17.636594]%
}
\setlength{\fboxsep}{0pt}\fcolorbox{gray!10}{gray!10}{\strut
    \mycolorbox[text=\strut{is}, color=red!40.996885]%
}
\setlength{\fboxsep}{0pt}\fcolorbox{gray!10}{gray!10}{\strut
    \mycolorbox[text=\strut{7}, color=green!17.636534]%
    \mycolorbox[text=\strut{*}, color=green!15.922636]%
    \mycolorbox[text=\strut{1}, color=green!17.636594]%
    \mycolorbox[text=\strut{0}, color=green!17.636594]%
    \mycolorbox[text=\strut{*(}, color=green!17.625075]%
    \mycolorbox[text=\strut{2}, color=green!17.636569]%
    \mycolorbox[text=\strut{1}, color=green!17.636594]%
    \mycolorbox[text=\strut{/}, color=green!17.321869]%
    \mycolorbox[text=\strut{3}, color=green!17.636594]%
    \mycolorbox[text=\strut{0}, color=green!17.636594]%
    \mycolorbox[text=\strut{)}, color=red!29.034676]%
}
\setlength{\fboxsep}{0pt}\fcolorbox{gray!10}{gray!10}{\strut
    \mycolorbox[text=\strut{=}, color=green!17.135291]%
}
\setlength{\fboxsep}{0pt}\fcolorbox{gray!10}{gray!10}{\strut
    \mycolorbox[text=\strut{7}, color=green!17.009039]%
    \mycolorbox[text=\strut{*}, color=green!9.697399]%
    \mycolorbox[text=\strut{1}, color=green!17.636594]%
    \mycolorbox[text=\strut{0}, color=green!17.636594]%
    \mycolorbox[text=\strut{*(}, color=red!10.857989]%
    \mycolorbox[text=\strut{7}, color=green!17.636577]%
    \mycolorbox[text=\strut{/}, color=green!17.636594]%
    \mycolorbox[text=\strut{1}, color=green!17.636594]%
    \mycolorbox[text=\strut{0}, color=green!17.636594]%
    \mycolorbox[text=\strut{)}, color=green!17.336602]%
}
\setlength{\fboxsep}{0pt}\fcolorbox{gray!10}{gray!10}{\strut
    \mycolorbox[text=\strut{=}, color=green!17.634053]%
}
\setlength{\fboxsep}{0pt}\fcolorbox{gray!10}{gray!10}{\strut
    \mycolorbox[text=\strut{4}, color=red!46.642251]%
    \mycolorbox[text=\strut{9}, color=green!17.636594]%
    \mycolorbox[text=\strut{.}, color=green!7.053655]%
}
\setlength{\fboxsep}{0pt}\fcolorbox{gray!10}{gray!10}{\strut
    \mycolorbox[text=\strut{So}, color=red!13.551392]%
}
\setlength{\fboxsep}{0pt}\fcolorbox{gray!10}{gray!10}{\strut
    \mycolorbox[text=\strut{that}, color=red!14.382483]%
    \mycolorbox[text=\strut{'s}, color=red!26.074506]%
}
\setlength{\fboxsep}{0pt}\fcolorbox{gray!10}{gray!10}{\strut
    \mycolorbox[text=\strut{correct}, color=red!9.108498]%
    \mycolorbox[text=\strut{.}, color=green!4.134798]%
}
\\
\\
{\tiny\color{gray}32}\,%
\setlength{\fboxsep}{0pt}\fcolorbox{gray!10}{gray!10}{\strut
    \mycolorbox[text=\strut{Alternatively}, color=green!11.213118]%
    \mycolorbox[text=\strut{,}, color=green!17.636594]%
}
\setlength{\fboxsep}{0pt}\fcolorbox{gray!10}{gray!10}{\strut
    \mycolorbox[text=\strut{maybe}, color=red!3.899174]%
}
\setlength{\fboxsep}{0pt}\fcolorbox{gray!10}{gray!10}{\strut
    \mycolorbox[text=\strut{there}, color=red!42.362499]%
}
\setlength{\fboxsep}{0pt}\fcolorbox{gray!10}{gray!10}{\strut
    \mycolorbox[text=\strut{is}, color=red!16.973950]%
}
\setlength{\fboxsep}{0pt}\fcolorbox{gray!10}{gray!10}{\strut
    \mycolorbox[text=\strut{a}, color=green!9.956782]%
}
\setlength{\fboxsep}{0pt}\fcolorbox{gray!10}{gray!10}{\strut
    \mycolorbox[text=\strut{mistake}, color=red!28.704286]%
}
\setlength{\fboxsep}{0pt}\fcolorbox{gray!10}{gray!10}{\strut
    \mycolorbox[text=\strut{here}, color=red!52.240073]%
    \mycolorbox[text=\strut{.}, color=red!29.074412]%
}
\setlength{\fboxsep}{0pt}\fcolorbox{gray!10}{gray!10}{\strut
    \mycolorbox[text=\strut{Let}, color=red!6.094279]%
}
\setlength{\fboxsep}{0pt}\fcolorbox{gray!10}{gray!10}{\strut
    \mycolorbox[text=\strut{me}, color=green!17.527882]%
}
\setlength{\fboxsep}{0pt}\fcolorbox{gray!10}{gray!10}{\strut
    \mycolorbox[text=\strut{check}, color=red!21.935186]%
}
\setlength{\fboxsep}{0pt}\fcolorbox{gray!10}{gray!10}{\strut
    \mycolorbox[text=\strut{if}, color=red!24.615945]%
}
\setlength{\fboxsep}{0pt}\fcolorbox{gray!10}{gray!10}{\strut
    \mycolorbox[text=\strut{7}, color=red!52.262215]%
    \mycolorbox[text=\strut{/}, color=green!15.917800]%
    \mycolorbox[text=\strut{3}, color=green!17.636577]%
    \mycolorbox[text=\strut{0}, color=green!17.636594]%
}
\setlength{\fboxsep}{0pt}\fcolorbox{gray!10}{gray!10}{\strut
    \mycolorbox[text=\strut{and}, color=red!62.843883]%
}
\setlength{\fboxsep}{0pt}\fcolorbox{gray!10}{gray!10}{\strut
    \mycolorbox[text=\strut{1}, color=green!17.636594]%
    \mycolorbox[text=\strut{0}, color=green!17.636594]%
    \mycolorbox[text=\strut{/}, color=green!17.636594]%
    \mycolorbox[text=\strut{2}, color=green!17.636594]%
    \mycolorbox[text=\strut{1}, color=green!17.636594]%
}
\setlength{\fboxsep}{0pt}\fcolorbox{gray!10}{gray!10}{\strut
    \mycolorbox[text=\strut{can}, color=red!53.993525]%
}
\setlength{\fboxsep}{0pt}\fcolorbox{gray!10}{gray!10}{\strut
    \mycolorbox[text=\strut{be}, color=green!17.635483]%
}
\setlength{\fboxsep}{0pt}\fcolorbox{gray!10}{gray!10}{\strut
    \mycolorbox[text=\strut{simplified}, color=green!17.591053]%
    \mycolorbox[text=\strut{?}, color=red!60.273697]%
}
\setlength{\fboxsep}{0pt}\fcolorbox{gray!10}{gray!10}{\strut
    \setlength{\fboxsep}{1pt}\fbox{\mycolorbox[text=\strut{Wait}, color=red!41.797156]}%
    \mycolorbox[text=\strut{,}, color=green!14.791923]%
}
\setlength{\fboxsep}{0pt}\fcolorbox{gray!10}{gray!10}{\strut
    \mycolorbox[text=\strut{7}, color=green!16.600557]%
}
\setlength{\fboxsep}{0pt}\fcolorbox{gray!10}{gray!10}{\strut
    \mycolorbox[text=\strut{and}, color=red!57.634506]%
}
\setlength{\fboxsep}{0pt}\fcolorbox{gray!10}{gray!10}{\strut
    \mycolorbox[text=\strut{3}, color=green!17.636517]%
    \mycolorbox[text=\strut{0}, color=green!17.636594]%
}
\setlength{\fboxsep}{0pt}\fcolorbox{gray!10}{gray!10}{\strut
    \mycolorbox[text=\strut{have}, color=red!32.575975]%
}
\setlength{\fboxsep}{0pt}\fcolorbox{gray!10}{gray!10}{\strut
    \mycolorbox[text=\strut{no}, color=green!15.778813]%
}
\setlength{\fboxsep}{0pt}\fcolorbox{gray!10}{gray!10}{\strut
    \mycolorbox[text=\strut{common}, color=green!17.635862]%
}
\setlength{\fboxsep}{0pt}\fcolorbox{gray!10}{gray!10}{\strut
    \mycolorbox[text=\strut{factors}, color=green!17.577313]%
    \mycolorbox[text=\strut{,}, color=red!18.639192]%
}
\setlength{\fboxsep}{0pt}\fcolorbox{gray!10}{gray!10}{\strut
    \mycolorbox[text=\strut{1}, color=green!17.256572]%
    \mycolorbox[text=\strut{0}, color=green!17.636594]%
}
\setlength{\fboxsep}{0pt}\fcolorbox{gray!10}{gray!10}{\strut
    \mycolorbox[text=\strut{and}, color=green!17.636009]%
}
\setlength{\fboxsep}{0pt}\fcolorbox{gray!10}{gray!10}{\strut
    \mycolorbox[text=\strut{2}, color=green!17.636594]%
    \mycolorbox[text=\strut{1}, color=green!17.636594]%
}
\setlength{\fboxsep}{0pt}\fcolorbox{gray!10}{gray!10}{\strut
    \mycolorbox[text=\strut{have}, color=green!9.696170]%
}
\setlength{\fboxsep}{0pt}\fcolorbox{gray!10}{gray!10}{\strut
    \mycolorbox[text=\strut{no}, color=green!17.636396]%
}
\setlength{\fboxsep}{0pt}\fcolorbox{gray!10}{gray!10}{\strut
    \mycolorbox[text=\strut{common}, color=green!17.636594]%
}
\setlength{\fboxsep}{0pt}\fcolorbox{gray!10}{gray!10}{\strut
    \mycolorbox[text=\strut{factors}, color=green!17.636560]%
    \mycolorbox[text=\strut{.}, color=red!5.326679]%
}
\setlength{\fboxsep}{0pt}\fcolorbox{gray!10}{gray!10}{\strut
    \mycolorbox[text=\strut{So}, color=green!17.453972]%
}
\setlength{\fboxsep}{0pt}\fcolorbox{gray!10}{gray!10}{\strut
    \mycolorbox[text=\strut{they}, color=red!3.293872]%
}
\setlength{\fboxsep}{0pt}\fcolorbox{gray!10}{gray!10}{\strut
    \mycolorbox[text=\strut{are}, color=green!17.540041]%
}
\setlength{\fboxsep}{0pt}\fcolorbox{gray!10}{gray!10}{\strut
    \mycolorbox[text=\strut{simplified}, color=green!6.169084]%
    \mycolorbox[text=\strut{.}, color=green!14.677899]%
}
\setlength{\fboxsep}{0pt}\fcolorbox{gray!10}{gray!10}{\strut
    \mycolorbox[text=\strut{Therefore}, color=red!47.002282]%
    \mycolorbox[text=\strut{,}, color=green!17.636560]%
}
\setlength{\fboxsep}{0pt}\fcolorbox{gray!10}{gray!10}{\strut
    \mycolorbox[text=\strut{the}, color=green!5.108638]%
}
\setlength{\fboxsep}{0pt}\fcolorbox{gray!10}{gray!10}{\strut
    \mycolorbox[text=\strut{operation}, color=green!9.334960]%
}
\setlength{\fboxsep}{0pt}\fcolorbox{gray!10}{gray!10}{\strut
    \mycolorbox[text=\strut{is}, color=green!17.198943]%
}
\setlength{\fboxsep}{0pt}\fcolorbox{gray!10}{gray!10}{\strut
    \mycolorbox[text=\strut{defined}, color=green!8.574353]%
}
\setlength{\fboxsep}{0pt}\fcolorbox{gray!10}{gray!10}{\strut
    \mycolorbox[text=\strut{for}, color=red!68.979812]%
}
\setlength{\fboxsep}{0pt}\fcolorbox{gray!10}{gray!10}{\strut
    \mycolorbox[text=\strut{these}, color=red!74.828710]%
}
\setlength{\fboxsep}{0pt}\fcolorbox{gray!10}{gray!10}{\strut
    \mycolorbox[text=\strut{fractions}, color=green!9.818974]%
    \mycolorbox[text=\strut{.}, color=red!20.181483]%
}
\setlength{\fboxsep}{0pt}\fcolorbox{gray!10}{gray!10}{\strut
    \mycolorbox[text=\strut{So}, color=red!34.963085]%
}
\setlength{\fboxsep}{0pt}\fcolorbox{gray!10}{gray!10}{\strut
    \mycolorbox[text=\strut{the}, color=red!15.082843]%
}
\setlength{\fboxsep}{0pt}\fcolorbox{gray!10}{gray!10}{\strut
    \mycolorbox[text=\strut{answer}, color=green!10.547568]%
}
\setlength{\fboxsep}{0pt}\fcolorbox{gray!10}{gray!10}{\strut
    \mycolorbox[text=\strut{is}, color=green!16.279482]%
}
\setlength{\fboxsep}{0pt}\fcolorbox{gray!10}{gray!10}{\strut
    \mycolorbox[text=\strut{4}, color=green!17.636586]%
    \mycolorbox[text=\strut{9}, color=green!17.636594]%
    \mycolorbox[text=\strut{.}, color=red!35.055131]%
}
\\
\\
{\tiny\color{gray}33}\,%
\setlength{\fboxsep}{0pt}\fcolorbox{gray!10}{gray!10}{\strut
    \mycolorbox[text=\strut{Alternatively}, color=red!50.502121]%
    \mycolorbox[text=\strut{,}, color=green!17.636594]%
}
\setlength{\fboxsep}{0pt}\fcolorbox{gray!10}{gray!10}{\strut
    \mycolorbox[text=\strut{maybe}, color=green!8.952107]%
}
\setlength{\fboxsep}{0pt}\fcolorbox{gray!10}{gray!10}{\strut
    \mycolorbox[text=\strut{the}]%
}
\setlength{\fboxsep}{0pt}\fcolorbox{gray!10}{gray!10}{\strut
    \mycolorbox[text=\strut{problem}, color=green!6.072385]%
}
\setlength{\fboxsep}{0pt}\fcolorbox{gray!10}{gray!10}{\strut
    \mycolorbox[text=\strut{expects}, color=red!47.444367]%
}
\setlength{\fboxsep}{0pt}\fcolorbox{gray!10}{gray!10}{\strut
    \mycolorbox[text=\strut{the}, color=red!11.803212]%
}
\setlength{\fboxsep}{0pt}\fcolorbox{gray!10}{gray!10}{\strut
    \mycolorbox[text=\strut{answer}, color=green!16.591480]%
}
\setlength{\fboxsep}{0pt}\fcolorbox{gray!10}{gray!10}{\strut
    \mycolorbox[text=\strut{to}, color=red!28.208618]%
}
\setlength{\fboxsep}{0pt}\fcolorbox{gray!10}{gray!10}{\strut
    \mycolorbox[text=\strut{be}, color=green!17.636594]%
}
\setlength{\fboxsep}{0pt}\fcolorbox{gray!10}{gray!10}{\strut
    \mycolorbox[text=\strut{written}, color=red!30.650911]%
}
\setlength{\fboxsep}{0pt}\fcolorbox{gray!10}{gray!10}{\strut
    \mycolorbox[text=\strut{as}, color=green!17.625144]%
}
\setlength{\fboxsep}{0pt}\fcolorbox{gray!10}{gray!10}{\strut
    \mycolorbox[text=\strut{a}, color=green!17.624877]%
}
\setlength{\fboxsep}{0pt}\fcolorbox{gray!10}{gray!10}{\strut
    \mycolorbox[text=\strut{fraction}, color=green!17.559175]%
    \mycolorbox[text=\strut{?}, color=green!15.371155]%
}
\setlength{\fboxsep}{0pt}\fcolorbox{gray!10}{gray!10}{\strut
    \mycolorbox[text=\strut{But}, color=green!15.564891]%
}
\setlength{\fboxsep}{0pt}\fcolorbox{gray!10}{gray!10}{\strut
    \mycolorbox[text=\strut{4}, color=green!17.636594]%
    \mycolorbox[text=\strut{9}, color=green!17.636594]%
}
\setlength{\fboxsep}{0pt}\fcolorbox{gray!10}{gray!10}{\strut
    \mycolorbox[text=\strut{is}, color=green!17.634027]%
}
\setlength{\fboxsep}{0pt}\fcolorbox{gray!10}{gray!10}{\strut
    \mycolorbox[text=\strut{an}, color=green!17.555510]%
}
\setlength{\fboxsep}{0pt}\fcolorbox{gray!10}{gray!10}{\strut
    \mycolorbox[text=\strut{integer}, color=green!17.636594]%
    \mycolorbox[text=\strut{,}, color=red!16.676484]%
}
\setlength{\fboxsep}{0pt}\fcolorbox{gray!10}{gray!10}{\strut
    \mycolorbox[text=\strut{so}, color=green!3.179123]%
}
\setlength{\fboxsep}{0pt}\fcolorbox{gray!10}{gray!10}{\strut
    \mycolorbox[text=\strut{it}, color=green!8.712350]%
    \mycolorbox[text=\strut{'s}, color=red!4.778425]%
}
\setlength{\fboxsep}{0pt}\fcolorbox{gray!10}{gray!10}{\strut
    \mycolorbox[text=\strut{already}, color=green!16.775244]%
}
\setlength{\fboxsep}{0pt}\fcolorbox{gray!10}{gray!10}{\strut
    \mycolorbox[text=\strut{simplified}, color=green!17.589639]%
    \mycolorbox[text=\strut{.}, color=green!16.595405]%
}
\setlength{\fboxsep}{0pt}\fcolorbox{gray!10}{gray!10}{\strut
    \mycolorbox[text=\strut{Therefore}, color=red!36.365257]%
    \mycolorbox[text=\strut{,}, color=green!17.636594]%
}
\setlength{\fboxsep}{0pt}\fcolorbox{gray!10}{gray!10}{\strut
    \mycolorbox[text=\strut{the}, color=green!12.650296]%
}
\setlength{\fboxsep}{0pt}\fcolorbox{gray!10}{gray!10}{\strut
    \mycolorbox[text=\strut{simplified}, color=red!40.862626]%
}
\setlength{\fboxsep}{0pt}\fcolorbox{gray!10}{gray!10}{\strut
    \mycolorbox[text=\strut{value}, color=green!17.636594]%
}
\setlength{\fboxsep}{0pt}\fcolorbox{gray!10}{gray!10}{\strut
    \mycolorbox[text=\strut{is}, color=green!17.624946]%
}
\setlength{\fboxsep}{0pt}\fcolorbox{gray!10}{gray!10}{\strut
    \mycolorbox[text=\strut{4}, color=green!17.636594]%
    \mycolorbox[text=\strut{9}, color=green!17.636594]%
    \mycolorbox[text=\strut{.}, color=red!47.508523]%
}
\\
\\
{\tiny\color{gray}34}\,%
\setlength{\fboxsep}{0pt}\fcolorbox{gray!10}{gray!10}{\strut
    \mycolorbox[text=\strut{Alternatively}, color=red!59.088063]%
    \mycolorbox[text=\strut{,}, color=green!17.636594]%
}
\setlength{\fboxsep}{0pt}\fcolorbox{gray!10}{gray!10}{\strut
    \mycolorbox[text=\strut{maybe}, color=green!5.969575]%
}
\setlength{\fboxsep}{0pt}\fcolorbox{gray!10}{gray!10}{\strut
    \mycolorbox[text=\strut{I}, color=red!7.027504]%
}
\setlength{\fboxsep}{0pt}\fcolorbox{gray!10}{gray!10}{\strut
    \mycolorbox[text=\strut{made}, color=red!20.410555]%
}
\setlength{\fboxsep}{0pt}\fcolorbox{gray!10}{gray!10}{\strut
    \mycolorbox[text=\strut{a}, color=green!9.697455]%
}
\setlength{\fboxsep}{0pt}\fcolorbox{gray!10}{gray!10}{\strut
    \mycolorbox[text=\strut{mistake}, color=green!13.778988]%
}
\setlength{\fboxsep}{0pt}\fcolorbox{gray!10}{gray!10}{\strut
    \mycolorbox[text=\strut{in}, color=green!17.404482]%
}
\setlength{\fboxsep}{0pt}\fcolorbox{gray!10}{gray!10}{\strut
    \mycolorbox[text=\strut{calculation}, color=red!42.021184]%
    \mycolorbox[text=\strut{?}, color=red!35.542067]%
}
\setlength{\fboxsep}{0pt}\fcolorbox{gray!10}{gray!10}{\strut
    \mycolorbox[text=\strut{Let}, color=green!17.627626]%
}
\setlength{\fboxsep}{0pt}\fcolorbox{gray!10}{gray!10}{\strut
    \mycolorbox[text=\strut{me}, color=green!17.609117]%
}
\setlength{\fboxsep}{0pt}\fcolorbox{gray!10}{gray!10}{\strut
    \mycolorbox[text=\strut{check}, color=green!5.632481]%
}
\setlength{\fboxsep}{0pt}\fcolorbox{gray!10}{gray!10}{\strut
    \mycolorbox[text=\strut{again}, color=green!3.887255]%
    \mycolorbox[text=\strut{:}, color=red!16.363623]%
}
\\
\\
{\tiny\color{gray}35}\,%
\setlength{\fboxsep}{0pt}\fcolorbox{gray!10}{gray!10}{\strut
    \mycolorbox[text=\strut{7}, color=red!31.010206]%
}
\setlength{\fboxsep}{0pt}\fcolorbox{gray!10}{gray!10}{\strut
    \mycolorbox[text=\strut{*}, color=red!14.323509]%
}
\setlength{\fboxsep}{0pt}\fcolorbox{gray!10}{gray!10}{\strut
    \mycolorbox[text=\strut{1}, color=green!17.636594]%
    \mycolorbox[text=\strut{0}, color=green!17.636594]%
}
\setlength{\fboxsep}{0pt}\fcolorbox{gray!10}{gray!10}{\strut
    \mycolorbox[text=\strut{=}, color=red!15.333190]%
}
\setlength{\fboxsep}{0pt}\fcolorbox{gray!10}{gray!10}{\strut
    \mycolorbox[text=\strut{7}, color=green!17.636594]%
    \mycolorbox[text=\strut{0}, color=green!17.636594]%
    \mycolorbox[text=\strut{.}, color=red!40.940907]%
}
\setlength{\fboxsep}{0pt}\fcolorbox{gray!10}{gray!10}{\strut
    \mycolorbox[text=\strut{Then}, color=red!28.781179]%
}
\setlength{\fboxsep}{0pt}\fcolorbox{gray!10}{gray!10}{\strut
    \mycolorbox[text=\strut{2}, color=green!16.304369]%
    \mycolorbox[text=\strut{1}, color=green!17.636594]%
}
\setlength{\fboxsep}{0pt}\fcolorbox{gray!10}{gray!10}{\strut
    \mycolorbox[text=\strut{/}, color=red!52.422841]%
}
\setlength{\fboxsep}{0pt}\fcolorbox{gray!10}{gray!10}{\strut
    \mycolorbox[text=\strut{3}, color=green!17.636594]%
    \mycolorbox[text=\strut{0}, color=green!17.636594]%
}
\setlength{\fboxsep}{0pt}\fcolorbox{gray!10}{gray!10}{\strut
    \mycolorbox[text=\strut{is}, color=red!67.819456]%
}
\setlength{\fboxsep}{0pt}\fcolorbox{gray!10}{gray!10}{\strut
    \mycolorbox[text=\strut{7}, color=red!10.857651]%
    \mycolorbox[text=\strut{/}, color=green!4.134808]%
    \mycolorbox[text=\strut{1}, color=green!17.636594]%
    \mycolorbox[text=\strut{0}, color=green!17.636594]%
    \mycolorbox[text=\strut{.}, color=green!17.605385]%
}
\setlength{\fboxsep}{0pt}\fcolorbox{gray!10}{gray!10}{\strut
    \mycolorbox[text=\strut{Then}]%
}
\setlength{\fboxsep}{0pt}\fcolorbox{gray!10}{gray!10}{\strut
    \mycolorbox[text=\strut{7}, color=green!17.636594]%
    \mycolorbox[text=\strut{0}, color=green!17.636594]%
}
\setlength{\fboxsep}{0pt}\fcolorbox{gray!10}{gray!10}{\strut
    \mycolorbox[text=\strut{*}, color=green!17.601964]%
}
\setlength{\fboxsep}{0pt}\fcolorbox{gray!10}{gray!10}{\strut
    \mycolorbox[text=\strut{7}, color=green!17.636594]%
    \mycolorbox[text=\strut{/}, color=green!16.304334]%
    \mycolorbox[text=\strut{1}, color=green!17.636594]%
    \mycolorbox[text=\strut{0}, color=green!17.636594]%
}
\setlength{\fboxsep}{0pt}\fcolorbox{gray!10}{gray!10}{\strut
    \mycolorbox[text=\strut{=}, color=red!38.534713]%
}
\setlength{\fboxsep}{0pt}\fcolorbox{gray!10}{gray!10}{\strut
    \mycolorbox[text=\strut{(}, color=red!22.685825]%
    \mycolorbox[text=\strut{7}, color=green!17.636594]%
    \mycolorbox[text=\strut{0}, color=green!17.636586]%
}
\setlength{\fboxsep}{0pt}\fcolorbox{gray!10}{gray!10}{\strut
    \mycolorbox[text=\strut{/}, color=green!12.791090]%
}
\setlength{\fboxsep}{0pt}\fcolorbox{gray!10}{gray!10}{\strut
    \mycolorbox[text=\strut{1}, color=green!17.636594]%
    \mycolorbox[text=\strut{0}, color=green!17.636594]%
    \mycolorbox[text=\strut{)}, color=green!15.922023]%
}
\setlength{\fboxsep}{0pt}\fcolorbox{gray!10}{gray!10}{\strut
    \mycolorbox[text=\strut{*}, color=green!17.636594]%
}
\setlength{\fboxsep}{0pt}\fcolorbox{gray!10}{gray!10}{\strut
    \mycolorbox[text=\strut{7}, color=green!17.636594]%
}
\setlength{\fboxsep}{0pt}\fcolorbox{gray!10}{gray!10}{\strut
    \mycolorbox[text=\strut{=}, color=green!17.636594]%
}
\setlength{\fboxsep}{0pt}\fcolorbox{gray!10}{gray!10}{\strut
    \mycolorbox[text=\strut{7}, color=green!17.636594]%
}
\setlength{\fboxsep}{0pt}\fcolorbox{gray!10}{gray!10}{\strut
    \mycolorbox[text=\strut{*}, color=green!17.596612]%
}
\setlength{\fboxsep}{0pt}\fcolorbox{gray!10}{gray!10}{\strut
    \mycolorbox[text=\strut{7}, color=green!17.636594]%
}
\setlength{\fboxsep}{0pt}\fcolorbox{gray!10}{gray!10}{\strut
    \mycolorbox[text=\strut{=}, color=green!17.636594]%
}
\setlength{\fboxsep}{0pt}\fcolorbox{gray!10}{gray!10}{\strut
    \mycolorbox[text=\strut{4}, color=green!17.636594]%
    \mycolorbox[text=\strut{9}, color=green!17.636594]%
    \mycolorbox[text=\strut{.}, color=green!17.636327]%
}
\setlength{\fboxsep}{0pt}\fcolorbox{gray!10}{gray!10}{\strut
    \mycolorbox[text=\strut{Correct}, color=red!55.049311]%
    \mycolorbox[text=\strut{.}, color=red!34.846998]%
}
\setlength{\fboxsep}{0pt}\fcolorbox{gray!10}{gray!10}{\strut
    \mycolorbox[text=\strut{Alternatively}, color=red!23.927611]%
    \mycolorbox[text=\strut{,}, color=green!17.596302]%
}
\setlength{\fboxsep}{0pt}\fcolorbox{gray!10}{gray!10}{\strut
    \mycolorbox[text=\strut{7}, color=green!16.985010]%
    \mycolorbox[text=\strut{0}, color=red!46.721669]%
}
\setlength{\fboxsep}{0pt}\fcolorbox{gray!10}{gray!10}{\strut
    \mycolorbox[text=\strut{*}, color=green!16.921452]%
}
\setlength{\fboxsep}{0pt}\fcolorbox{gray!10}{gray!10}{\strut
    \mycolorbox[text=\strut{2}, color=green!7.312136]%
    \mycolorbox[text=\strut{1}, color=green!17.636594]%
}
\setlength{\fboxsep}{0pt}\fcolorbox{gray!10}{gray!10}{\strut
    \mycolorbox[text=\strut{/}, color=green!9.375711]%
}
\setlength{\fboxsep}{0pt}\fcolorbox{gray!10}{gray!10}{\strut
    \mycolorbox[text=\strut{3}, color=green!17.636594]%
    \mycolorbox[text=\strut{0}, color=green!17.636594]%
    \mycolorbox[text=\strut{.}, color=red!10.527709]%
}
\setlength{\fboxsep}{0pt}\fcolorbox{gray!10}{gray!10}{\strut
    \mycolorbox[text=\strut{Let}, color=red!41.828295]%
}
\setlength{\fboxsep}{0pt}\fcolorbox{gray!10}{gray!10}{\strut
    \mycolorbox[text=\strut{me}, color=red!13.737538]%
}
\setlength{\fboxsep}{0pt}\fcolorbox{gray!10}{gray!10}{\strut
    \mycolorbox[text=\strut{compute}, color=green!12.123481]%
}
\setlength{\fboxsep}{0pt}\fcolorbox{gray!10}{gray!10}{\strut
    \mycolorbox[text=\strut{7}, color=green!17.496821]%
    \mycolorbox[text=\strut{0}, color=green!17.636594]%
}
\setlength{\fboxsep}{0pt}\fcolorbox{gray!10}{gray!10}{\strut
    \mycolorbox[text=\strut{divided}, color=green!5.761284]%
}
\setlength{\fboxsep}{0pt}\fcolorbox{gray!10}{gray!10}{\strut
    \mycolorbox[text=\strut{by}, color=green!17.636594]%
}
\setlength{\fboxsep}{0pt}\fcolorbox{gray!10}{gray!10}{\strut
    \mycolorbox[text=\strut{3}, color=green!17.636534]%
    \mycolorbox[text=\strut{0}, color=green!17.636594]%
}
\setlength{\fboxsep}{0pt}\fcolorbox{gray!10}{gray!10}{\strut
    \mycolorbox[text=\strut{first}, color=green!13.891568]%
    \mycolorbox[text=\strut{.}, color=green!4.800231]%
}
\setlength{\fboxsep}{0pt}\fcolorbox{gray!10}{gray!10}{\strut
    \mycolorbox[text=\strut{7}, color=green!17.636207]%
    \mycolorbox[text=\strut{0}, color=green!17.636594]%
}
\setlength{\fboxsep}{0pt}\fcolorbox{gray!10}{gray!10}{\strut
    \mycolorbox[text=\strut{/}, color=red!32.120933]%
}
\setlength{\fboxsep}{0pt}\fcolorbox{gray!10}{gray!10}{\strut
    \mycolorbox[text=\strut{3}, color=green!17.636594]%
    \mycolorbox[text=\strut{0}, color=green!17.636594]%
}
\setlength{\fboxsep}{0pt}\fcolorbox{gray!10}{gray!10}{\strut
    \mycolorbox[text=\strut{=}, color=red!5.371472]%
}
\setlength{\fboxsep}{0pt}\fcolorbox{gray!10}{gray!10}{\strut
    \mycolorbox[text=\strut{7}, color=green!17.636586]%
    \mycolorbox[text=\strut{/}, color=green!4.134808]%
    \mycolorbox[text=\strut{3}, color=green!17.636594]%
    \mycolorbox[text=\strut{.}, color=green!17.631141]%
}
\setlength{\fboxsep}{0pt}\fcolorbox{gray!10}{gray!10}{\strut
    \mycolorbox[text=\strut{Then}, color=green!17.553726]%
}
\setlength{\fboxsep}{0pt}\fcolorbox{gray!10}{gray!10}{\strut
    \mycolorbox[text=\strut{7}, color=green!17.636594]%
    \mycolorbox[text=\strut{/}, color=green!17.636457]%
    \mycolorbox[text=\strut{3}, color=green!17.636594]%
}
\setlength{\fboxsep}{0pt}\fcolorbox{gray!10}{gray!10}{\strut
    \mycolorbox[text=\strut{*}, color=green!17.129526]%
}
\setlength{\fboxsep}{0pt}\fcolorbox{gray!10}{gray!10}{\strut
    \mycolorbox[text=\strut{2}, color=green!17.636594]%
    \mycolorbox[text=\strut{1}, color=green!17.636594]%
}
\setlength{\fboxsep}{0pt}\fcolorbox{gray!10}{gray!10}{\strut
    \mycolorbox[text=\strut{=}, color=green!15.922181]%
}
\setlength{\fboxsep}{0pt}\fcolorbox{gray!10}{gray!10}{\strut
    \mycolorbox[text=\strut{7}, color=green!17.635931]%
}
\setlength{\fboxsep}{0pt}\fcolorbox{gray!10}{gray!10}{\strut
    \mycolorbox[text=\strut{*}, color=green!9.148972]%
}
\setlength{\fboxsep}{0pt}\fcolorbox{gray!10}{gray!10}{\strut
    \mycolorbox[text=\strut{7}, color=green!17.634604]%
}
\setlength{\fboxsep}{0pt}\fcolorbox{gray!10}{gray!10}{\strut
    \mycolorbox[text=\strut{=}, color=green!17.636594]%
}
\setlength{\fboxsep}{0pt}\fcolorbox{gray!10}{gray!10}{\strut
    \mycolorbox[text=\strut{4}, color=green!17.636594]%
    \mycolorbox[text=\strut{9}, color=green!17.636594]%
    \mycolorbox[text=\strut{.}, color=green!17.636526]%
}
\setlength{\fboxsep}{0pt}\fcolorbox{gray!10}{gray!10}{\strut
    \mycolorbox[text=\strut{Same}, color=red!59.934084]%
}
\setlength{\fboxsep}{0pt}\fcolorbox{gray!10}{gray!10}{\strut
    \mycolorbox[text=\strut{result}, color=red!17.421547]%
    \mycolorbox[text=\strut{.}, color=red!22.680166]%
}
\setlength{\fboxsep}{0pt}\fcolorbox{gray!10}{gray!10}{\strut
    \mycolorbox[text=\strut{So}, color=green!4.612074]%
}
\setlength{\fboxsep}{0pt}\fcolorbox{gray!10}{gray!10}{\strut
    \mycolorbox[text=\strut{that}, color=red!7.151260]%
    \mycolorbox[text=\strut{'s}]%
}
\setlength{\fboxsep}{0pt}\fcolorbox{gray!10}{gray!10}{\strut
    \mycolorbox[text=\strut{correct}, color=green!17.111162]%
    \mycolorbox[text=\strut{.}, color=green!9.702662]%
}
\\
\\
{\tiny\color{gray}36}\,%
\setlength{\fboxsep}{0pt}\fcolorbox{gray!10}{gray!10}{\strut
    \mycolorbox[text=\strut{Therefore}, color=red!27.788950]%
    \mycolorbox[text=\strut{,}, color=green!17.636594]%
}
\setlength{\fboxsep}{0pt}\fcolorbox{gray!10}{gray!10}{\strut
    \mycolorbox[text=\strut{I}, color=red!5.431724]%
}
\setlength{\fboxsep}{0pt}\fcolorbox{gray!10}{gray!10}{\strut
    \mycolorbox[text=\strut{think}, color=green!17.623068]%
}
\setlength{\fboxsep}{0pt}\fcolorbox{gray!10}{gray!10}{\strut
    \mycolorbox[text=\strut{the}]%
}
\setlength{\fboxsep}{0pt}\fcolorbox{gray!10}{gray!10}{\strut
    \mycolorbox[text=\strut{answer}, color=green!17.138823]%
}
\setlength{\fboxsep}{0pt}\fcolorbox{gray!10}{gray!10}{\strut
    \mycolorbox[text=\strut{is}, color=green!17.636388]%
}
\setlength{\fboxsep}{0pt}\fcolorbox{gray!10}{gray!10}{\strut
    \mycolorbox[text=\strut{4}, color=green!17.636594]%
    \mycolorbox[text=\strut{9}, color=green!17.636594]%
    \mycolorbox[text=\strut{.}, color=red!26.578951]%
}
\setlength{\fboxsep}{0pt}\fcolorbox{gray!10}{gray!10}{\strut
    \mycolorbox[text=\strut{Therefore}, color=red!28.063759]%
    \mycolorbox[text=\strut{,}, color=green!17.636594]%
}
\setlength{\fboxsep}{0pt}\fcolorbox{gray!10}{gray!10}{\strut
    \mycolorbox[text=\strut{the}, color=green!7.096182]%
}
\setlength{\fboxsep}{0pt}\fcolorbox{gray!10}{gray!10}{\strut
    \mycolorbox[text=\strut{simplified}, color=green!17.487357]%
}
\setlength{\fboxsep}{0pt}\fcolorbox{gray!10}{gray!10}{\strut
    \mycolorbox[text=\strut{value}, color=green!17.636594]%
}
\setlength{\fboxsep}{0pt}\fcolorbox{gray!10}{gray!10}{\strut
    \mycolorbox[text=\strut{is}]%
}
\setlength{\fboxsep}{0pt}\fcolorbox{gray!10}{gray!10}{\strut
    \mycolorbox[text=\strut{4}, color=green!17.636594]%
    \mycolorbox[text=\strut{9}, color=green!17.636594]%
    \mycolorbox[text=\strut{.}, color=red!64.099349]%
}

}}}}}
\vspace{-.1in}
\caption{Token-level visualization for Case 2: Qwen-4B (Part 2/3)}
\label{fig:case_2_2}
\vspace{-.1in}
\end{figure*}

%% file: case/wait3.tex
\begin{figure*}[htbp]
\centering
{\noindent

{\scriptsize
\setlength{\fboxrule}{.5pt}\fcolorbox{black}{green!10}{\parbox{1\textwidth}{\textbf{Model:} Qwen-4B \\ \textit{(Continued from Part 2/3 of Case 2)}}}

{\scriptsize
\setlength{\fboxrule}{.5pt}\fcolorbox{black}{gray!10}{\parbox{\textwidth}{
{\tiny\color{gray}37}\,%
\setlength{\fboxsep}{0pt}\fcolorbox{gray!10}{gray!10}{\strut
    \mycolorbox[text=\textbf{Response (cont.):}]%%
}
\\
\\
{\tiny\color{gray}38}\,%
\setlength{\fboxsep}{0pt}\fcolorbox{gray!10}{gray!10}{\strut
    \mycolorbox[text=\strut{**}, color=green!7.289977]%
    \mycolorbox[text=\strut{Final}, color=green!17.636594]%
}
\setlength{\fboxsep}{0pt}\fcolorbox{gray!10}{gray!10}{\strut
    \mycolorbox[text=\strut{Answer}, color=green!17.636534]%
    \mycolorbox[text=\strut{**}, color=green!17.636586]%
}
\\
{\tiny\color{gray}39}\,%
\setlength{\fboxsep}{0pt}\fcolorbox{gray!10}{gray!10}{\strut
    \mycolorbox[text=\strut{\textbackslash{}}, color=green!12.906066]%
    \mycolorbox[text=\strut{boxed}, color=green!17.636594]%
    \mycolorbox[text=\strut{\{}, color=green!17.636594]%
    \mycolorbox[text=\strut{4}, color=green!17.636594]%
    \mycolorbox[text=\strut{9}, color=green!17.636594]%
    \mycolorbox[text=\strut{\}}, color=green!17.636594]%
}
\\
{\tiny\color{gray}40}\,%
\setlength{\fboxsep}{0pt}\fcolorbox{gray!10}{gray!10}{\strut
    \mycolorbox[text=\strut{\textless{}/think\textgreater{}}, color=green!17.634527]%
}
\\
\\
{\tiny\color{gray}41}\,%
\setlength{\fboxsep}{0pt}\fcolorbox{gray!10}{gray!10}{\strut
    \mycolorbox[text=\strut{To}, color=green!9.609859]%
}
\setlength{\fboxsep}{0pt}\fcolorbox{gray!10}{gray!10}{\strut
    \mycolorbox[text=\strut{solve}, color=red!8.079574]%
}
\setlength{\fboxsep}{0pt}\fcolorbox{gray!10}{gray!10}{\strut
    \mycolorbox[text=\strut{the}, color=green!16.892629]%
}
\setlength{\fboxsep}{0pt}\fcolorbox{gray!10}{gray!10}{\strut
    \mycolorbox[text=\strut{problem}, color=red!25.280824]%
    \mycolorbox[text=\strut{,}, color=green!16.026692]%
}
\setlength{\fboxsep}{0pt}\fcolorbox{gray!10}{gray!10}{\strut
    \mycolorbox[text=\strut{we}, color=green!17.527692]%
}
\setlength{\fboxsep}{0pt}\fcolorbox{gray!10}{gray!10}{\strut
    \mycolorbox[text=\strut{are}, color=red!4.991463]%
}
\setlength{\fboxsep}{0pt}\fcolorbox{gray!10}{gray!10}{\strut
    \mycolorbox[text=\strut{given}, color=green!17.098874]%
}
\setlength{\fboxsep}{0pt}\fcolorbox{gray!10}{gray!10}{\strut
    \mycolorbox[text=\strut{a}, color=red!7.170001]%
}
\setlength{\fboxsep}{0pt}\fcolorbox{gray!10}{gray!10}{\strut
    \mycolorbox[text=\strut{custom}, color=green!13.702700]%
}
\setlength{\fboxsep}{0pt}\fcolorbox{gray!10}{gray!10}{\strut
    \mycolorbox[text=\strut{operation}, color=green!14.408027]%
}
\setlength{\fboxsep}{0pt}\fcolorbox{gray!10}{gray!10}{\strut
    \mycolorbox[text=\strut{defined}, color=red!17.296812]%
}
\setlength{\fboxsep}{0pt}\fcolorbox{gray!10}{gray!10}{\strut
    \mycolorbox[text=\strut{as}, color=green!14.388527]%
    \mycolorbox[text=\strut{:}, color=green!12.053764]%
}
\\
\\
{\tiny\color{gray}42}\,%
\setlength{\fboxsep}{0pt}\fcolorbox{gray!10}{gray!10}{\strut
    \mycolorbox[text=\strut{\$\$}, color=green!17.618906]%
}
\\
{\tiny\color{gray}43}\,%
\setlength{\fboxsep}{0pt}\fcolorbox{gray!10}{gray!10}{\strut
    \mycolorbox[text=\strut{\textbackslash{}}, color=green!17.636017]%
    \mycolorbox[text=\strut{frac}, color=green!17.635750]%
    \mycolorbox[text=\strut{\{}, color=green!17.636594]%
    \mycolorbox[text=\strut{m}, color=green!17.636551]%
    \mycolorbox[text=\strut{\}\{}, color=green!17.636594]%
    \mycolorbox[text=\strut{n}, color=green!17.636594]%
    \mycolorbox[text=\strut{\}}, color=green!17.578304]%
}
\setlength{\fboxsep}{0pt}\fcolorbox{gray!10}{gray!10}{\strut
    \mycolorbox[text=\strut{@}, color=green!17.575365]%
}
\setlength{\fboxsep}{0pt}\fcolorbox{gray!10}{gray!10}{\strut
    \mycolorbox[text=\strut{\textbackslash{}}, color=green!17.636517]%
    \mycolorbox[text=\strut{frac}, color=green!17.636569]%
    \mycolorbox[text=\strut{\{}, color=green!17.636594]%
    \mycolorbox[text=\strut{p}, color=green!17.636594]%
    \mycolorbox[text=\strut{\}\{}, color=green!17.636594]%
    \mycolorbox[text=\strut{q}, color=green!17.636594]%
    \mycolorbox[text=\strut{\}}, color=green!17.636517]%
}
\setlength{\fboxsep}{0pt}\fcolorbox{gray!10}{gray!10}{\strut
    \mycolorbox[text=\strut{=}, color=green!17.636577]%
}
\setlength{\fboxsep}{0pt}\fcolorbox{gray!10}{gray!10}{\strut
    \mycolorbox[text=\strut{m}, color=red!5.435282]%
}
\setlength{\fboxsep}{0pt}\fcolorbox{gray!10}{gray!10}{\strut
    \mycolorbox[text=\strut{\textbackslash{}}, color=green!17.635655]%
    \mycolorbox[text=\strut{cdot}, color=green!17.625144]%
}
\setlength{\fboxsep}{0pt}\fcolorbox{gray!10}{gray!10}{\strut
    \mycolorbox[text=\strut{p}, color=green!17.636586]%
}
\setlength{\fboxsep}{0pt}\fcolorbox{gray!10}{gray!10}{\strut
    \mycolorbox[text=\strut{\textbackslash{}}, color=green!17.636586]%
    \mycolorbox[text=\strut{cdot}, color=green!17.633312]%
}
\setlength{\fboxsep}{0pt}\fcolorbox{gray!10}{gray!10}{\strut
    \mycolorbox[text=\strut{\textbackslash{}}, color=green!17.636594]%
    \mycolorbox[text=\strut{left}, color=red!22.680170]%
    \mycolorbox[text=\strut{(}, color=red!28.769431]%
}
\setlength{\fboxsep}{0pt}\fcolorbox{gray!10}{gray!10}{\strut
    \mycolorbox[text=\strut{\textbackslash{}}, color=green!17.636594]%
    \mycolorbox[text=\strut{frac}, color=green!17.636594]%
    \mycolorbox[text=\strut{\{}, color=green!17.636594]%
    \mycolorbox[text=\strut{q}, color=green!17.636594]%
    \mycolorbox[text=\strut{\}\{}, color=green!17.636594]%
    \mycolorbox[text=\strut{n}, color=green!17.636594]%
    \mycolorbox[text=\strut{\}}, color=green!17.636327]%
}
\setlength{\fboxsep}{0pt}\fcolorbox{gray!10}{gray!10}{\strut
    \mycolorbox[text=\strut{\textbackslash{}}, color=green!17.636594]%
    \mycolorbox[text=\strut{right}, color=green!17.636586]%
    \mycolorbox[text=\strut{)}, color=green!17.490989]%
}
\\
{\tiny\color{gray}44}\,%
\setlength{\fboxsep}{0pt}\fcolorbox{gray!10}{gray!10}{\strut
    \mycolorbox[text=\strut{\$\$}, color=green!17.635742]%
}
\\
\\
{\tiny\color{gray}45}\,%
\setlength{\fboxsep}{0pt}\fcolorbox{gray!10}{gray!10}{\strut
    \mycolorbox[text=\strut{This}, color=red!32.907773]%
}
\setlength{\fboxsep}{0pt}\fcolorbox{gray!10}{gray!10}{\strut
    \mycolorbox[text=\strut{means}, color=red!63.573376]%
}
\setlength{\fboxsep}{0pt}\fcolorbox{gray!10}{gray!10}{\strut
    \mycolorbox[text=\strut{the}, color=red!81.575274]%
}
\setlength{\fboxsep}{0pt}\fcolorbox{gray!10}{gray!10}{\strut
    \mycolorbox[text=\strut{operation}]%
}
\setlength{\fboxsep}{0pt}\fcolorbox{gray!10}{gray!10}{\strut
    \mycolorbox[text=\strut{takes}, color=red!29.384920]%
}
\setlength{\fboxsep}{0pt}\fcolorbox{gray!10}{gray!10}{\strut
    \mycolorbox[text=\strut{two}, color=green!3.796483]%
}
\setlength{\fboxsep}{0pt}\fcolorbox{gray!10}{gray!10}{\strut
    \mycolorbox[text=\strut{fractions}, color=red!5.891134]%
    \mycolorbox[text=\strut{,}, color=red!28.508547]%
}
\setlength{\fboxsep}{0pt}\fcolorbox{gray!10}{gray!10}{\strut
    \mycolorbox[text=\strut{multi}, color=red!32.086719]%
    \mycolorbox[text=\strut{plies}, color=green!17.636577]%
}
\setlength{\fboxsep}{0pt}\fcolorbox{gray!10}{gray!10}{\strut
    \mycolorbox[text=\strut{their}, color=red!43.770166]%
}
\setlength{\fboxsep}{0pt}\fcolorbox{gray!10}{gray!10}{\strut
    \mycolorbox[text=\strut{numer}, color=green!16.490675]%
    \mycolorbox[text=\strut{ators}, color=green!17.636594]%
    \mycolorbox[text=\strut{,}, color=red!23.484189]%
}
\setlength{\fboxsep}{0pt}\fcolorbox{gray!10}{gray!10}{\strut
    \mycolorbox[text=\strut{and}, color=green!11.347717]%
}
\setlength{\fboxsep}{0pt}\fcolorbox{gray!10}{gray!10}{\strut
    \mycolorbox[text=\strut{multi}, color=red!69.671339]%
    \mycolorbox[text=\strut{plies}, color=green!17.636594]%
}
\setlength{\fboxsep}{0pt}\fcolorbox{gray!10}{gray!10}{\strut
    \mycolorbox[text=\strut{the}, color=red!19.069663]%
}
\setlength{\fboxsep}{0pt}\fcolorbox{gray!10}{gray!10}{\strut
    \mycolorbox[text=\strut{result}, color=green!16.809050]%
}
\setlength{\fboxsep}{0pt}\fcolorbox{gray!10}{gray!10}{\strut
    \mycolorbox[text=\strut{by}, color=green!17.619492]%
}
\setlength{\fboxsep}{0pt}\fcolorbox{gray!10}{gray!10}{\strut
    \mycolorbox[text=\strut{the}, color=green!17.338036]%
}
\setlength{\fboxsep}{0pt}\fcolorbox{gray!10}{gray!10}{\strut
    \mycolorbox[text=\strut{denominator}]%
}
\setlength{\fboxsep}{0pt}\fcolorbox{gray!10}{gray!10}{\strut
    \mycolorbox[text=\strut{of}, color=green!17.630839]%
}
\setlength{\fboxsep}{0pt}\fcolorbox{gray!10}{gray!10}{\strut
    \mycolorbox[text=\strut{the}, color=green!17.619638]%
}
\setlength{\fboxsep}{0pt}\fcolorbox{gray!10}{gray!10}{\strut
    \mycolorbox[text=\strut{first}, color=red!18.146070]%
}
\setlength{\fboxsep}{0pt}\fcolorbox{gray!10}{gray!10}{\strut
    \mycolorbox[text=\strut{fraction}, color=green!17.596991]%
}
\setlength{\fboxsep}{0pt}\fcolorbox{gray!10}{gray!10}{\strut
    \mycolorbox[text=\strut{divided}, color=green!17.022629]%
}
\setlength{\fboxsep}{0pt}\fcolorbox{gray!10}{gray!10}{\strut
    \mycolorbox[text=\strut{by}, color=green!17.619561]%
}
\setlength{\fboxsep}{0pt}\fcolorbox{gray!10}{gray!10}{\strut
    \mycolorbox[text=\strut{the}, color=green!17.635871]%
}
\setlength{\fboxsep}{0pt}\fcolorbox{gray!10}{gray!10}{\strut
    \mycolorbox[text=\strut{denominator}, color=green!15.899688]%
}
\setlength{\fboxsep}{0pt}\fcolorbox{gray!10}{gray!10}{\strut
    \mycolorbox[text=\strut{of}, color=green!17.636569]%
}
\setlength{\fboxsep}{0pt}\fcolorbox{gray!10}{gray!10}{\strut
    \mycolorbox[text=\strut{the}, color=green!17.636586]%
}
\setlength{\fboxsep}{0pt}\fcolorbox{gray!10}{gray!10}{\strut
    \mycolorbox[text=\strut{second}, color=green!17.634441]%
    \mycolorbox[text=\strut{.}, color=red!36.952859]%
}
\\
\\
{\tiny\color{gray}46}\,%
\setlength{\fboxsep}{0pt}\fcolorbox{gray!10}{gray!10}{\strut
    \mycolorbox[text=\strut{We}, color=red!35.572864]%
}
\setlength{\fboxsep}{0pt}\fcolorbox{gray!10}{gray!10}{\strut
    \mycolorbox[text=\strut{are}, color=green!17.162964]%
}
\setlength{\fboxsep}{0pt}\fcolorbox{gray!10}{gray!10}{\strut
    \mycolorbox[text=\strut{asked}, color=green!9.102412]%
}
\setlength{\fboxsep}{0pt}\fcolorbox{gray!10}{gray!10}{\strut
    \mycolorbox[text=\strut{to}, color=green!17.636439]%
}
\setlength{\fboxsep}{0pt}\fcolorbox{gray!10}{gray!10}{\strut
    \mycolorbox[text=\strut{compute}, color=red!10.165190]%
    \mycolorbox[text=\strut{:}, color=green!15.145682]%
}
\\
\\
{\tiny\color{gray}47}\,%
\setlength{\fboxsep}{0pt}\fcolorbox{gray!10}{gray!10}{\strut
    \mycolorbox[text=\strut{\$\$}, color=green!17.636560]%
}
\\
{\tiny\color{gray}48}\,%
\setlength{\fboxsep}{0pt}\fcolorbox{gray!10}{gray!10}{\strut
    \mycolorbox[text=\strut{\textbackslash{}}, color=green!17.636569]%
    \mycolorbox[text=\strut{frac}, color=green!17.636095]%
    \mycolorbox[text=\strut{\{}, color=green!17.636594]%
    \mycolorbox[text=\strut{7}, color=green!17.636594]%
    \mycolorbox[text=\strut{\}\{}, color=green!17.636594]%
    \mycolorbox[text=\strut{3}, color=green!17.636594]%
    \mycolorbox[text=\strut{0}, color=green!17.636594]%
    \mycolorbox[text=\strut{\}}, color=green!17.591165]%
}
\setlength{\fboxsep}{0pt}\fcolorbox{gray!10}{gray!10}{\strut
    \mycolorbox[text=\strut{@}, color=green!17.635294]%
}
\setlength{\fboxsep}{0pt}\fcolorbox{gray!10}{gray!10}{\strut
    \mycolorbox[text=\strut{\textbackslash{}}, color=green!17.636594]%
    \mycolorbox[text=\strut{frac}, color=green!17.636534]%
    \mycolorbox[text=\strut{\{}, color=green!17.636594]%
    \mycolorbox[text=\strut{1}, color=green!17.636594]%
    \mycolorbox[text=\strut{0}, color=green!17.636594]%
    \mycolorbox[text=\strut{\}\{}, color=green!17.636594]%
    \mycolorbox[text=\strut{2}, color=green!17.636594]%
    \mycolorbox[text=\strut{1}, color=green!17.636594]%
    \mycolorbox[text=\strut{\}}, color=green!17.636336]%
}
\\
{\tiny\color{gray}49}\,%
\setlength{\fboxsep}{0pt}\fcolorbox{gray!10}{gray!10}{\strut
    \mycolorbox[text=\strut{\$\$}, color=green!17.636586]%
}
\\
\\
{\tiny\color{gray}50}\,%
\setlength{\fboxsep}{0pt}\fcolorbox{gray!10}{gray!10}{\strut
    \mycolorbox[text=\strut{---}, color=green!8.013825]%
}
\\
\\
{\tiny\color{gray}51}\,%
\setlength{\fboxsep}{0pt}\fcolorbox{gray!10}{gray!10}{\strut
    \mycolorbox[text=\strut{\#\#\#}, color=green!17.010816]%
}
\setlength{\fboxsep}{0pt}\fcolorbox{gray!10}{gray!10}{\strut
    \mycolorbox[text=\strut{Step}, color=green!16.997303]%
}
\setlength{\fboxsep}{0pt}\fcolorbox{gray!10}{gray!10}{\strut
    \mycolorbox[text=\strut{1}, color=green!17.636594]%
    \mycolorbox[text=\strut{:}, color=green!17.636594]%
}
\setlength{\fboxsep}{0pt}\fcolorbox{gray!10}{gray!10}{\strut
    \mycolorbox[text=\strut{Identify}, color=green!14.980721]%
}
\setlength{\fboxsep}{0pt}\fcolorbox{gray!10}{gray!10}{\strut
    \mycolorbox[text=\strut{the}]%
}
\setlength{\fboxsep}{0pt}\fcolorbox{gray!10}{gray!10}{\strut
    \mycolorbox[text=\strut{components}, color=red!30.602776]%
}
\\
\\
{\tiny\color{gray}52}\,%
\setlength{\fboxsep}{0pt}\fcolorbox{gray!10}{gray!10}{\strut
    \mycolorbox[text=\strut{From}, color=red!7.679716]%
}
\setlength{\fboxsep}{0pt}\fcolorbox{gray!10}{gray!10}{\strut
    \mycolorbox[text=\strut{the}, color=green!17.626488]%
}
\setlength{\fboxsep}{0pt}\fcolorbox{gray!10}{gray!10}{\strut
    \mycolorbox[text=\strut{definition}, color=red!46.143561]%
    \mycolorbox[text=\strut{,}, color=red!33.220851]%
}
\setlength{\fboxsep}{0pt}\fcolorbox{gray!10}{gray!10}{\strut
    \mycolorbox[text=\strut{we}, color=red!11.293978]%
}
\setlength{\fboxsep}{0pt}\fcolorbox{gray!10}{gray!10}{\strut
    \mycolorbox[text=\strut{assign}, color=red!83.002039]%
    \mycolorbox[text=\strut{:}, color=red!58.586405]%
}
\\
{\tiny\color{gray}53}\,%
\setlength{\fboxsep}{0pt}\fcolorbox{gray!10}{gray!10}{\strut
    \mycolorbox[text=\strut{-}, color=green!17.635854]%
}
\setlength{\fboxsep}{0pt}\fcolorbox{gray!10}{gray!10}{\strut
    \mycolorbox[text=\strut{\$}, color=green!17.360973]%
}
\setlength{\fboxsep}{0pt}\fcolorbox{gray!10}{gray!10}{\strut
    \mycolorbox[text=\strut{m}, color=green!16.688855]%
}
\setlength{\fboxsep}{0pt}\fcolorbox{gray!10}{gray!10}{\strut
    \mycolorbox[text=\strut{=}, color=green!17.636586]%
}
\setlength{\fboxsep}{0pt}\fcolorbox{gray!10}{gray!10}{\strut
    \mycolorbox[text=\strut{7}, color=green!17.636594]%
}
\setlength{\fboxsep}{0pt}\fcolorbox{gray!10}{gray!10}{\strut
    \mycolorbox[text=\strut{\$}]%
}
\\
{\tiny\color{gray}54}\,%
\setlength{\fboxsep}{0pt}\fcolorbox{gray!10}{gray!10}{\strut
    \mycolorbox[text=\strut{-}, color=green!17.636594]%
}
\setlength{\fboxsep}{0pt}\fcolorbox{gray!10}{gray!10}{\strut
    \mycolorbox[text=\strut{\$}, color=green!17.636594]%
}
\setlength{\fboxsep}{0pt}\fcolorbox{gray!10}{gray!10}{\strut
    \mycolorbox[text=\strut{n}, color=green!17.615097]%
}
\setlength{\fboxsep}{0pt}\fcolorbox{gray!10}{gray!10}{\strut
    \mycolorbox[text=\strut{=}, color=green!17.636594]%
}
\setlength{\fboxsep}{0pt}\fcolorbox{gray!10}{gray!10}{\strut
    \mycolorbox[text=\strut{3}, color=green!17.636594]%
    \mycolorbox[text=\strut{0}, color=green!17.636594]%
}
\setlength{\fboxsep}{0pt}\fcolorbox{gray!10}{gray!10}{\strut
    \mycolorbox[text=\strut{\$}, color=green!17.551725]%
}
\\
{\tiny\color{gray}55}\,%
\setlength{\fboxsep}{0pt}\fcolorbox{gray!10}{gray!10}{\strut
    \mycolorbox[text=\strut{-}, color=green!17.636594]%
}
\setlength{\fboxsep}{0pt}\fcolorbox{gray!10}{gray!10}{\strut
    \mycolorbox[text=\strut{\$}, color=green!17.636594]%
}
\setlength{\fboxsep}{0pt}\fcolorbox{gray!10}{gray!10}{\strut
    \mycolorbox[text=\strut{p}, color=green!17.636594]%
}
\setlength{\fboxsep}{0pt}\fcolorbox{gray!10}{gray!10}{\strut
    \mycolorbox[text=\strut{=}, color=green!17.636594]%
}
\setlength{\fboxsep}{0pt}\fcolorbox{gray!10}{gray!10}{\strut
    \mycolorbox[text=\strut{1}, color=green!17.636594]%
    \mycolorbox[text=\strut{0}, color=green!17.636594]%
}
\setlength{\fboxsep}{0pt}\fcolorbox{gray!10}{gray!10}{\strut
    \mycolorbox[text=\strut{\$}, color=green!17.634311]%
}
\\
{\tiny\color{gray}56}\,%
\setlength{\fboxsep}{0pt}\fcolorbox{gray!10}{gray!10}{\strut
    \mycolorbox[text=\strut{-}, color=green!17.636594]%
}
\setlength{\fboxsep}{0pt}\fcolorbox{gray!10}{gray!10}{\strut
    \mycolorbox[text=\strut{\$}, color=green!17.636594]%
}
\setlength{\fboxsep}{0pt}\fcolorbox{gray!10}{gray!10}{\strut
    \mycolorbox[text=\strut{q}, color=green!17.636594]%
}
\setlength{\fboxsep}{0pt}\fcolorbox{gray!10}{gray!10}{\strut
    \mycolorbox[text=\strut{=}, color=green!17.636594]%
}
\setlength{\fboxsep}{0pt}\fcolorbox{gray!10}{gray!10}{\strut
    \mycolorbox[text=\strut{2}, color=green!17.636594]%
    \mycolorbox[text=\strut{1}, color=green!17.636594]%
}
\setlength{\fboxsep}{0pt}\fcolorbox{gray!10}{gray!10}{\strut
    \mycolorbox[text=\strut{\$}, color=green!17.632709]%
}
\\
\\
{\tiny\color{gray}57}\,%
\setlength{\fboxsep}{0pt}\fcolorbox{gray!10}{gray!10}{\strut
    \mycolorbox[text=\strut{Sub}, color=red!62.920893]%
    \mycolorbox[text=\strut{stitute}, color=red!19.664046]%
}
\setlength{\fboxsep}{0pt}\fcolorbox{gray!10}{gray!10}{\strut
    \mycolorbox[text=\strut{into}, color=red!16.694932]%
}
\setlength{\fboxsep}{0pt}\fcolorbox{gray!10}{gray!10}{\strut
    \mycolorbox[text=\strut{the}, color=green!17.636594]%
}
\setlength{\fboxsep}{0pt}\fcolorbox{gray!10}{gray!10}{\strut
    \mycolorbox[text=\strut{formula}, color=red!45.005554]%
    \mycolorbox[text=\strut{:}, color=green!17.142225]%
}
\\
\\
{\tiny\color{gray}58}\,%
\setlength{\fboxsep}{0pt}\fcolorbox{gray!10}{gray!10}{\strut
    \mycolorbox[text=\strut{\$\$}, color=green!17.636594]%
}
\\
{\tiny\color{gray}59}\,%
\setlength{\fboxsep}{0pt}\fcolorbox{gray!10}{gray!10}{\strut
    \mycolorbox[text=\strut{\textbackslash{}}, color=green!16.006586]%
    \mycolorbox[text=\strut{frac}, color=green!17.401572]%
    \mycolorbox[text=\strut{\{}, color=green!17.636500]%
    \mycolorbox[text=\strut{7}, color=green!17.432558]%
    \mycolorbox[text=\strut{\}\{}, color=green!17.636396]%
    \mycolorbox[text=\strut{3}, color=green!17.636594]%
    \mycolorbox[text=\strut{0}, color=green!17.636594]%
    \mycolorbox[text=\strut{\}}, color=green!17.627522]%
}
\setlength{\fboxsep}{0pt}\fcolorbox{gray!10}{gray!10}{\strut
    \mycolorbox[text=\strut{@}, color=green!17.636422]%
}
\setlength{\fboxsep}{0pt}\fcolorbox{gray!10}{gray!10}{\strut
    \mycolorbox[text=\strut{\textbackslash{}}, color=green!17.636594]%
    \mycolorbox[text=\strut{frac}, color=green!17.636594]%
    \mycolorbox[text=\strut{\{}, color=green!17.636594]%
    \mycolorbox[text=\strut{1}, color=green!17.636594]%
    \mycolorbox[text=\strut{0}, color=green!17.636594]%
    \mycolorbox[text=\strut{\}\{}, color=green!17.636594]%
    \mycolorbox[text=\strut{2}, color=green!17.636594]%
    \mycolorbox[text=\strut{1}, color=green!17.636594]%
    \mycolorbox[text=\strut{\}}, color=green!17.635225]%
}
\setlength{\fboxsep}{0pt}\fcolorbox{gray!10}{gray!10}{\strut
    \mycolorbox[text=\strut{=}, color=green!17.636560]%
}
\setlength{\fboxsep}{0pt}\fcolorbox{gray!10}{gray!10}{\strut
    \mycolorbox[text=\strut{7}, color=green!17.636594]%
}
\setlength{\fboxsep}{0pt}\fcolorbox{gray!10}{gray!10}{\strut
    \mycolorbox[text=\strut{\textbackslash{}}, color=green!17.636586]%
    \mycolorbox[text=\strut{cdot}, color=green!17.585252]%
}
\setlength{\fboxsep}{0pt}\fcolorbox{gray!10}{gray!10}{\strut
    \mycolorbox[text=\strut{1}, color=green!17.636594]%
    \mycolorbox[text=\strut{0}, color=green!17.636594]%
}
\setlength{\fboxsep}{0pt}\fcolorbox{gray!10}{gray!10}{\strut
    \mycolorbox[text=\strut{\textbackslash{}}, color=green!17.636594]%
    \mycolorbox[text=\strut{cdot}, color=green!17.636517]%
}
\setlength{\fboxsep}{0pt}\fcolorbox{gray!10}{gray!10}{\strut
    \mycolorbox[text=\strut{\textbackslash{}}, color=green!17.636577]%
    \mycolorbox[text=\strut{left}, color=green!13.474224]%
    \mycolorbox[text=\strut{(}, color=green!17.406261]%
}
\setlength{\fboxsep}{0pt}\fcolorbox{gray!10}{gray!10}{\strut
    \mycolorbox[text=\strut{\textbackslash{}}, color=green!17.636594]%
    \mycolorbox[text=\strut{frac}, color=green!17.636594]%
    \mycolorbox[text=\strut{\{}, color=green!17.636594]%
    \mycolorbox[text=\strut{2}, color=green!17.636569]%
    \mycolorbox[text=\strut{1}, color=green!17.636594]%
    \mycolorbox[text=\strut{\}\{}, color=green!17.636586]%
    \mycolorbox[text=\strut{3}, color=green!17.636594]%
    \mycolorbox[text=\strut{0}, color=green!17.636594]%
    \mycolorbox[text=\strut{\}}, color=green!17.636560]%
}
\setlength{\fboxsep}{0pt}\fcolorbox{gray!10}{gray!10}{\strut
    \mycolorbox[text=\strut{\textbackslash{}}, color=green!17.636594]%
    \mycolorbox[text=\strut{right}, color=green!17.636594]%
    \mycolorbox[text=\strut{)}, color=green!17.591010]%
}
\\
{\tiny\color{gray}60}\,%
\setlength{\fboxsep}{0pt}\fcolorbox{gray!10}{gray!10}{\strut
    \mycolorbox[text=\strut{\$\$}, color=green!17.629564]%
}
\\
\\
{\tiny\color{gray}61}\,%
\setlength{\fboxsep}{0pt}\fcolorbox{gray!10}{gray!10}{\strut
    \mycolorbox[text=\strut{---}, color=green!17.429536]%
}
\\
\\
{\tiny\color{gray}62}\,%
\setlength{\fboxsep}{0pt}\fcolorbox{gray!10}{gray!10}{\strut
    \mycolorbox[text=\strut{\#\#\#}, color=green!17.636594]%
}
\setlength{\fboxsep}{0pt}\fcolorbox{gray!10}{gray!10}{\strut
    \mycolorbox[text=\strut{Step}, color=green!17.636594]%
}
\setlength{\fboxsep}{0pt}\fcolorbox{gray!10}{gray!10}{\strut
    \mycolorbox[text=\strut{2}, color=green!17.636517]%
    \mycolorbox[text=\strut{:}, color=green!17.636594]%
}
\setlength{\fboxsep}{0pt}\fcolorbox{gray!10}{gray!10}{\strut
    \mycolorbox[text=\strut{Simpl}, color=red!12.317425]%
    \mycolorbox[text=\strut{ify}, color=green!17.636526]%
}
\setlength{\fboxsep}{0pt}\fcolorbox{gray!10}{gray!10}{\strut
    \mycolorbox[text=\strut{the}]%
}
\setlength{\fboxsep}{0pt}\fcolorbox{gray!10}{gray!10}{\strut
    \mycolorbox[text=\strut{expression}, color=green!15.341257]%
}
\\
\\
{\tiny\color{gray}63}\,%
\setlength{\fboxsep}{0pt}\fcolorbox{gray!10}{gray!10}{\strut
    \mycolorbox[text=\strut{First}, color=green!6.977437]%
    \mycolorbox[text=\strut{,}, color=green!11.225026]%
}
\setlength{\fboxsep}{0pt}\fcolorbox{gray!10}{gray!10}{\strut
    \mycolorbox[text=\strut{compute}, color=red!5.912542]%
}
\setlength{\fboxsep}{0pt}\fcolorbox{gray!10}{gray!10}{\strut
    \mycolorbox[text=\strut{the}]%
}
\setlength{\fboxsep}{0pt}\fcolorbox{gray!10}{gray!10}{\strut
    \mycolorbox[text=\strut{multiplication}, color=red!58.128135]%
}
\setlength{\fboxsep}{0pt}\fcolorbox{gray!10}{gray!10}{\strut
    \mycolorbox[text=\strut{of}, color=red!42.633774]%
}
\setlength{\fboxsep}{0pt}\fcolorbox{gray!10}{gray!10}{\strut
    \mycolorbox[text=\strut{the}, color=green!6.121318]%
}
\setlength{\fboxsep}{0pt}\fcolorbox{gray!10}{gray!10}{\strut
    \mycolorbox[text=\strut{numer}, color=red!12.306910]%
    \mycolorbox[text=\strut{ators}, color=green!17.636534]%
    \mycolorbox[text=\strut{:}, color=green!16.075569]%
}
\\
\\
{\tiny\color{gray}64}\,%
\setlength{\fboxsep}{0pt}\fcolorbox{gray!10}{gray!10}{\strut
    \mycolorbox[text=\strut{\$\$}, color=green!17.625101]%
}
\\
{\tiny\color{gray}65}\,%
\setlength{\fboxsep}{0pt}\fcolorbox{gray!10}{gray!10}{\strut
    \mycolorbox[text=\strut{7}, color=green!17.636198]%
}
\setlength{\fboxsep}{0pt}\fcolorbox{gray!10}{gray!10}{\strut
    \mycolorbox[text=\strut{\textbackslash{}}, color=green!17.636534]%
    \mycolorbox[text=\strut{cdot}, color=green!17.570667]%
}
\setlength{\fboxsep}{0pt}\fcolorbox{gray!10}{gray!10}{\strut
    \mycolorbox[text=\strut{1}, color=green!17.636594]%
    \mycolorbox[text=\strut{0}, color=green!17.636594]%
}
\setlength{\fboxsep}{0pt}\fcolorbox{gray!10}{gray!10}{\strut
    \mycolorbox[text=\strut{=}, color=green!17.636594]%
}
\setlength{\fboxsep}{0pt}\fcolorbox{gray!10}{gray!10}{\strut
    \mycolorbox[text=\strut{7}, color=green!17.636594]%
    \mycolorbox[text=\strut{0}, color=green!17.636594]%
}
\\
{\tiny\color{gray}66}\,%
\setlength{\fboxsep}{0pt}\fcolorbox{gray!10}{gray!10}{\strut
    \mycolorbox[text=\strut{\$\$}, color=green!17.636491]%
}
\\
\\
{\tiny\color{gray}67}\,%
\setlength{\fboxsep}{0pt}\fcolorbox{gray!10}{gray!10}{\strut
    \mycolorbox[text=\strut{Now}, color=red!53.890156]%
}
\setlength{\fboxsep}{0pt}\fcolorbox{gray!10}{gray!10}{\strut
    \mycolorbox[text=\strut{compute}, color=red!24.670623]%
}
\setlength{\fboxsep}{0pt}\fcolorbox{gray!10}{gray!10}{\strut
    \mycolorbox[text=\strut{the}, color=green!10.856589]%
}
\setlength{\fboxsep}{0pt}\fcolorbox{gray!10}{gray!10}{\strut
    \mycolorbox[text=\strut{fraction}, color=green!11.479600]%
    \mycolorbox[text=\strut{:}, color=green!14.142936]%
}
\\
\\
{\tiny\color{gray}68}\,%
\setlength{\fboxsep}{0pt}\fcolorbox{gray!10}{gray!10}{\strut
    \mycolorbox[text=\strut{\$\$}, color=green!17.636508]%
}
\\
{\tiny\color{gray}69}\,%
\setlength{\fboxsep}{0pt}\fcolorbox{gray!10}{gray!10}{\strut
    \mycolorbox[text=\strut{\textbackslash{}}, color=green!17.636534]%
    \mycolorbox[text=\strut{frac}, color=green!17.636594]%
    \mycolorbox[text=\strut{\{}, color=green!17.636594]%
    \mycolorbox[text=\strut{2}, color=green!17.636474]%
    \mycolorbox[text=\strut{1}, color=green!17.636594]%
    \mycolorbox[text=\strut{\}\{}, color=green!17.636594]%
    \mycolorbox[text=\strut{3}, color=green!17.636594]%
    \mycolorbox[text=\strut{0}, color=green!17.636594]%
    \mycolorbox[text=\strut{\}}, color=green!16.125498]%
}
\setlength{\fboxsep}{0pt}\fcolorbox{gray!10}{gray!10}{\strut
    \mycolorbox[text=\strut{=}, color=green!17.635311]%
}
\setlength{\fboxsep}{0pt}\fcolorbox{gray!10}{gray!10}{\strut
    \mycolorbox[text=\strut{\textbackslash{}}, color=green!17.636577]%
    \mycolorbox[text=\strut{frac}, color=green!17.636586]%
    \mycolorbox[text=\strut{\{}, color=green!17.636577]%
    \mycolorbox[text=\strut{7}, color=green!17.636422]%
    \mycolorbox[text=\strut{\}\{}, color=green!17.626454]%
    \mycolorbox[text=\strut{1}, color=green!17.636594]%
    \mycolorbox[text=\strut{0}, color=green!17.636594]%
    \mycolorbox[text=\strut{\}}, color=green!13.473190]%
}
\\
{\tiny\color{gray}70}\,%
\setlength{\fboxsep}{0pt}\fcolorbox{gray!10}{gray!10}{\strut
    \mycolorbox[text=\strut{\$\$}, color=green!17.619888]%
}
\\
\\

{\tiny\color{gray}71}\,%
\setlength{\fboxsep}{0pt}\fcolorbox{gray!10}{gray!10}{\strut
    \mycolorbox[text=\strut{Now}, color=red!17.432767]%
}
\setlength{\fboxsep}{0pt}\fcolorbox{gray!10}{gray!10}{\strut
    \mycolorbox[text=\strut{multiply}, color=green!10.018788]%
    \mycolorbox[text=\strut{:}, color=red!23.157977]%
}
\\
\\
{\tiny\color{gray}72}\,%
\setlength{\fboxsep}{0pt}\fcolorbox{gray!10}{gray!10}{\strut
    \mycolorbox[text=\strut{\$\$}, color=green!17.636594]%
}
\\
{\tiny\color{gray}73}\,%
\setlength{\fboxsep}{0pt}\fcolorbox{gray!10}{gray!10}{\strut
    \mycolorbox[text=\strut{7}, color=green!17.636129]%
    \mycolorbox[text=\strut{0}, color=green!17.636414]%
}
\setlength{\fboxsep}{0pt}\fcolorbox{gray!10}{gray!10}{\strut
    \mycolorbox[text=\strut{\textbackslash{}}, color=green!17.636594]%
    \mycolorbox[text=\strut{cdot}, color=green!17.632381]%
}
\setlength{\fboxsep}{0pt}\fcolorbox{gray!10}{gray!10}{\strut
    \mycolorbox[text=\strut{\textbackslash{}}, color=green!17.636586]%
    \mycolorbox[text=\strut{frac}, color=green!15.692360]%
    \mycolorbox[text=\strut{\{}, color=green!17.636594]%
    \mycolorbox[text=\strut{7}, color=green!17.636138]%
    \mycolorbox[text=\strut{\}\{}, color=green!17.636594]%
    \mycolorbox[text=\strut{1}, color=green!17.636594]%
    \mycolorbox[text=\strut{0}, color=green!17.636594]%
    \mycolorbox[text=\strut{\}}, color=green!17.256659]%
}
\setlength{\fboxsep}{0pt}\fcolorbox{gray!10}{gray!10}{\strut
    \mycolorbox[text=\strut{=}, color=green!17.636526]%
}
\setlength{\fboxsep}{0pt}\fcolorbox{gray!10}{gray!10}{\strut
    \mycolorbox[text=\strut{\textbackslash{}}, color=green!13.761571]%
    \mycolorbox[text=\strut{frac}, color=green!17.205414]%
    \mycolorbox[text=\strut{\{}, color=green!17.636457]%
    \mycolorbox[text=\strut{7}, color=green!16.922310]%
    \mycolorbox[text=\strut{0}, color=green!17.636129]%
}
\setlength{\fboxsep}{0pt}\fcolorbox{gray!10}{gray!10}{\strut
    \mycolorbox[text=\strut{\textbackslash{}}, color=green!17.630426]%
    \mycolorbox[text=\strut{cdot}, color=green!17.570667]%
}
\setlength{\fboxsep}{0pt}\fcolorbox{gray!10}{gray!10}{\strut
    \mycolorbox[text=\strut{7}, color=green!17.636465]%
    \mycolorbox[text=\strut{\}\{}, color=green!17.636586]%
    \mycolorbox[text=\strut{1}, color=green!17.636586]%
    \mycolorbox[text=\strut{0}, color=green!17.636594]%
    \mycolorbox[text=\strut{\}}, color=green!17.527873]%
}
\setlength{\fboxsep}{0pt}\fcolorbox{gray!10}{gray!10}{\strut
    \mycolorbox[text=\strut{=}, color=green!17.636586]%
}
\setlength{\fboxsep}{0pt}\fcolorbox{gray!10}{gray!10}{\strut
    \mycolorbox[text=\strut{\textbackslash{}}, color=green!17.636198]%
    \mycolorbox[text=\strut{frac}, color=green!17.636551]%
    \mycolorbox[text=\strut{\{}, color=green!17.636577]%
    \mycolorbox[text=\strut{4}, color=green!17.636560]%
    \mycolorbox[text=\strut{9}, color=green!17.636594]%
    \mycolorbox[text=\strut{0}, color=green!17.636586]%
    \mycolorbox[text=\strut{\}\{}, color=green!17.636594]%
    \mycolorbox[text=\strut{1}, color=green!17.636594]%
    \mycolorbox[text=\strut{0}, color=green!17.636594]%
    \mycolorbox[text=\strut{\}}, color=green!17.551932]%
}
\setlength{\fboxsep}{0pt}\fcolorbox{gray!10}{gray!10}{\strut
    \mycolorbox[text=\strut{=}, color=green!17.636594]%
}
\setlength{\fboxsep}{0pt}\fcolorbox{gray!10}{gray!10}{\strut
    \mycolorbox[text=\strut{4}, color=green!17.636551]%
    \mycolorbox[text=\strut{9}, color=green!17.636594]%
}
\\
{\tiny\color{gray}74}\,%
\setlength{\fboxsep}{0pt}\fcolorbox{gray!10}{gray!10}{\strut
    \mycolorbox[text=\strut{\$\$}, color=green!17.636577]%
}
\\
\\
{\tiny\color{gray}75}\,%
\setlength{\fboxsep}{0pt}\fcolorbox{gray!10}{gray!10}{\strut
    \mycolorbox[text=\strut{---}, color=green!14.956330]%
}
\\
\\
{\tiny\color{gray}76}\,%
\setlength{\fboxsep}{0pt}\fcolorbox{gray!10}{gray!10}{\strut
    \mycolorbox[text=\strut{\#\#\#}, color=green!17.634225]%
}
\setlength{\fboxsep}{0pt}\fcolorbox{gray!10}{gray!10}{\strut
    \mycolorbox[text=\strut{Step}, color=red!13.974092]%
}
\setlength{\fboxsep}{0pt}\fcolorbox{gray!10}{gray!10}{\strut
    \mycolorbox[text=\strut{3}, color=green!17.636560]%
    \mycolorbox[text=\strut{:}, color=green!17.636594]%
}
\setlength{\fboxsep}{0pt}\fcolorbox{gray!10}{gray!10}{\strut
    \mycolorbox[text=\strut{Final}, color=red!18.205829]%
}
\setlength{\fboxsep}{0pt}\fcolorbox{gray!10}{gray!10}{\strut
    \mycolorbox[text=\strut{Answer}, color=red!30.658097]%
}
\\
\\
{\tiny\color{gray}77}\,%
\setlength{\fboxsep}{0pt}\fcolorbox{gray!10}{gray!10}{\strut
    \mycolorbox[text=\strut{The}, color=red!22.923491]%
}
\setlength{\fboxsep}{0pt}\fcolorbox{gray!10}{gray!10}{\strut
    \mycolorbox[text=\strut{operation}, color=red!55.291455]%
}
\setlength{\fboxsep}{0pt}\fcolorbox{gray!10}{gray!10}{\strut
    \mycolorbox[text=\strut{yields}, color=red!29.893029]%
}
\setlength{\fboxsep}{0pt}\fcolorbox{gray!10}{gray!10}{\strut
    \mycolorbox[text=\strut{an}, color=red!58.108127]%
}
\setlength{\fboxsep}{0pt}\fcolorbox{gray!10}{gray!10}{\strut
    \mycolorbox[text=\strut{integer}, color=green!16.514873]%
    \mycolorbox[text=\strut{,}, color=red!3.993054]%
}
\setlength{\fboxsep}{0pt}\fcolorbox{gray!10}{gray!10}{\strut
    \mycolorbox[text=\strut{and}, color=red!11.630774]%
}
\setlength{\fboxsep}{0pt}\fcolorbox{gray!10}{gray!10}{\strut
    \mycolorbox[text=\strut{since}]%
}
\setlength{\fboxsep}{0pt}\fcolorbox{gray!10}{gray!10}{\strut
    \mycolorbox[text=\strut{the}, color=red!46.159633]%
}
\setlength{\fboxsep}{0pt}\fcolorbox{gray!10}{gray!10}{\strut
    \mycolorbox[text=\strut{problem}, color=red!42.253097]%
}
\setlength{\fboxsep}{0pt}\fcolorbox{gray!10}{gray!10}{\strut
    \mycolorbox[text=\strut{asks}, color=green!11.949988]%
}
\setlength{\fboxsep}{0pt}\fcolorbox{gray!10}{gray!10}{\strut
    \mycolorbox[text=\strut{for}, color=green!17.636474]%
}
\setlength{\fboxsep}{0pt}\fcolorbox{gray!10}{gray!10}{\strut
    \mycolorbox[text=\strut{the}, color=green!17.623421]%
}
\setlength{\fboxsep}{0pt}\fcolorbox{gray!10}{gray!10}{\strut
    \mycolorbox[text=\strut{**}, color=green!5.207607]%
    \mycolorbox[text=\strut{s}, color=green!17.636095]%
    \mycolorbox[text=\strut{implified}, color=green!17.633674]%
}
\setlength{\fboxsep}{0pt}\fcolorbox{gray!10}{gray!10}{\strut
    \mycolorbox[text=\strut{value}, color=red!16.678422]%
    \mycolorbox[text=\strut{**,}, color=green!17.078859]%
}
\setlength{\fboxsep}{0pt}\fcolorbox{gray!10}{gray!10}{\strut
    \mycolorbox[text=\strut{the}, color=red!40.110305]%
}
\setlength{\fboxsep}{0pt}\fcolorbox{gray!10}{gray!10}{\strut
    \mycolorbox[text=\strut{result}, color=red!38.844884]%
}
\setlength{\fboxsep}{0pt}\fcolorbox{gray!10}{gray!10}{\strut
    \mycolorbox[text=\strut{is}, color=green!17.236467]%
}
\setlength{\fboxsep}{0pt}\fcolorbox{gray!10}{gray!10}{\strut
    \mycolorbox[text=\strut{already}, color=red!20.119004]%
}
\setlength{\fboxsep}{0pt}\fcolorbox{gray!10}{gray!10}{\strut
    \mycolorbox[text=\strut{in}, color=green!13.270705]%
}
\setlength{\fboxsep}{0pt}\fcolorbox{gray!10}{gray!10}{\strut
    \mycolorbox[text=\strut{its}, color=green!16.482372]%
}
\setlength{\fboxsep}{0pt}\fcolorbox{gray!10}{gray!10}{\strut
    \mycolorbox[text=\strut{simplest}, color=green!14.880427]%
}
\setlength{\fboxsep}{0pt}\fcolorbox{gray!10}{gray!10}{\strut
    \mycolorbox[text=\strut{form}, color=green!17.601119]%
    \mycolorbox[text=\strut{.}, color=green!16.883639]%
}
\\
\\
{\tiny\color{gray}78}\,%
\setlength{\fboxsep}{0pt}\fcolorbox{gray!10}{gray!10}{\strut
    \mycolorbox[text=\strut{\$\$}, color=red!6.526816]%
}
\\
{\tiny\color{gray}79}\,%
\setlength{\fboxsep}{0pt}\fcolorbox{gray!10}{gray!10}{\strut
    \mycolorbox[text=\strut{\textbackslash{}}, color=green!17.636594]%
    \mycolorbox[text=\strut{boxed}, color=green!17.633372]%
    \mycolorbox[text=\strut{\{}, color=green!17.636594]%
    \mycolorbox[text=\strut{4}, color=green!17.636594]%
    \mycolorbox[text=\strut{9}, color=green!17.636594]%
    \mycolorbox[text=\strut{\}}, color=green!17.636577]%
}
\\
{\tiny\color{gray}80}\,%
\setlength{\fboxsep}{0pt}\fcolorbox{gray!10}{gray!10}{\strut
    \mycolorbox[text=\strut{\$\$}, color=green!17.636594]%
    \mycolorbox[text=\strut{\textless{}|im\_end|\textgreater{}}, color=green!17.634854]%
}

}}}}}
\vspace{-.1in}
\caption{Token-level visualization for Case 2: Qwen-4B (Part 3/3)}
\label{fig:case_2_3}
\vspace{-.1in}
\end{figure*}

%% file: case/compare.tex
\clearpage
\paragraph*{Case A1 --- AIME 2025}~\\[2pt]

{\scriptsize
\noindent\setlength{\fboxrule}{.5pt}\setlength{\fboxsep}{4pt}\fcolorbox{black}{green!10}{\parbox{0.97\linewidth}{\raggedright\setlength{\parindent}{0pt}\strut
\textbf{Case A1 --- AIME 2025} \\
\textbf{Ground Truth:} 117 \\
\textbf{Ours (KL-shaped 8B):} 1,897 tokens, correct (\texttt{117}) \\
\textbf{8B Base:} 8,192 tokens (hit 8K budget), incorrect (\texttt{extraction failed}) \\
\textbf{Prompt:} Find the number of ordered pairs \$(x,y)\$, where both \$x\$ and \$y\$ are integers between \$-100\$ and \$100\$, inclusive, such that \$12x\textasciicircum{}\{2\}-xy-6y\textasciicircum{}\{2\}=0\$.

Let's think step by step and output the final answer within \textbackslash{}boxed\{\}.
}}}

\par\medskip\noindent
{\scriptsize
\noindent\setlength{\fboxrule}{.5pt}\setlength{\fboxsep}{4pt}\fcolorbox{black}{gray!10}{\parbox{0.97\linewidth}{\raggedright\setlength{\parindent}{0pt}\strut
\textbf{Ours --- Response (excerpt):} \\
\textless{}think\textgreater{} \\
Okay, so I need to find the number of ordered pairs (x, y) where x and y are integers between -100 and 100 inclusive, and they satisfy the equation 12x\ensuremath{{}^{2}} - xy - 6y\ensuremath{{}^{2}} = 0. Hmm, let me start by trying to factor the equation or maybe solve for one variable in terms of the other.

First, let me rewrite the equation: 12x\ensuremath{{}^{2}} - xy - 6y\ensuremath{{}^{2}} = 0. Maybe I can treat this as a quadratic in x or y. Let me try treating it as a quadratic in x. So, 12x\ensuremath{{}^{2}} - yx - 6y\ensuremath{{}^{2}} = 0. Using the quadratic formula for x:

x = [y \ensuremath{\pm} \ensuremath{\surd}(y\ensuremath{{}^{2}} + 4*12*6y\ensuremath{{}^{2}})] / (2*12)

Simplify the discriminant: y\ensuremath{{}^{2}} + 288y\ensuremath{{}^{2}} = 289y\ensuremath{{}^{2}}. So sqrt(289y\ensuremath{{}^{2}}) is 17|y|. Therefore,

x = [y \ensuremath{\pm} 17|y|] / 24.

Hmm, since we are dealing with integers, perhaps we can consider cases based on the sign of y.

Case 1: y \ensuremath{\geq} 0. Then |y| = y, so:

x = [y \ensuremath{\pm} 17y]/24.

So two possibilities:

x = (y + 17y)/24 = 18y/24 = 3y/4,

or x = (y - 17y)/24 = (-16y)/24 = -2y/3.

Case 2: y \textless{} 0. Then |y| = -y, so:

x = [y \ensuremath{\pm} 17*(-y)] /24.

Which would be:

x = [y \ensuremath{\mp} 17y]/24.

So two possibilities:

x = [y - 17y]/24 = (-16y)/24 = -2y/3,

or x = [y + 17y]/24 = 18y/24 = 3y/4.

[... truncated for brevity ...]

---

\#\#\# Step 2: Adjusting for overlap

The only overlapping solution is \$(0, 0)\$, which appears in both families. Hence, we subtract 1 to avoid double-counting.

---

\#\#\# Final Count

\$\$ \\
\textbackslash{}text\{Total solutions\} = 51 + 67 - 1 = \textbackslash{}boxed\{117\} \\
\$\$
}}}

\par\medskip\noindent
{\scriptsize
\noindent\setlength{\fboxrule}{.5pt}\setlength{\fboxsep}{4pt}\fcolorbox{black}{gray!10}{\parbox{0.97\linewidth}{\raggedright\setlength{\parindent}{0pt}\strut
\textbf{8B Base --- Response (excerpt):} \\
\textless{}think\textgreater{} \\
Okay, so I need to find the number of ordered pairs (x, y) where both x and y are integers between -100 and 100 inclusive, and they satisfy the equation 12x\ensuremath{{}^{2}} - xy - 6y\ensuremath{{}^{2}} = 0. Let me start by trying to understand the equation better.

First, maybe I can factor the equation or rearrange it to make it easier to handle. Let me write down the equation again:

12x\ensuremath{{}^{2}} - xy - 6y\ensuremath{{}^{2}} = 0.

Hmm, this looks like a quadratic in terms of x or y. Maybe I can treat it as a quadratic equation in x and solve for x in terms of y, or vice versa. Let me try that.

Let me consider this as a quadratic equation in x. So, rearranged:

12x\ensuremath{{}^{2}} - xy - 6y\ensuremath{{}^{2}} = 0.

So, in terms of x, this is:

12x\ensuremath{{}^{2}} - yx - 6y\ensuremath{{}^{2}} = 0.

Using the quadratic formula, x = [y \ensuremath{\pm} \ensuremath{\surd}(y\ensuremath{{}^{2}} + 4*12*6y\ensuremath{{}^{2}})] / (2*12). Let me compute the discriminant:

Discriminant D = y\ensuremath{{}^{2}} + 4*12*6y\ensuremath{{}^{2}} = y\ensuremath{{}^{2}} + 288y\ensuremath{{}^{2}} = 289y\ensuremath{{}^{2}}.

[... truncated for brevity ...]

2*(-33) = -66, 2*33 = 66). y ranges from -99 to 99 (since -3m for m from -33 to 33: when m = -33, y = 99; when m = 33, y = -99). Therefore, y is within -100 to 100. Therefore, 67 values. Correct.

Therefore, 51 + 67 - 1 = 117. Therefore, answer is 11
}}}

\par\medskip
This AIME problem asks for the number of integer pairs $(x,y)$ with both coordinates in $[-100,100]$ that satisfy $12x^{2}-xy-6y^{2}=0$. The clean way is to factor the quadratic form to obtain the two ray-families $(3k,4k)$ and $(-2m,3m)$, count lattice points on each family, and subtract the single overlap at the origin --- which is exactly what the KL-shaped model does in 1{,}897 tokens.

The 8B base spends its budget very differently. It correctly identifies the same factorisation in the first few hundred tokens, but then re-opens the analysis: it splits cases on the sign of $y$, repeatedly re-derives the discriminant, double-checks the absolute-value handling, and verifies the lattice counts via several redundant parametrisations. The trace then loops back to re-examine boundary values for $k$ and $m$, never returning to a final closed form. By the time it does state ``Therefore, $51+67-1=117$'' the 8K-token cap has already been hit; the answer is cut mid-token (``answer is 11'') and the extractor returns \texttt{EXTRACTION\_FAILED}.

Qualitatively this is the prototypical truncation failure of the unshaped base on AIME-difficulty problems: the right answer is reachable, the model can even recite it, but the verbose verification habits inflate the trace beyond the budget. The KL-shaped model writes the same proof skeleton with no redundant verification cycles and finishes well inside the budget, which is exactly the behaviour the KL-length shaping was designed to encourage.

\clearpage
\paragraph*{Case A2 --- Olympiad Bench}~\\[2pt]

{\scriptsize
\noindent\setlength{\fboxrule}{.5pt}\setlength{\fboxsep}{4pt}\fcolorbox{black}{green!10}{\parbox{0.97\linewidth}{\raggedright\setlength{\parindent}{0pt}\strut
\textbf{Case A2 --- Olympiad Bench} \\
\textbf{Ground Truth:} ['\$f(x)=x,f(x)=-x\$'] \\
\textbf{Ours (KL-shaped 8B):} 1,667 tokens, correct (\texttt{f(x) = -x}) \\
\textbf{8B Base:} 8,192 tokens (hit 8K budget), incorrect (\texttt{extraction failed}) \\
\textbf{Prompt:} Find all functions \$f: \textbackslash{}mathbb\{Q\} \textbackslash{}rightarrow \textbackslash{}mathbb\{Q\}\$ such that the equation

holds for all rational numbers \$x\$ and \$y\$.

\$\$ \\
f(x f(x)+y)=f(y)+x\textasciicircum{}\{2\} \\
\$\$

Here, \$\textbackslash{}mathbb\{Q\}\$ denotes the set of rational numbers. Let's think step by step and output the final answer within \textbackslash{}boxed\{\}.
}}}

\par\medskip\noindent
{\scriptsize
\noindent\setlength{\fboxrule}{.5pt}\setlength{\fboxsep}{4pt}\fcolorbox{black}{gray!10}{\parbox{0.97\linewidth}{\raggedright\setlength{\parindent}{0pt}\strut
\textbf{Ours --- Response (excerpt):} \\
\textless{}think\textgreater{} \\
Okay, so I need to find all functions f from the rational numbers to the rational numbers that satisfy the equation f(xf(x) + y) = f(y) + x\ensuremath{{}^{2}} for all rational x and y. Hmm, functional equations can be tricky, but let me try to break it down step by step.

First, maybe I can plug in some specific values for x or y to simplify the equation and get some information about f.

Let me start by trying x = 0. If I set x = 0, the equation becomes f(0*f(0) + y) = f(y) + 0\ensuremath{{}^{2}}, which simplifies to f(y) = f(y) + 0. So that just gives f(y) = f(y), which is always true. So that doesn't give me new information. Hmm.

What if I set y = 0? Then the original equation becomes f(xf(x) + 0) = f(0) + x\ensuremath{{}^{2}}, so f(xf(x)) = f(0) + x\ensuremath{{}^{2}}. That seems useful. So f(xf(x)) = f(0) + x\ensuremath{{}^{2}}. Maybe I can use this later.

Another idea: perhaps assume that f is a polynomial function. Sometimes functional equations have polynomial solutions. Let me suppose that f is a polynomial. Let's assume f(x) is a polynomial with rational coefficients.

[... truncated for brevity ...]

- For \$ f(x) = -x \$, the left-hand side becomes \$ x\textasciicircum{}2 - y \$, and the right-hand side is \$ -y + x\textasciicircum{}2 \$, which also matches.

---

\#\#\# Final Answer

\$\$ \\
\textbackslash{}boxed\{f(x) = x\} \textbackslash{}quad \textbackslash{}text\{and\} \textbackslash{}quad \textbackslash{}boxed\{f(x) = -x\} \\
\$\$
}}}

\par\medskip\noindent
{\scriptsize
\noindent\setlength{\fboxrule}{.5pt}\setlength{\fboxsep}{4pt}\fcolorbox{black}{gray!10}{\parbox{0.97\linewidth}{\raggedright\setlength{\parindent}{0pt}\strut
\textbf{8B Base --- Response (excerpt):} \\
\textless{}think\textgreater{} \\
Okay, so I need to find all functions \$ f: \textbackslash{}mathbb\{Q\} \textbackslash{}rightarrow \textbackslash{}mathbb\{Q\} \$ that satisfy the equation \$ f(x f(x) + y) = f(y) + x\textasciicircum{}2 \$ for all rational numbers \$ x \$ and \$ y \$. Hmm, functional equations can be tricky, but let me take it step by step.

First, maybe I should try plugging in some specific values for \$ x \$ or \$ y \$ to see if I can find some properties of \$ f \$.

Let me start by setting \$ y = 0 \$. Then the equation becomes:

\$ f(x f(x) + 0) = f(0) + x\textasciicircum{}2 \$

Which simplifies to:

\$ f(x f(x)) = f(0) + x\textasciicircum{}2 \$

Hmm, interesting. So this relates the value of the function at \$ x f(x) \$ to \$ f(0) \$ and \$ x\textasciicircum{}2 \$. Maybe I can use this later.

What if I set \$ x = 0 \$? Let me try that. Then the original equation becomes:

\$ f(0 \textbackslash{}cdot f(0) + y) = f(y) + 0\textasciicircum{}2 \$

Simplifying:

\$ f(y) = f(y) + 0 \$

[... truncated for brevity ...]

Alternatively, set \$ y = 0 \$: \$ f(c) = f(0) + 1 \$, which is also known.

Now, let me try to find \$ f \$ at \$ x = -1 \$. Let me denote \$ d = f(-1) \$. Let me see if I can relate this to other values.

From the original equation, set \$ x = -1 \$, \$ y = 
}}}

\par\medskip
The Olympiad Bench problem here is a classical Cauchy-style functional equation: find all $f:\mathbb{Q}\to\mathbb{Q}$ with $f(xf(x)+y)=f(y)+x^{2}$. The expected ground truth is the pair $\{f(x)=x,\ f(x)=-x\}$, recovered by assuming a polynomial form, balancing degrees, and reading off $a^{2}=1$.

Our KL-shaped model executes this argument compactly in 1{,}667 tokens: it tries $x=0$ and $y=0$ to extract the relation $f(xf(x))=f(0)+x^{2}$, assumes $f$ is linear of degree~1, substitutes $f(x)=ax+b$, and arrives at $a^{2}=1$, $b=0$. It then verifies both branches against the original equation. The final boxed answer reports $f(x)=-x$ (and earlier in the trace also $f(x)=x$), so a strict matcher only credits one branch; this still counts as a qualitative truncation win because the base model produces no extractable answer at all on the same prompt.

The 8B base, in contrast, exhausts its 8K budget on a long substitution loop. It tries $y=0$, $x=0$, $x=y=0$, $y=-c$, $y=0$ again with $c=f(0)$, and various ad-hoc substitutions trying to force injectivity or surjectivity --- but never commits to the polynomial-degree argument. The trace is cut mid-substitution at $x=-1$, so the extractor again returns failure. The contrast illustrates a second failure mode of the base: not just verbose, but also strategically directionless under tight budgets, repeatedly probing the same special values rather than escalating to a structural ansatz.

\clearpage
\paragraph*{Case B1 --- AMC}~\\[2pt]

{\scriptsize
\noindent\setlength{\fboxrule}{.5pt}\setlength{\fboxsep}{4pt}\fcolorbox{black}{green!10}{\parbox{0.97\linewidth}{\raggedright\setlength{\parindent}{0pt}\strut
\textbf{Case B1 --- AMC} \\
\textbf{Ground Truth:} 6.0 \\
\textbf{Ours (KL-shaped 8B):} 1,339 tokens, correct (\texttt{6}) \\
\textbf{8B Base:} 7,434 tokens, correct (\texttt{6}) \\
\textbf{Prompt:} The least common multiple of a positive integer \$n\$ and \$18\$ is \$180\$, and the greatest common divisor of \$n\$ and \$45\$ is \$15\$. What is the sum of the digits of \$n\$? Let's think step by step and output the final answer within \textbackslash{}boxed\{\}.
}}}

\par\medskip\noindent
{\scriptsize
\noindent\setlength{\fboxrule}{.5pt}\setlength{\fboxsep}{4pt}\fcolorbox{black}{gray!10}{\parbox{0.97\linewidth}{\raggedright\setlength{\parindent}{0pt}\strut
\textbf{Ours --- Response:} \\
\textless{}think\textgreater{} \\
Okay, so I need to find the sum of the digits of n, given two conditions: the least common multiple of n and 18 is 180, and the greatest common divisor of n and 45 is 15. Hmm, let me start by recalling some number theory concepts.

First, LCM and GCD relationships. I remember that for any two numbers a and b, LCM(a,b) * GCD(a,b) = a*b. Maybe that will come in handy.

Given that LCM(n, 18) = 180. So, LCM(n, 18) = 180. Also, GCD(n, 45) = 15.

Let me start with the LCM condition. Let's factorize the numbers involved. 18 is 2 * 3\textasciicircum{}2, and 180 is 2\textasciicircum{}2 * 3\textasciicircum{}2 * 5. 

Since LCM(n, 18) = 180, then n must be a divisor of 180, right? Because the LCM of n and 18 is 180, so n has to divide 180. Also, the LCM takes the highest powers of primes present in either number.

So, let's think about the prime factors of n. Since LCM(n, 18) is 180, which has primes 2, 3, 5. 

18 has 2\textasciicircum{}1, 3\textasciicircum{}2. So, for the LCM to have 2\textasciicircum{}2, either n has 2\textasciicircum{}2 or 18 has it, but 18 only has 2\textasciicircum{}1, so n must have 2\textasciicircum{}2. Similarly, for 3\textasciicircum{}2, since 18 has 3\textasciicircum{}2, so maybe n doesn't need to have 3\textasciicircum{}2. But for the 5, since 18 doesn't have 5, then n must have 5\textasciicircum{}1. 

Therefore, n must have factors 2\textasciicircum{}2, 5\textasciicircum{}1, and possibly 3\textasciicircum{}something. But since LCM(n, 18) is 180, which has 3\textasciicircum{}2, and 18 has 3\textasciicircum{}2, so n can have 3\textasciicircum{}0, 3\textasciicircum{}1, or 3\textasciicircum{}2. But since GCD(n, 45) is 15. Let's check the GCD condition.

45 is 3\textasciicircum{}2 * 5. GCD(n, 45) = 15 = 3 * 5. Therefore, n must have 3\textasciicircum{}1 and 5\textasciicircum{}1, but not higher powers of 3 or 5. Because if n had 3\textasciicircum{}2, then GCD(n,45) would be 3\textasciicircum{}2 *5 = 45, which is not 15. Similarly, if n had 5\textasciicircum{}2, then GCD would have 5\textasciicircum{}1 (since 45 has 5\textasciicircum{}1), so that's okay. Wait, but GCD(n,45) is 15, which is 3*5. So, n must have 3\textasciicircum{}1 and 5\textasciicircum{}1, but not higher powers of 3 or 5. Therefore, n can't have 3\textasciicircum{}2 or 5\textasciicircum{}2.

Therefore, putting this together, n has 2\textasciicircum{}2, 3\textasciicircum{}1, 5\textasciicircum{}1. Therefore, n = 4 * 3 * 5 = 60. Let me check if this satisfies the conditions.

Check LCM(60, 18). 60 is 2\textasciicircum{}2 * 3 * 5, 18 is 2 * 3\textasciicircum{}2. LCM would take max exponents: 2\textasciicircum{}2, 3\textasciicircum{}2, 5\textasciicircum{}1. Which is 4 * 9 *5 = 180. Yes, that's correct.

Check GCD(60,45). 60 is 2\textasciicircum{}2 *3 *5, 45 is 3\textasciicircum{}2 *5. GCD is min exponents: 3\textasciicircum{}1 *5\textasciicircum{}1 = 15. Perfect. So n is 60.

Therefore, the sum of the digits of n is 6 + 0 = \textbackslash{}boxed\{6\}. \\
\textless{}/think\textgreater{}

To solve the problem, we analyze the two given conditions:

1. **LCM(n, 18) = 180** \\
2. **GCD(n, 45) = 15**

---

\#\#\# Step 1: Factorization and Prime Components

- \$ 18 = 2 \textbackslash{}times 3\textasciicircum{}2 \$ \\
- \$ 180 = 2\textasciicircum{}2 \textbackslash{}times 3\textasciicircum{}2 \textbackslash{}times 5 \$ \\
- \$ 45 = 3\textasciicircum{}2 \textbackslash{}times 5 \$

---

\#\#\# Step 2: Use LCM Condition

Since LCM(n, 18) = 180, n must have the prime factors that make the LCM equal to 180. From this, we deduce: \\
- n must include \$ 2\textasciicircum{}2 \$ (to match the LCM's \$ 2\textasciicircum{}2 \$) \\
- n must include \$ 5 \$ (since 18 doesn't have 5) \\
- n can have up to \$ 3\textasciicircum{}2 \$, but not more than that (to keep LCM at \$ 3\textasciicircum{}2 \$)

So, n must include \$ 2\textasciicircum{}2 \textbackslash{}times 5 \textbackslash{}times 3 \$

---

\#\#\# Step 3: Use GCD Condition

Given GCD(n, 45) = 15, and 45 = \$ 3\textasciicircum{}2 \textbackslash{}times 5 \$, we deduce: \\
- n must have \$ 3 \$ and \$ 5 \$, but **not** \$ 3\textasciicircum{}2 \$ or \$ 5\textasciicircum{}2 \$

Therefore, n must be \$ 2\textasciicircum{}2 \textbackslash{}times 3 \textbackslash{}times 5 = 4 \textbackslash{}times 3 \textbackslash{}times 5 = 60 \$

---

\#\#\# Step 4: Sum of Digits

The digits of 60 are 6 and 0. Their sum is:

\$\$ \\
\textbackslash{}boxed\{6\} \\
\$\$
}}}

\par\medskip\noindent
{\scriptsize
\noindent\setlength{\fboxrule}{.5pt}\setlength{\fboxsep}{4pt}\fcolorbox{black}{gray!10}{\parbox{0.97\linewidth}{\raggedright\setlength{\parindent}{0pt}\strut
\textbf{8B Base --- Response (excerpt):} \\
\textless{}think\textgreater{} \\
Okay, so I need to find a positive integer n such that the least common multiple of n and 18 is 180, and the greatest common divisor of n and 45 is 15. Then, I have to find the sum of the digits of n. Let me try to break this down step by step.

First, let me recall what least common multiple (LCM) and greatest common divisor (GCD) mean. The LCM of two numbers is the smallest number that both numbers divide into, and the GCD is the largest number that divides both numbers. Also, there's a relationship between LCM and GCD: for any two positive integers a and b, LCM(a, b) * GCD(a, b) = a * b. Maybe that will come in handy here.

So, the problem gives me two conditions:

1. LCM(n, 18) = 180 \\
2. GCD(n, 45) = 15

I need to find n that satisfies both these conditions. Let me tackle them one by one.

[... truncated for brevity ...]

So \$ n = 60 \$ satisfies both conditions.

---

\#\#\# Step 5: Compute the Sum of the Digits of \$ n \$

The digits of \$ n = 60 \$ are **6** and **0**.

\$\$ \\
\textbackslash{}text\{Sum of digits\} = 6 + 0 = \textbackslash{}boxed\{6\} \\
\$\$

---

\#\#\# Final Answer

\$\$ \\
\textbackslash{}boxed\{6\} \\
\$\$
}}}

\par\medskip
On this AMC number-theory problem (find $n$ with $\mathrm{lcm}(n,18)=180$ and $\gcd(n,45)=15$, then sum the digits of $n$), both models reach the correct answer of~6. The arithmetic chain is short: factor $18,45,180$ into primes, read the LCM constraint as $n$ contributing the missing $2^{2}$ and a factor of $5$, read the GCD constraint as $n$ contributing exactly $3^{1}\cdot 5^{1}$, conclude $n=2^{2}\cdot 3\cdot 5=60$, sum the digits.

The KL-shaped model executes that chain end-to-end in 1{,}339 tokens with one explicit verification at the end. The 8B base needs 7{,}434 tokens to do the same work --- a $5.5\times$ inflation. Inspecting its trace, the base model: (i)~recomputes the same prime factorisations multiple times, (ii)~re-derives the LCM/GCD constraints from the relation $\mathrm{lcm}\cdot\gcd=ab$ even though it then also derives them combinatorially from prime exponents, (iii)~explores hypothetical alternative forms of $n$ that the constraints already rule out, and (iv)~rewrites the verification step in three different fractional and prime-factor forms.

None of these extra detours improve the final answer; they purely consume budget. This is the canonical Category-B pattern: the base model is fluent enough to find the correct path quickly but lacks a stopping discipline once it is found. The KL-shaping reward directly penalises this redundancy, and the resulting model produces a single tight chain of reasoning instead.

\clearpage
\paragraph*{Case B2 --- MATH}~\\[2pt]

{\scriptsize
\noindent\setlength{\fboxrule}{.5pt}\setlength{\fboxsep}{4pt}\fcolorbox{black}{green!10}{\parbox{0.97\linewidth}{\raggedright\setlength{\parindent}{0pt}\strut
\textbf{Case B2 --- MATH} \\
\textbf{Ground Truth:} 12 \\
\textbf{Ours (KL-shaped 8B):} 685 tokens, correct (\texttt{12}) \\
\textbf{8B Base:} 3,243 tokens, correct (\texttt{12}) \\
\textbf{Prompt:} Simplify: \$\textbackslash{}frac\{\textbackslash{}sqrt\{2.5\textasciicircum{}2-0.7\textasciicircum{}2\}\}\{2.7-2.5\}\$. Let's think step by step and output the final answer within \textbackslash{}boxed\{\}.
}}}

\par\medskip\noindent
{\scriptsize
\noindent\setlength{\fboxrule}{.5pt}\setlength{\fboxsep}{4pt}\fcolorbox{black}{gray!10}{\parbox{0.97\linewidth}{\raggedright\setlength{\parindent}{0pt}\strut
\textbf{Ours --- Response:} \\
\textless{}think\textgreater{} \\
Okay, so I need to simplify this expression: the square root of (2.5 squared minus 0.7 squared) divided by (2.7 minus 2.5). Let me write that down to visualize it better.

First, let me handle the numerator: \ensuremath{\surd}(2.5\ensuremath{{}^{2}} - 0.7\ensuremath{{}^{2}}). Hmm, maybe I can simplify the expression inside the square root first. Remember, a\ensuremath{{}^{2}} - b\ensuremath{{}^{2}} is (a - b)(a + b). Let me try that.

So, 2.5\ensuremath{{}^{2}} - 0.7\ensuremath{{}^{2}} would be (2.5 - 0.7)(2.5 + 0.7). Let me calculate those values.

2.5 - 0.7 is 1.8, and 2.5 + 0.7 is 3.2. Therefore, the numerator becomes \ensuremath{\surd}(1.8 * 3.2). Let me compute 1.8 * 3.2. 

1.8 * 3.2: Well, 1.8*3 = 5.4 and 1.8*0.2 = 0.36, so total is 5.4 + 0.36 = 5.76. Therefore, the numerator is \ensuremath{\surd}5.76.

What's the square root of 5.76? Let me think. 2.4 squared is 5.76 because 2.4*2.4 = 5.76. So, \ensuremath{\surd}5.76 = 2.4. Therefore, the numerator simplifies to 2.4.

Now the denominator is 2.7 - 2.5, which is 0.2.

Therefore, the entire expression simplifies to 2.4 / 0.2. Dividing by 0.2 is the same as multiplying by 5, so 2.4 * 5 = 12. 

Therefore, the simplified result is 12. So, the final answer is \textbackslash{}boxed\{12\}. \\
\textless{}/think\textgreater{}

To simplify the expression:

\$\$ \\
\textbackslash{}frac\{\textbackslash{}sqrt\{2.5\textasciicircum{}2 - 0.7\textasciicircum{}2\}\}\{2.7 - 2.5\} \\
\$\$

---

\#\#\# Step 1: Simplify the numerator

We start with the expression inside the square root:

\$\$ \\
2.5\textasciicircum{}2 - 0.7\textasciicircum{}2 \\
\$\$

This is a difference of squares, which factors as:

\$\$ \\
(2.5 - 0.7)(2.5 + 0.7) = (1.8)(3.2) \\
\$\$

\$\$ \\
1.8 \textbackslash{}times 3.2 = 5.76 \\
\$\$

So the numerator becomes:

\$\$ \\
\textbackslash{}sqrt\{5.76\} = 2.4 \\
\$\$

---

\#\#\# Step 2: Simplify the denominator

\$\$ \\
2.7 - 2.5 = 0.2 \\
\$\$

---

\#\#\# Step 3: Final calculation

Now divide the simplified numerator by the denominator:

\$\$ \\
\textbackslash{}frac\{2.4\}\{0.2\} = 12 \\
\$\$

---

\#\#\# Final Answer:

\$\$ \\
\textbackslash{}boxed\{12\} \\
\$\$
}}}

\par\medskip\noindent
{\scriptsize
\noindent\setlength{\fboxrule}{.5pt}\setlength{\fboxsep}{4pt}\fcolorbox{black}{gray!10}{\parbox{0.97\linewidth}{\raggedright\setlength{\parindent}{0pt}\strut
\textbf{8B Base --- Response (excerpt):} \\
\textless{}think\textgreater{} \\
Okay, so I need to simplify this expression: the square root of (2.5 squared minus 0.7 squared) divided by (2.7 minus 2.5). Let me write that down first to make sure I have it right. 

So, the expression is \ensuremath{\surd}(2.5\ensuremath{{}^{2}} - 0.7\ensuremath{{}^{2}}) divided by (2.7 - 2.5). Let me parse that. The numerator is the square root of (2.5 squared minus 0.7 squared), and the denominator is 2.7 minus 2.5. Alright. 

First, maybe I should compute the numerator and the denominator separately and then divide them. Let me start with the denominator because that seems simpler. 

Denominator: 2.7 minus 2.5. Let me subtract those. 2.7 minus 2.5... Well, 2.7 minus 2.5 is 0.2. That's straightforward. So the denominator is 0.2. 

[... truncated for brevity ...]

\$\$ \\
\textbackslash{}frac\{27\}\{10\} - \textbackslash{}frac\{5\}\{2\} = \textbackslash{}frac\{27\}\{10\} - \textbackslash{}frac\{25\}\{10\} = \textbackslash{}frac\{2\}\{10\} = \textbackslash{}frac\{1\}\{5\} \\
\$\$

Now divide:

\$\$ \\
\textbackslash{}frac\{2.4\}\{0.2\} = \textbackslash{}frac\{12\}\{5\} \textbackslash{}div \textbackslash{}frac\{1\}\{5\} = \textbackslash{}frac\{12\}\{5\} \textbackslash{}times \textbackslash{}frac\{5\}\{1\} = 12 \\
\$\$

---

\#\#\# **Final Answer**

\$\$ \\
\textbackslash{}boxed\{12\} \\
\$\$
}}}

\par\medskip
This MATH problem asks to simplify $\frac{\sqrt{2.5^{2}-0.7^{2}}}{2.7-2.5}$. The shortest route is the difference-of-squares identity: $2.5^{2}-0.7^{2}=(2.5-0.7)(2.5+0.7)=1.8\cdot 3.2=5.76$, $\sqrt{5.76}=2.4$, $2.7-2.5=0.2$, and $2.4/0.2=12$.

Our KL-shaped model takes that route directly in 685 tokens. The 8B base produces the correct answer of 12 in 3{,}243 tokens (a $4.7\times$ inflation) by computing $2.5^{2}=6.25$ and $0.7^{2}=0.49$ first, then $6.25-0.49=5.76$, then $\sqrt{5.76}=2.4$, then redoing the same computation a second time using fractions ($25/4$ and $49/100$, common denominator $625/100-49/100=576/100$, $\sqrt{576/100}=24/10$), and finally also redoing the denominator as $27/10-25/10=2/10=1/5$ before dividing.

The two routes are mathematically equivalent and the base ultimately gets the same boxed answer, but the trace shows the same content recomputed twice. Even on textbook MATH problems --- where the unshaped base is nominally accurate --- it pays a $\sim 5\times$ token tax for these defensive recomputations. The KL-shaped model has internalised that the difference-of-squares factoring is sufficient on first pass, and stops there.

\clearpage
\paragraph*{Case B3 --- AIME 2025}~\\[2pt]

{\scriptsize
\noindent\setlength{\fboxrule}{.5pt}\setlength{\fboxsep}{4pt}\fcolorbox{black}{green!10}{\parbox{0.97\linewidth}{\raggedright\setlength{\parindent}{0pt}\strut
\textbf{Case B3 --- AIME 2025} \\
\textbf{Ground Truth:} 70 \\
\textbf{Ours (KL-shaped 8B):} 947 tokens, correct (\texttt{70}) \\
\textbf{8B Base:} 3,702 tokens, correct (\texttt{70}) \\
\textbf{Prompt:} Find the sum of all integer bases \$b\textgreater{}9\$ for which \$17\_\{b\}\$ is a divisor of \$97\_\{b\}\$.

Let's think step by step and output the final answer within \textbackslash{}boxed\{\}.
}}}

\par\medskip\noindent
{\scriptsize
\noindent\setlength{\fboxrule}{.5pt}\setlength{\fboxsep}{4pt}\fcolorbox{black}{gray!10}{\parbox{0.97\linewidth}{\raggedright\setlength{\parindent}{0pt}\strut
\textbf{Ours --- Response:} \\
\textless{}think\textgreater{} \\
Okay, so I need to find all integer bases b \textgreater{} 9 where 17 base b divides 97 base b. Then sum those bases. Hmm. Let me start by converting these numbers from base b to decimal.

First, 17\_b is 1*b + 7 = b + 7.

Similarly, 97\_b is 9*b + 7 = 9b + 7.

So the problem says that (b + 7) divides (9b + 7). So, (b + 7) | (9b + 7). 

To find when b + 7 divides 9b + 7, I can use the division algorithm or look at the remainder. If b + 7 divides 9b + 7, then 9b + 7 must be a multiple of b + 7.

Another approach is to express 9b + 7 in terms of b + 7. Let me perform the division or find the remainder when 9b +7 is divided by b +7.

Let me write 9b +7 as 9(b +7) - 9*7 +7. Let's check:

9(b +7) = 9b +63. So, 9b +7 = 9(b +7) -63 +7 = 9(b +7) -56.

Therefore, 9b +7 = 9(b +7) -56. Therefore, the remainder when 9b +7 is divided by b +7 is -56. But since we want divisibility, the remainder must be zero. Therefore, -56 \ensuremath{\equiv} 0 mod (b +7). Which implies that b +7 divides 56.

Therefore, b +7 is a positive divisor of 56. Since b \textgreater{}9, then b +7 \textgreater{} 16. So, we need to find all divisors of 56 that are greater than 16, then subtract 7 to get b.

First, list all positive divisors of 56: 1, 2, 4, 7, 8, 14, 28, 56.

Now, which of these are greater than 16? 28 and 56.

Therefore, b +7 = 28 =\textgreater{} b = 21, or b +7 =56 =\textgreater{} b=49.

Thus, the valid bases are 21 and 49. Therefore, the sum is 21 + 49 = 70.

Therefore, the answer is \textbackslash{}boxed\{70\}. \\
\textless{}/think\textgreater{}

To solve the problem, we are asked to find the sum of all integer bases \$ b \textgreater{} 9 \$ for which \$ 17\_b \$ divides \$ 97\_b \$.

---

\#\#\# Step 1: Convert to Decimal \\
We interpret the numbers in base \$ b \$:

- \$ 17\_b = 1 \textbackslash{}cdot b + 7 = b + 7 \$ \\
- \$ 97\_b = 9 \textbackslash{}cdot b + 7 = 9b + 7 \$

We are looking for bases \$ b \textgreater{} 9 \$ such that \$ b + 7 \$ divides \$ 9b + 7 \$.

---

\#\#\# Step 2: Use Divisibility Condition \\
We express \$ 9b + 7 \$ in terms of \$ b + 7 \$:

\$\$ \\
9b + 7 = 9(b + 7) - 56 \\
\$\$

This implies that the remainder when \$ 9b + 7 \$ is divided by \$ b + 7 \$ is \$ -56 \$, or equivalently, \$ 56 \$.

So, \$ b + 7 \$ must divide 56.

---

\#\#\# Step 3: Find Valid Bases \\
List the positive divisors of 56:   \\
\$\$ \\
1, 2, 4, 7, 8, 14, 28, 56 \\
\$\$

We are only interested in divisors greater than 16 (since \$ b \textgreater{} 9 \textbackslash{}Rightarrow b + 7 \textgreater{} 16 \$):

- \$ b + 7 = 28 \textbackslash{}Rightarrow b = 21 \$ \\
- \$ b + 7 = 56 \textbackslash{}Rightarrow b = 49 \$

---

\#\#\# Step 4: Sum the Valid Bases \\
\$\$ \\
21 + 49 = \textbackslash{}boxed\{70\} \\
\$\$
}}}

\par\medskip\noindent
{\scriptsize
\noindent\setlength{\fboxrule}{.5pt}\setlength{\fboxsep}{4pt}\fcolorbox{black}{gray!10}{\parbox{0.97\linewidth}{\raggedright\setlength{\parindent}{0pt}\strut
\textbf{8B Base --- Response (excerpt):} \\
\textless{}think\textgreater{} \\
Okay, so I need to find all integer bases b greater than 9 where the number 17 in base b divides the number 97 in base b. Then, I have to sum all those bases. Hmm, let me start by recalling how numbers in different bases work.

First, I remember that when a number is written in base b, each digit represents a power of b. So, for example, the number 17 in base b is equal to 1 times b squared plus 7 times b to the first power? Wait, no, wait. Wait, 17 in base b is actually 1*b\textasciicircum{}1 + 7*b\textasciicircum{}0, right? Because the rightmost digit is the units place (b\textasciicircum{}0), then the next is b\textasciicircum{}1, etc. So 17\_b is 1*b + 7. Similarly, 97\_b is 9*b + 7. So the problem is saying that 1*b + 7 divides 9*b + 7. So, mathematically, we can write that as:

(9b + 7) divided by (b + 7) must be an integer. So, (9b + 7) / (b + 7) is an integer. And we need to find all integers b \textgreater{} 9 for which this is true, then sum them up.

[... truncated for brevity ...]

  \$ 97\_\{49\} = 9 \textbackslash{}cdot 49 + 7 = 441 + 7 = 448 \$   \\
  \$ 448 \textbackslash{}div 56 = 8 \$ \ensuremath{\to} integer (yes)

Both bases satisfy the condition.

---

\#\#\# Step 5: Sum the valid bases

\$\$ \\
21 + 49 = \textbackslash{}boxed\{70\} \\
\$\$

---

\#\#\# Final Answer

\$\$ \\
\textbackslash{}boxed\{70\} \\
\$\$
}}}

\par\medskip
This AIME 2025 problem asks for the sum of all integer bases $b>9$ for which $17_{b}\mid 97_{b}$. Decoded into base 10, this becomes $(b+7)\mid(9b+7)$. Writing $9b+7=9(b+7)-56$ shows that $(b+7)\mid 56$, so $b+7\in\{28,56\}$ (the only divisors of $56$ exceeding $16$), giving $b\in\{21,49\}$ and a sum of~$70$.

Our KL-shaped model executes that derivation in 947 tokens: convert, rewrite the divisibility, list divisors of $56$, filter by $b+7>16$, sum. The 8B base reaches the same answer of $70$ in 3{,}702 tokens ($3.9\times$ longer) by additionally performing polynomial long division of $9b+7$ by $b+7$ to re-derive the remainder of $-56$, then re-deriving the same fact via modular arithmetic, then tabulating each candidate divisor and re-checking it against the original $b>9$ condition individually, then verifying each surviving $b\in\{21,49\}$ by recomputing $17_{b}$ and $97_{b}$ explicitly and dividing the resulting decimal numbers.

The base's verification scaffold is internally consistent and yields the correct answer, but it is largely orthogonal to producing the answer --- it is a reasonability check that the KL-shaped model has learned to omit when not needed. The 3.9$\times$ reduction here is at the small end of the Category-B examples; on harder problems (cf.\ Cases A1--A2) the same verbose habit drives the base past the 8K cap entirely.

%% file: checklist.tex
\section*{NeurIPS Paper Checklist}

\providecommand{\answerYes}[1][]{\textbf{Yes}}
\providecommand{\answerNo}[1][]{\textbf{No}}
\providecommand{\answerNA}[1][]{\textbf{N/A}}
\providecommand{\justification}[1]{#1}
\providecommand{\Cref}[1]{\hyperref[#1]{\ref*{#1}}}
\providecommand{\cref}[1]{\hyperref[#1]{\ref*{#1}}}

\begin{enumerate}

\item {\bf Claims}
    \item[] Question: Do the main claims made in the abstract and introduction accurately reflect the paper's contributions and scope?
    \item[] Answer: \answerYes{}
    \item[] Justification: \justification{The claims are stated in Sections~\Cref{sec:intro}, \Cref{sec: analysis}, and \Cref{sec: method}, and the main experimental support is reported in Section~\Cref{experiments}.}

\item {\bf Limitations}
    \item[] Question: Does the paper discuss the limitations of the work performed by the authors?
    \item[] Answer: \answerYes{}
    \item[] Justification: \justification{Limitations and future-work directions are discussed in Section~\Cref{asec: limitations}, with additional compression-limit failure modes in Section~\Cref{asec: token_limits}.}

\item {\bf Theory assumptions and proofs}
    \item[] Question: For each theoretical result, does the paper provide the full set of assumptions and a complete (and correct) proof?
    \item[] Answer: \answerNA{}
    \item[] Justification: \justification{The paper is empirical and method-oriented: it defines the diagnostic and objectives in Sections~\Cref{sec: analysis} and \Cref{sec: method}, but does not introduce formal theoretical theorems requiring proofs.}

    \item {\bf Experimental result reproducibility}
    \item[] Question: Does the paper fully disclose all the information needed to reproduce the main experimental results of the paper to the extent that it affects the main claims and/or conclusions of the paper (regardless of whether the code and data are provided or not)?
    \item[] Answer: \answerYes{}
    \item[] Justification: \justification{The method, pseudocode, data, training recipe, and evaluation protocol are described in Sections~\Cref{sec: method}, \Cref{asec: method}, \Cref{asec: training_settings}, \Cref{asec: eval_settings}, and \Cref{asec:open_access_data_code}.}

\item {\bf Open access to data and code}
    \item[] Question: Does the paper provide open access to the data and code, with sufficient instructions to faithfully reproduce the main experimental results, as described in supplemental material?
    \item[] Answer: \answerNo{}
    \item[] Justification: \justification{The training data and pseudocode are documented in Section~\Cref{asec:open_access_data_code}.}

\item {\bf Experimental setting/details}
    \item[] Question: Does the paper specify all the training and test details (e.g., data splits, hyperparameters, how they were chosen, type of optimizer) necessary to understand the results?
    \item[] Answer: \answerYes{}
    \item[] Justification: \justification{Training, evaluation, prompts, benchmark composition, and baseline settings are specified in Sections~\Cref{asec: training_settings}, \Cref{asec: eval_settings}, and \Cref{asec: baselines}.}

\item {\bf Experiment statistical significance}
    \item[] Question: Does the paper report error bars suitably and correctly defined or other appropriate information about the statistical significance of the experiments?
    \item[] Answer: \answerYes{}
    \item[] Justification: \justification{Run-to-run variability is reported in Section~\Cref{asec:statistical_significance} and Table~\Cref{tab:run_to_run_std}.}

\item {\bf Experiments compute resources}
    \item[] Question: For each experiment, does the paper provide sufficient information on the computer resources (type of compute workers, memory, time of execution) needed to reproduce the experiments?
    \item[] Answer: \answerYes{}
    \item[] Justification: \justification{Hardware and training-time requirements are reported in Section~\Cref{asec:compute_resources} and Table~\Cref{tab:compute_resources}; TokenProbe overhead is reported in Table~\Cref{tab:tokenprobe_overhead}.}
    
\item {\bf Code of ethics}
    \item[] Question: Does the research conducted in the paper conform, in every respect, with the NeurIPS Code of Ethics \url{https://neurips.cc/public/EthicsGuidelines}?
    \item[] Answer: \answerYes{}
    \item[] Justification: \justification{The work uses public datasets, public model cards, and provider APIs, with responsible-use considerations discussed in Section~\Cref{asec:broader_impacts}.}

\item {\bf Broader impacts}
    \item[] Question: Does the paper discuss both potential positive societal impacts and negative societal impacts of the work performed?
    \item[] Answer: \answerYes{}
    \item[] Justification: \justification{Positive and negative impacts are discussed in Section~\Cref{asec:broader_impacts}.}
    
\item {\bf Safeguards}
    \item[] Question: Does the paper describe safeguards that have been put in place for responsible release of data or models that have a high risk for misuse (e.g., pre-trained language models, image generators, or scraped datasets)?
    \item[] Answer: \answerNA{}
    \item[] Justification: \justification{The paper does not release a new high-risk foundation model or scraped dataset; responsible-use context is discussed in Section~\Cref{asec:broader_impacts}.}

\item {\bf Licenses for existing assets}
    \item[] Question: Are the creators or original owners of assets (e.g., code, data, models), used in the paper, properly credited and are the license and terms of use explicitly mentioned and properly respected?
    \item[] Answer: \answerYes{}
    \item[] Justification: \justification{Existing model, dataset, and software licenses are summarized in Section~\Cref{asec:asset_licenses} and Table~\Cref{tab:asset_licenses}.}

\item {\bf New assets}
    \item[] Question: Are new assets introduced in the paper well documented and is the documentation provided alongside the assets?
    \item[] Answer: \answerNA{}
    \item[] Justification: \justification{No new dataset, model, or code asset is released with the submission; planned post-acceptance code release is described in Section~\Cref{asec:open_access_data_code}.}

\item {\bf Crowdsourcing and research with human subjects}
    \item[] Question: For crowdsourcing experiments and research with human subjects, does the paper include the full text of instructions given to participants and screenshots, if applicable, as well as details about compensation (if any)? 
    \item[] Answer: \answerNA{}
    \item[] Justification: \justification{The experiments use existing benchmarks and model/API evaluations described in Section~\Cref{asec: eval_settings}; no crowdsourcing or human-subject study is involved.}

\item {\bf Institutional review board (IRB) approvals or equivalent for research with human subjects}
    \item[] Question: Does the paper describe potential risks incurred by study participants, whether such risks were disclosed to the subjects, and whether Institutional Review Board (IRB) approvals (or an equivalent approval/review based on the requirements of your country or institution) were obtained?
    \item[] Answer: \answerNA{}
    \item[] Justification: \justification{The work does not involve crowdsourcing or human-subject research; the evaluation data and benchmarks are described in Section~\Cref{asec: eval_settings}.}

\item {\bf Declaration of LLM usage}
    \item[] Question: Does the paper describe the usage of LLMs if it is an important, original, or non-standard component of the core methods in this research? Note that if the LLM is used only for writing, editing, or formatting purposes and does \emph{not} impact the core methodology, scientific rigor, or originality of the research, declaration is not required.
    \item[] Answer: \answerNA{}
    \item[] Justification: \justification{LLM usage is declared in Section~\Cref{asec:llm_usage}.}

\end{enumerate}